\documentclass{article}

     \PassOptionsToPackage{numbers, compress}{natbib}

     \usepackage[preprint]{neurips_2023}

\usepackage[utf8]{inputenc} 
\usepackage[T1]{fontenc}    
\usepackage[hidelinks]{hyperref}       
\usepackage{url}            
\usepackage{booktabs}       
\usepackage{amsfonts}       
\usepackage{nicefrac}       
\usepackage{microtype}      
\usepackage{xcolor}         

\usepackage{subfig}

\usepackage{multirow}

\usepackage{algorithm}
\usepackage{algpseudocode}
\algdef{S}[FOR]{ForEach}[1]{\algorithmicforeach\ #1\ \algorithmicdo}

\usepackage{amssymb}
\usepackage{amsmath}
\usepackage{mathtools}
\DeclareMathAlphabet{\mathcal}{OMS}{cmsy}{m}{n}

\newcommand {\ydarnew}[1]{}
\newcommand {\presubmissioncomment}[1]{}
\newcommand {\selfcomment}[1]{}
\newcommand {\ydar}[1]{}
\newcommand\lolo[1]{}
\newcommand\richb[1]{}

\title{Frozen Overparameterization:\\A Double Descent Perspective on Transfer Learning of Deep Neural Networks}

\author{%
Yehuda Dar \\
  Ben-Gurion University \\
  \texttt{ydar@bgu.ac.il} \\
  \And 
  Lorenzo Luzi \\
  Rice University \\
  \texttt{enzo@rice.edu} \\
  \And 
  Richard G.~Baraniuk\\
  Rice University \\
  \texttt{richb@rice.edu} \\
}

\begin{document}

\maketitle

\begin{abstract}
We study the generalization behavior of transfer learning of deep neural networks (DNNs). 
We adopt the overparameterization perspective --- featuring interpolation of the training data (i.e., approximately zero train error) and the double descent phenomenon --- to explain the delicate effect of the transfer learning setting on generalization performance. 
We study how the generalization behavior of transfer learning is affected by the dataset size in the source and target tasks, the number of transferred layers that are kept frozen in the target DNN training, and the similarity between the source and target tasks. 
We show that the test error evolution during the target DNN training has a more significant double descent effect when the target training dataset is sufficiently large.
In addition, a larger source training dataset can yield a slower target DNN training. \ydarnew{removed from arxiv v2: ``with a lower double descent peak."} 
Moreover, we demonstrate that the number of frozen layers can determine whether the transfer learning is effectively underparameterized or overparameterized and, in turn, this may induce a freezing-wise double descent phenomenon that determines the relative success or failure of learning. 
Also, we show that the double descent phenomenon may make a transfer from a less related source task better than a transfer from a more related source task.
We establish our results using image classification experiments with the ResNet, DenseNet and the vision transformer (ViT) architectures. 
\end{abstract}

\section{Introduction}
\label{sec:intro}
Transfer learning is a common practice for training deep neural networks (DNNs) based on pre-trained DNNs (e.g., \cite{pan2009survey,bengio2012deep,shin2016deep,long2017deep,wang2020fewshotlearningsurvey}).
The learning from scratch of the vast number of parameters in modern DNNs requires considerable amounts of training data and computational resources. Both data and compute power are often lacking, a key issue that can be addressed via transfer learning. 
The implementation of transfer learning requires a series of design choices such as the number of layers to transfer from a given pre-trained (source) DNN and how much to allow them to change in the training of the target task. Specifically, the pre-trained transferred layers can be set fixed (``frozen'') or serve as initialization in a short (fine tuning) or long training process.  
These design choices are pivotal and affect the generalization performance in intricate ways that are still far from being sufficiently understood \cite{yosinski2014how,raghu2019transfusion,kornblith2019better,ericsson2021how,mensink2021factors}, leaving the common practice to extensively rely on trial-and-error.

Modern DNNs are \textit{overparameterized} models, namely, they have much more learnable parameters than training data samples; moreover, such models are usually trained to interpolate (i.e., perfectly fit, having training error numerically zero) their training data \cite{zhang2017understanding,belkin2019reconciling,zhang2021understanding,dar2021farewell}.
Despite interpolating their training data, DNNs can generalize excellently to (test) data beyond their training dataset; this contradicts the conventional machine learning guideline that associates overfitting with poor generalization performance.

\begin{figure*}[t]
	\begin{center}
   \subfloat[]{\includegraphics[width=0.487\textwidth]{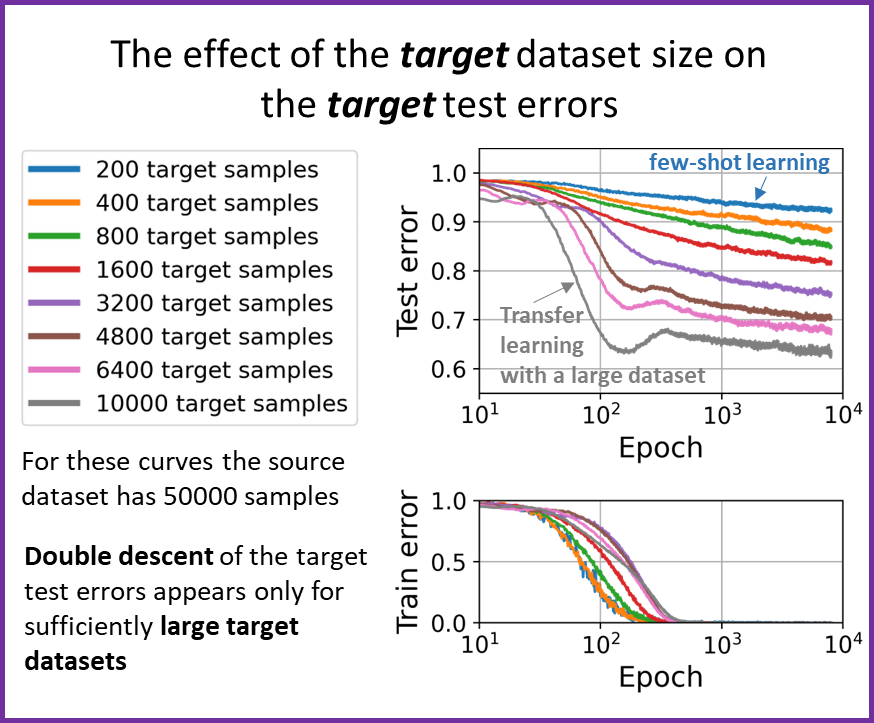}
\label{fig:transfer_learning_double_descent_target_dataset_size}}
   \subfloat[]{\includegraphics[width=0.49\textwidth]{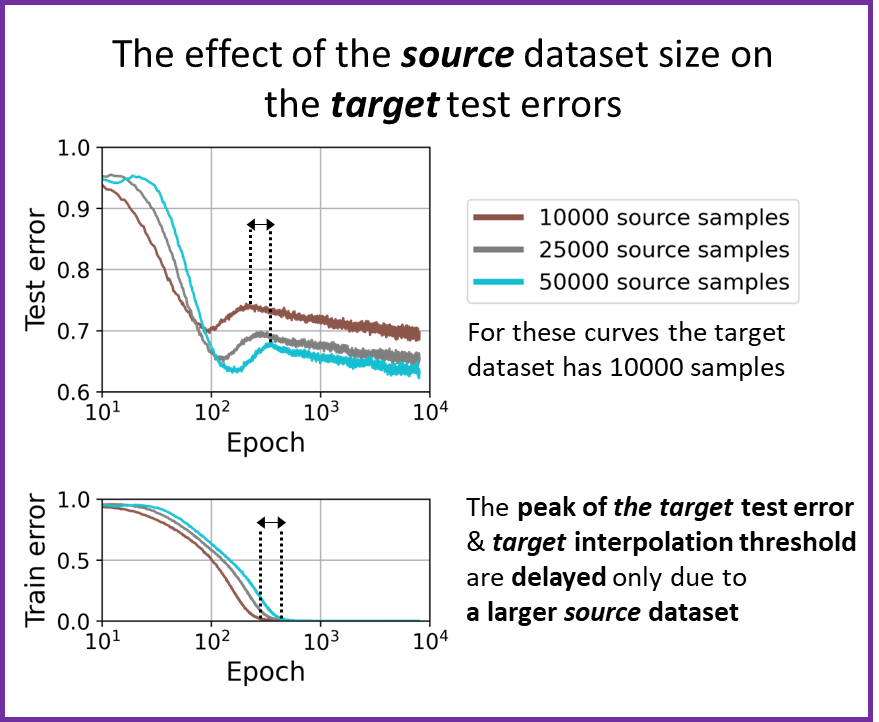}
\label{fig:transfer_learning_double_descent_source_dataset_size}}
\caption{Double descent in transfer learning of a ResNet-18 for classification of CIFAR-100 (all the 100 classes, target data has 20\% label noise). The pre-trained source model is a ResNet-18 for classification of 100 classes from Tiny ImageNet (image size 32x32x3, source data does not have label noise). This figure shows how the training evolution of the errors of the target model is affected by (a) the target training dataset size for the same pre-trained source model, and (b) the source training dataset size for the same target dataset. The legends specify the total size (i.e., number of samples for all the 100 classes together) of the examined training datasets. See more details in Section \ref{sec:double descent in transfer learning}.}
		\label{fig:transfer_learning_double_descent_dataset_size}
	\end{center}
\end{figure*}

Many overparameterized models have test errors that follow a \textit{double descent} shape when examined with respect to the learned model complexity (e.g., number of learnable parameters) \cite{belkin2019reconciling}. This double descent behavior has two parts: In the underparameterized regime, where the learned model cannot interpolate its training data, the test error follows the ``U-shape'' of the classical bias--variance tradeoff; in the overparameterized regime, where the learned model interpolates its training data, the test error can decrease as the learned model complexity increases (e.g., more learnable parameters). 

The double descent phenomenon in DNNs has received a detailed empirical analysis in  \cite{nakkiran2019deep}. Specifically, evaluating the test and train errors (i.e., the relative portion of wrong classification in the test and train datasets, respectively) for a range of DNN widths and in each of the many training epochs shows two kinds of double descent phenomena of the test error: (i) as a function of the model width, and (ii) as a function of the training epoch; both have a peaking test error around or towards the entrance to the interpolation regime where the train error is zero (or numerically close to zero). These double descent phenomena are usually more noticeable when the data is noisy (e.g., label noise in classification problems). 
The results in \cite{nakkiran2019deep} consider diverse learning settings and, yet, all of them are for learning DNNs from scratch.

In this paper we go considerably beyond the scope of \cite{nakkiran2019deep} and study the generalization and double descent phenomenon in transfer learning. 
Our first contribution is the characterization of the transfer learning settings where the double descent phenomenon is likely to emerge. 
As in learning from scratch, label noise in the training dataset is an important promoter of double descent\footnote{Our results show that epoch-wise double descent can emerge in DenseNet and ViT training from noiseless datasets, whereas label noise is indeed crucial for epoch-wise double descent in ResNet training.}; yet, in transfer learning, such label noise needs to exist only in the target dataset and not necessarily in the source dataset. Moreover, 
{\textbf{double descent in transfer learning requires a sufficiently large target dataset}} (Fig.~\ref{fig:transfer_learning_double_descent_target_dataset_size}); this implies that double descent is less likely to appear in few-shot learning settings (i.e., when a very limited number of training samples are available for the target task \cite{wang2020fewshotlearningsurvey}). 
Also, we show that the size of the \textbf{\textit{source} training dataset can determine the speed of which the transfer learning of the target task achieves interpolation of the target dataset} (Fig.~\ref{fig:transfer_learning_double_descent_source_dataset_size}).  



Our second contribution is in showing how freezing the transferred pre-trained layers affects the generalization behavior. 
Freezing layers determines the number of learnable parameters in the target DNN and, hence, it is a unique way of transfer learning to control the parameterization level without changing the overall DNN architecture. 
In transfer learning with a large target dataset, we show that freezing (relatively many) layers can eliminate the epoch-wise double descent that exists when no or moderate freezing is applied. Moreover, we identify a new kind of double descent phenomenon when examining the test error as a function of the number of frozen layers (for a fixed training epoch); we call it {\textbf{freezing-wise double descent}}.


Our third contribution is the examination of the effect of the similarity between the source and target tasks on the generalization behavior, especially in cases with double descent behaviors. Higher task similarity is usually expected to yield better generalization in transfer learning. Interestingly, our results exemplify that {\textbf{transfer from a less related task may generalize better or on par with transfer from a more related task}}. 
Hence, when double descent is likely to emerge, a good utilization of a pre-trained model from a similar task requires to be extra careful in choosing the training duration and number of frozen layers.

\subsection{Related Work}
\label{subsec:related work}

Previous works provide analytical theories for the double descent phenomenon in transfer learning of simple models such as linear \cite{dar2020double,dar2021common} and two-layer nonlinear networks \cite{gerace2022probing}. This work is the first to examine the double descent phenomenon in transfer learning of DNNs; specifically, we provide new insights that cannot be obtained in simple models and are directly relevant to the deep learning practice. 


More broadly, this work also relates to the ongoing research efforts to elucidate the conditions for a successful transfer learning of DNNs \cite{yosinski2014how,raghu2019transfusion,kornblith2019better,ericsson2021how,mensink2021factors}. Compared to these works, (i) we are the first to examine the double descent phenomenon in transfer learning of DNNs, and (ii) our explicit examination of overparameterization provides new insights on the effects of source and target dataset sizes on the generalization behavior and training speed in transfer learning.  

\subsection{Paper Outline}

This paper is organized as follows. In Section \ref{sec:The Transfer Learning Settings} we provide the details of the  transfer learning settings and the classification problems. 
In Section \ref{sec:double descent in transfer learning} we show the effect of the source and target dataset sizes on the emergence of the double descent phenomenon in transfer learning. 
In Section \ref{sec:Freezing Layers} we demonstrate the effect of freezing layers on the double descent phenomenon. 
In Section \ref{sec:Task Similarity} we examine transfer from tasks at different similarity levels. 
We conclude the paper in Section \ref{sec:Conclusion}. 
Additional experimental details and results are provided in the Appendices.

\section{The Transfer Learning Settings}
\label{sec:The Transfer Learning Settings}

\begin{table}[b]
  \centering
\small{
\begin{tabular}{ *{2}{|c}|*{2}{|c}| } 
\hline

\multicolumn{2}{|c||}{Source task} & \multicolumn{2}{c|}{Target task} \\
\cline{1-4}
\# classes & Dataset & \# classes & Dataset  \\
\hline
\hline
40 & Tiny Imagenet (first 40 classes) & 40 & CIFAR-100 (40 classes, 2 per superclass) \\ 
\hline
40 & \shortstack{Tiny Imagenet\\(40 classes similar to target classes)} & 40 & CIFAR-100 (40 classes, 2 per superclass) \\ \hline
100 & Tiny Imagenet (first 100 classes) & 100 & CIFAR-100 (all 100 classes) \\ \hline
200 & Tiny Imagenet (all 200 classes) & 10 & CIFAR-10 \\ \hline
\hline
\end{tabular}
}
  \caption{The source and target classification tasks that are examined in the main paper. See Appendix \ref{appendix:sec:Additional Details on the Examined Classification Problems} for more details and the complete list of evaluated tasks (that also includes the Food-101 dataset). }
  \label{tab:source and target tasks}
\end{table}

This paper studies transfer learning of DNNs in the ResNet-18 \cite{he2016deep}, DenseNet-121 \cite{huang2017densely}, and the Vision Transformer (ViT) \cite{dosovitskiy2021image} architectures (see Appendix \ref{appendix:sec:Additional Details on the Examined DNN Architectures} for more details on the architectures). 
We consider image classification problems that are defined using data from the CIFAR-10, CIFAR-100 \cite{krizhevsky2009learning}, Food-101 \cite{bossard2014food}, Tiny ImageNet \cite{le2015tiny} datasets. The pairs of source-target tasks that are examined in the main paper are listed in Table \ref{tab:source and target tasks}. 
All input images are of 32x32x3 pixels size, except for the Tiny ImageNet dataset for which we examine also the 64x64x3 pixels size in the experiments for ResNet and DenseNet. In transfer learning settings where the source and target tasks have the same input size and number of classes, the last layer has the same dimension in the source and target DNNs and therefore can be transferred; otherwise, the last layer is randomly initialized. See Appendix \ref{appendix:sec:Additional Details on the Examined Classification Problems} for more details.

All the source DNNs were trained on datasets without any (artificially added) label noise and for 4000 training epochs with the Adam optimizer, batch size 128,  learning rate 0.0001 for source tasks with 10 classes, and learning rate 0.00005 for source tasks with 40 or 100 classes.

In contrast to the relatively large training dataset of the source tasks, the target tasks have limited data for training (a challenge that often motivates using transfer learning in practice). We will consider a variety of dataset sizes, ranging from the few-shot learning setting (e.g., less than 10 training samples per class) and up to larger datasets (e.g., with 100 samples per class) that are still considerably smaller than the datasets in the source DNN training. Due to the relatively small size of the target dataset, we define the batch size to be twice (or x4.5 for target tasks with 100 classes) the number of training samples per class, but not more than 128 samples or 25\% of the entire dataset. The training is for 8000 epochs using the Adam optimizer with a constant learning rate of 0.00001.  Similar to the studies of double descent in the learning of DNNs from scratch \cite{nakkiran2019deep,somepalli2022can}, we use long training and low learning rate in order to evaluate the ``natural'' generalization behavior in a smooth manner; e.g., in contrast to using learning rate schedulers that may affect the epoch-wise error curves and interfere with our goal of studying the existence or lack of the double descent phenomenon.

We will give particular attention to generalization trends that stem by using different dataset sizes and different number of frozen layers (as defined in Section \ref{sec:Freezing Layers}).

\begin{figure*}[t]
  \centering
  \includegraphics[width=0.85\textwidth]{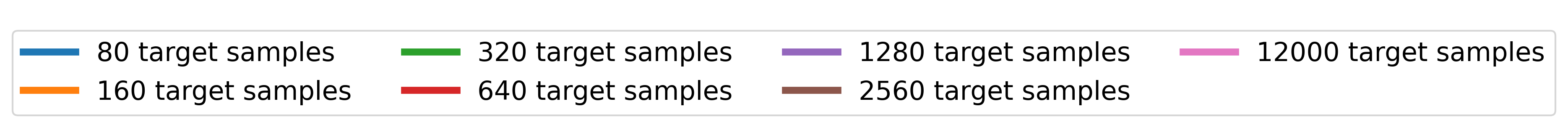}
  \\[-3ex]
  \subfloat[Noiseless]{
   \includegraphics[width=0.49\textwidth]{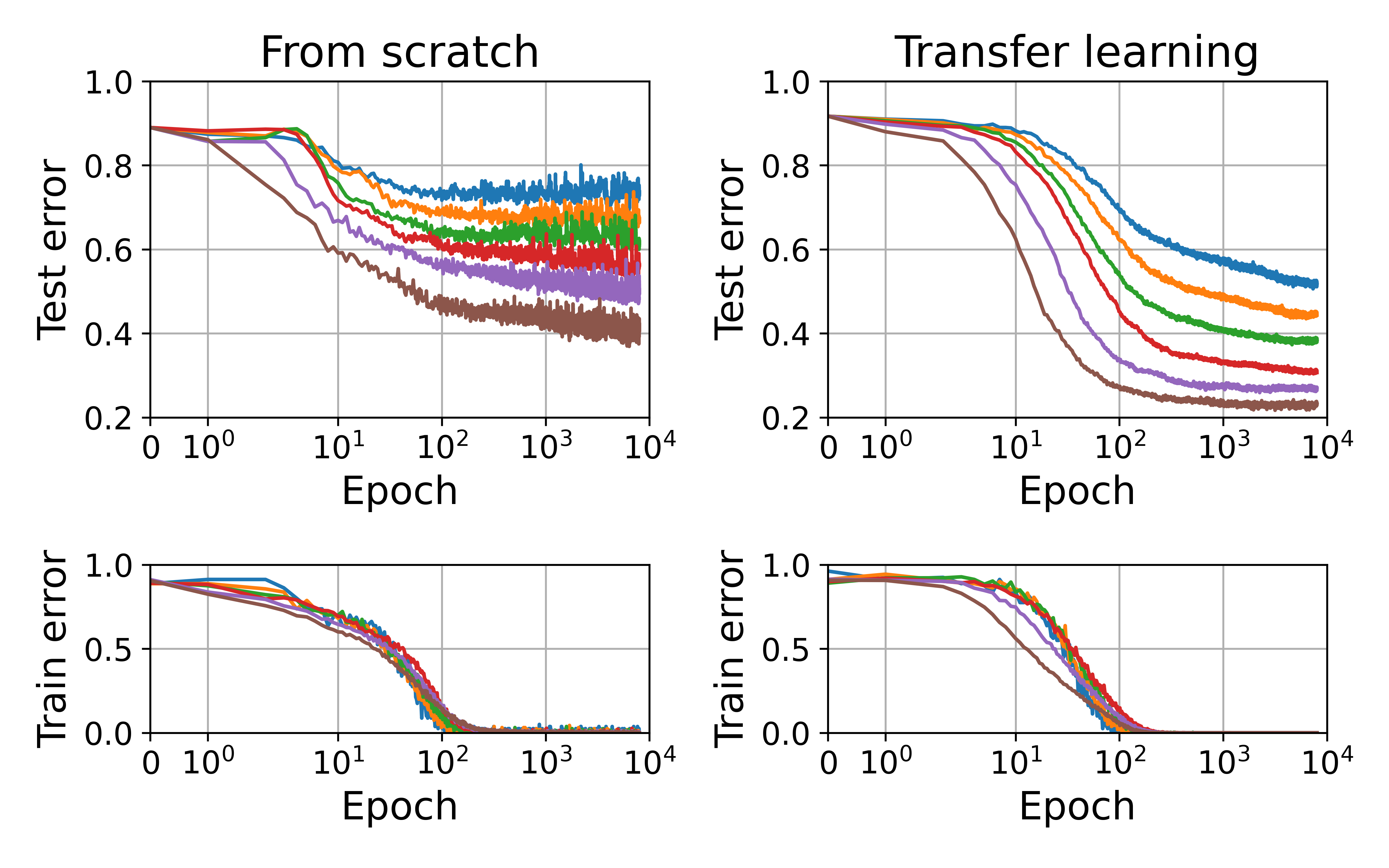}
   \label{fig:test_error_vs_epoch_for_several_dataset_sizes_40class_resnetW64_noiseless}}
   \subfloat[20\% label noise in target dataset]{\includegraphics[width=0.49\textwidth]{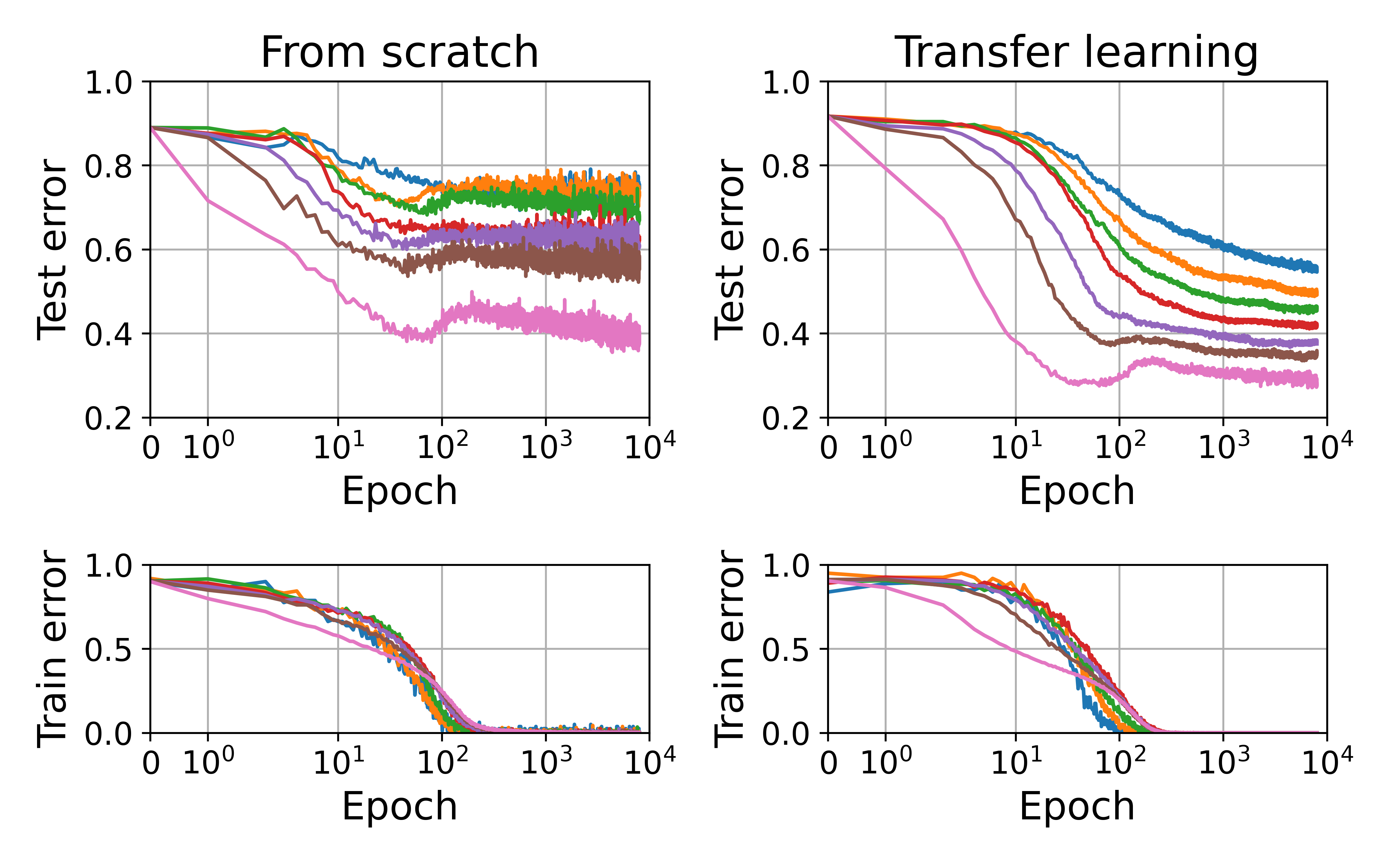}
\label{fig:test_error_vs_epoch_for_several_dataset_sizes_40class_resnetW64_srcNoiselessTgtNoisy0p2}}
   \caption{Evaluation of ResNet-18 training for the CIFAR-10 target task. The transfer learning is from the source task of 200 Tiny ImageNet classes (input image size 64x64x3) with source dataset of 100k training samples. Each curve color corresponds to another size of the target dataset. Note that the double descent emerges in transfer learning only for noisy (target) datasets that are sufficiently large.}
   \label{fig:test_error_vs_epoch_for_several_dataset_sizes_40class_resnetW64}
\end{figure*}

\section{When Does Double Descent Emerge in Transfer Learning?}
\label{sec:double descent in transfer learning}

Label noise is known as a promoter of significant double descent phenomena in DNNs \cite{nakkiran2019deep}. Accordingly, in learning the target DNN from scratch, the test error follows a double descent trend along the training epochs when the target dataset has label noise (Fig.~\ref{fig:test_error_vs_epoch_for_several_dataset_sizes_40class_resnetW64}); interestingly, we observe double descent for DenseNet and ViT also for some of the noiseless datasets (see, e.g., the black and brown test error curves for target dataset of 640 samples in Fig.~\ref{subfig:different_source_dataset_size_densenet_srcNoiselessTgtNOISELESS-src200classtgtCIFAR10class_tinyimagenet64} and more results, including for ViT, in Appendix \ref{appendix:subsec:Additional Experimental Results for Section 3.1}). 
\lolo{I think that the first sentence of this first paragraph seems a bit out of place.}

In this section we consider the effect of sizes of the target and source datasets on the emergence of the double descent phenomenon. For a start, we consider transfer learning where the entire target DNN is trained from its initialization with the pre-trained source parameters (excluding the last layer if its dimension mismatches due to different number of classes and/or input size). Later, in Section \ref{sec:Freezing Layers} we will also examine transfer learning with frozen layers.
\lolo{It might be better to combine these two paragraphs together}

\subsection{Double Descent Can Emerge in Transfer Learning, but is Less Likely in Few-Shot Learning}
\label{subsec:effect of target dataset size}

The test error in transfer learning is more likely to follow an epoch-wise double descent if the target dataset is \textit{sufficiently large} and has label noise (see Fig.~\ref{fig:test_error_vs_epoch_for_several_dataset_sizes_40class_resnetW64} and the additional results in Appendix \ref{appendix:subsec:Additional Experimental Results for Section 3.1}). 
Transfer learning is often motivated by limited availability of data for training and, therefore, the lack of double descent in the case of a small target dataset is of great importance. 
Specifically, in few-shot learning the target dataset is extremely small and, thus, the occurrence of double descent is even less probable.
The lack of a double descent peak of the test error implies that the generalization performance is less sensitive to the exact duration of training and stopping criterion.

Fig.~\ref{fig:test_error_vs_epoch_for_several_dataset_sizes_40class_resnetW64} also shows that whereas learning from scratch from a small target dataset can induce double descent, the corresponding transfer learning setting with the same target dataset does not have a double descent behavior. This demonstrates the ability of transfer learning to regularize (i.e., reduce or eliminate) the double descent peak in the test error. Specifically, \textbf{transfer learning provides stronger implicit regularization for a smaller target dataset}.



\subsection{Larger Source Datasets Can Delay Interpolation in the Target DNN}
\label{subsec:effect of source dataset size}

\begin{figure*}[t]
  \centering
  \includegraphics[width=0.85\textwidth]{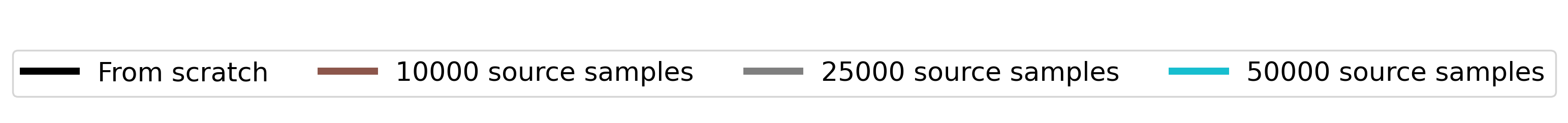}
  \\[-1ex]
  \subfloat[ResNet]{\label{appendix:fig:src_different_dataset_size_100Classes_srcImagenetF100classtgtCIFAR100all_srcNoiselessTgtNoisy0p2_resnet}
   \includegraphics[width=0.245\textwidth]{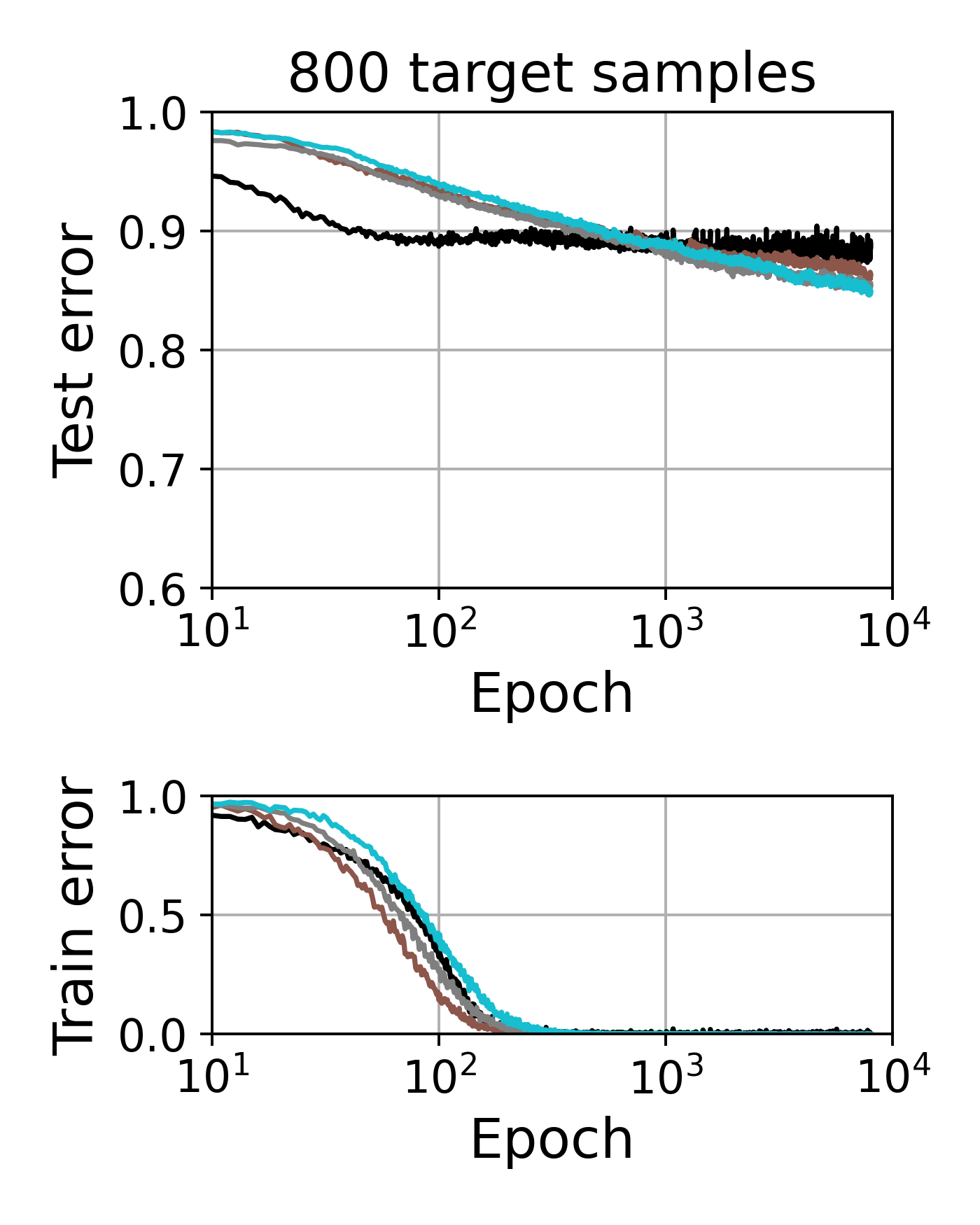}
      \includegraphics[width=0.245\textwidth]{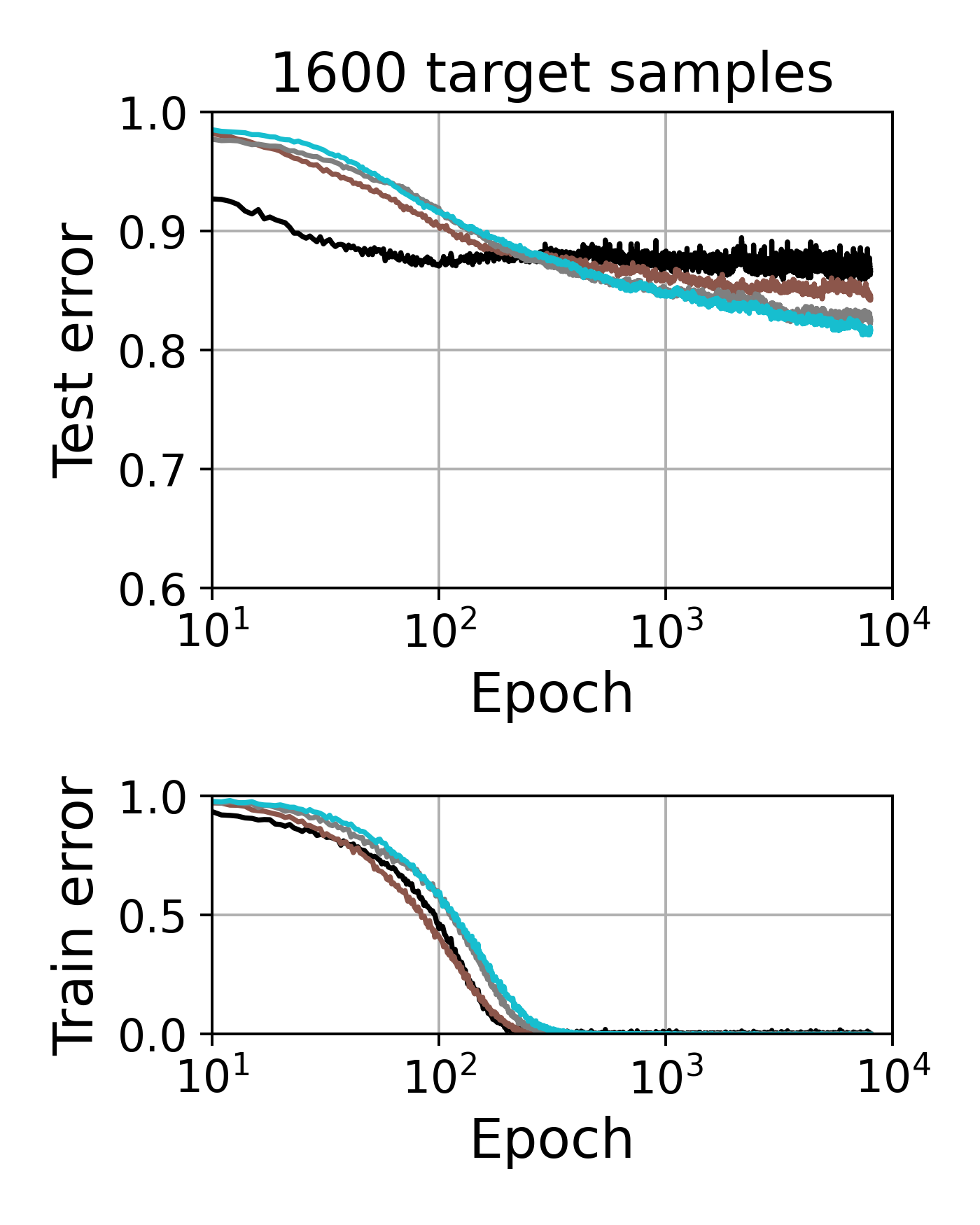}
   \includegraphics[width=0.245\textwidth]{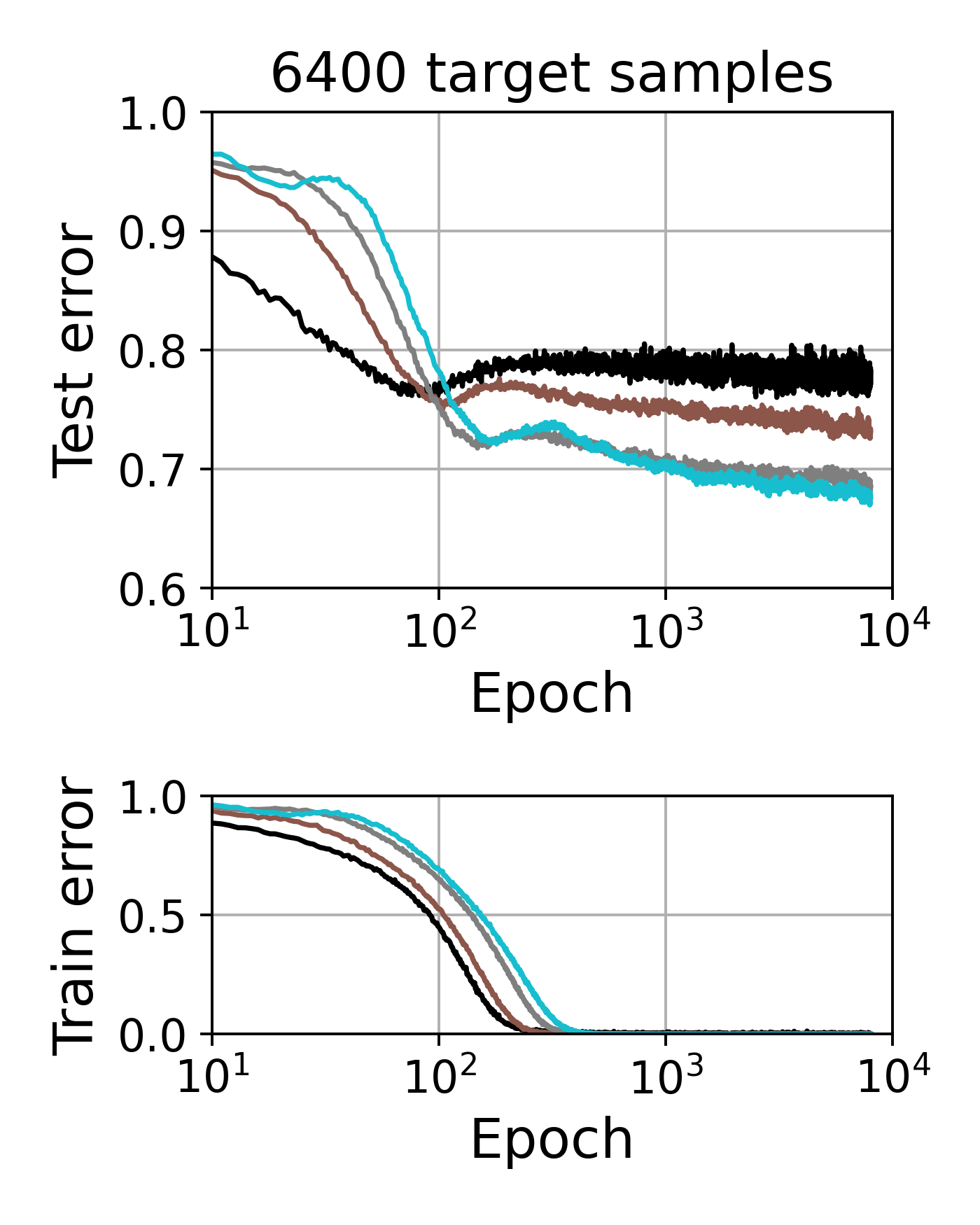}
   \includegraphics[width=0.245\textwidth]{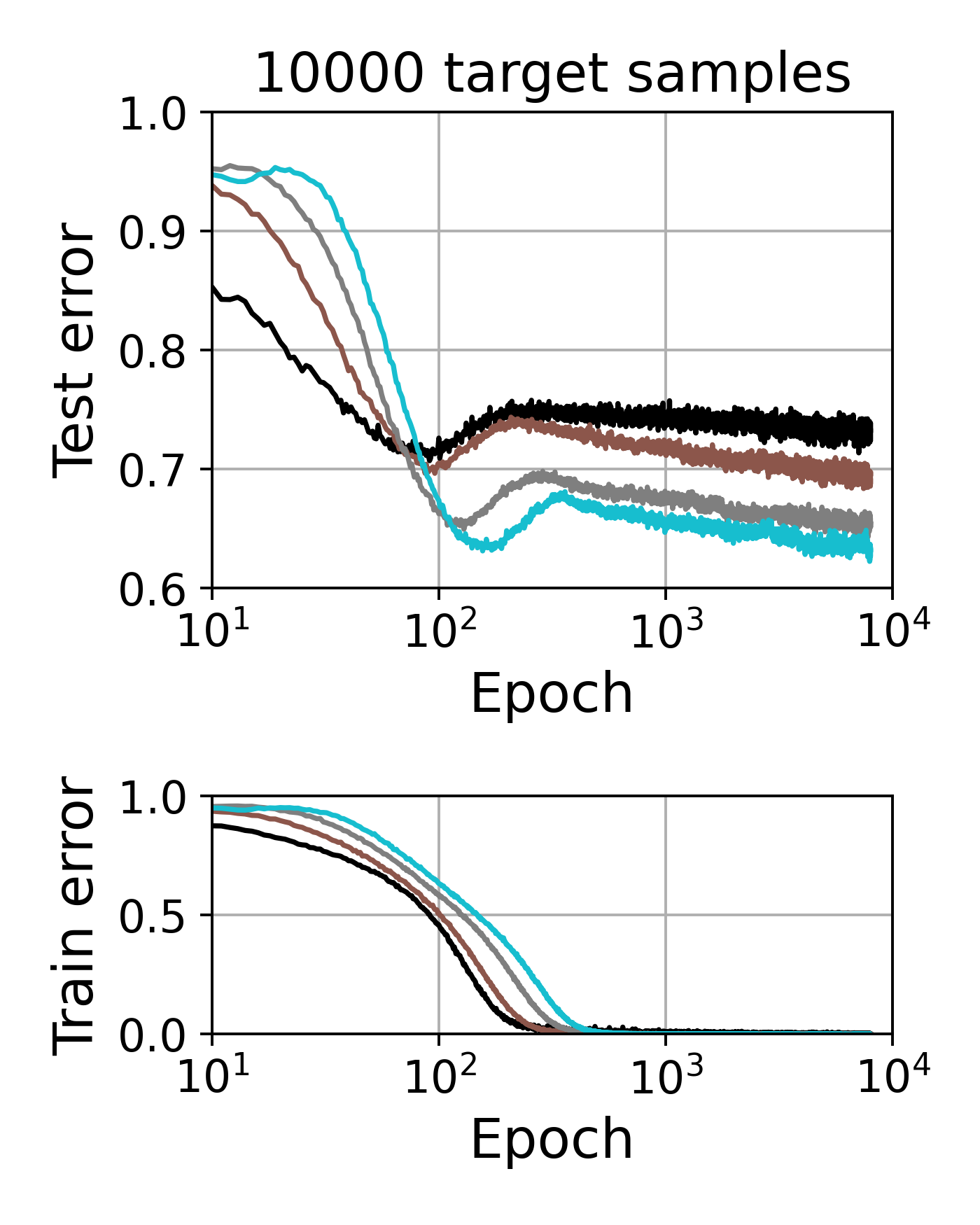} }
   \\\subfloat[DenseNet]{\label{appendix:fig:src_different_dataset_size_100Classes_srcImagenetF100classtgtCIFAR100all_srcNoiselessTgtNoisy0p2_densenet}
   \includegraphics[width=0.245\textwidth]{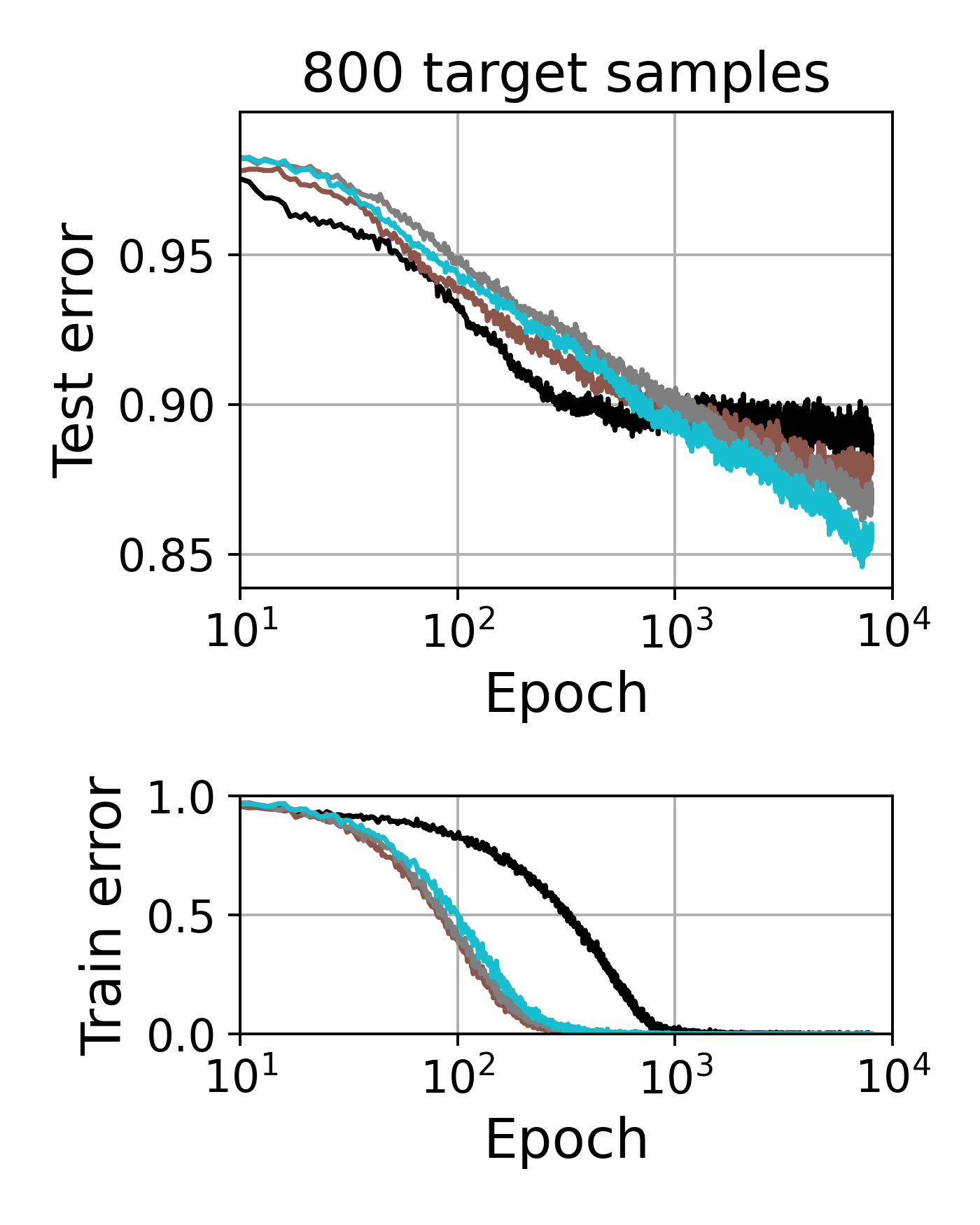}
   \includegraphics[width=0.245\textwidth]{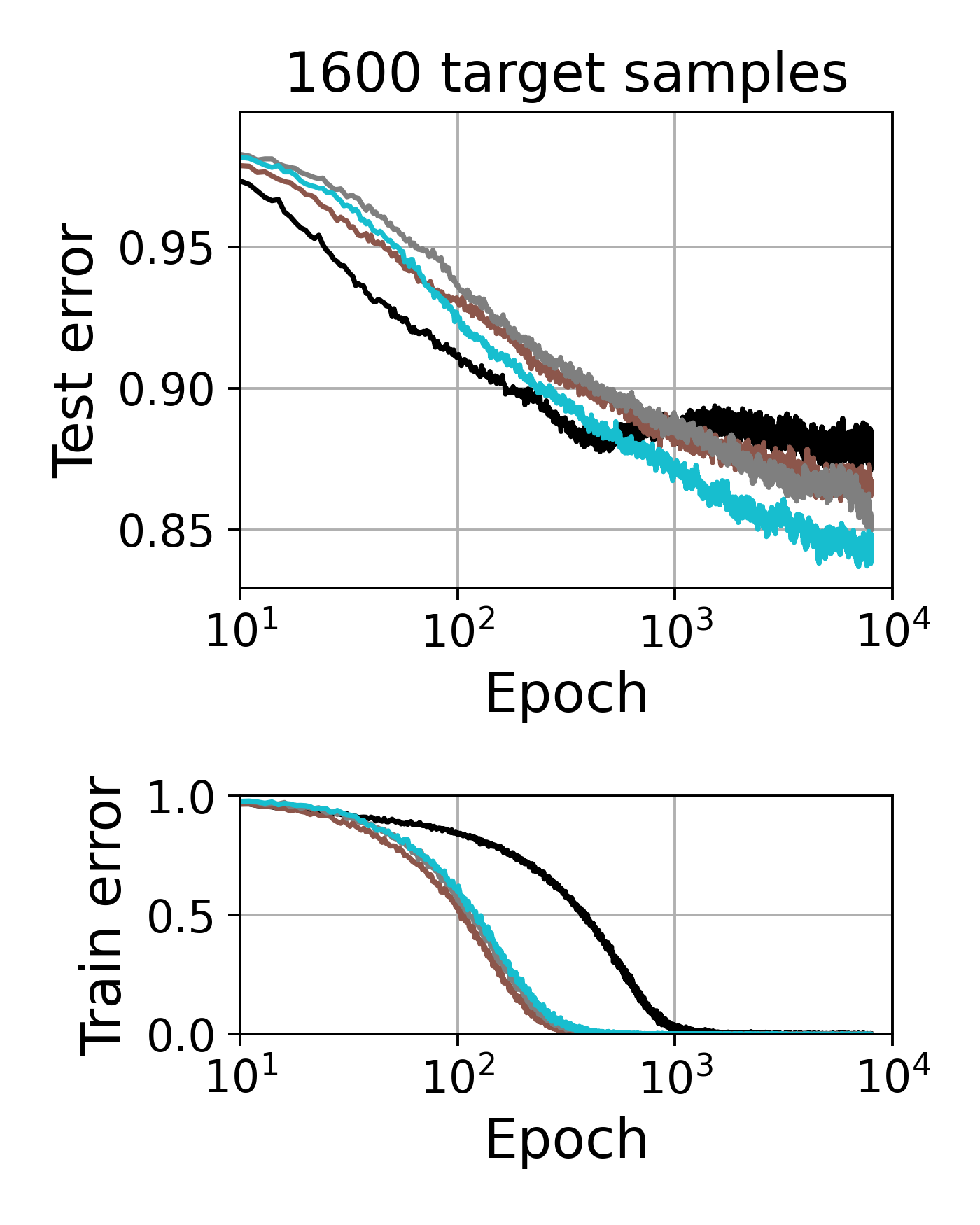}
   \includegraphics[width=0.245\textwidth]{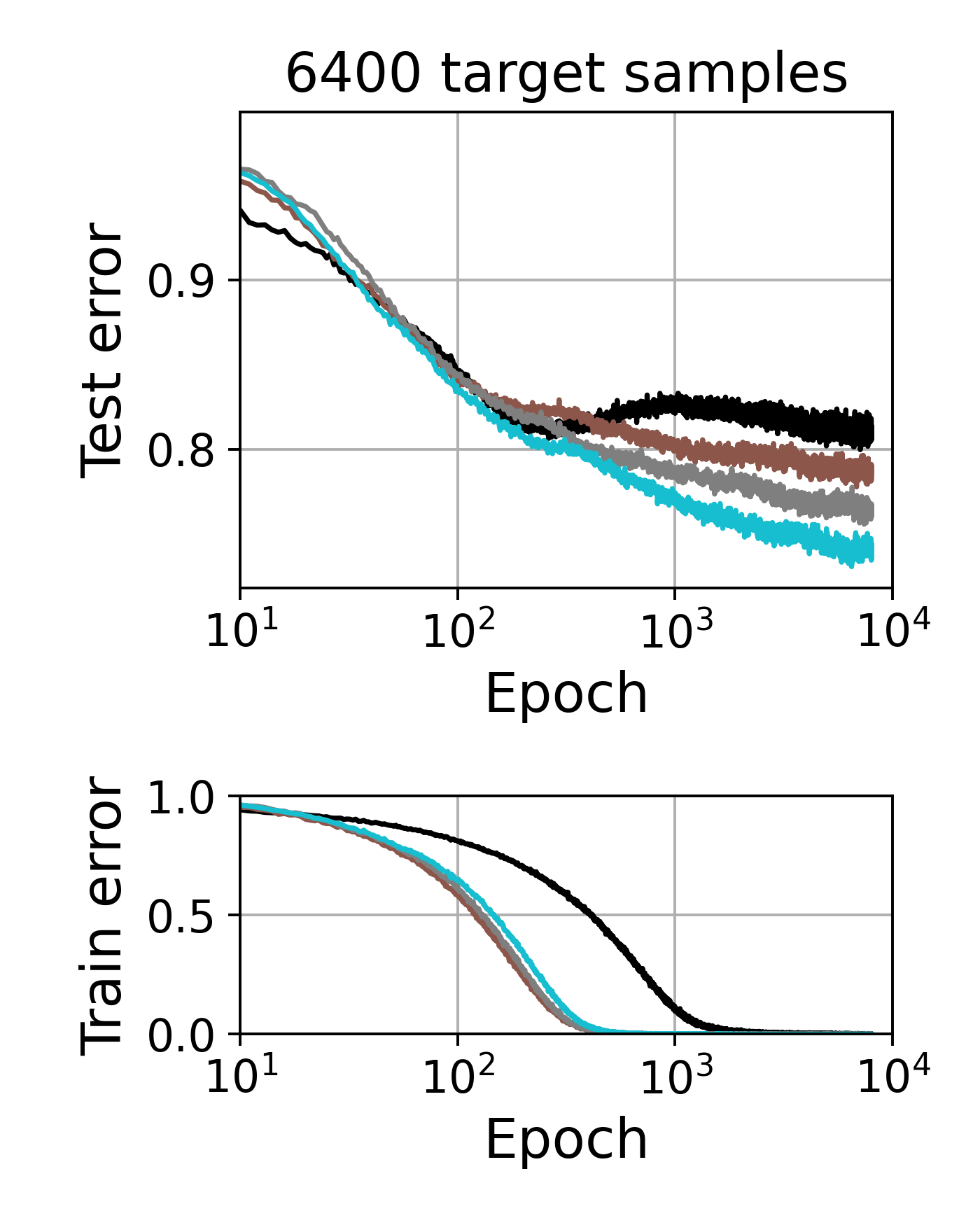}
   \includegraphics[width=0.245\textwidth]{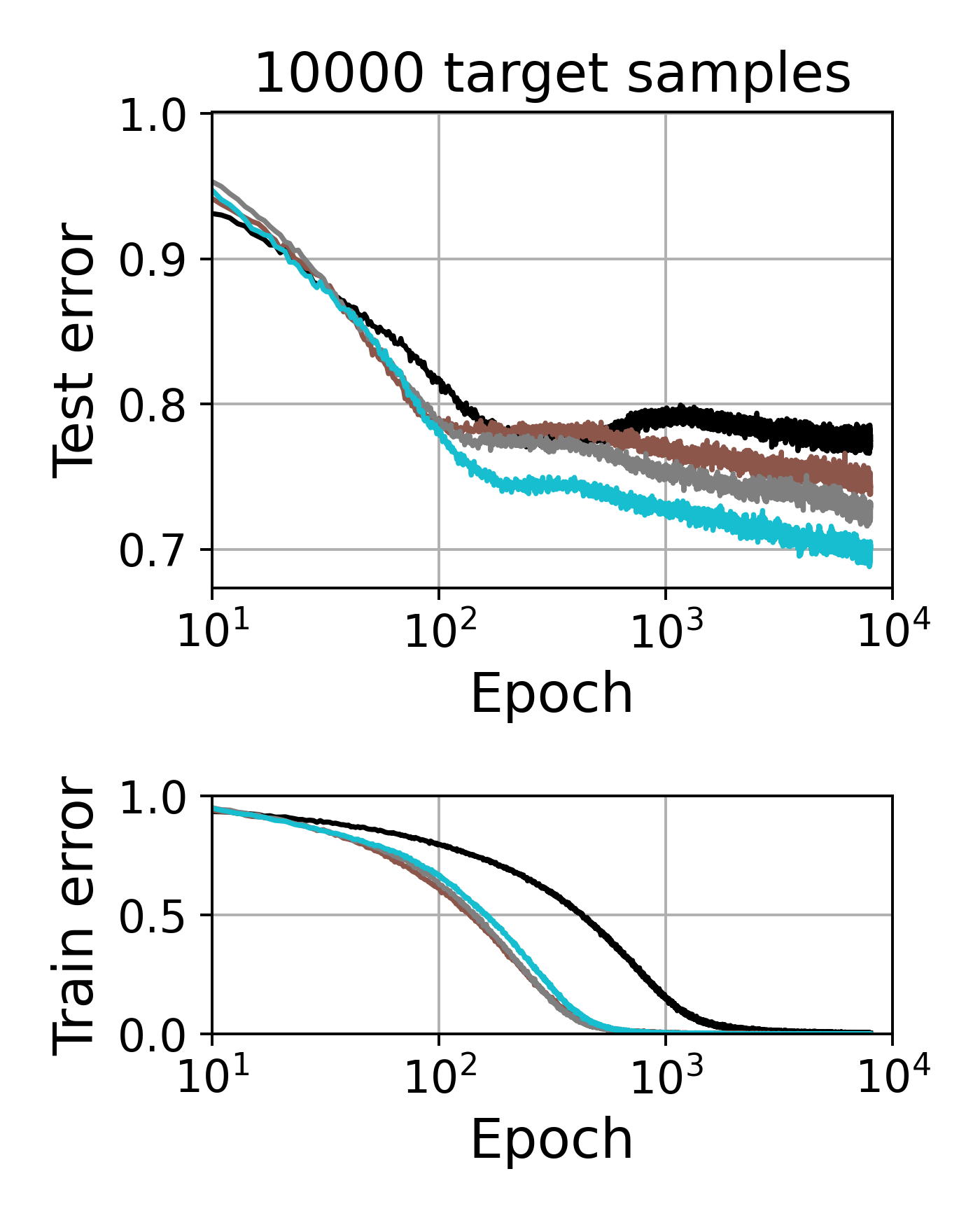}}
   \caption{The effect of the source dataset size on the target model training. Evaluation of training for a target classification task of \textbf{CIFAR-100} (all 100 classes)  with 20\% label noise in the target dataset. The transfer learning is from the source task of 100 Tiny ImageNet classes (input image size 32x32x3). Each curve color corresponds to another size of the source dataset.  }
   \label{fig:src_different_dataset_size_100Classes_srcImagenetF100classtgtCIFAR100all_srcNoiselessTgtNoisy0p2}
\end{figure*}

Now we turn to examine the possible effect of the size of the \textit{source} training dataset on the evolution of the errors in the transfer learning of the \textit{target} DNN. 
Recall that the source DNN is trained to interpolate its (noiseless) source dataset, and then it is utilized as an initialization for the target DNN training from its (target) dataset. 
Here the experiment is designed as follows: For a given target training dataset, we examine several transfer learning options for the target DNN --- the difference among these options is in the size of the source dataset that the pre-trained source model was trained on. 
Our results (Figs.~\ref{fig:transfer_learning_double_descent_source_dataset_size}, \ref{fig:src_different_dataset_size_100Classes_srcImagenetF100classtgtCIFAR100all_srcNoiselessTgtNoisy0p2}, \ref{fig:source_dataset_size_as_regularization}, and the additional results in Appendix \ref{appendix:subsec:Additional Experimental Results for Section 3.2}) demonstrate that \textbf{a larger source dataset can provide a stronger implicit regularization effect} on the target DNN training. Put differently, a larger source dataset can reduce the effective complexity of the target DNN.  

The lower complexity of the target DNN due to stronger implicit regularization by a larger source dataset can be reflected by one or more of the following: 
\begin{enumerate}
    \item \textbf{Slower training for larger source datasets.} Namely, a larger source dataset can delay the arrival of the target DNN to interpolation of its own dataset. For ResNet, this behavior is significant and can be observed for both small and large target datasets (see Figs.~\ref{fig:transfer_learning_double_descent_source_dataset_size}, \ref{appendix:fig:src_different_dataset_size_100Classes_srcImagenetF100classtgtCIFAR100all_srcNoiselessTgtNoisy0p2_resnet}, \ref{subfig:different_source_dataset_size_resnet_srcNoiselessTgtNoisy0p2-src200classtgtCIFAR10class_tinyimagenet64}, \ref{fig:src_different_dataset_size_100Classes_srcImagenetF100classtgtFood100all_srcNoiselessTgtNoisy0p2}, \ref{fig:src_different_dataset_size_10Classes_srcCIFAR10classtgtFoodF10_srcNoiselessTgtNoisy0p2}). For DenseNet and ViT, this behavior is weaker and can be observed for the relatively small target datasets (see Figs.~\ref{appendix:fig:src_different_dataset_size_100Classes_srcImagenetF100classtgtCIFAR100all_srcNoiselessTgtNoisy0p2_densenet}, \ref{subfig:different_source_dataset_size_densenet_srcNoiselessTgtNOISELESS-src200classtgtCIFAR10class_tinyimagenet64}, \ref{fig:src_different_dataset_size_100Classes_srcImagenetF100classtgtCIFAR100all_srcNoiselessTgtNoisy0p2_ViT_appendix}). 

    To explain why a larger source dataset does not always slow down the target DNN training (especially for large target datasets), let us examine the speed of learning the target DNN from scratch. Typically, transfer learning becomes more similar to learning the target DNN from scratch for a smaller source dataset and a larger target dataset. Hence, the speed of learning from scratch may affect the relative speed of transfer learning from a small source dataset compared to a large source dataset. This effect depends on whether learning from scratch of the target DNN is faster or slower than the transfer learning options; e.g., note in Figs.~\ref{fig:src_different_dataset_size_100Classes_srcImagenetF100classtgtCIFAR100all_srcNoiselessTgtNoisy0p2}, \ref{fig:source_dataset_size_as_regularization} that the train error curve of learning from scratch (the black curve) can be to the left, to the right, or in between the train error curves of the examined transfer learning settings (the cyan, gray and brown curves). 
    Accordingly, if learning from scratch is faster than the transfer learning alternatives, it further contributes to speed up transfer learning from a smaller source dataset; this behavior extends the implicit regularization effect of transfer learning and explains the increased training speed differences between the cyan, gray and brown curves for the large target datasets in Figs.~\ref{appendix:fig:src_different_dataset_size_100Classes_srcImagenetF100classtgtCIFAR100all_srcNoiselessTgtNoisy0p2_resnet}, \ref{fig:src_different_dataset_size_100Classes_srcImagenetF100classtgtFood100all_srcNoiselessTgtNoisy0p2}, \ref{fig:src_different_dataset_size_10Classes_srcCIFAR10classtgtFoodF10_srcNoiselessTgtNoisy0p2}. 
    However, if learning from scratch is slower than the transfer learning alternatives, it slows down transfer learning from a smaller source dataset; this  contradicts the implicit regularization effect of transfer learning and explains the reduced training speed differences (or sometimes a different convergence order) between the cyan, gray and brown curves for the large target datasets in Figs.~\ref{appendix:fig:src_different_dataset_size_100Classes_srcImagenetF100classtgtCIFAR100all_srcNoiselessTgtNoisy0p2_densenet}, \ref{fig:source_dataset_size_as_regularization}, \ref{fig:src_different_dataset_size_100Classes_srcImagenetF100classtgtCIFAR100all_srcNoiselessTgtNoisy0p2_ViT_appendix}, \ref{fig:src_different_dataset_size_100Classes_srcImagenetF100classtgtCIFAR100all_srcNoiselessTgtNoisy0p2_narrowResNet}.
    This situation where learning from scratch is slower than the transfer learning alternatives occurs more when the last layer cannot be transferred (Fig.~\ref{fig:source_dataset_size_as_regularization}), for particular architectures such as ViT (Fig.~\ref{fig:src_different_dataset_size_100Classes_srcImagenetF100classtgtCIFAR100all_srcNoiselessTgtNoisy0p2_ViT_appendix}), or for lower parameterization levels (e.g., narrower ResNets as in Fig.~\ref{fig:src_different_dataset_size_100Classes_srcImagenetF100classtgtCIFAR100all_srcNoiselessTgtNoisy0p2_narrowResNet}). 
    
    To conclude, slower training for larger source datasets consistently appears for relatively small target datasets, but for relatively large target datasets the behavior is determined also by the relative speed of learning from scratch.

    \item \textbf{Reduced (or eliminated) double descent peak of the target test error due to a larger source dataset.} This can be observed in Figs.~\ref{fig:source_dataset_size_as_regularization}, \ref{appendix:fig:src_different_dataset_size_100Classes_srcImagenetF100classtgtCIFAR100all_srcNoiselessTgtNoisy0p2_resnetW24} for the ResNet and DenseNet architectures with relatively large target datasets. 
    Interestingly, the reduced double descent peak due to a larger source dataset is less observed in our results than the training slow-down effect.  

\end{enumerate}

For learning from scratch, slower arrival to interpolation and reduced double descent peak are both shown in \cite{nakkiran2019deep} as possible consequences of regularization mechanisms such as data augmentation and weight decay. Accordingly, our observations support our hypothesis that transfer learning with a larger source dataset can induce a stronger implicit regularization effect on the target DNN training.

\begin{figure*}[t]
  \centering
  \includegraphics[width=0.95\linewidth]{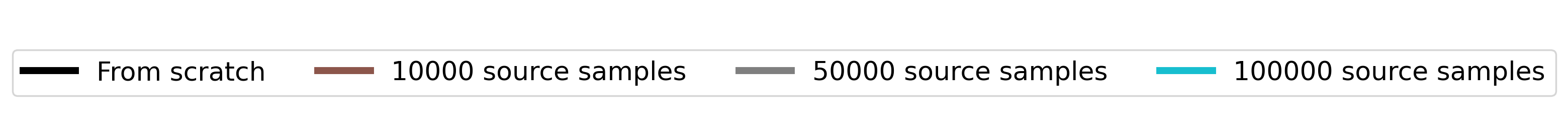}
  \\[-1ex]
    \subfloat[ResNet-18 (target dataset with 20\% label noise)]{
    \includegraphics[width=0.24\linewidth]{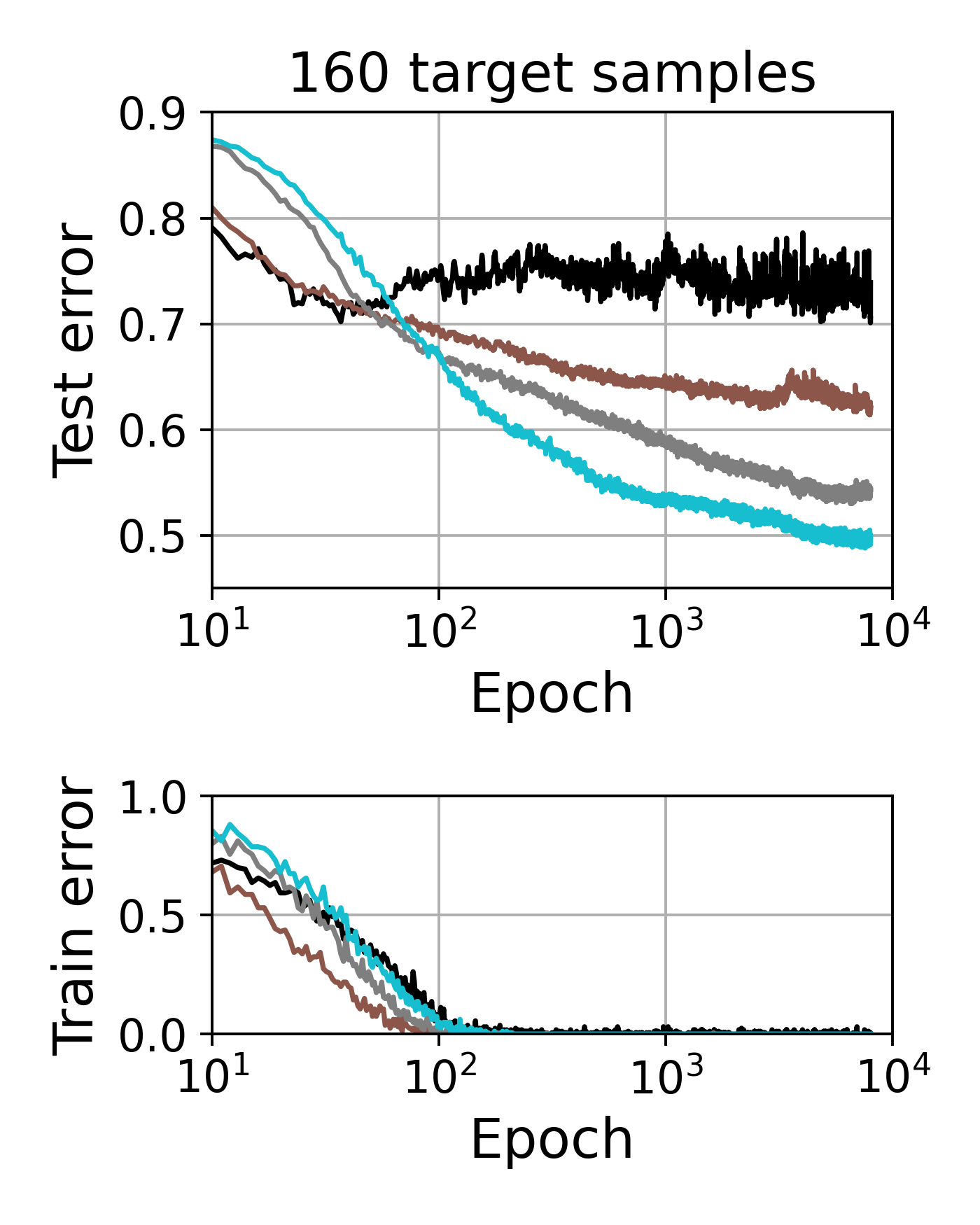}
        \includegraphics[width=0.24\linewidth]{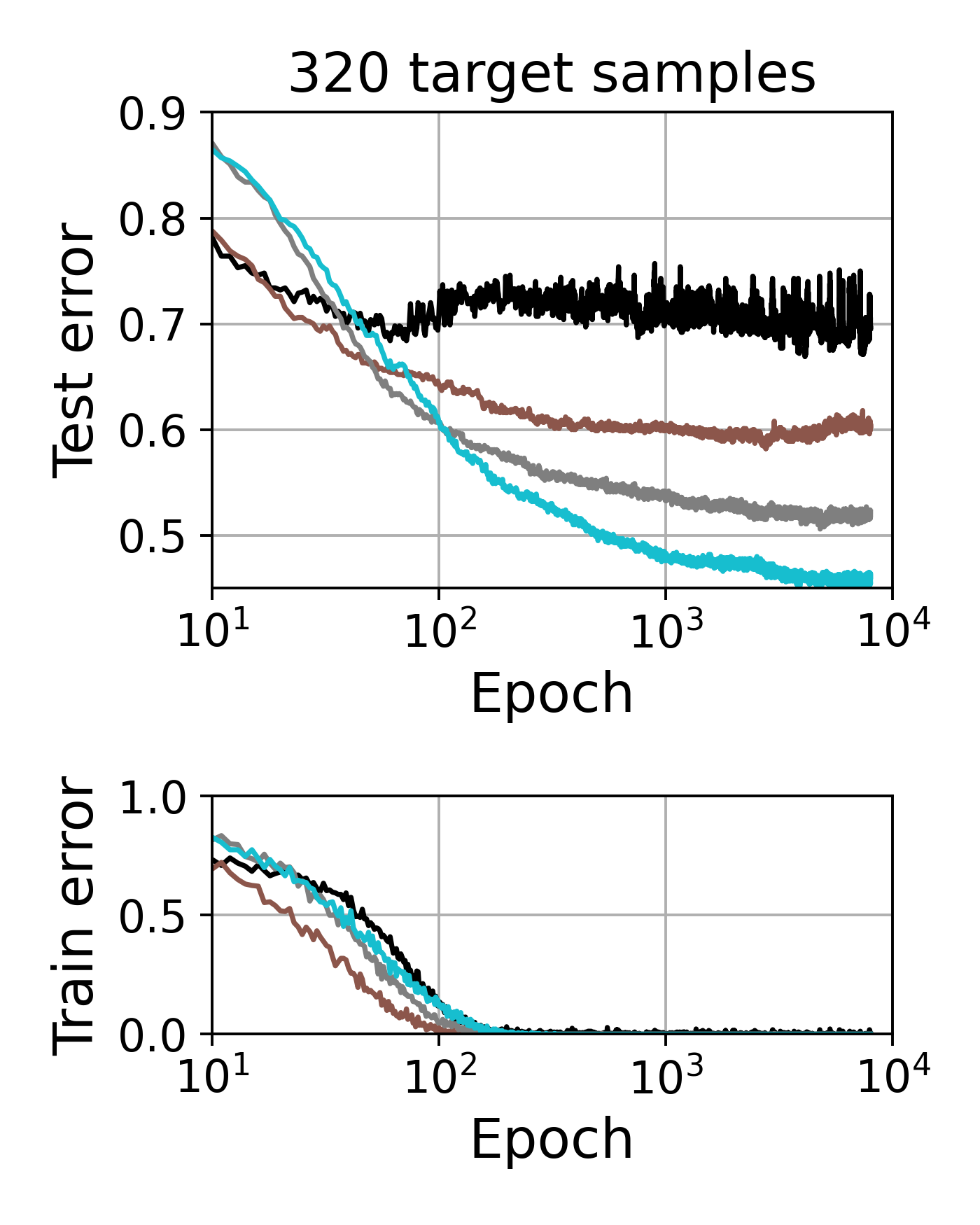}
            \includegraphics[width=0.24\linewidth]{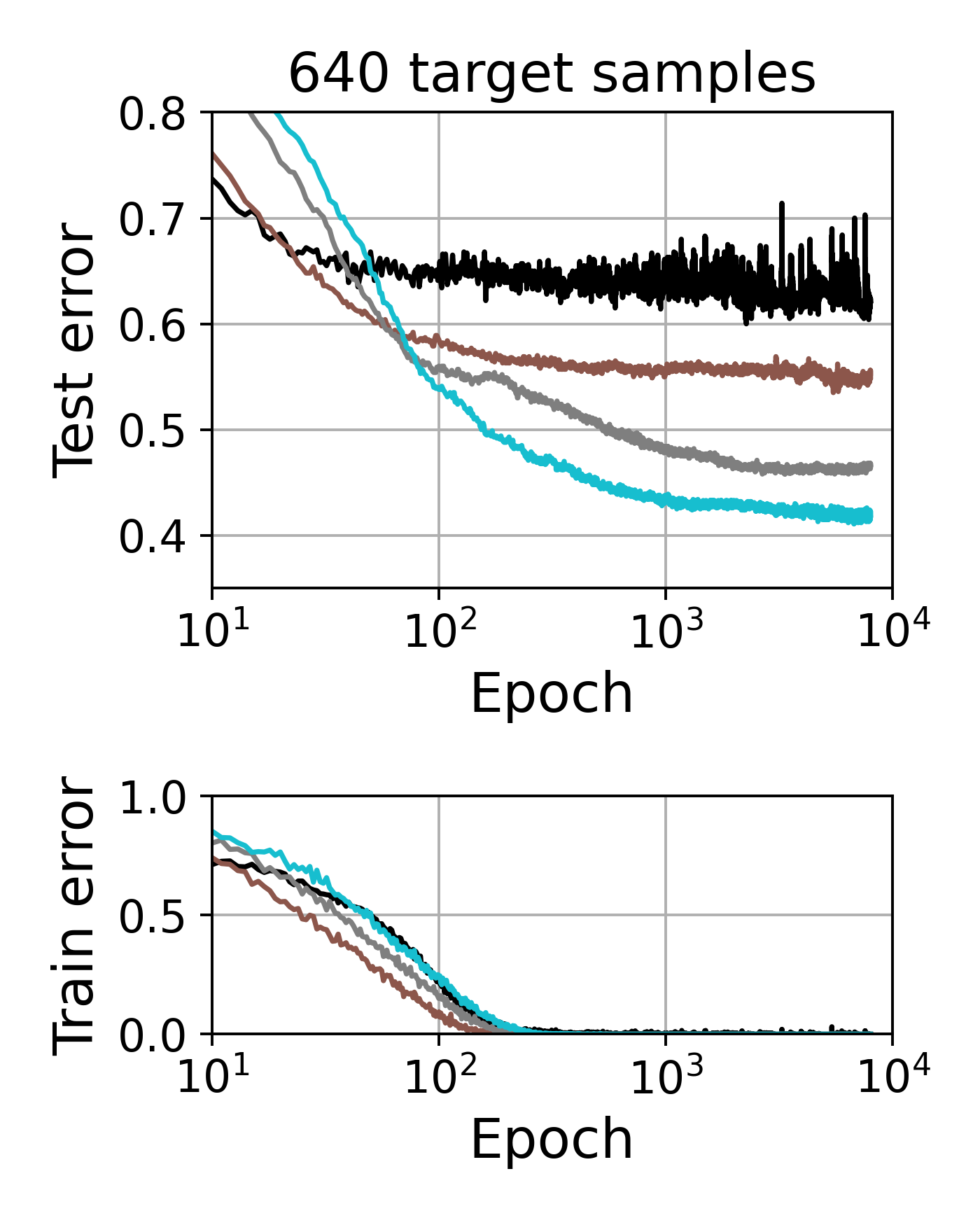}
                \includegraphics[width=0.24\linewidth]{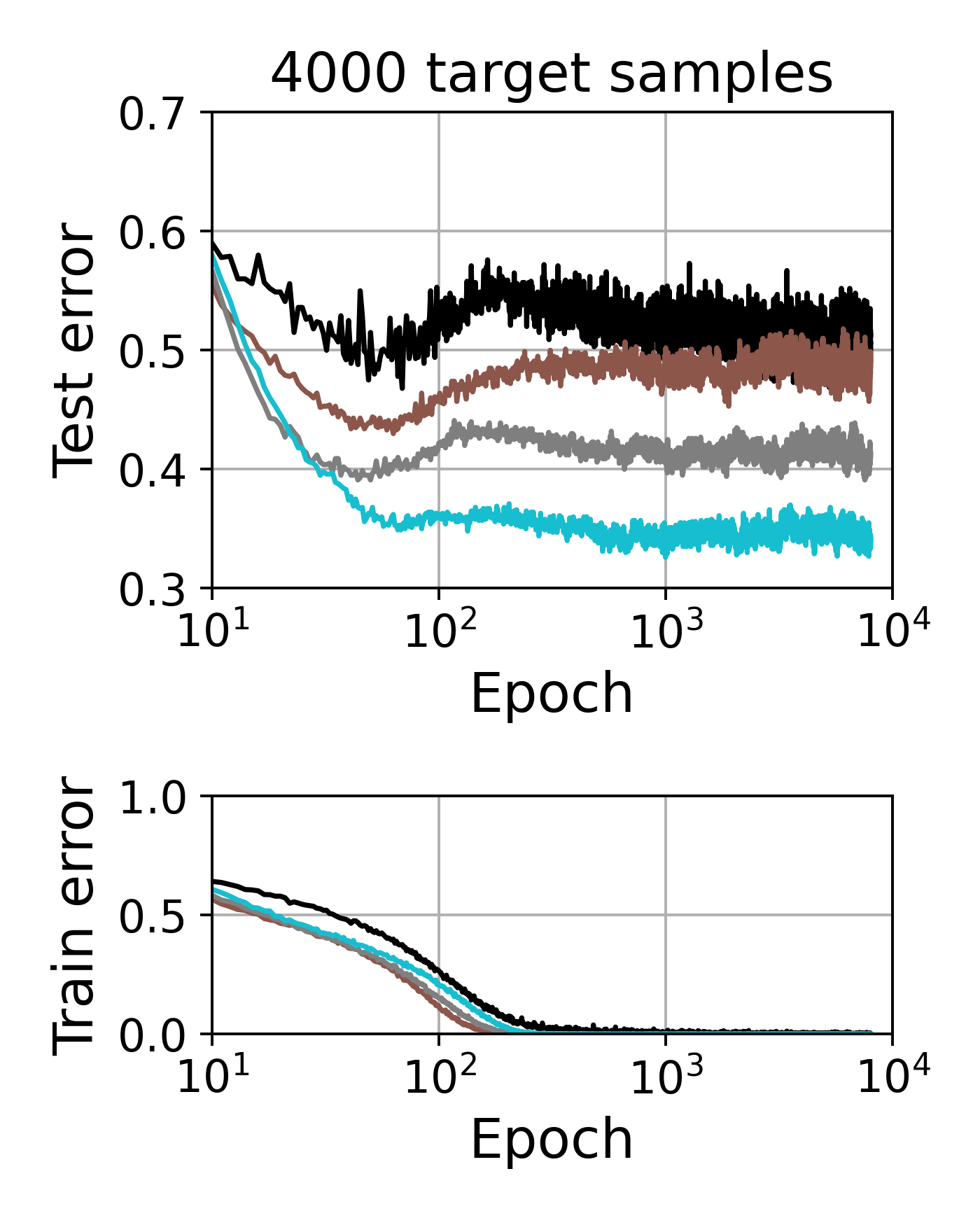}
    \label{subfig:different_source_dataset_size_resnet_srcNoiselessTgtNoisy0p2-src200classtgtCIFAR10class_tinyimagenet64} }
  \\
  \subfloat[DenseNet (noiseless target dataset)]{
  \includegraphics[width=0.24\linewidth]{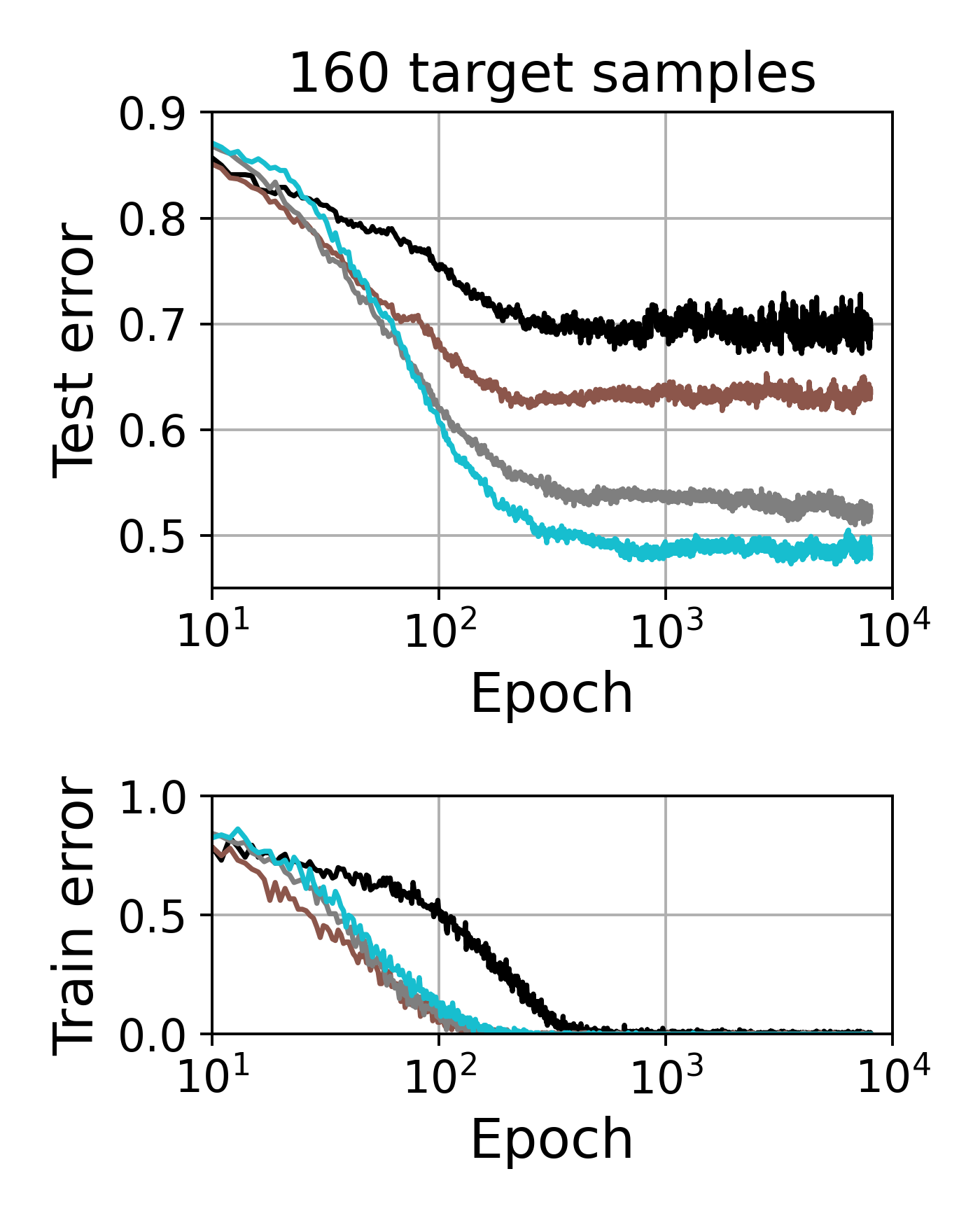}
  \includegraphics[width=0.24\linewidth]{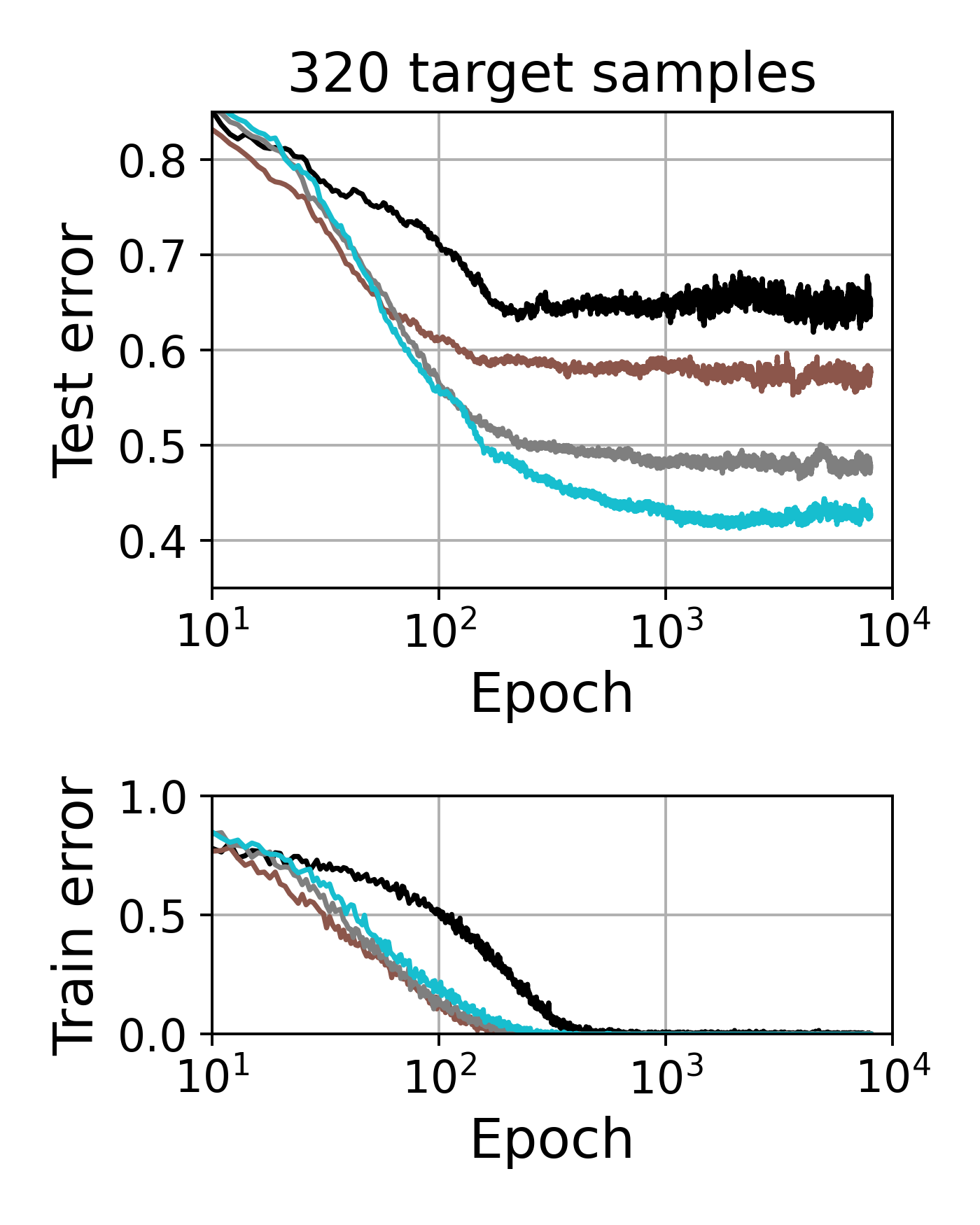}
    \includegraphics[width=0.24\linewidth]{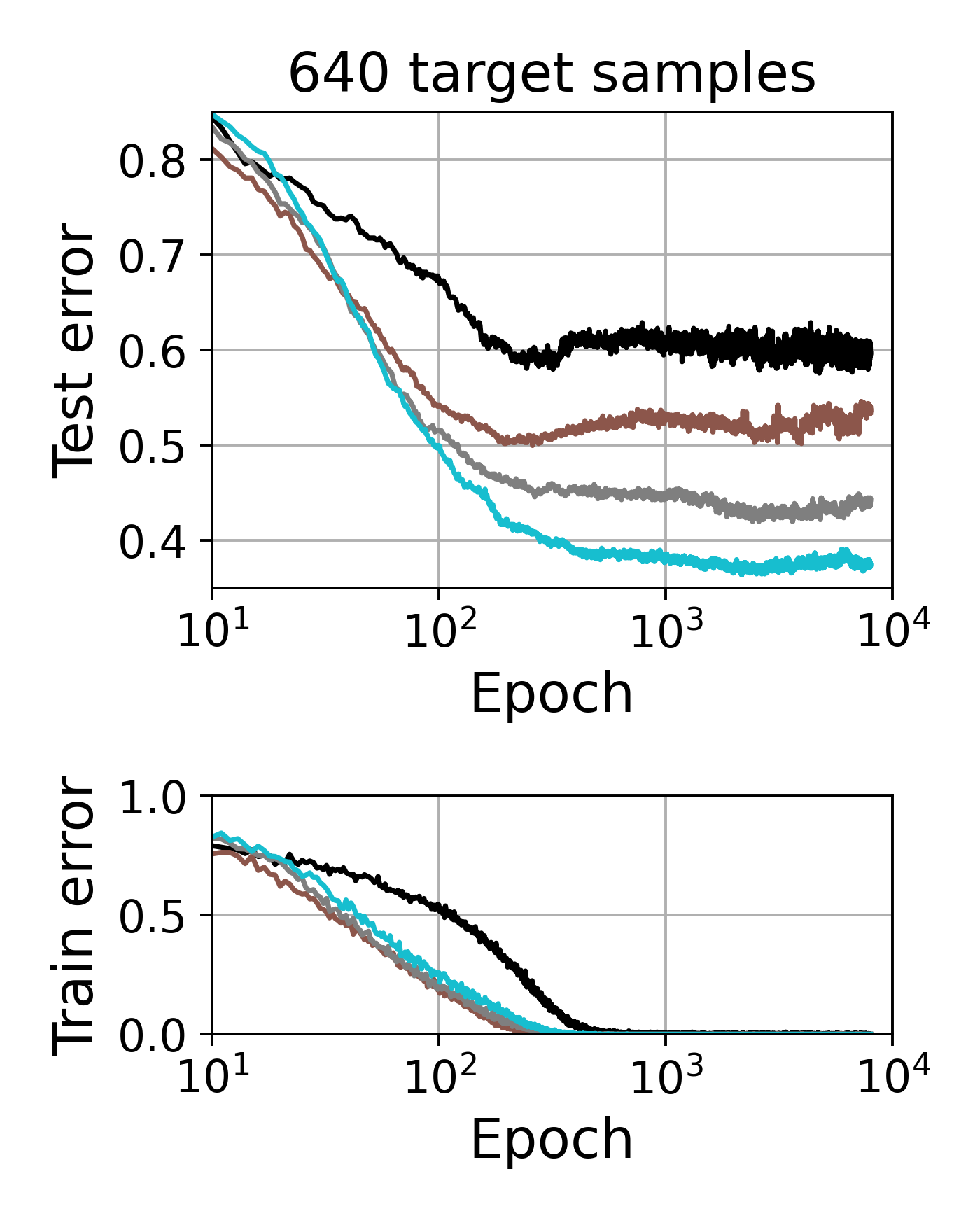}
        \includegraphics[width=0.24\linewidth]{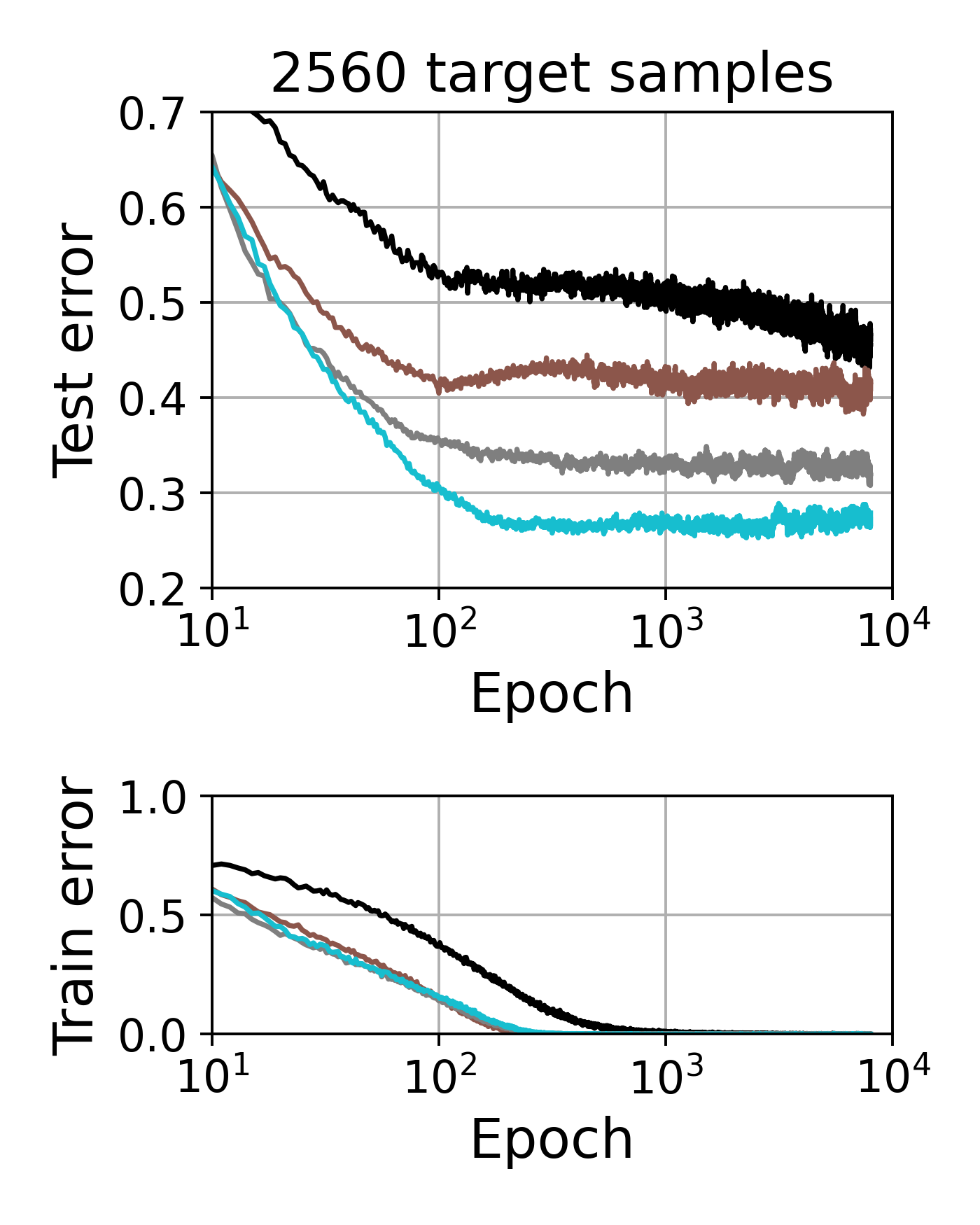}
    \label{subfig:different_source_dataset_size_densenet_srcNoiselessTgtNOISELESS-src200classtgtCIFAR10class_tinyimagenet64} }
  \caption{Evaluation of transfer learning for the CIFAR-10 target task from the Tiny ImageNet source task (200 classes, input image size 64x64x3). Each curve color corresponds to another size of the source dataset. Each subfigure corresponds to another size of the target dataset.}
  \label{fig:source_dataset_size_as_regularization}
\end{figure*}

\section{The Effect of Freezing Transferred Layers on Overparameterization and Double Descent}
\label{sec:Freezing Layers} 

A prevalent transfer learning strategy is to take layers from the pre-trained (source) DNN and setting them fixed (i.e., ``freezing'' them) in the target DNN without further adjustments or fine tuning. This approach reduces the number of learnable parameters in the target DNN and, by that, determines its \textit{effective} parameterization level.

Popular transfer learning implementations may significantly vary in the relative portion of the DNN that is kept frozen. One typical design is to freeze the entire DNN but the last (top) layer; in contrast, another typical design is to freeze only several early layers of the DNN. 
In order to conduct transfer learning by partially freezing the network, we need to understand how the generalization performance and its evolution during training are affected by the freezing.

\subsection{The Examined Freezing Modes}
\label{subsec:The Examined Freezing Modes}
For the ResNet-18 model, we consider six levels of freezing in transfer learning. The levels are denoted numerically from `Frozen-0' to `Frozen-5'. The Frozen-0 level refers to using the entire pre-trained DNN as initialization of the target DNN without freezing any of the layers. The Frozen-5 level refers to freezing the entire DNN except for the last fully-connected layer (this last layer is trained via fine tuning the values transferred from the co-located layer of the DNN). In all of these freezing levels, the last layer is randomly initialized if it cannot be transferred (due to differences in the number of classes and/or input image size). The definition of all the examined freezing levels are described in Table \ref{tab:Freezing levels for Resnet-18} (more details on the ResNet-18 layout are provided in Appendix \ref{appendix:sec:Additional Details on the Examined DNN Architectures}). Corresponding freezing levels are defined for the DenseNet architecture, see Appendix \ref{appendix:sec:Additional Details on the Examined DNN Architectures}.

For the ViT architecture we consider 8 levels of freezing: from `Frozen-0' (no freezing at all) to `Frozen-7' (the entire DNN but the last fully-connected layer is frozen). `Frozen-1' has the first (patch embedding) layer frozen. `Frozen-2' to `Frozen-7' correspond to the gradual freezing of the 6 transformer blocks (each of these blocks includes an attention module, etc.) A more detailed definition of the ViT freezing levels is provided in Appendix \ref{appendix:sec:Additional Details on the Examined DNN Architectures}.

\begin{table}[t]
  \centering
\begin{tabular}{ |c|*{6}{|c}||c| } 
\hline
Level  & \multicolumn{6}{c||}{Frozen ResNet-18 components} & \shortstack{\# frozen\\layers} \\
\cline{2-7}
  & \shortstack{Conv\\layer 1~} & \shortstack{ResNet\\block 1~} & \shortstack{ResNet\\block 2~} & \shortstack{ResNet\\block 3~} & \shortstack{ResNet\\block 4~} & \shortstack{FC\\layer}  &  \\
\hline
Frozen-0 &  &  & & & & &0 \\ 
\hline
Frozen-1 & \checkmark &  & & & & & 1 \\ 
\hline
Frozen-2 & \checkmark & \checkmark & & & & & 5 \\ 
\hline
Frozen-3 & \checkmark & \checkmark & \checkmark & & & & 9 \\ 
\hline
Frozen-4 & \checkmark & \checkmark & \checkmark & \checkmark & & & 13 \\ 
\hline
Frozen-5 & \checkmark & \checkmark & \checkmark & \checkmark & \checkmark & & 17 \\ 
\hline
\end{tabular}
  \caption{Definition of the utilized freezing levels for ResNet-18. Each row specifies the frozen ResNet components in a particular freezing level. Recall that each of the ResNet-18 blocks is consisted of 4 convolutional layers.}
  \label{tab:Freezing levels for Resnet-18}
\end{table}

\subsection{Freezing Layers Can Eliminate the Epoch-wise Double Descent for Large Target Datasets}
\label{subsec:Freezing Layers Can Eliminate Double Descent for Large Target Datasets}

In Section \ref{subsec:effect of target dataset size} we showed that, in transfer learning with no frozen layers (i.e., the `Frozen-0' option), double descent  emerges  when the target dataset is sufficiently large. 
Here we consider transfer learning with frozen layers. Naturally, freezing layers in the target DNN effectively reduces its parameterization level (simply because less parameters are learned). Accordingly, freezing many layers can make the target DNN effectively underparameterized, namely, being far from interpolating the target training dataset. Moreover, the test error of such underparameterized DNN does not follow an epoch-wise double descent shape during training. 

\begin{figure*}[t]
  \centering
    \includegraphics[width=\linewidth]{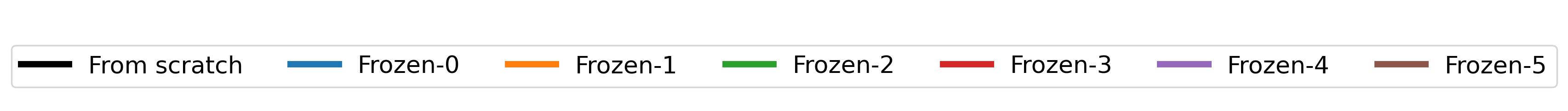}
   \\[-1ex]
  \subfloat[]{\includegraphics[width=0.24\linewidth]{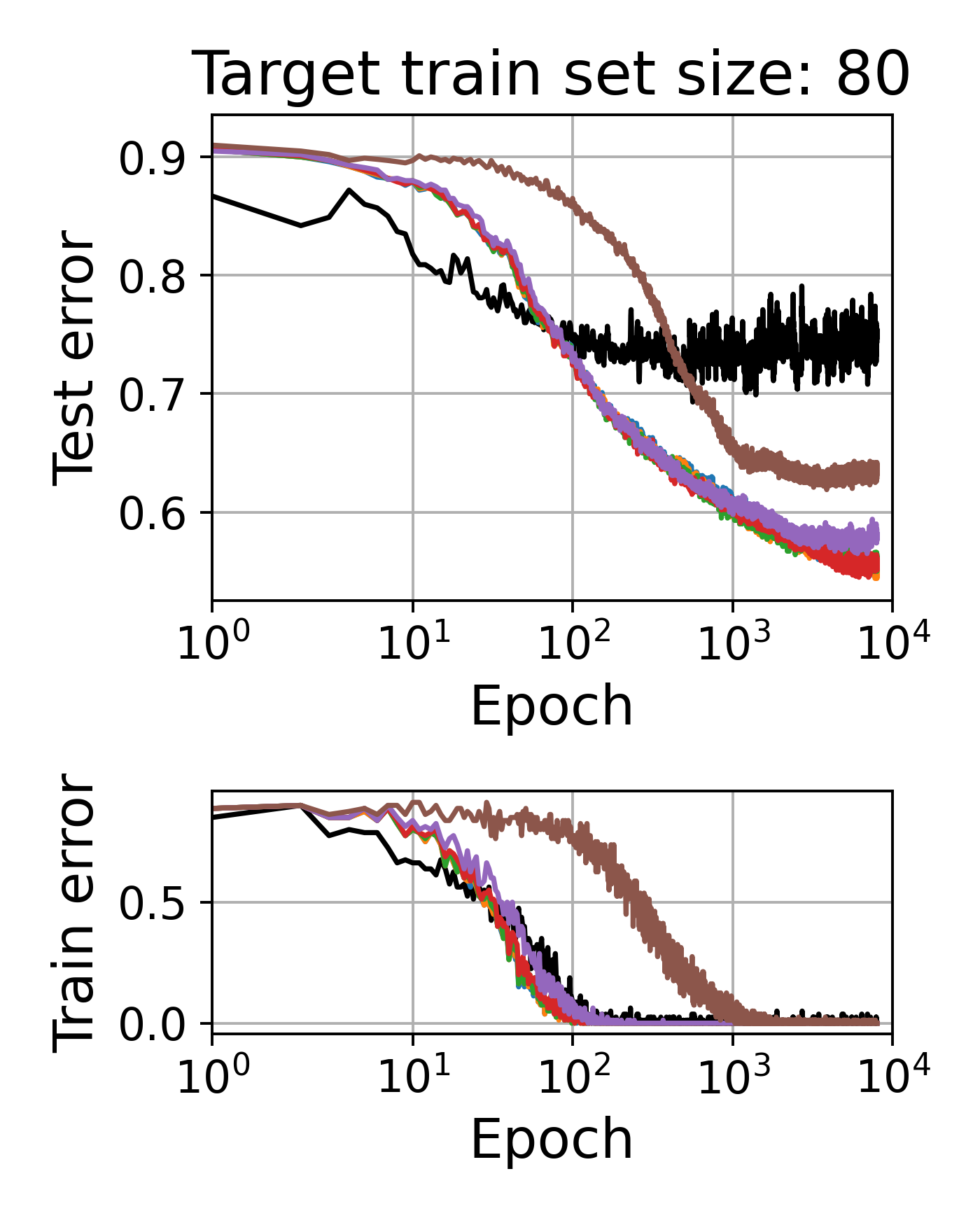}\label{fig:resnet_1d_error_diagrams_freezing_vs_epochs_tgtdataset80}}
  \subfloat[]{\includegraphics[width=0.24\linewidth]{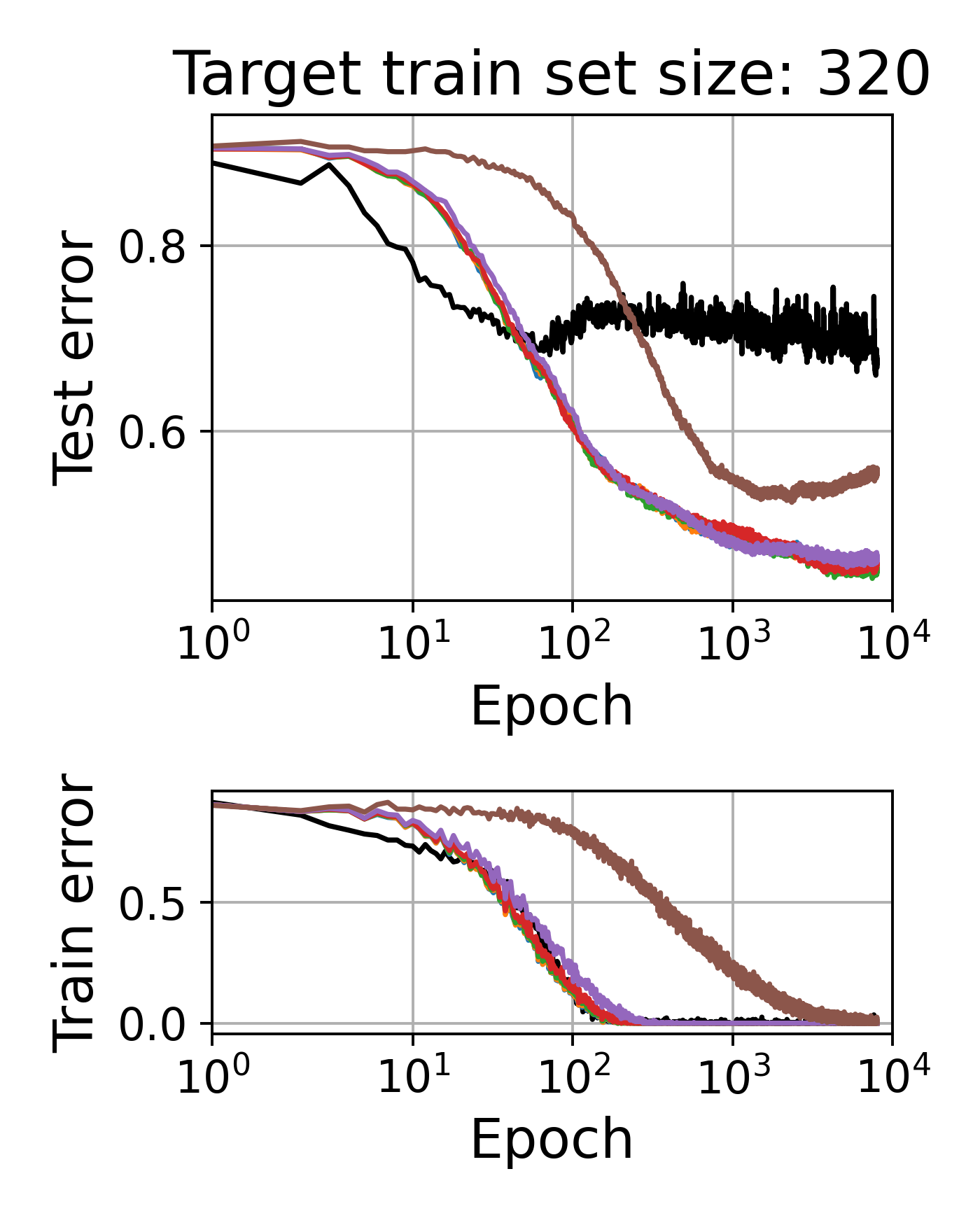}\label{fig:resnet_1d_error_diagrams_freezing_vs_epochs_tgtdataset320}}
  \subfloat[]{\includegraphics[width=0.24\linewidth]{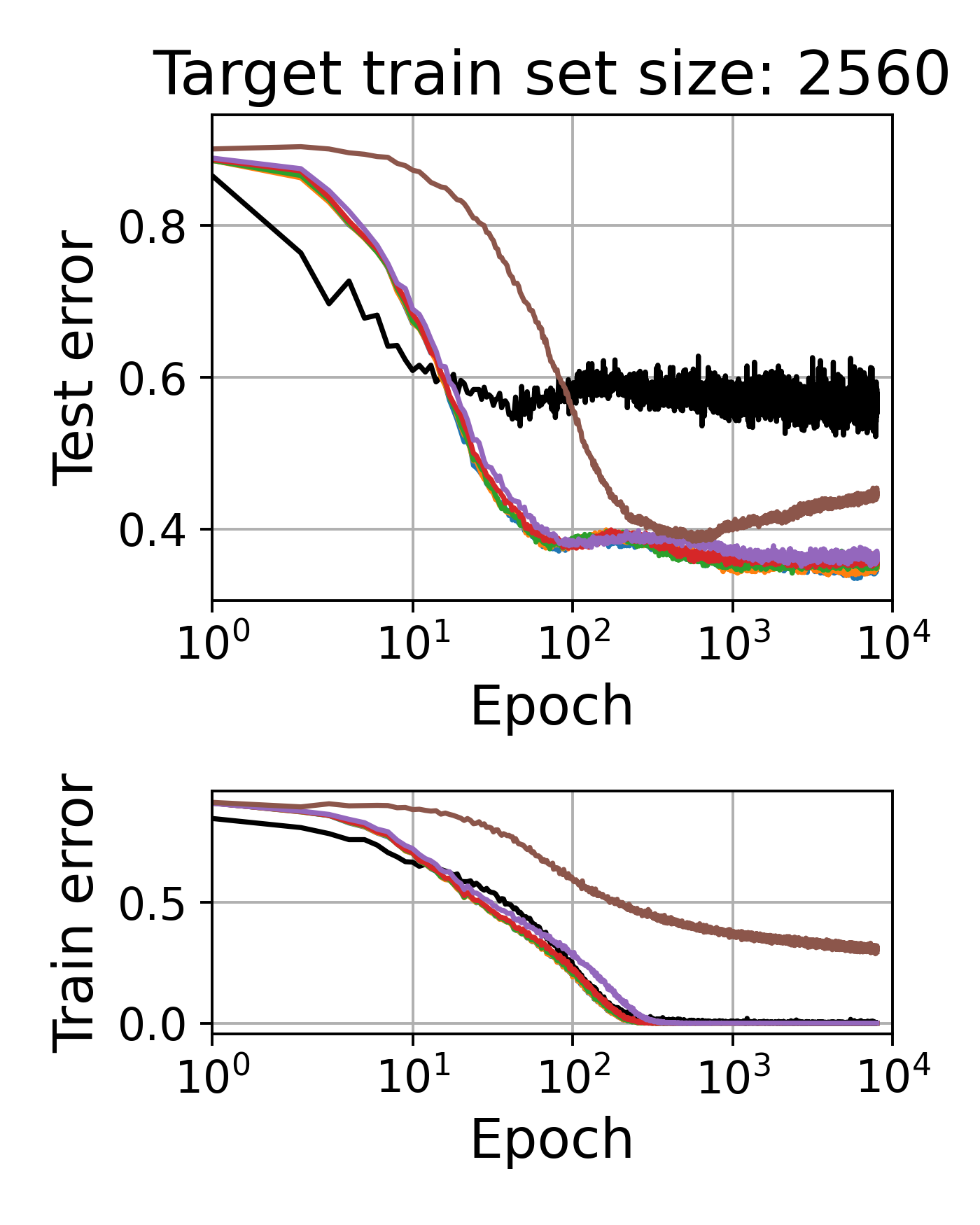}\label{fig:resnet_1d_error_diagrams_freezing_vs_epochs_tgtdataset2560}}
  \subfloat[]{\includegraphics[width=0.24\linewidth]{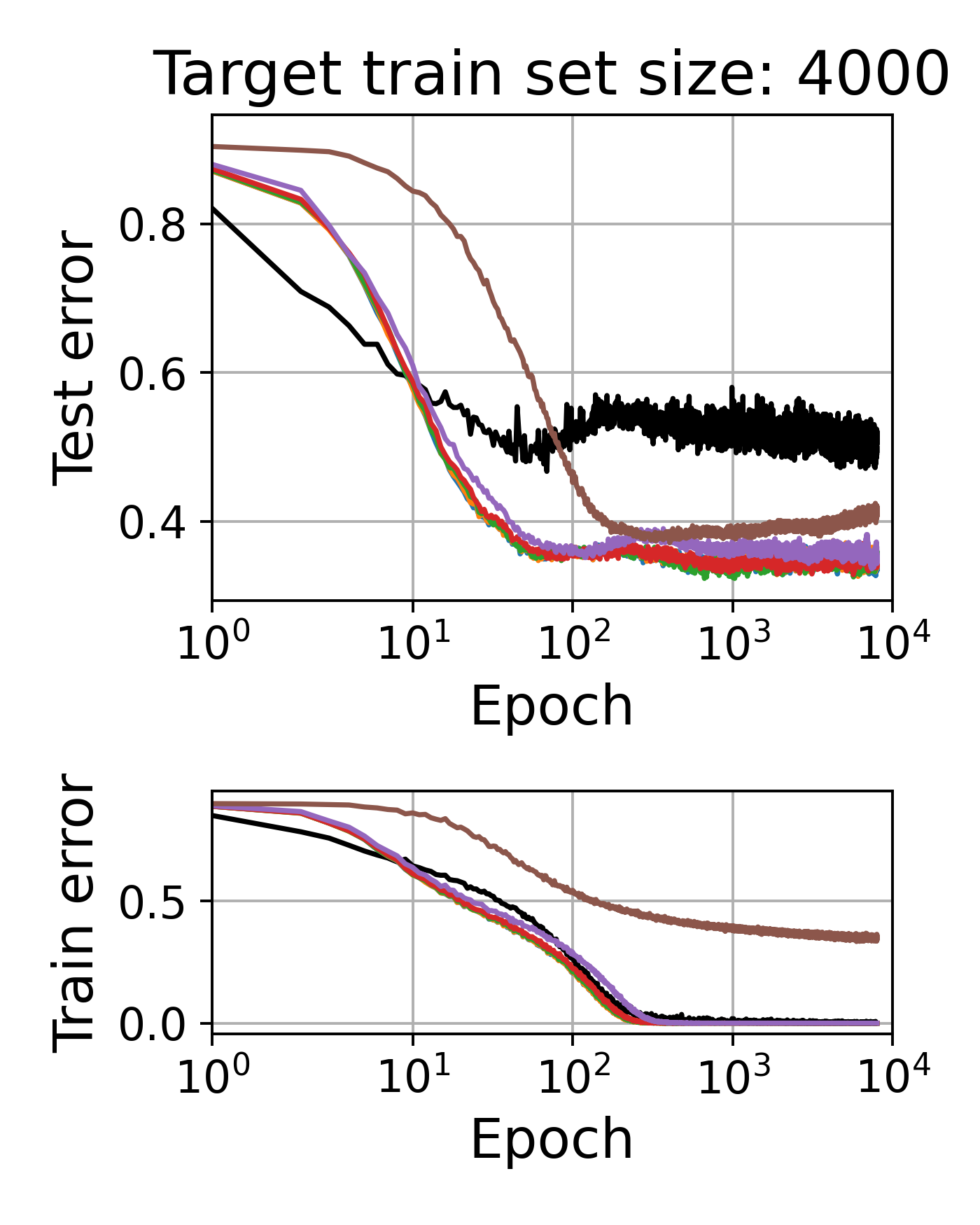}\label{fig:resnet_1d_error_diagrams_freezing_vs_epochs_tgtdataset4000}}
  \caption{Evaluation of transfer learning at various levels of freezing and comparison to learning from scratch. The target task is CIFAR-10 with 20\% label noise in target datasets. The subfigures differ in the target dataset size. The architecture is ResNet-18. The transfer learning are from the Tiny ImageNet source task (200 classes, input image size 64x64x3). For better visibility, the value range of the  vertical axis may differ among the subfigures.}
  \label{fig:resnet_1d_error_diagrams_freezing_vs_epochs}
\end{figure*}

This elimination of a double descent behavior due to freezing many layers is evident in Figs.~\ref{fig:resnet_1d_error_diagrams_freezing_vs_epochs_tgtdataset2560}-\ref{fig:resnet_1d_error_diagrams_freezing_vs_epochs_tgtdataset4000}, specifically, see the error curves of the `Frozen-5' setting (the brown curves) for sufficiently large target datasets. This behavior is further supported by the additional results in Appendix \ref{appendix:subsec:Additional Experimental Results for Section 4.2}. 

\begin{figure*}[t]
  \centering
  \subfloat[]{\includegraphics[width=0.33\linewidth]{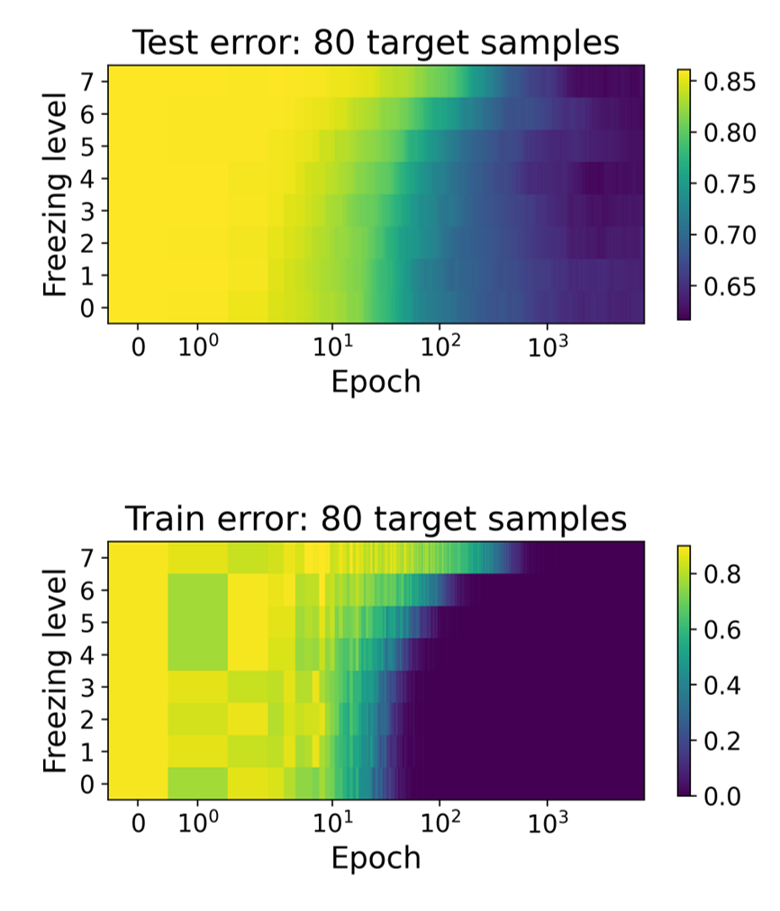}\label{fig:ViT_2d_error_diagrams_freezing_vs_epochs_tgtdataset80}}
  \subfloat[]{\includegraphics[width=0.33\linewidth]{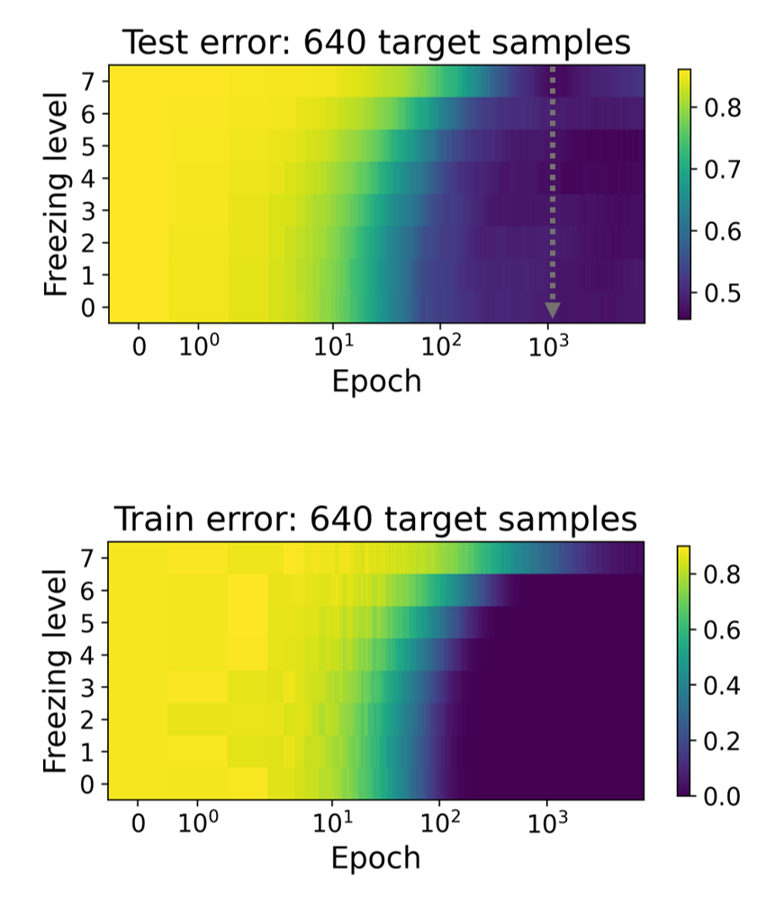}\label{fig:ViT_2d_error_diagrams_freezing_vs_epochs_tgtdataset640}}
  \subfloat[]{\includegraphics[width=0.33\linewidth]{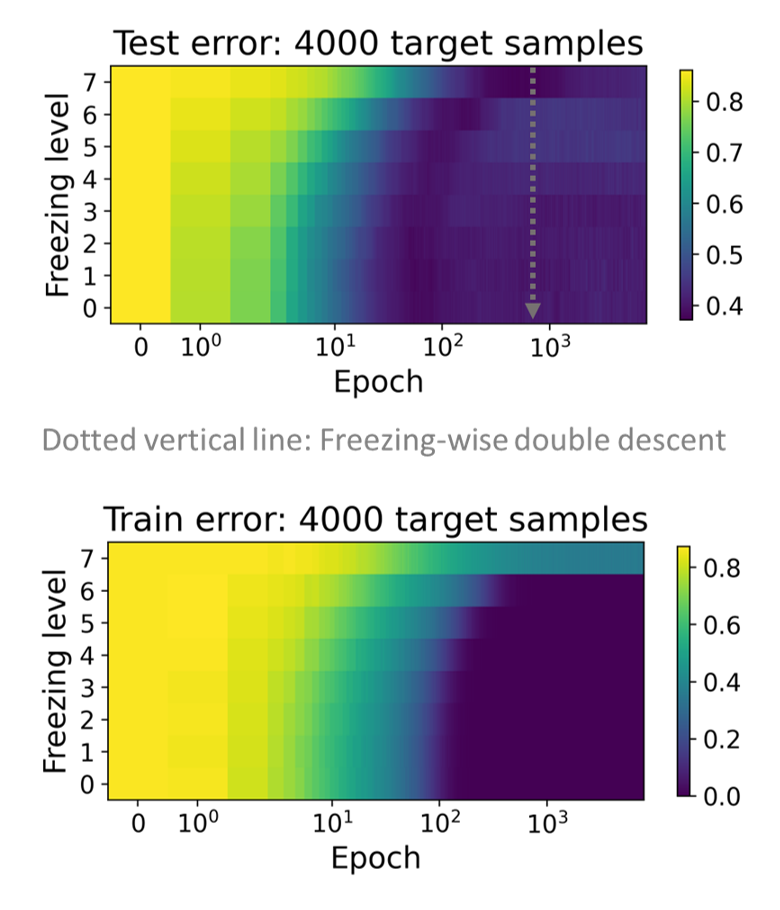}\label{fig:ViT_2d_error_diagrams_freezing_vs_epochs_tgtdataset4000}}
  \caption{Evaluation of transfer learning errors at various freezing levels and training epochs. The target task is CIFAR-10 with 20\% label noise in target datasets. The subfigures (a), (b), (c) differ in the target dataset size. The dotted gray line exemplifies the trajectory of the freezing-wize double descent. The architecture is ViT. The transfer learning are from the Tiny ImageNet source task (200 classes, input image size 32x32x3).}
  \label{fig:ViT_2d_error_diagrams_freezing_vs_epochs}
\end{figure*}

\subsection{Freezing-wise Double Descent}
\label{subsec:Freezing-wise Double Descent}

As the number of frozen layers determines the effective parameterization level of the target DNN, it also determines whether the learned DNN interpolates the target dataset or not. This leads to the question whether a double descent phenomenon can emerge as a function of the freezing level. 
Indeed, a freezing-wise double descent phenomenon\footnote{In this paper, we use the term \textit{double descent phenomenon} to denote any generalization behavior that its test error peaks around the interpolation threshold of its training data . This definition also includes test error curves that do not have the ``first'' descent in the underparameterized regime.} can occur in transfer learning, as we will show next. 

To observe the freezing-wise double descent we need sufficient resolution in the freezing level grid; accordingly, the eight freezing levels that we use for ViT are useful for observing the contour of the interpolation threshold in the 2D error diagrams in Fig.~\ref{fig:ViT_2d_error_diagrams_freezing_vs_epochs} (specifically, the interpolation threshold is the contour where the train error becomes numerically zero, i.e., this is the boundary between the dark blue and the lighter colors in Fig.~\ref{fig:ViT_2d_error_diagrams_freezing_vs_epochs}). 
The vertical gray dotted lines in Figs.~\ref{fig:ViT_2d_error_diagrams_freezing_vs_epochs_tgtdataset640}-\ref{fig:ViT_2d_error_diagrams_freezing_vs_epochs_tgtdataset4000} exemplify trajectories of the freezing-wise double descent, for a specific epoch, with the following aspects:  
\begin{itemize}
    \item The target model at the maximal freezing level (Frozen-7) does not interpolate the target dataset (its train error is significantly greater than zero) and the corresponding test error is relatively good.
    \item The target error peaks around the interpolation threshold (the Frozen-6 level) and then decreases when the freezing level is further reduced.
    \item For a relatively large target dataset (e.g., 4000 target samples in Fig.~\ref{fig:ViT_2d_error_diagrams_freezing_vs_epochs_tgtdataset4000}) the test error does not decrease enough due to increased overparameterization (as induced by less frozen layers) to beat the underparameterized performance with maximal freezing (Frozen-7). 
    \item For a medium-sized target dataset (e.g., 640 target samples in Fig.~\ref{fig:ViT_2d_error_diagrams_freezing_vs_epochs_tgtdataset640}) the test error decreases significantly and arrives (e.g., at the Frozen-4 level) to a test error that competes or outperforms the underparameterized option of maximal freezing (Frozen-7). Even though the best performance is obtained at the overparameterized regime of freezing, note that the minimal freezing level (Frozen-0) is not necessarily the best. 
\end{itemize}

Further, the results in Fig.~\ref{fig:ViT_2d_error_diagrams_freezing_vs_epochs} show that for sufficiently small target dataset the best freezing mode corresponds to interpolation of the target dataset (see Fig.~\ref{fig:ViT_2d_error_diagrams_freezing_vs_epochs_tgtdataset80}). Additional results are provided in Appendix \ref{appendix:subsec:Additional Experimental Results for Section 4.3}.

\section{Double Descent Can Make the Less Related Task Better to Transfer From}
\label{sec:Task Similarity}

The similarity between the source and target tasks naturally affects the generalization performance in transfer learning. In this section we examine whether transfer from a more related task is necessarily better than transfer from a less related task. 

\begin{figure*}[t]
  \centering
  \includegraphics[width=0.6\linewidth]{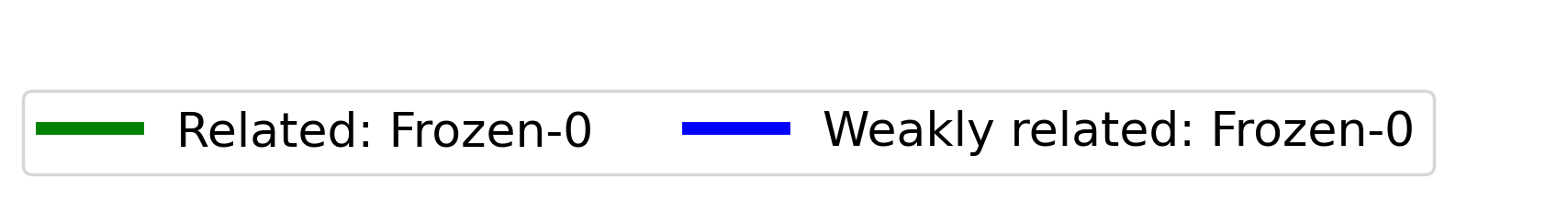}
  \\[-1ex]
    \subfloat[ResNet-18 (target dataset with 20\% label noise)]{
    \includegraphics[width=0.24\linewidth]{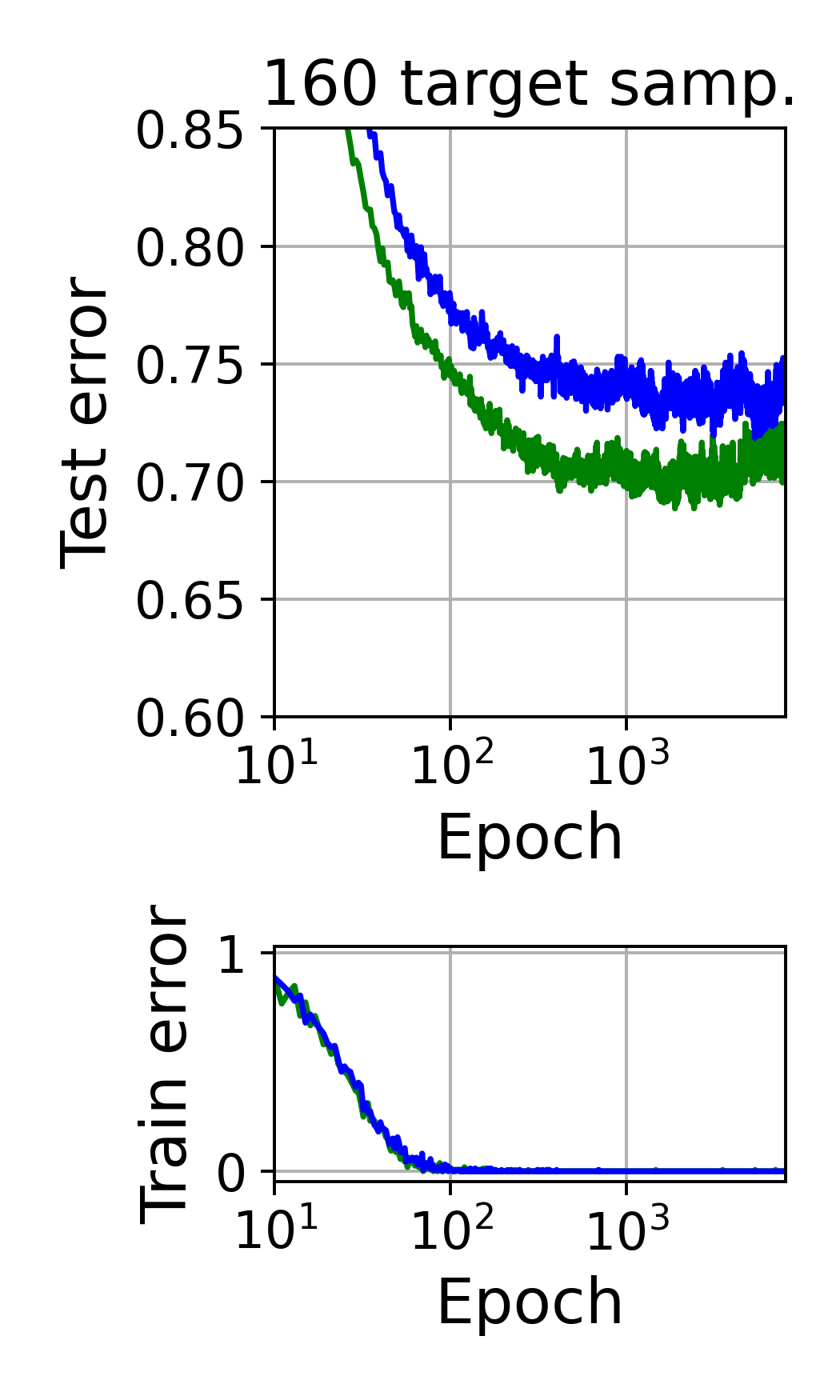}
            \includegraphics[width=0.24\linewidth]{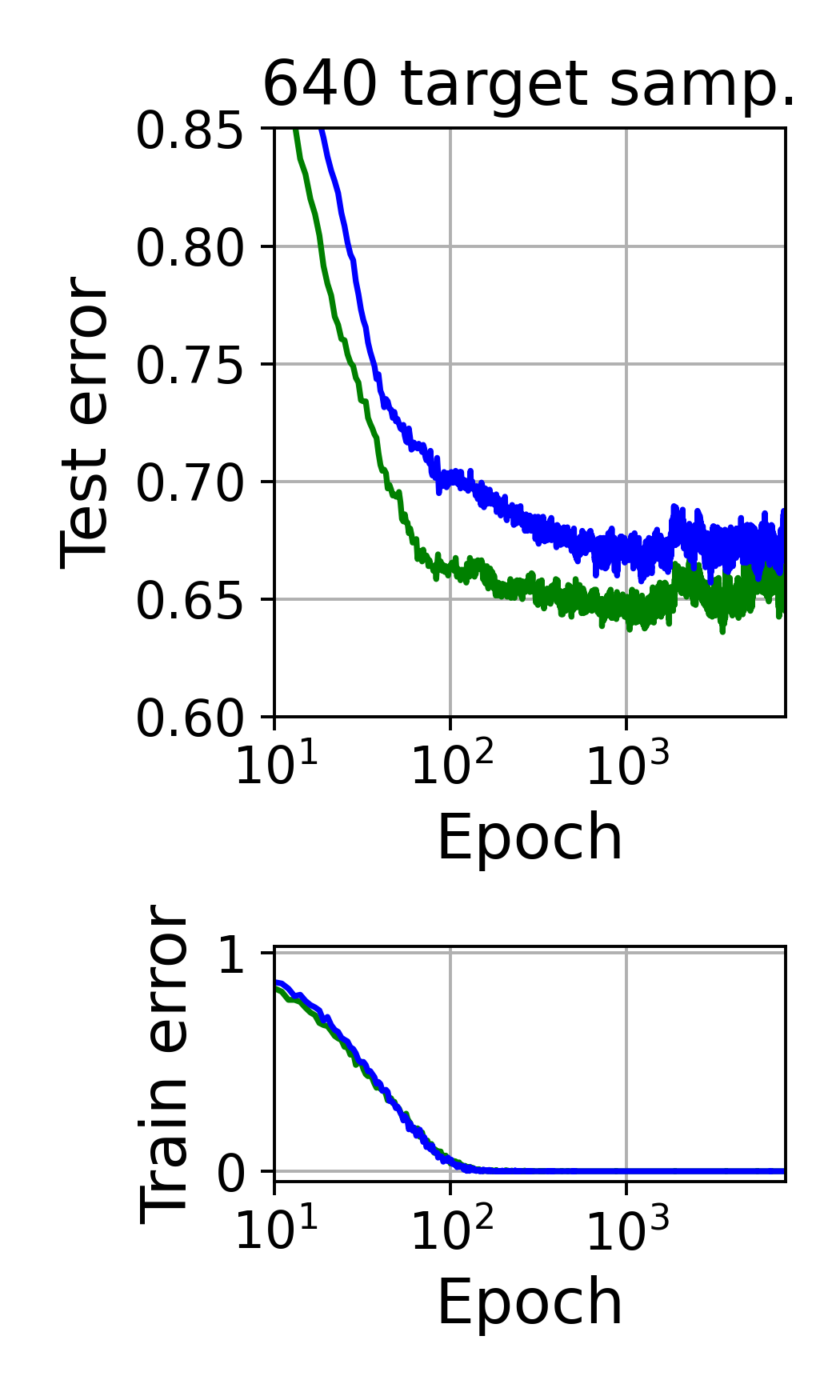}
                \includegraphics[width=0.24\linewidth]{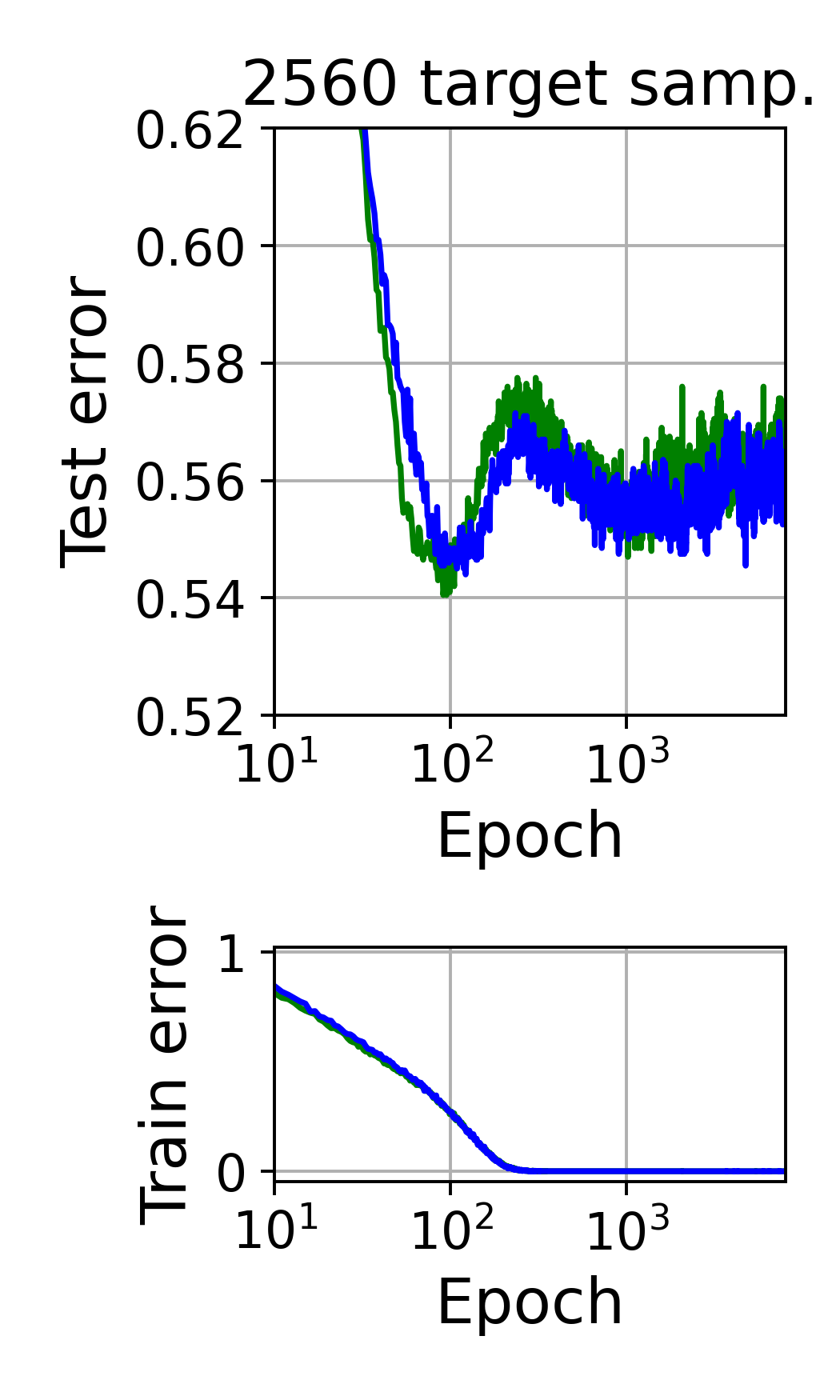}
                    \includegraphics[width=0.24\linewidth]{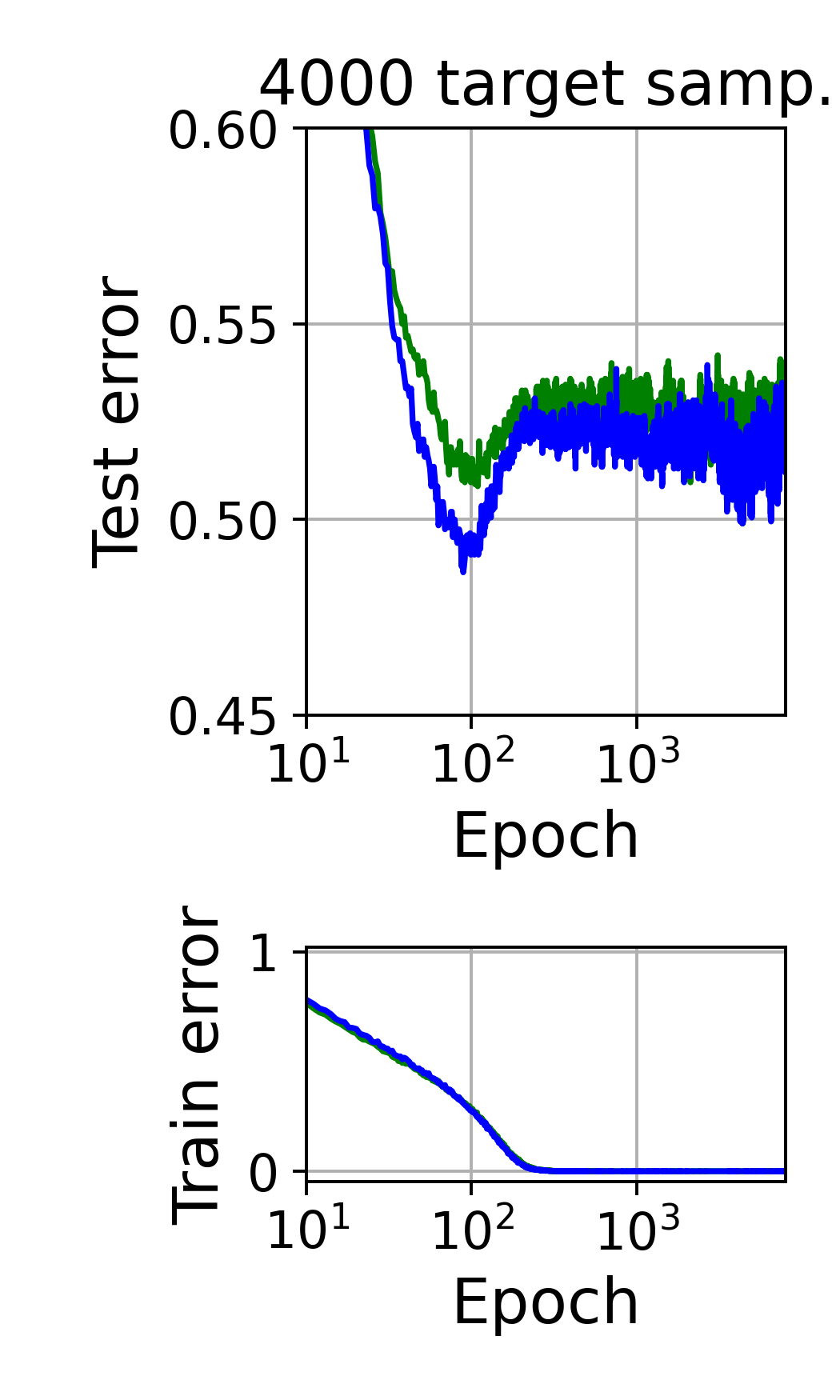}
    \label{subfig:task_similarity_resnet_tinyimagenet64First40Classes-vs-tinyimagenet64Similar40Classes_srcNoiselessTgtNoisy0p2} }
  \caption{Transfer from a more related task can be less beneficial due to double descent. The target task includes 40 classes from CIFAR-100. The comparison is between transfer learning from a related task (40 classes from Tiny ImageNet that are conceptually related to the target task classes) and transfer learning from a weakly related task (40 classes from Tiny ImageNet that are arbitrarily chosen). Each subfigure corresponds to another size of the target dataset. For large target datasets, transfer from a more related task can be less beneficial. See Appendix \ref{appendix:sec:Additional Experimental Results for Section 5} for a detailed description of the experiment setting.}
  \label{fig:task_similarity}
\end{figure*}

 Our results in Fig.~\ref{subfig:task_similarity_resnet_tinyimagenet64First40Classes-vs-tinyimagenet64Similar40Classes_srcNoiselessTgtNoisy0p2}  show that the double descent phenomenon induces cases where \textbf{transfer from a less related task can generalize on par or better than a transfer from a more related task} (see the results for larger target datasets in Fig.~\ref{subfig:task_similarity_resnet_tinyimagenet64First40Classes-vs-tinyimagenet64Similar40Classes_srcNoiselessTgtNoisy0p2}). These examples contradict the basic intuition that it is better to transfer from a more similar task.  See additional results in Appendix \ref{appendix:sec:Additional Experimental Results for Section 5}.

In general, a particular learning setting whose test error follows a double descent shape has a degraded generalization performance in the interval of epochs around the double descent peak of its test error. Moreover, the compared learning settings may have their double descent peaks at different epochs. These reasons may cause the seemingly more beneficial learning setting to be outperformed by its competing setting. 
In our case, this implies that transfer learning from a more related task can be less beneficial than transfer learning from a less related task.


\section{Conclusion}
\label{sec:Conclusion} 
In this work we have developed and validated a new approach to research of the generalization performance of transfer learning. 
We have explicitly considered overparameterization concepts such as the double descent phenomenon and interpolation of the training data. 
Our findings elucidate how the generalization behavior of transfer learning is affected by the source and target dataset sizes, the number of frozen layers, and the similarity between the source and target tasks. 
We believe that our perspective and insights open a new direction to explore generalization in transfer learning in various problems and settings.

\section*{Acknowledgments}
This work was supported by NSF grants CCF-1911094, IIS-1838177, and IIS-1730574; ONR grants N00014-18-12571, N00014-20-1-2534, and MURI N00014-20-1-2787;
AFOSR grant FA9550-22-1-0060; and a Vannevar Bush Faculty Fellowship, ONR grant N00014-18-1-2047.

\bibliography{deep_transfer_learning_references}
\bibliographystyle{IEEEtran}

\appendix

\renewcommand\thefigure{\thesection.\arabic{figure}}    
\setcounter{figure}{0}  

\renewcommand\thetable{\thesection.\arabic{table}}    
\setcounter{table}{0} 

\counterwithin*{figure}{section}
\counterwithin*{table}{section}

\section*{Appendices}
\label{sec:Appendix Outline}

These appendices support the main paper as follows. Appendix \ref{appendix:sec:Additional Details on the Examined DNN Architectures} provides additional details on the DNN architectures that we examine in this paper. 
Appendix \ref{appendix:sec:Additional Details on the Examined Classification Problems} provides additional details on the classification problems that we consider in this paper. 
Appendices \ref{appendix:sec:Additional Experimental Results for Section 3}, \ref{appendix:sec:Additional Experimental Results for Section 4}, \ref{appendix:sec:Additional Experimental Results for Section 5} include additional experimental results that further support Sections \ref{sec:double descent in transfer learning}, \ref{sec:Freezing Layers}, \ref{sec:Task Similarity} in the main text, respectively. 
\ydarnew{arxiv v2: compute resources and limitations are removed for the arxiv version}

The indexing of figures in the Appendices below is prefixed with the letter of the corresponding Appendix. Other references correspond to the main paper. 


\section{Additional Details on the Examined DNN Architectures}
\label{appendix:sec:Additional Details on the Examined DNN Architectures}
In this paper we examine three DNN architectures: ResNet-18, DenseNet-121 and ViT. In this Appendix we describe the relevant details of the examined architectures, in addition to the details already provided in Sections \ref{sec:The Transfer Learning Settings} and \ref{sec:Freezing Layers} (specifically, note that the technical details of the training optimization are provided in Section \ref{sec:The Transfer Learning Settings}  and not here). 

\subsection{ResNet-18}
\label{appendix:sec:Additional Details on the Examined ResNet}
The examined ResNet-18 has the common structure and dimensions according to the number of classes in the addressed classification problem. We use the same ResNet-18 form as in \cite{nakkiran2019deep} that is based on the implementation from \url{https://github.com/kuangliu/pytorch-cifar}. Namely, this is the Preactivation ResNet-18 \ydar{did this ResNet type appear in the original ResNet paper cite{he2016deep}?} with the following structure: 
\begin{itemize}
    \item Convolutional layer
    \item 4 ResNet-18 blocks, each of these ResNet blocks includes a sequence of batch norm, ReLU, convolution, batch norm, ReLU, convolution.
    \item average pooling
    \item fully connected layer
\end{itemize}
The dimensions of the layers are as in \cite{nakkiran2019deep} for the standard width parameter $k=64$.

There is a total of 18 layers with learnable parameters; specifically, the first and the last layers, and four learnable layers in each of the ResNet blocks.
Accordingly, as explained in Section \ref{sec:Freezing Layers} , we consider 6 levels of freezing in ResNet-18: 
\begin{itemize}
    \item Frozen-0: No frozen layers at all, i.e., the entire target DNN can be trained from its initialization using the pre-trained source DNN.
    \item Frozen-1: Only the first convolutional layer is frozen with its pre-trained parameters in the target DNN training. 
    \item Frozen-2: Only the first convolutional layer and the first Resnet-18 block are frozen.
    \item Frozen-3: Only the first convolutional layer and the first two Resnet-18 blocks are frozen.
    \item Frozen-4: Only the first convolutional layer and the first three Resnet-18 blocks are frozen.
    \item Frozen-5: Only the first convolutional layer and the four Resnet-18 blocks are frozen. Namely, only the last fully-connected layer in the target DNN can be trained. 
\end{itemize}

\subsection{DenseNet}
\label{appendix:sec:Additional Details on the Examined DenseNet}
We consider a DenseNet-121 architecture in the same form and dimensions (specifically, with growth rate 12) as in \cite{somepalli2022can}. 
The examined DenseNet has the following structure: 
\begin{itemize}
    \item Convolutional layer
    \item 4 DenseNet-121 blocks, each of these DenseNet blocks includes several layers.
    \item average pooling
    \item fully connected layer.
\end{itemize}
Although the architecture (e.g., the connection forms) and number of layers are different between ResNet and DenseNet blocks,  the high-level structure (i.e., a conv layer, 4 blocks, average pooling, and FC layer) in DenseNet and ResNet is similar. Accordingly, we define the following 6 freezing levels in DenseNet: 
\begin{itemize}
    \item Frozen-0: No frozen layers at all, i.e., the entire target DNN can be trained from its initialization using the pre-trained source DNN.
    \item Frozen-1: Only the first convolutional layer is frozen with its pre-trained parameters in the target DNN training. 
    \item Frozen-2: Only the first convolutional layer and the first DenseNet-121 block are frozen.
    \item Frozen-3: Only the first convolutional layer and the first two DenseNet-121 blocks are frozen.
    \item Frozen-4: Only the first convolutional layer and the first three DenseNet-121 blocks are frozen.
    \item Frozen-5: Only the first convolutional layer and the four DenseNet-121 blocks are frozen. Namely, only the last fully-connected layer in the target DNN can be trained. 
\end{itemize}

\subsection{Vision Transformer (ViT)}
\label{appendix:sec:Additional Details on the Examined ViT}
We consider a ViT architecture in the same form and dimensions as in \cite{somepalli2022can}, namely, a patch size of $4\times4$ pixels, 6 transformer layers (each of these is actually a block of several layers), and 8 heads. This ViT architecture allows us to extensively examine its generalization behavior at various settings. 
The implementation of the ViT is based on the code from \url{https://github.com/lucidrains/vit-pytorch}. 

The structures and layers of the examined ViT can be described as follows: 
\begin{itemize}
    \item Patch embedding
    \item 6 transformer layers, each of these layers is actually a block of layers that includes norm layers, a multi-head attention sub-block, and an MLP
    \item an MLP head.
\end{itemize}
As explained in Section \ref{sec:Freezing Layers} , we consider 8 levels of freezing in ViT: 
\begin{itemize}
    \item Frozen-0: No frozen layers at all, i.e., the entire target DNN can be trained from its initialization using the pre-trained source DNN.
    \item Frozen-1: Only the patch embedding layer is frozen with its pre-trained parameters in the target DNN training. 
    \item Frozen-2: Only the patch embedding layer and the first transformer layer are frozen.
    \item Frozen-$r$ for $r\in\{3,\dots,6\}$: Only the patch embedding layer and the first $r-1$ transformer layers are frozen.    
    \item Frozen-7: Only the patch embedding layer and the six transformer layers are frozen. Namely, only the MLP head at the end of the target DNN can be trained. 
\end{itemize}


\section{Additional Details on the Examined Classification Tasks}
\label{appendix:sec:Additional Details on the Examined Classification Problems}

In this paper we consider classification problems that are defined based on the CIFAR-10, CIFAR-100 \cite{krizhevsky2009learning}, Food-101 \cite{bossard2014food}, and Tiny ImageNet \cite{le2015tiny} datasets. Table \ref{tab:source and target tasks - complete list - appendix} provides the complete list of source-target task pairs that are examined in the experiments in the main paper and appendices (some of these source-target task pairs were considered only in part of the experiments). In this Appendix we provide additional details that were not included in Section \ref{sec:The Transfer Learning Settings}  of the main paper due to the limited space.

\begin{table}[t]
  \centering
\small{
\begin{tabular}{ *{2}{|c}|*{2}{|c}| } 
\hline

\multicolumn{2}{|c||}{Source task} & \multicolumn{2}{c|}{Target task} \\
\cline{1-4}
\# classes & Dataset & \# classes & Dataset  \\
\hline
\hline
10 & CIFAR-10 (all 10 classes) & 10 & Food-101 (first 10 classes)\\ 
\hline
40 & Tiny Imagenet (first 40 classes) & 40 & CIFAR-100 (40 classes, 2 per superclass) \\ 
\hline
40 & \shortstack{Tiny Imagenet\\(40 classes similar to target classes)} & 40 & CIFAR-100 (40 classes, 2 per superclass) \\ \hline
100 & Tiny Imagenet (first 100 classes) & 100 & CIFAR-100 (all 100 classes) \\ \hline
100 & Tiny Imagenet (first 100 classes) & 100 & Food-101 (first 100 classes) \\ \hline
200 & Tiny Imagenet (all 200 classes) & 10 & CIFAR-10 \\ \hline
\hline
\end{tabular}
}
  \caption{The source and target classification tasks that are examined in the main paper and appendices.}
  \label{tab:source and target tasks - complete list - appendix}
\end{table}

All input images are of 32x32x3 pixels size, except for the Tiny ImageNet dataset for which we examine also the 64x64x3 pixels size in the experiments for ResNet
and DenseNet. 

As shown in Table \ref{tab:source and target tasks - complete list - appendix} we examine transfer learning between tasks with the same or with different number of classes. The number of classes determines the output dimension of the last fully-connected layer in the DNN. The input image size determines the input dimension of the last fully-connected layer in ResNet and DenseNet. \ydarnew{In ViT, the input image size affects the dimension of an earlier layer (due to the different number of patches for a different image size where the patch size is fixed), and therefore we consider ViT only in transfer between tasks with the same input image size.} Accordingly, our transfer learning settings between tasks with the same number of classes and the same input image size enables to transfer all the source DNN layers to the target DNN; otherwise (i.e., if the source and target tasks do not have the same number of classes and/or input image size), the last fully connected layer cannot be transferred and hence randomly initialized in the target DNN whereas the other layers are transferred from the source DNN. 

Table \ref{tab:source and target tasks - complete list - appendix} shows that some of the examined classification tasks are defined by considering all the available classes in the relevant dataset (for example, a classification task among all the 200 classes in Tiny ImageNet, a classification task among all the 100 classes in CIFAR-100, etc.) Yet, we also find it useful to define classification tasks based on subsets of standard datasets (for example, a classification task among 100 classes from Tiny ImageNet); such classification tasks are useful for examining transfer learning settings between source and target tasks with the same number of classes, and for examining transfer from source tasks at different similarity levels to the target task (as will be explained next). 

Some of the examined tasks are defined simply by considering the first classes in the dataset, namely, the classes with the lowest numerical class labels (for example, the first 100 classes in Tiny ImageNet are those corresponding to labels from 0 to 99). 

We also define a 40-class target task based on CIFAR-100, specifically, by considering two classes from each of the 20 CIFAR-100 superclasses as desribed in Table \ref{tab:CIFAR-100 superclasses} (recall that CIFAR-100 is defined with an inner partitioning of its 100 classes into 20 superclasses of 5 classes each). For this target task, we examine the transfer learning from two source tasks that differ in their similarity to the target task: 
\begin{itemize}
    \item Classification among the first 40 classes in Tiny ImageNet (i.e., the ImageNet classes with lables from 0 to 39)
    \item Classification among the 40 classes from Tiny ImageNet that were manually selected according to their conceptual similarity to the 40 CIFAR classes of the target task. These classes are: fountain, dugong, goldfish, jellyfish, brain coral, cauliflower, plate, frying pan, banana, bell pepper, computer keyboard, refrigerator, space heater, wooden spoon, fly, cockroach, lion, bighorn, water tower, obelisk, lakeside, cliff, orangutan, gazelle, king penguin, koala, snail, scorpion, academic gown, sunglasses, European fire salamander, tailed frog, sea slug, lesser panda, broom, barrel, limousine, convertible, go-kart, beach wagon. A comparison of this list to the CIFAR classes in Table \ref{tab:CIFAR-100 superclasses} shows the matching in the conceptual category of the classes between the two tasks, although for some of the classes the selection was relatively arbitrary due to lack of similar classes. 
\end{itemize}

\ydarnew{if we include the results for the mixed food-imagenet datasets, they should be also explained here and mentioned in the table }

\begin{table*}
  \centering
\begin{tabular}{ |l||l| } 
\hline
Superclass  & Classes \\
\hline
\hline
aquatic mammals &  \textbf{whale}, \textbf{dolphin}, seal, beaver, otter \\ 
\hline
fish &  \textbf{ray}, \textbf{trout}, shark, flatfish, aquarium fish \\ 
\hline
flowers &  \textbf{sunflower}, \textbf{orchid}, tulip, rose, poppy \\ 
\hline
food containers &  \textbf{bottle}, \textbf{bowl}, can, plate,  cup \\ 
\hline
fruit and vegetables &  \textbf{orange}, \textbf{pear}, mushroom, apple, sweet pepper \\
\hline
household electrical devices &  \textbf{clock}, \textbf{television}, lamp, keyboard, telephone \\ 
\hline
household furniture &  \textbf{wardrobe}, \textbf{table}, chair, couch, bed \\ 
\hline
insects &  \textbf{bee}, \textbf{beetle}, butterfly, cockroach, caterpillar \\ 
\hline
large carnivores & \textbf{leopard}, \textbf{bear}, lion, wolf, tiger \\ 
\hline
large man-made outdoor things &  \textbf{skyscraper}, \textbf{bridge}, house, castle,  road \\ 
\hline
large natural outdoor scenes &  \textbf{sea}, \textbf{cloud}, mountain, forest,  plain \\ 
\hline
large omnivores and herbivores & \textbf{cattle}, \textbf{elephant}, camel, chimpanzee, kangaroo \\ 
\hline
medium-sized mammals & \textbf{raccoon}, \textbf{possum}, skunk, porcupine, fox \\ 
\hline
non-insect invertebrates & \textbf{lobster}, \textbf{worm}, snail, crab, spider \\ 
\hline
people &   \textbf{girl}, \textbf{man}, boy, baby, woman \\ 
\hline
reptiles &  \textbf{crocodile}, \textbf{snake}, dinosaur, turtle, lizard \\ 
\hline
small mammals & \textbf{shrew}, \textbf{hamster}, rabbit, mouse, squirrel \\ 
\hline
trees & \textbf{maple tree}, \textbf{pine tree}, palm tree, oak tree,  willow tree \\ 
\hline
vehicles 1 & \textbf{train}, \textbf{bus}, bicycle, pickup truck, motorcycle \\ 
\hline
vehicles 2 &  \textbf{lawn mower}, \textbf{tractor}, streetcar, rocket, tank \\ 
\hline
\end{tabular}
  \caption{The CIFAR-100 superclasses and classes. The classes in bold text compose the \textit{target} 40-class task of CIFAR-100 in this paper}
  \label{tab:CIFAR-100 superclasses}
\end{table*}

\section{Additional Results for Section 3}
\label{appendix:sec:Additional Experimental Results for Section 3}

\subsection{Additional Results for Section 3.1}
\label{appendix:subsec:Additional Experimental Results for Section 3.1}

\begin{figure*}[t]
  \centering
  \includegraphics[width=0.85\textwidth]{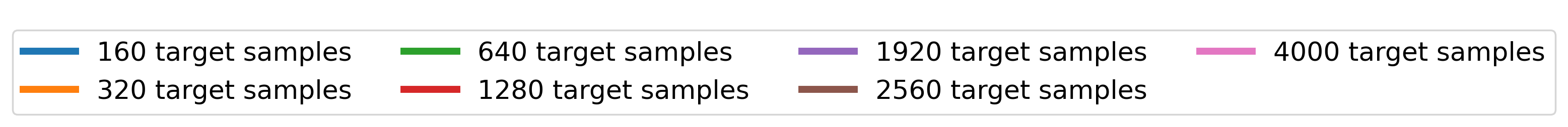}
  \\[-3ex]
  \subfloat[Noiseless]{
   \includegraphics[width=0.49\textwidth]{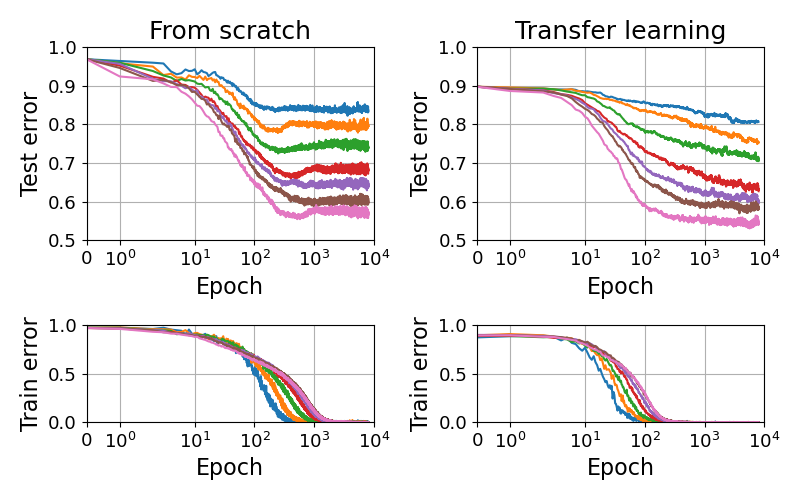}
   \label{fig:test_error_vs_epoch_for_several_dataset_sizes_vit_BSP20_src20k_tinyimagenet32_src40SimilarClasstgtCIFAR40class_tgtNOISELESS}}
   \subfloat[20\% label noise in target dataset]{\includegraphics[width=0.49\textwidth]{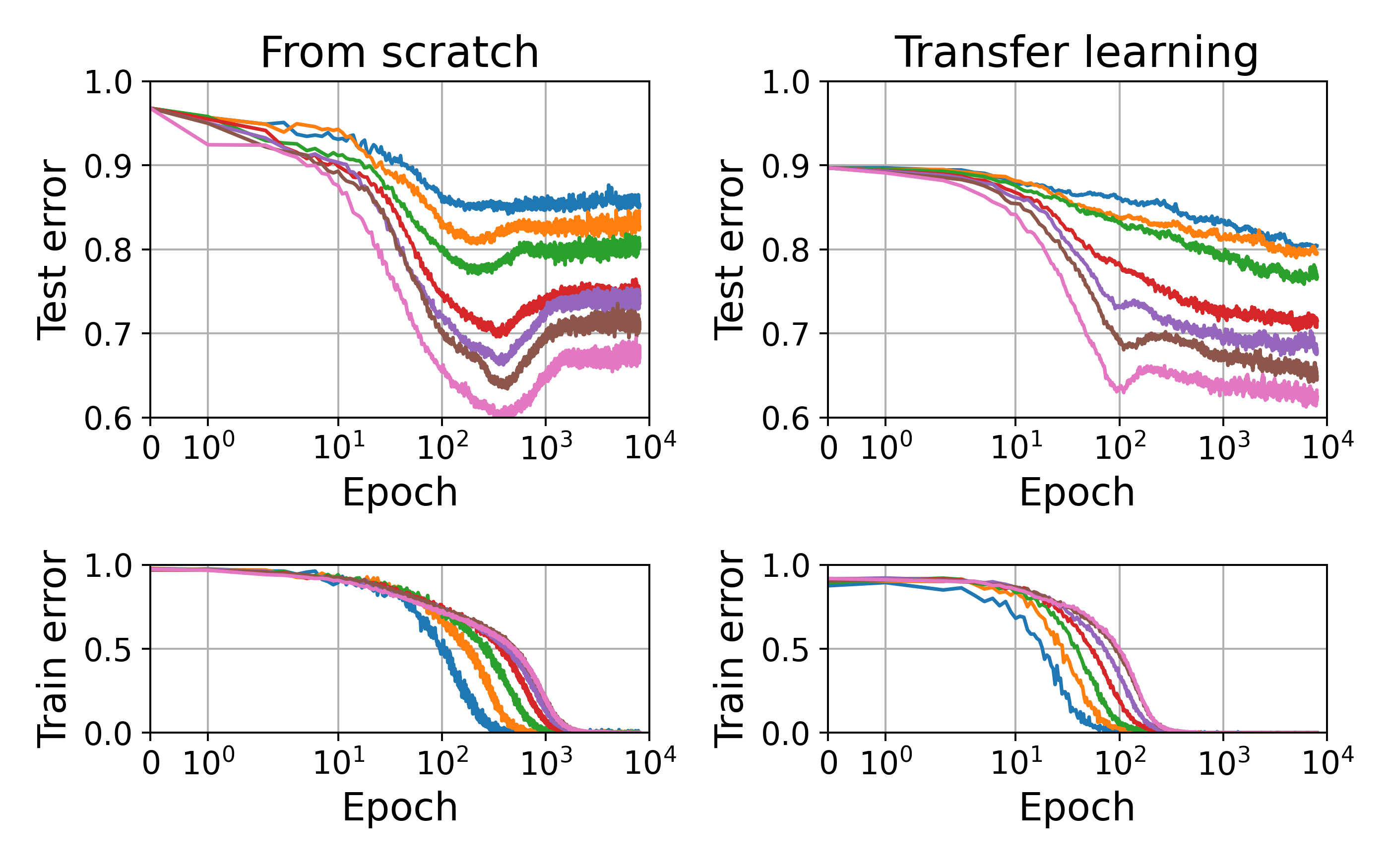}
\label{fig:test_error_vs_epoch_for_several_dataset_sizes_vit_BSP20_src20k_tinyimagenet32_src40SimilarClasstgtCIFAR40class}}
   \caption{Evaluation of \textbf{ViT} training for a target classification task of 40 classes from CIFAR-100. The transfer learning is from the source task of 40 Tiny ImageNet classes that are similar to the target task classes (input image size 32x32x3) with source dataset of 20k training samples. Each curve color corresponds to another size of the target dataset. For better visibility, the value range of the vertical axis differs between the subfigures of the noiseless and noisy settings.}
   \label{fig:test_error_vs_epoch_for_several_dataset_sizes_vit_BSP20_src20k_tinyimagenet32_src40SimilarClasstgtCIFAR40class}
\end{figure*}

\begin{figure*}[t]
  \centering
  \includegraphics[width=0.85\textwidth]{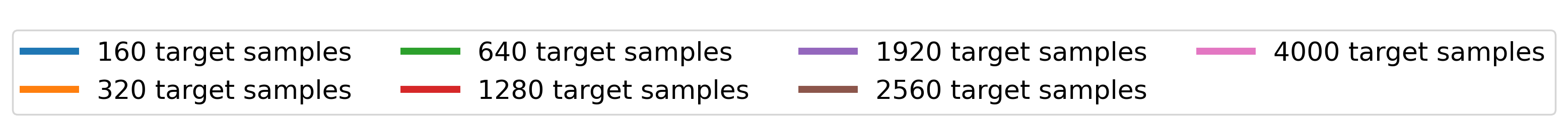}
  \\[-3ex]
  \subfloat[Noiseless]{
   \includegraphics[width=0.49\textwidth]{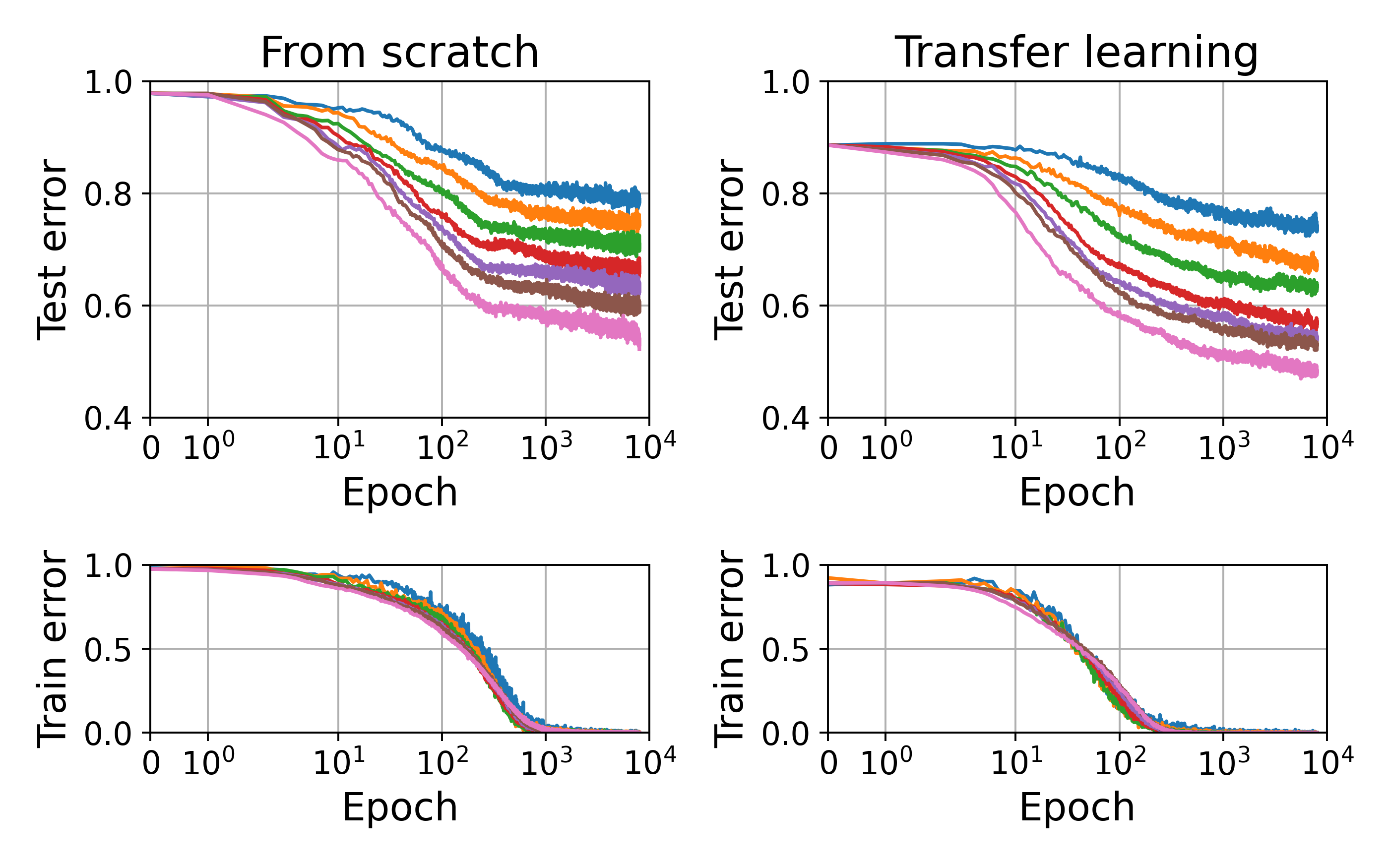}
   \label{fig:test_error_vs_epoch_for_several_dataset_sizes_densenet_BSP20_src20k_tinyimagenet32_src40SimilarClasstgtCIFAR40class_tgtNOISELESS}}
   \subfloat[20\% label noise in target dataset]{\includegraphics[width=0.49\textwidth]{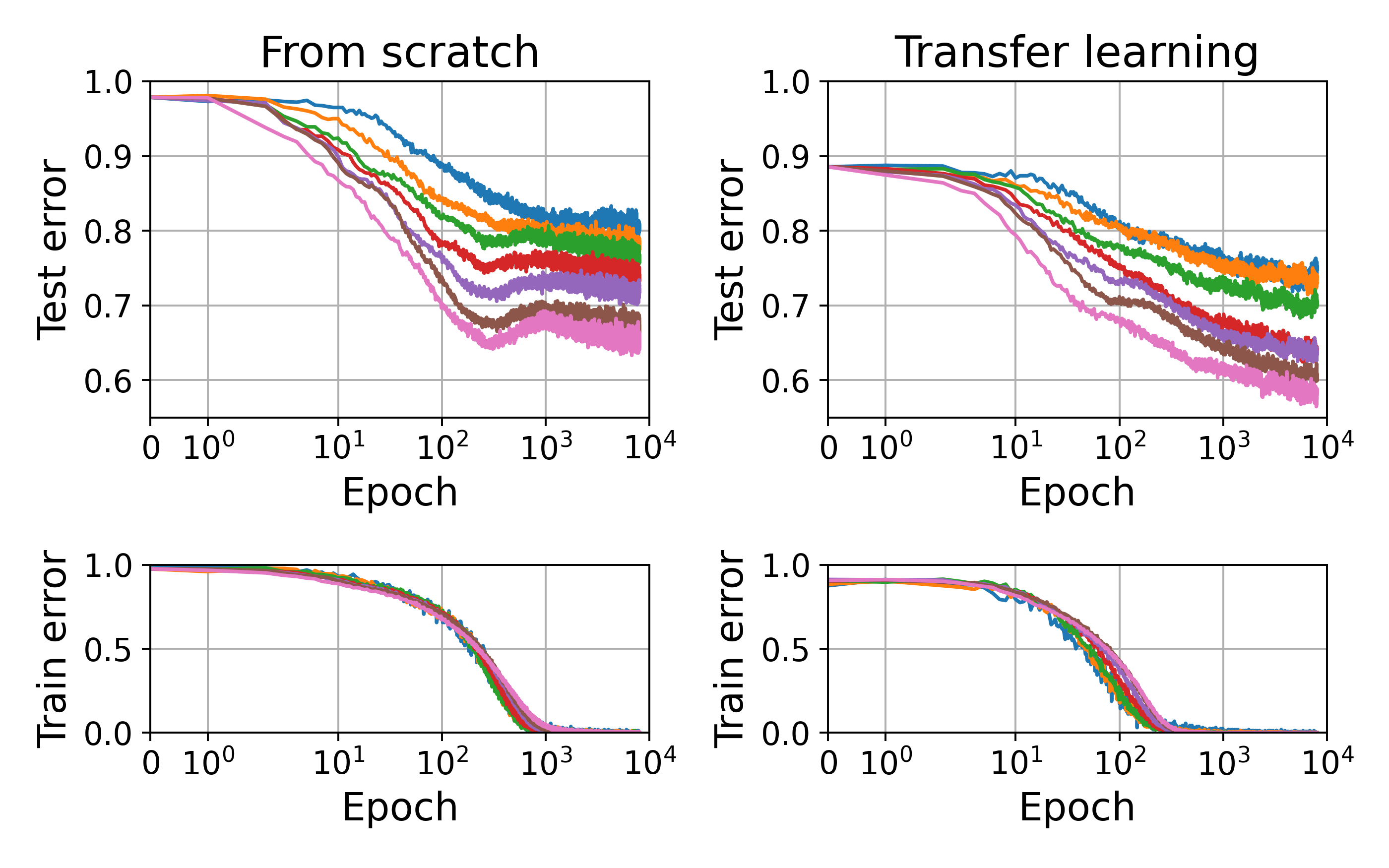}
\label{fig:test_error_vs_epoch_for_several_dataset_sizes_densenet_BSP20_src20k_tinyimagenet32_src40SimilarClasstgtCIFAR40class}}
   \caption{Evaluation of \textbf{DenseNet} training for a target classification task of 40 classes from CIFAR-100. The transfer learning is from the source task of 40 Tiny ImageNet classes that are similar to the target task classes (input image size 32x32x3) with source dataset of 20k training samples. Each curve color corresponds to another size of the target dataset. For better visibility, the value range of the vertical axis differs between the subfigures of the noiseless and noisy settings.}
   \label{fig:test_error_vs_epoch_for_several_dataset_sizes_densenet_BSP20_src20k_tinyimagenet32_src40SimilarClasstgtCIFAR40class}
\end{figure*}

\begin{figure*}[t]
  \centering
  \includegraphics[width=0.85\textwidth]{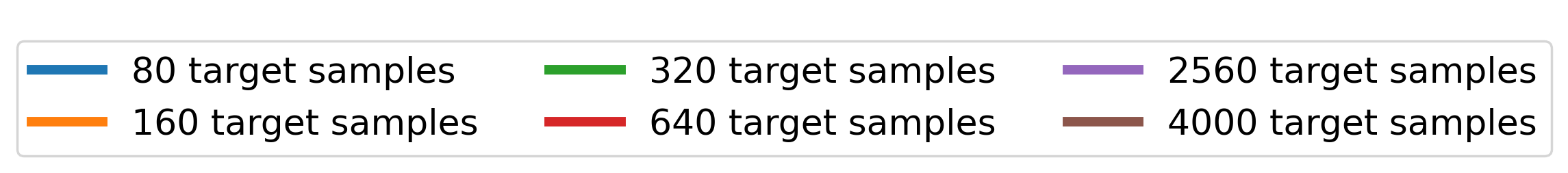}
  \\[-3ex]
  \subfloat[Noiseless]{
   \includegraphics[width=0.49\textwidth]{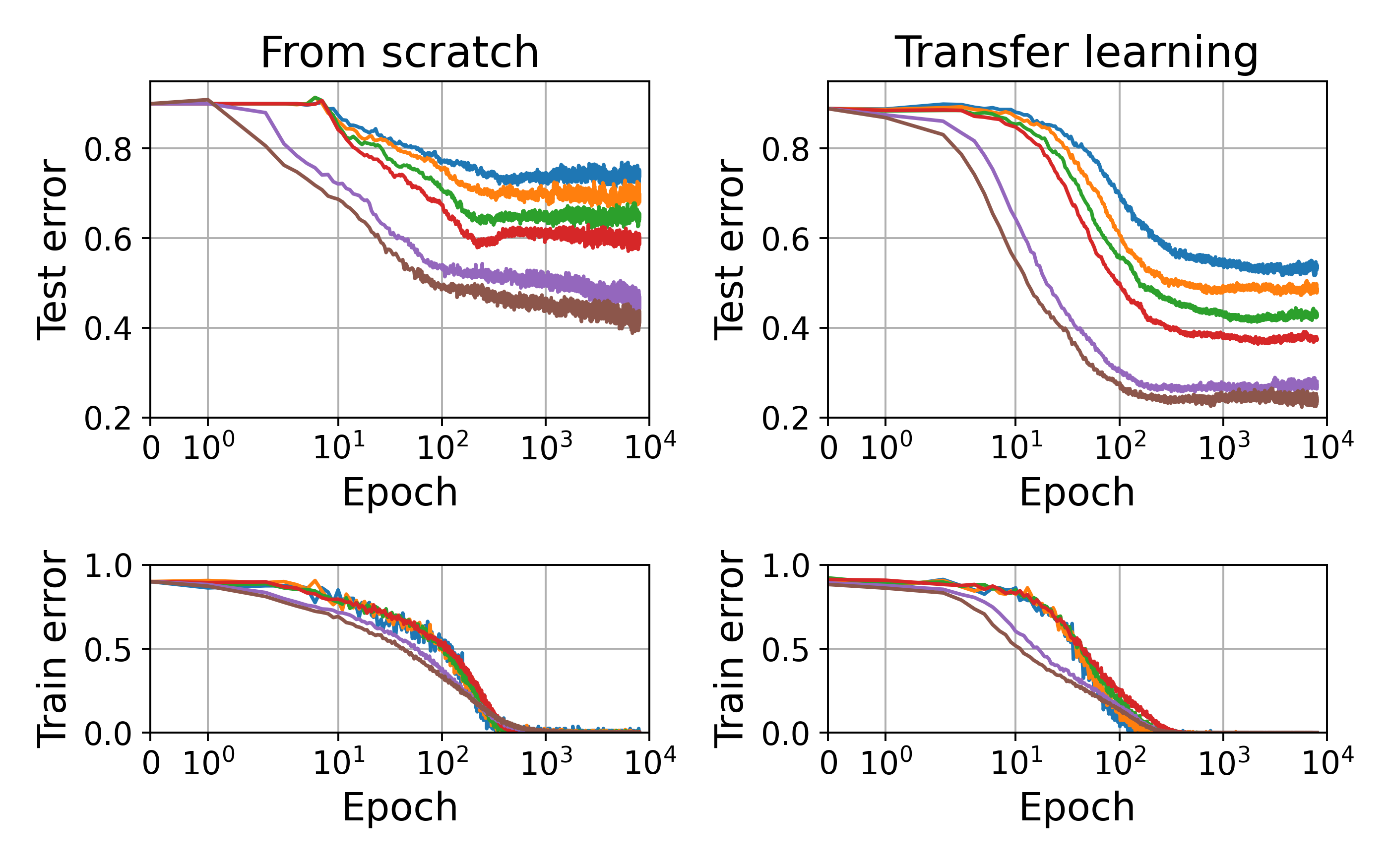}
   \label{fig:test_error_vs_epoch_for_several_dataset_sizes_densenet_BSP5_src100k_tinyimagenet64200C_tgtNoiselessCIFAR10}}
   \subfloat[20\% label noise in target dataset]{\includegraphics[width=0.49\textwidth]{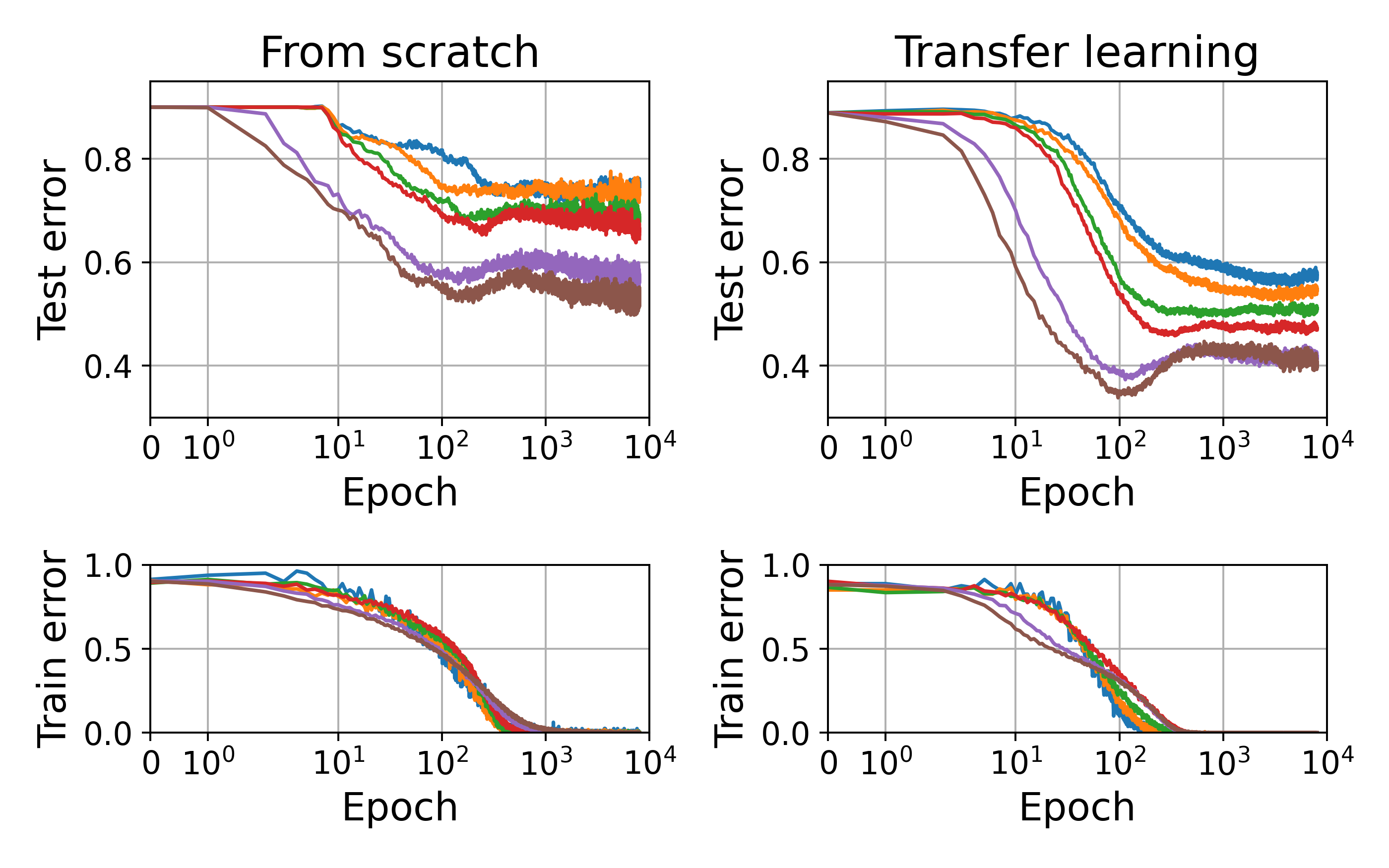}
\label{fig:test_error_vs_epoch_for_several_dataset_sizes_densenet_BSP5_src100k_tinyimagenet64200C_tgtNoisy0p2CIFAR10class}}
   \caption{Evaluation of \textbf{DenseNet} training for the CIFAR-10 target task. The transfer learning is from the source task of 200 Tiny ImageNet classes (input image size 64x64x3) with source dataset of 100k training samples. Each curve color corresponds to another size of the target dataset. For better visibility, the value range of the vertical axis differs between the subfigures of the noiseless and noisy settings.}
   \label{fig:test_error_vs_epoch_for_several_dataset_sizes_densenet_src100k_tinyimagenet64_200C_tgtNoisy0p2CIFAR10}
\end{figure*}

\begin{figure*}[t]
  \centering
  \includegraphics[width=0.85\textwidth]{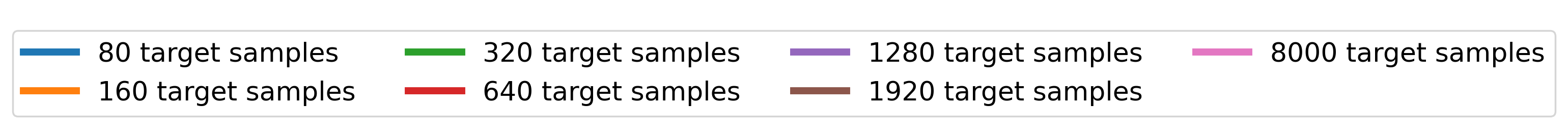}
  \\[-3ex]
  \subfloat[Noiseless]{
   \includegraphics[width=0.49\textwidth]{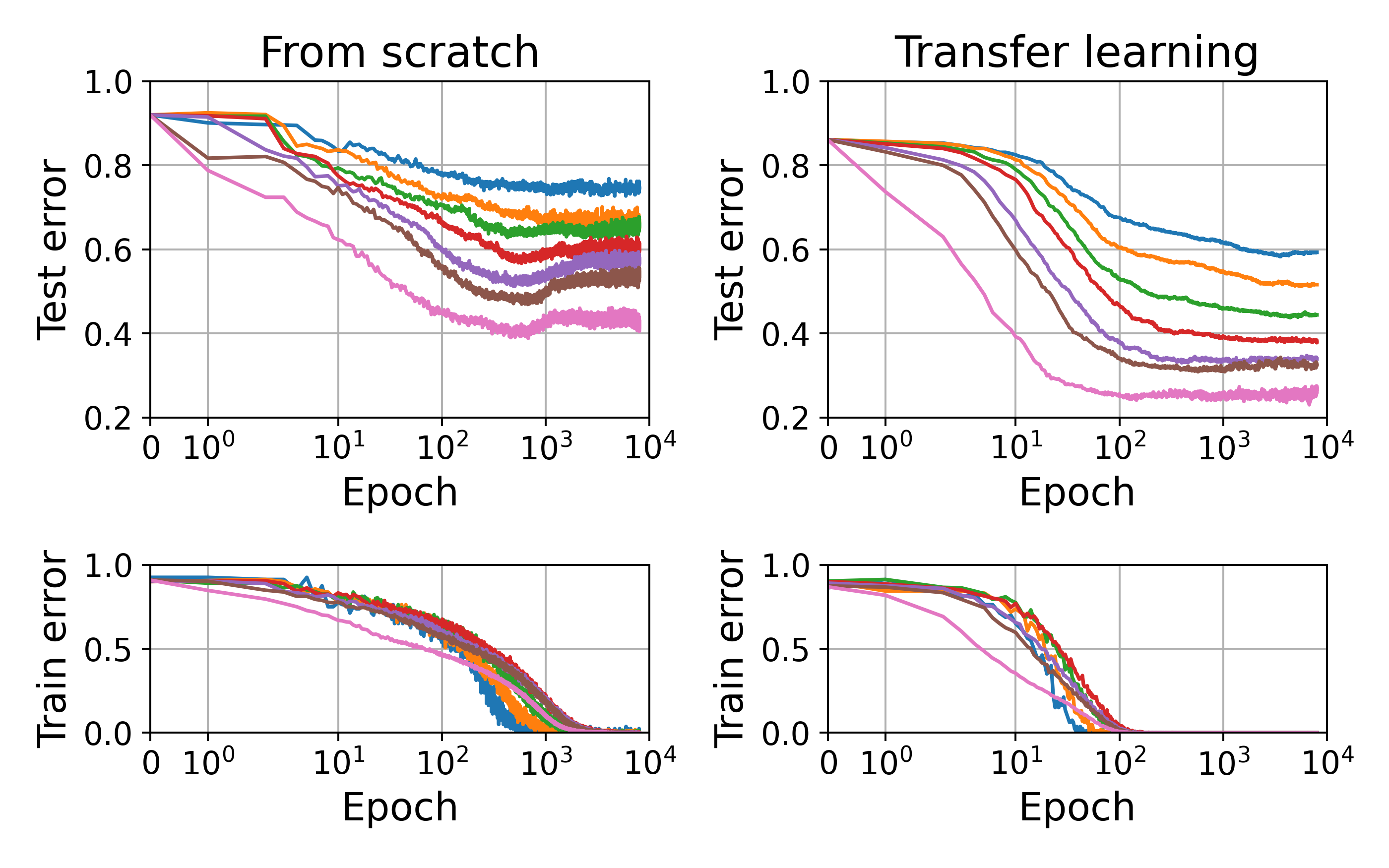}
   \label{fig:test_error_vs_epoch_for_several_dataset_sizes_vit_BSP5_src100k_tinyimagenet32_200C_tgtNoiselessCIFAR10class}}
   \subfloat[20\% label noise in target dataset]{\includegraphics[width=0.49\textwidth]{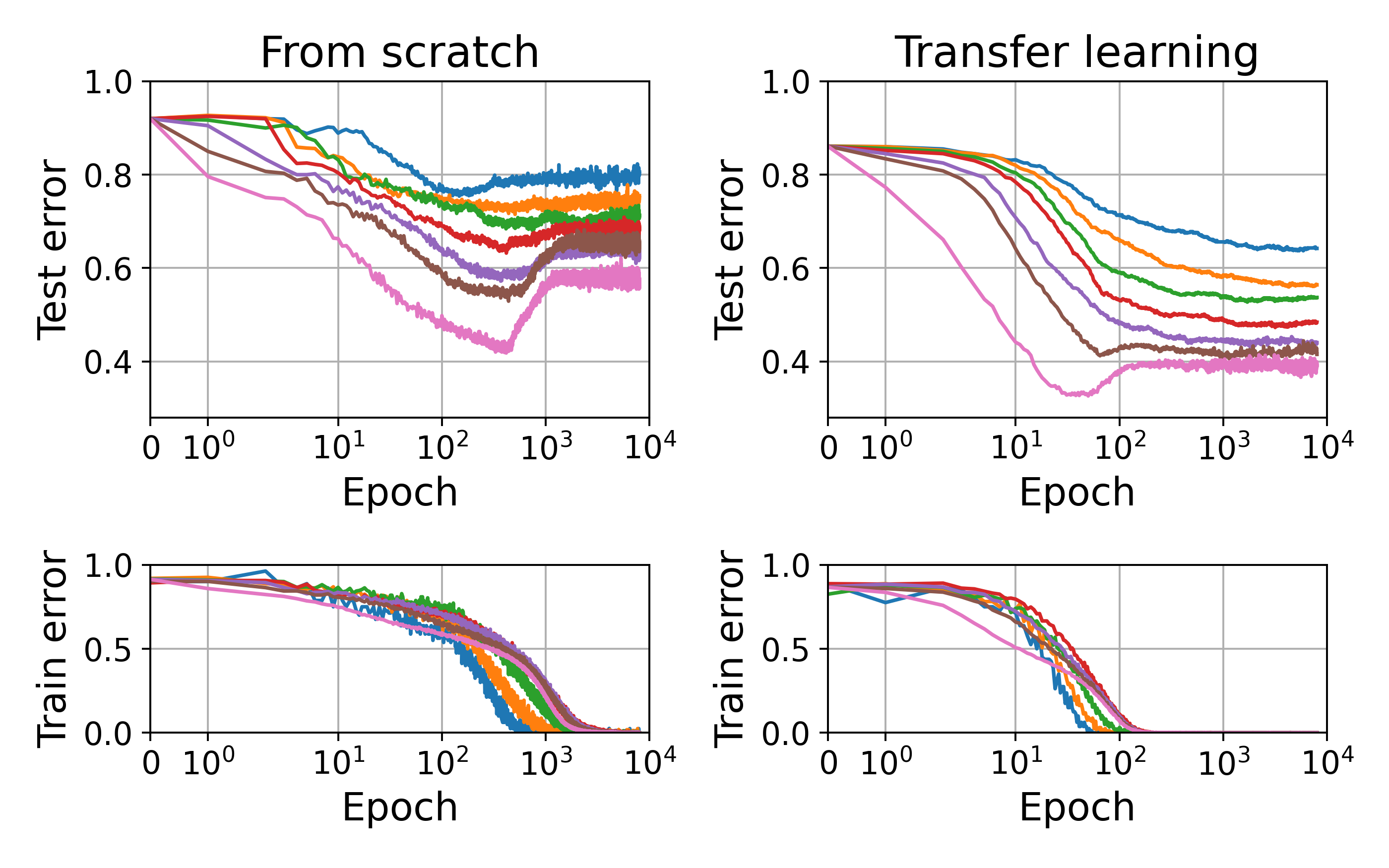}
\label{fig:test_error_vs_epoch_for_several_dataset_sizes_vit_BSP5_src100k_tinyimagenet32_200C_tgtNoisy0p2CIFAR10class}}
   \caption{Evaluation of \textbf{ViT} training for the CIFAR-10 target task. The transfer learning is from the source task of 200 Tiny ImageNet classes (input image size 32x32x3) with source dataset of 100k training samples. Each curve color corresponds to another size of the target dataset. For better visibility, the value range of the vertical axis differs between the subfigures of the noiseless and noisy settings.}
   \label{fig:test_error_vs_epoch_for_several_dataset_sizes_vit_src100k_tinyimagenet32_200C_tgtNoisy0p2CIFAR10}
\end{figure*}

 In Section \ref{subsec:effect of target dataset size} we have examined the evolution of the error curves in the learning from scratch and transfer learning for a variety of \textit{target} dataset sizes.
In addition to the results in Figures 1a and 2 for ResNet, Figures  \ref{fig:test_error_vs_epoch_for_several_dataset_sizes_vit_BSP20_src20k_tinyimagenet32_src40SimilarClasstgtCIFAR40class}, \ref{fig:test_error_vs_epoch_for_several_dataset_sizes_densenet_BSP20_src20k_tinyimagenet32_src40SimilarClasstgtCIFAR40class} show that a significant double descent emerges as the target dataset is sufficiently large (and with label noise) also for ViT and DenseNet in 40-class settings.

In the main text, Fig.~2 shows the results for learning a ResNet-18 for CIFAR-10 as the target task and transfer from the entire Tiny ImageNet (image size 64x64x3) as the source task . Here, Fig.~\ref{fig:test_error_vs_epoch_for_several_dataset_sizes_densenet_src100k_tinyimagenet64_200C_tgtNoisy0p2CIFAR10} shows the corresponding results for the same source and target tasks but for the DenseNet architecture.  Fig.~\ref{fig:test_error_vs_epoch_for_several_dataset_sizes_vit_src100k_tinyimagenet32_200C_tgtNoisy0p2CIFAR10} shows the results for the ViT (for the ViT the image size of the source task is 32x32x3). These additional results further support our finding that double descent, or a related behavior where the test error increases at the interpolation threshold but not necessarily decreases afterwards (e.g., due to insufficient overparameterization w.r.t.~the training dataset size), are more likely to emerge in transfer learning as the target dataset is larger and noisy.

\begin{figure*}[t]
  \centering
  \includegraphics[width=0.85\textwidth]{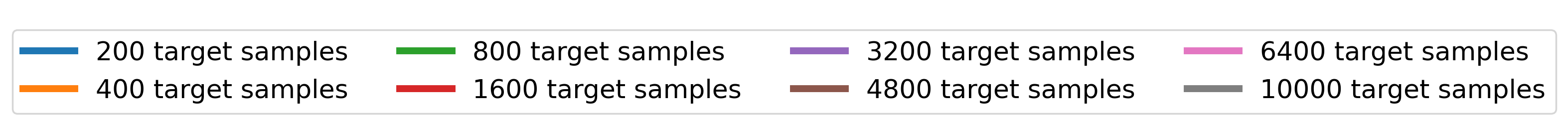}
  \\[-3ex]
  \subfloat[``Noiseless'' (no artificially added label noise)]{
   \includegraphics[width=0.49\textwidth]{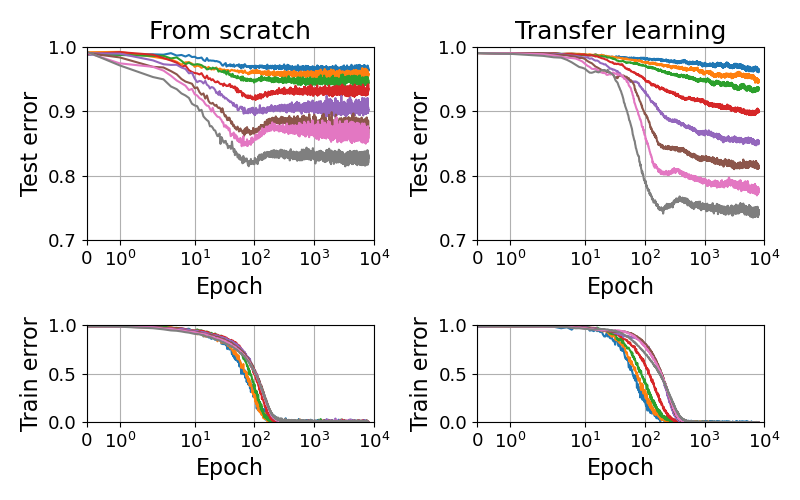}
   \label{fig:test_error_vs_epoch_for_several_dataset_sizes_100classImagenet100x32ToFood100_srcNoiselessTgtNoiseless}}
   \subfloat[20\% (artificially added) label noise in target dataset]{\includegraphics[width=0.49\textwidth]{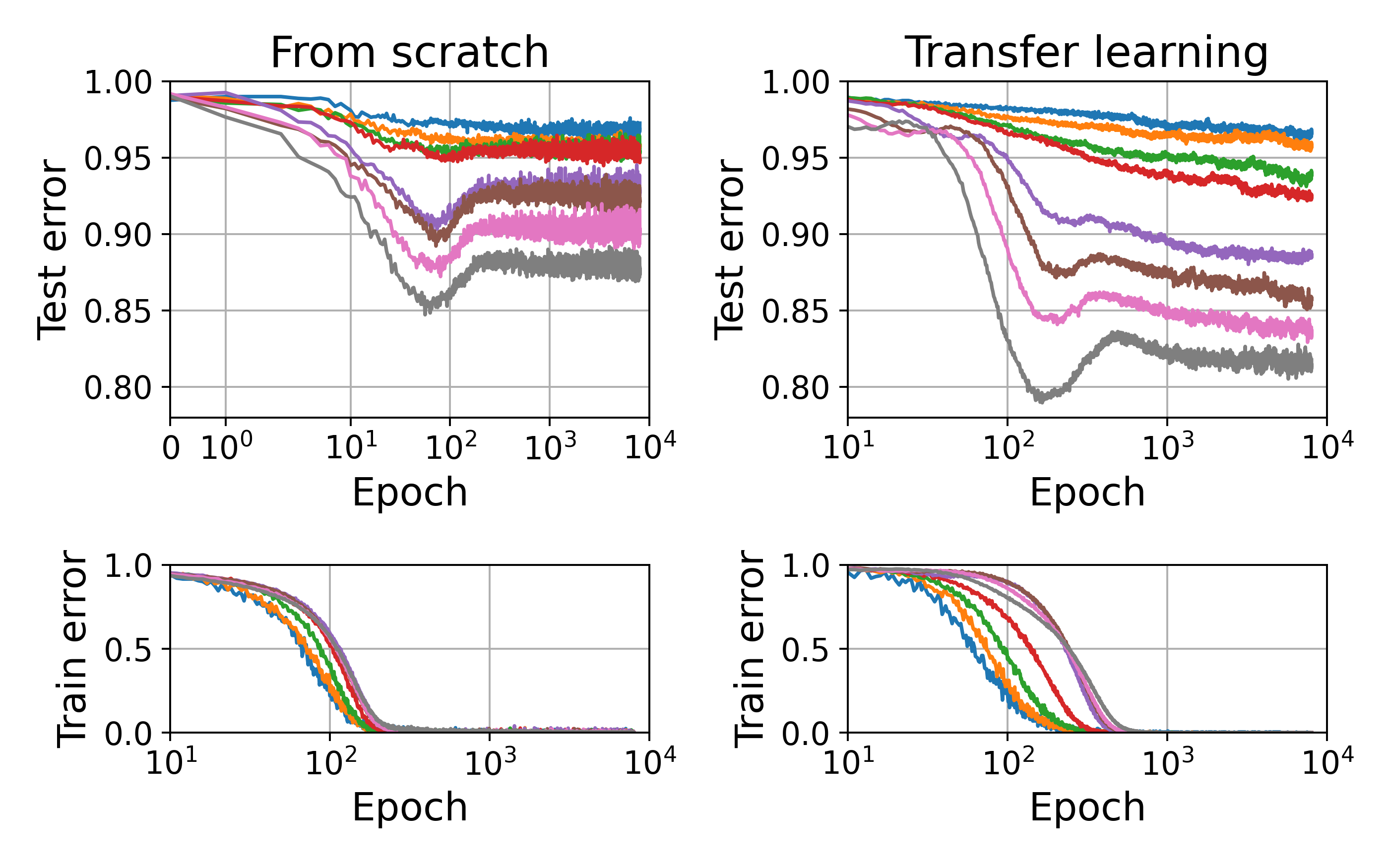}
\label{fig:test_error_vs_epoch_for_several_dataset_sizes_100classImagenet100x32ToFood100_srcNoiselessTgtNoisy0p2}}
   \caption{Evaluation of \textbf{ResNet} training for a target classification task of \textbf{100 classes from Food-101}. The transfer learning is from the source task of 100 Tiny ImageNet classes (input image size 32x32x3) with source dataset of 50k training samples. Each curve color corresponds to another size of the target dataset. }
   \label{fig:test_error_vs_epoch_for_several_dataset_sizes_resnet_src50k_tinyimagenet32_100C_tgtFood100}
\end{figure*}

Figure \ref{fig:test_error_vs_epoch_for_several_dataset_sizes_resnet_src50k_tinyimagenet32_100C_tgtFood100} examines a target task based on the Food-101 dataset and again shows for ResNet that double descent emerges for larger target datasets. The Food-101 dataset is known to have some label noise \cite{bossard2014food} and, therefore, in Fig.~\ref{fig:test_error_vs_epoch_for_several_dataset_sizes_resnet_src50k_tinyimagenet32_100C_tgtFood100} we observe double descent behavior also in the ``noiseless'' (i.e., without artificially added label noise) setting.

\ydarnew{maybe transfer learning of ViT with some of the first layers (at least the patch embedding) would behave nicer with double descent curves (the larger embedding dimension did not resolve the problem!); check the results of the experiment vit-tgtCIFAR10-srcImagenet100k-srcNoiselessTgtNoisy0p2-frozen3-BSP5-tgtarray2-y}

\subsection{Additional Results for Section 3.2}
\label{appendix:subsec:Additional Experimental Results for Section 3.2}

\begin{figure*}[t]
  \centering
  \includegraphics[width=0.85\textwidth]{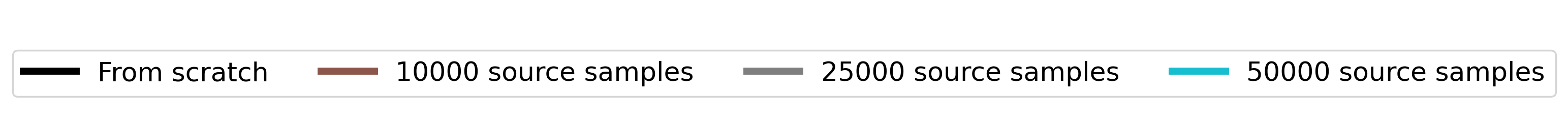}
  \\[-3ex]
  \subfloat[]{
   \includegraphics[width=0.245\textwidth]{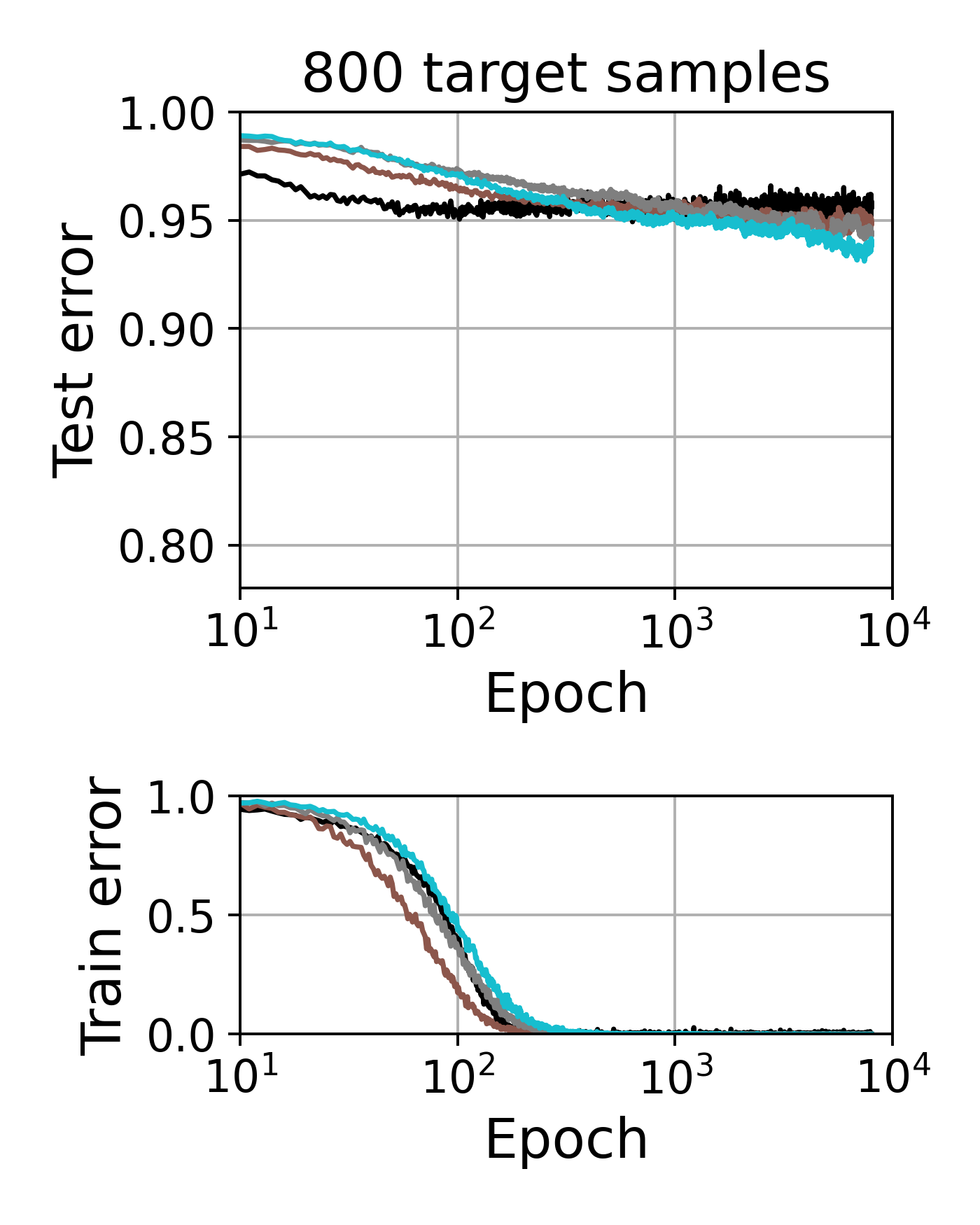}
   \label{fig:src_different_dataset_size_100Classes_srcImagenetF100classtgtFood100all_srcNoiselessTgtNoisy0p2_dataset800_w}}
   \subfloat[]{
   \includegraphics[width=0.245\textwidth]{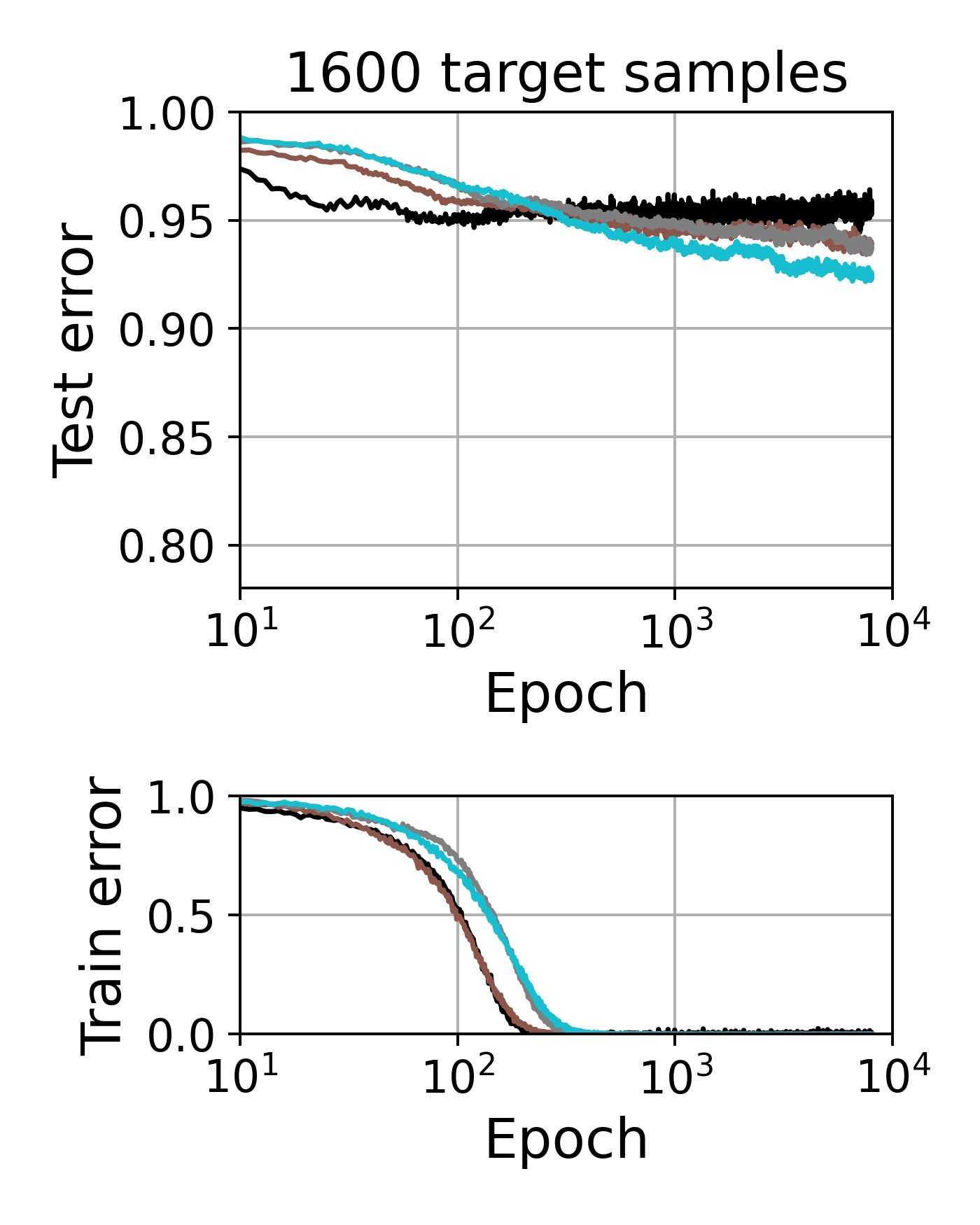}
   \label{fig:src_different_dataset_size_100Classes_srcImagenetF100classtgtFood100all_srcNoiselessTgtNoisy0p2_dataset1600_w}}
      \subfloat[]{
   \includegraphics[width=0.245\textwidth]{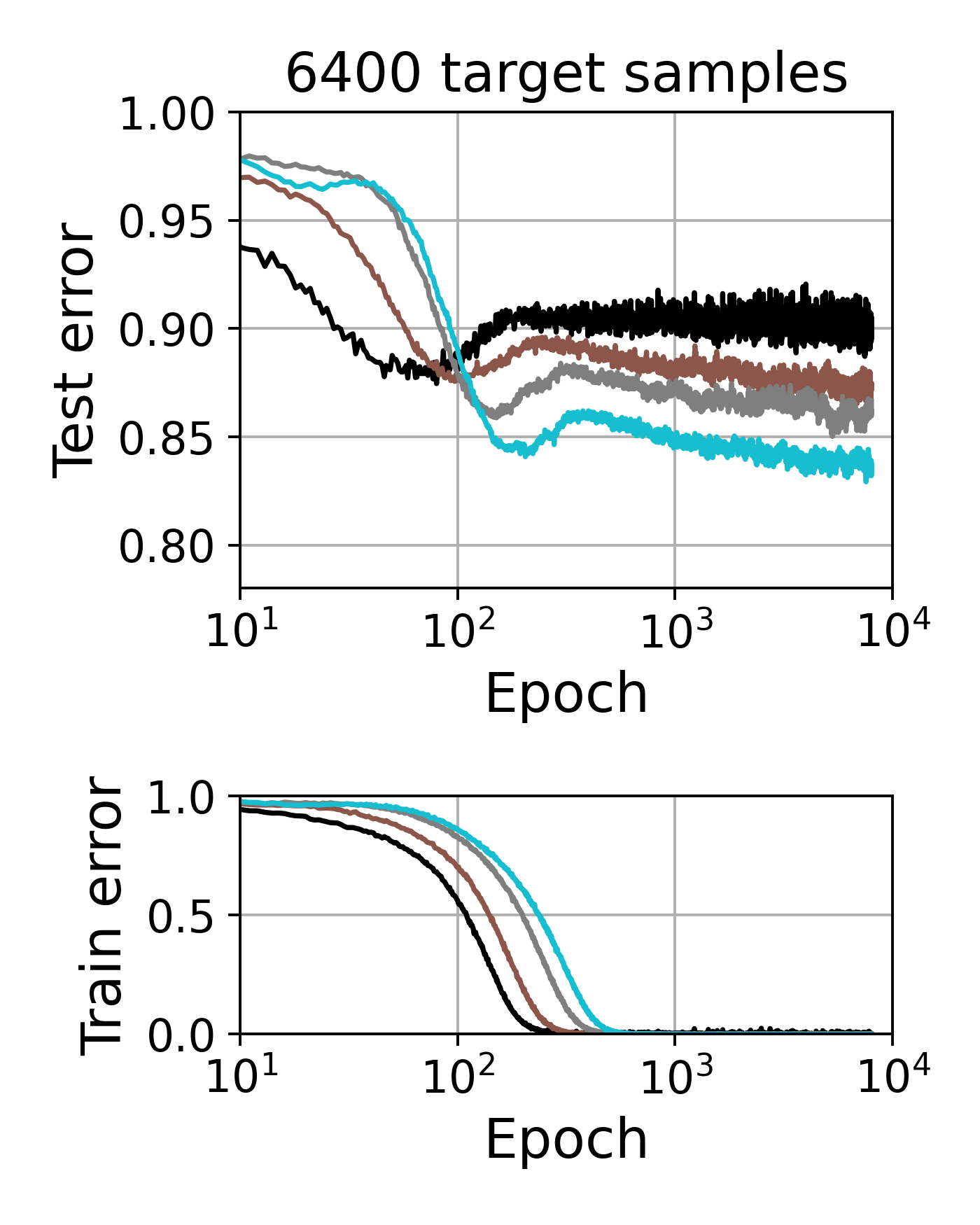}
   \label{fig:src_different_dataset_size_100Classes_srcImagenetF100classtgtFood100all_srcNoiselessTgtNoisy0p2_dataset6400_w}}
         \subfloat[]{
   \includegraphics[width=0.245\textwidth]{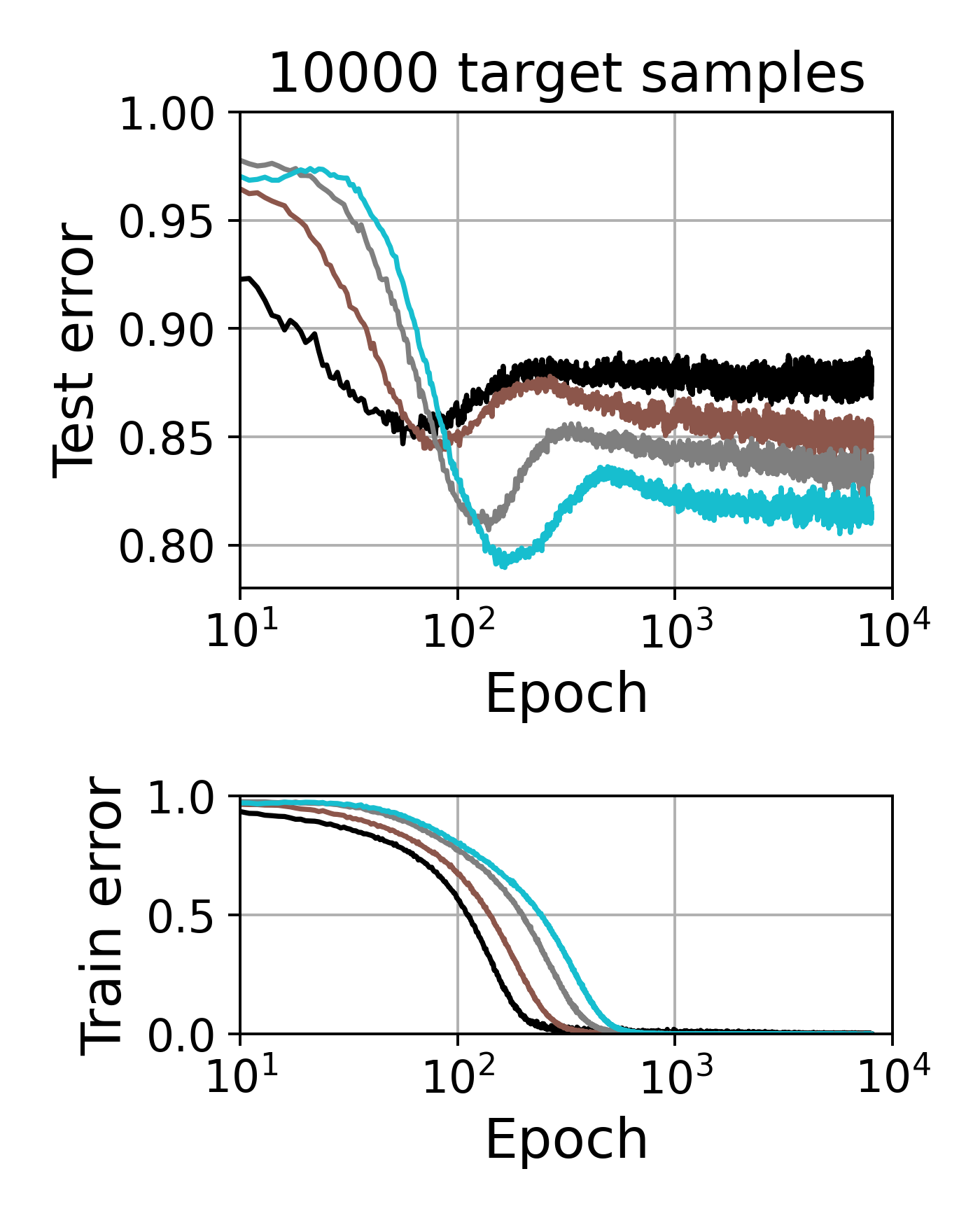}
   \label{fig:src_different_dataset_size_100Classes_srcImagenetF100classtgtFood100all_srcNoiselessTgtNoisy0p2_dataset10000_w}}
   \caption{The effect of the source dataset size on the target model training. Evaluation of \textbf{ResNet} training for a target classification task of \textbf{100 classes from Food-101} with 20\% label noise in the target dataset. The transfer learning is from the source task of 100 Tiny ImageNet classes (input image size 32x32x3). Each curve color corresponds to another size of the source dataset. }
   \label{fig:src_different_dataset_size_100Classes_srcImagenetF100classtgtFood100all_srcNoiselessTgtNoisy0p2}
\end{figure*}

\begin{figure*}[t]
  \centering
  \includegraphics[width=0.85\textwidth]{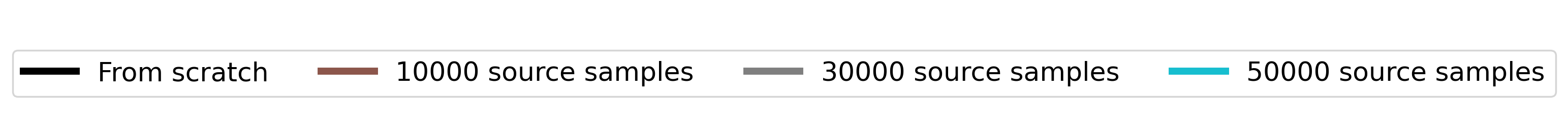}
  \\[-3ex]
  \subfloat[]{
   \includegraphics[width=0.245\textwidth]{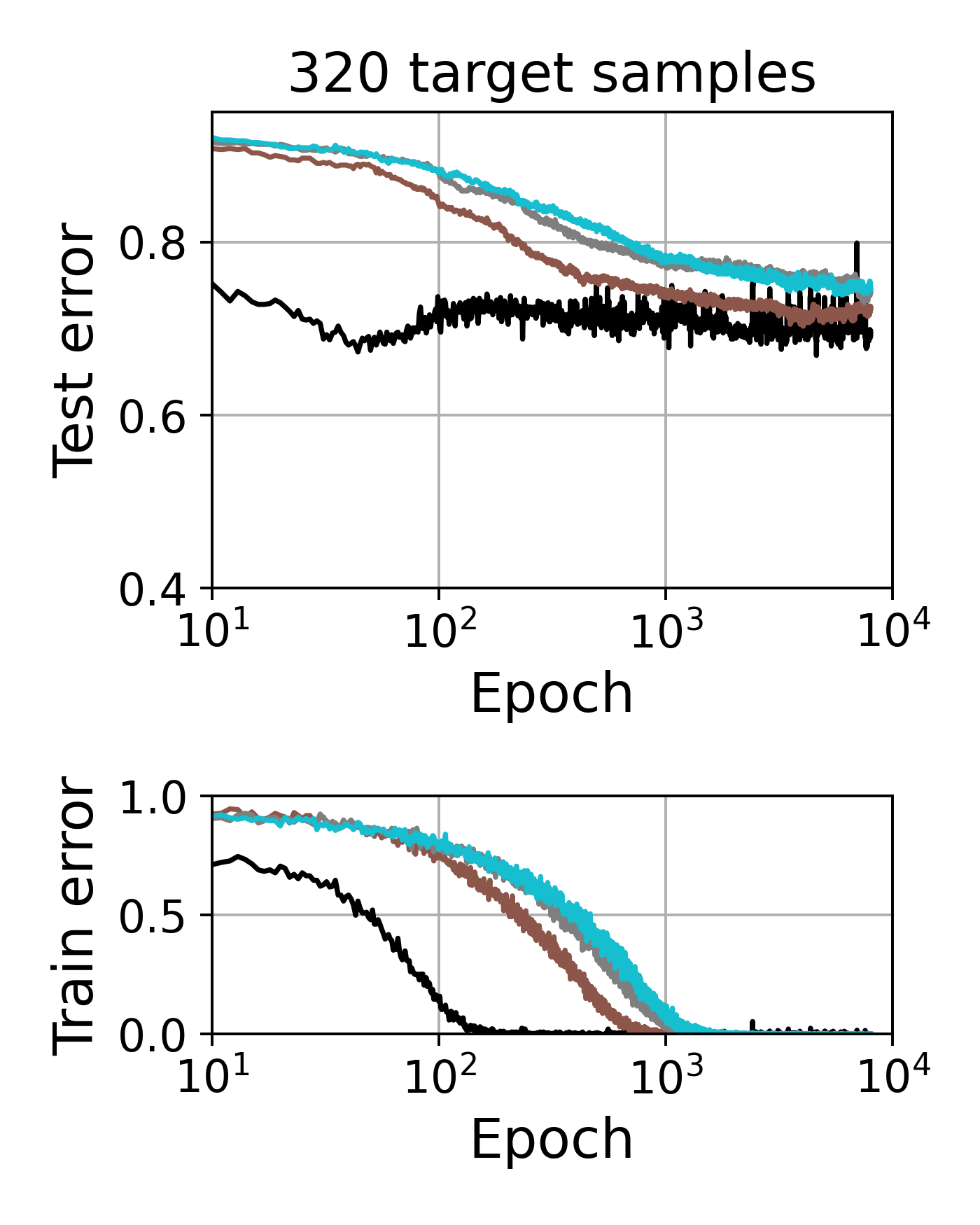}
   \label{fig:src_different_dataset_size_10Classes_srcCIFAR10classtgtFoodF10_srcNoiselessTgtNoisy0p2_dataset320_w}}
   \subfloat[]{
   \includegraphics[width=0.245\textwidth]{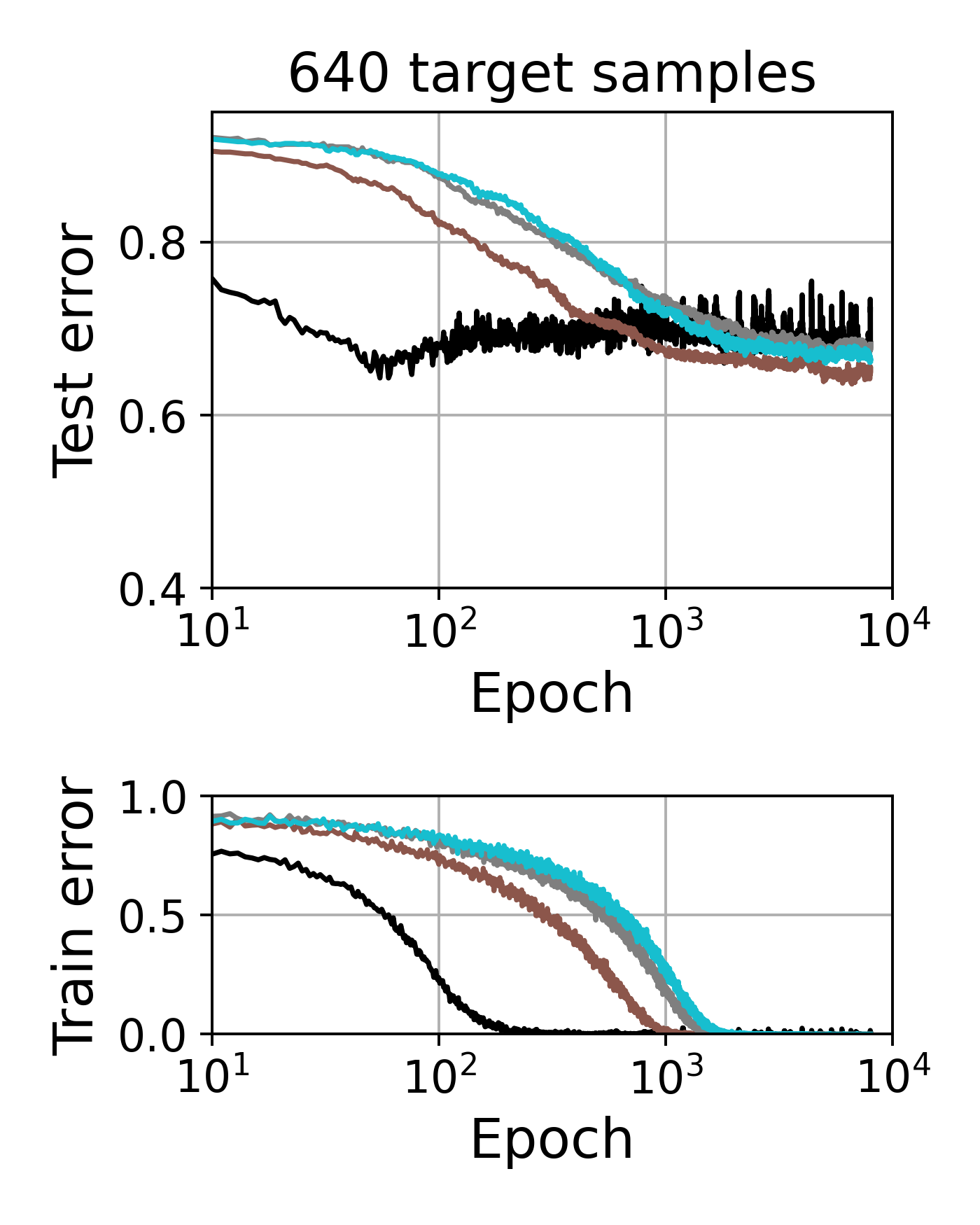}
   \label{fig:src_different_dataset_size_10Classes_srcCIFAR10classtgtFoodF10_srcNoiselessTgtNoisy0p2_dataset640_w}}
      \subfloat[]{
   \includegraphics[width=0.245\textwidth]{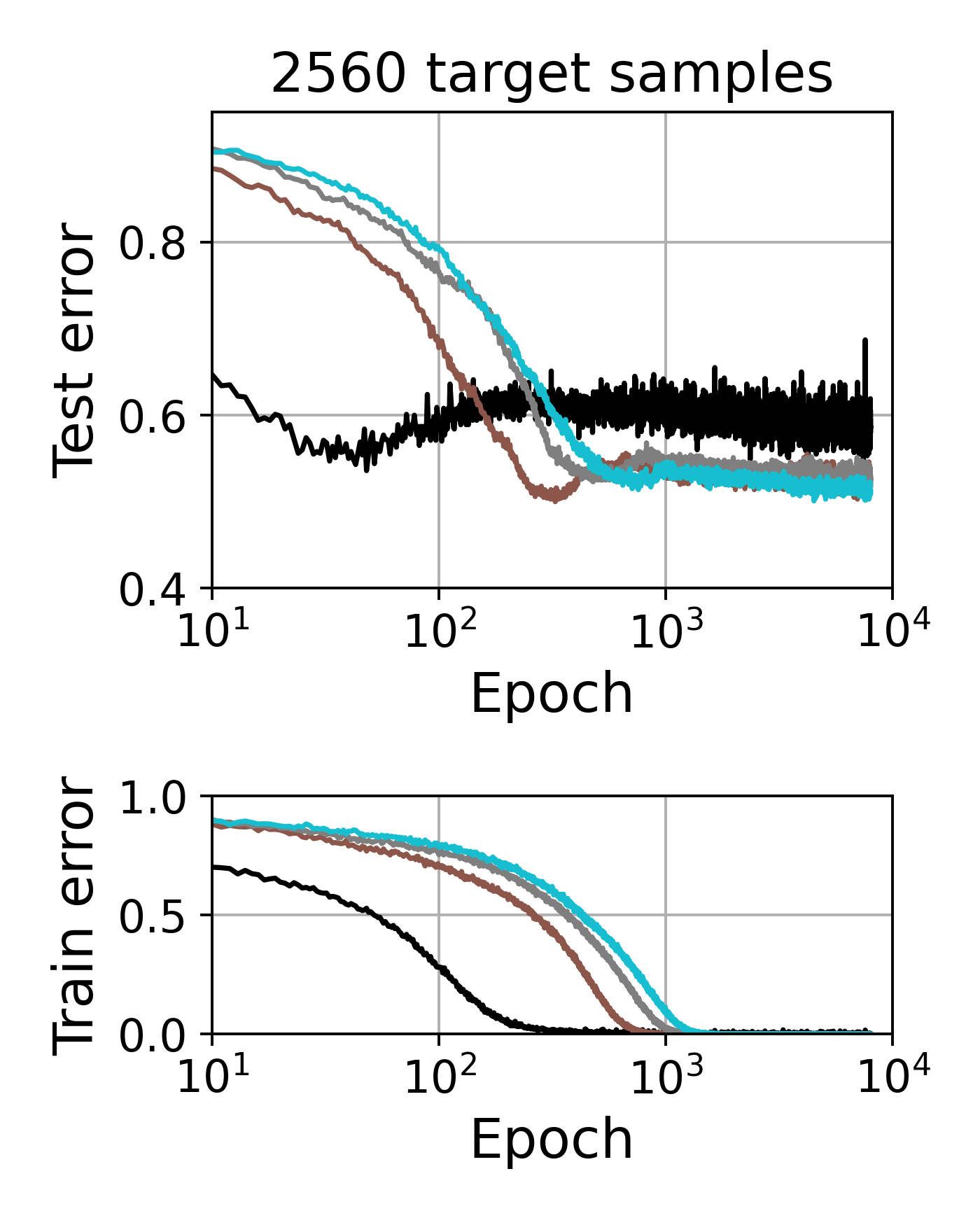}
   \label{fig:src_different_dataset_size_10Classes_srcCIFAR10classtgtFoodF10_srcNoiselessTgtNoisy0p2_dataset2560_w}}
         \subfloat[]{
   \includegraphics[width=0.245\textwidth]{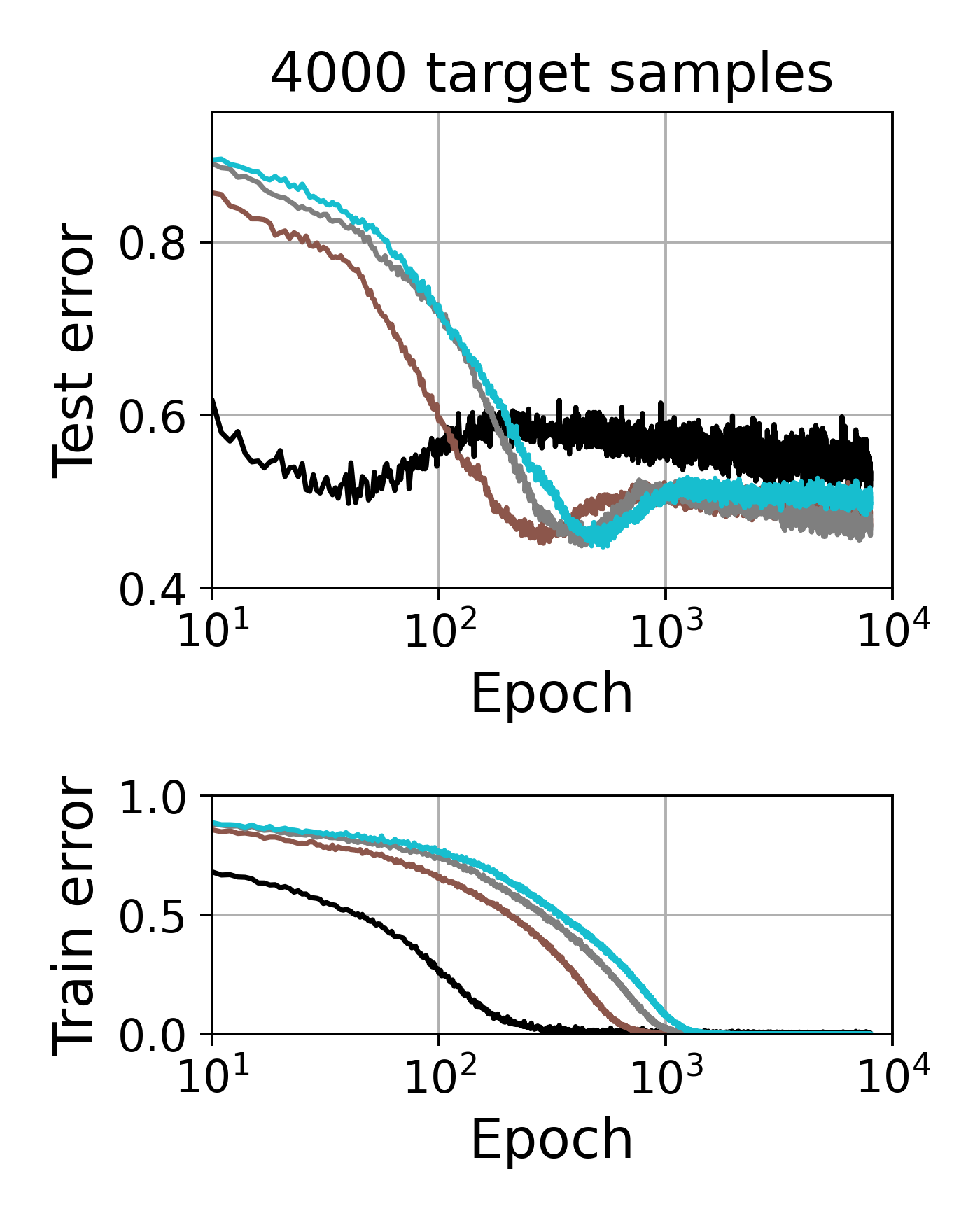}
   \label{fig:src_different_dataset_size_10Classes_srcCIFAR10classtgtFoodF10_srcNoiselessTgtNoisy0p2_dataset4000_w}}
   \caption{The effect of the source dataset size on the target model training. Evaluation of \textbf{ResNet} training for a target classification task of \textbf{10 classes from Food-101} (input image size 32x32x3) with 20\% label noise in the target dataset. The transfer learning is from the source task of CIFAR-10. }
   \label{fig:src_different_dataset_size_10Classes_srcCIFAR10classtgtFoodF10_srcNoiselessTgtNoisy0p2}
\end{figure*}

\begin{figure*}[t]
  \centering
  \includegraphics[width=0.85\textwidth]{figures/legend_src_different_dataset_size_100Classes_srcImagenetF100classtgtCIFAR100all_srcNoiselessTgtNoisy0p2_wScratch.png}
  \\[-1ex]
\subfloat[ViT]{\label{appendix:fig:src_different_dataset_size_100Classes_srcImagenetF100classtgtCIFAR100all_srcNoiselessTgtNoisy0p2_vit}
   \includegraphics[width=0.245\textwidth]{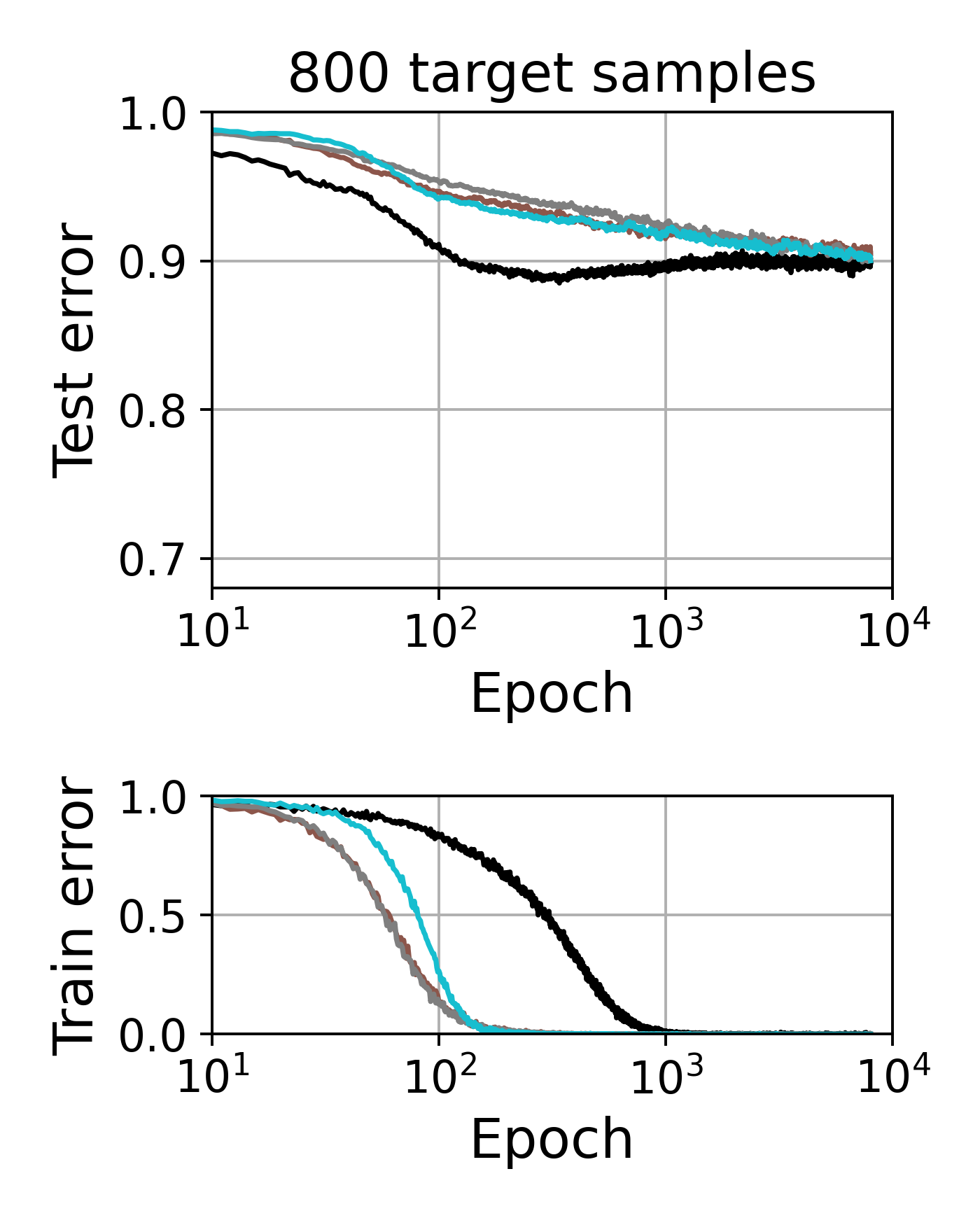}
      \includegraphics[width=0.245\textwidth]{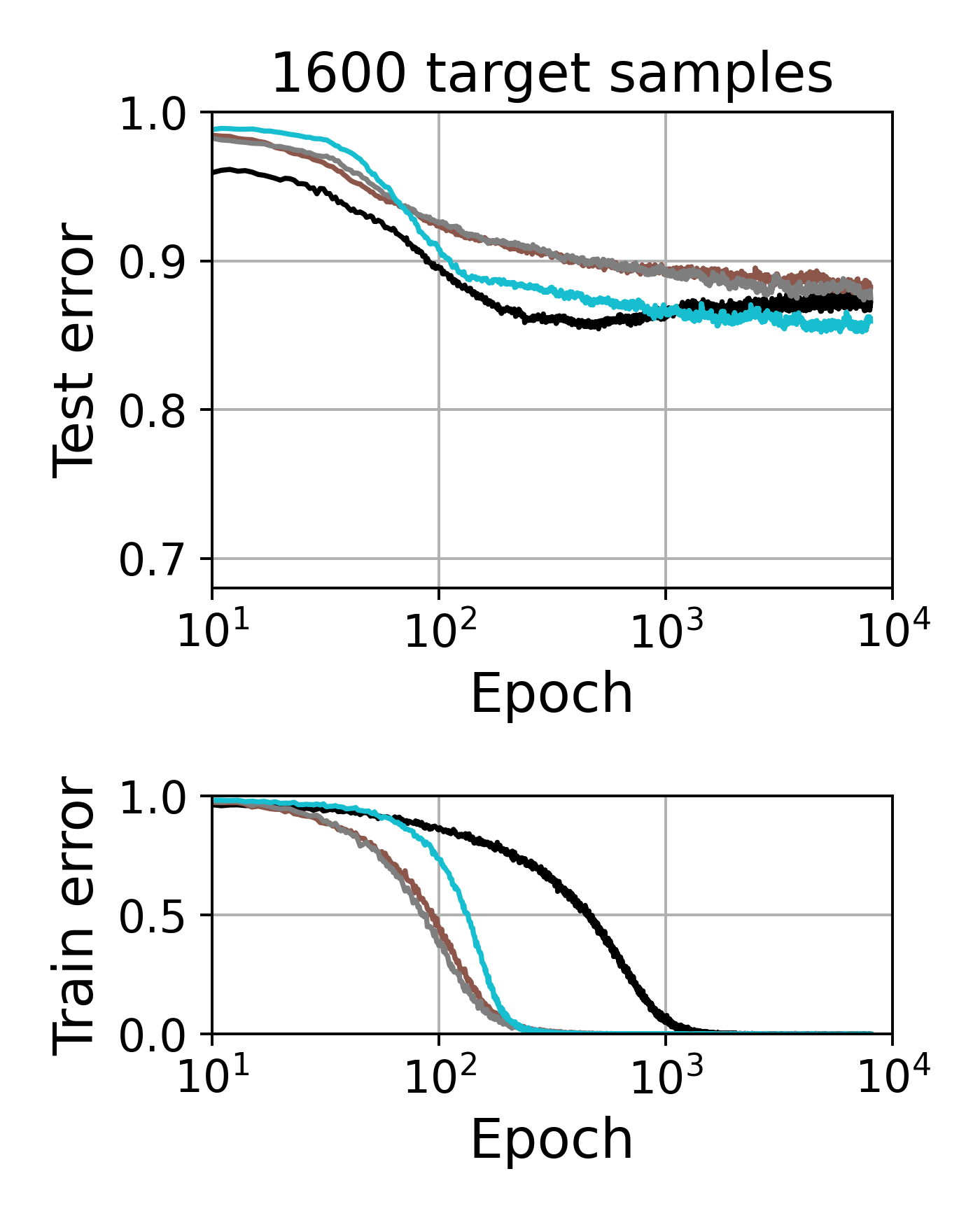}
   \includegraphics[width=0.245\textwidth]{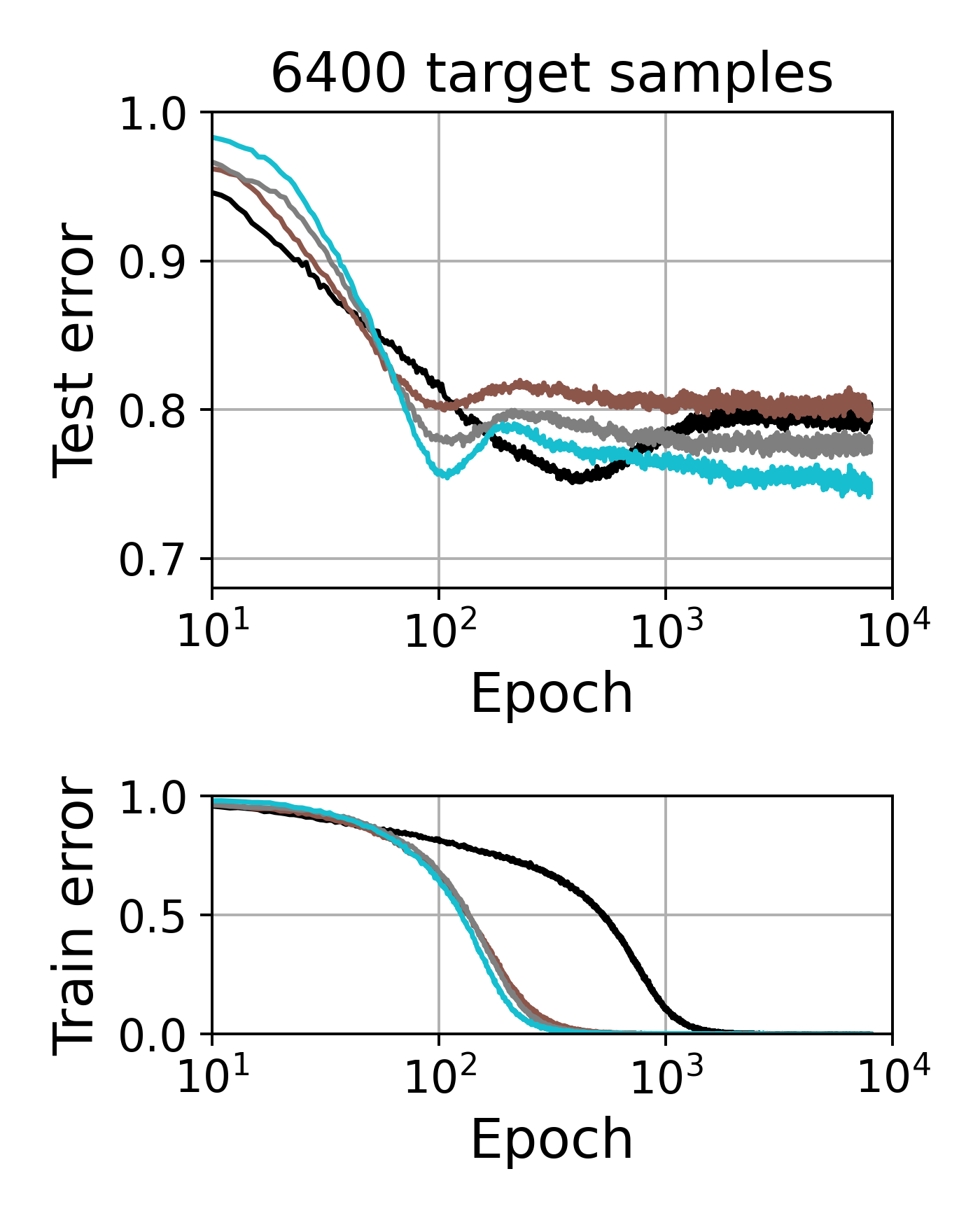}
   \includegraphics[width=0.245\textwidth]{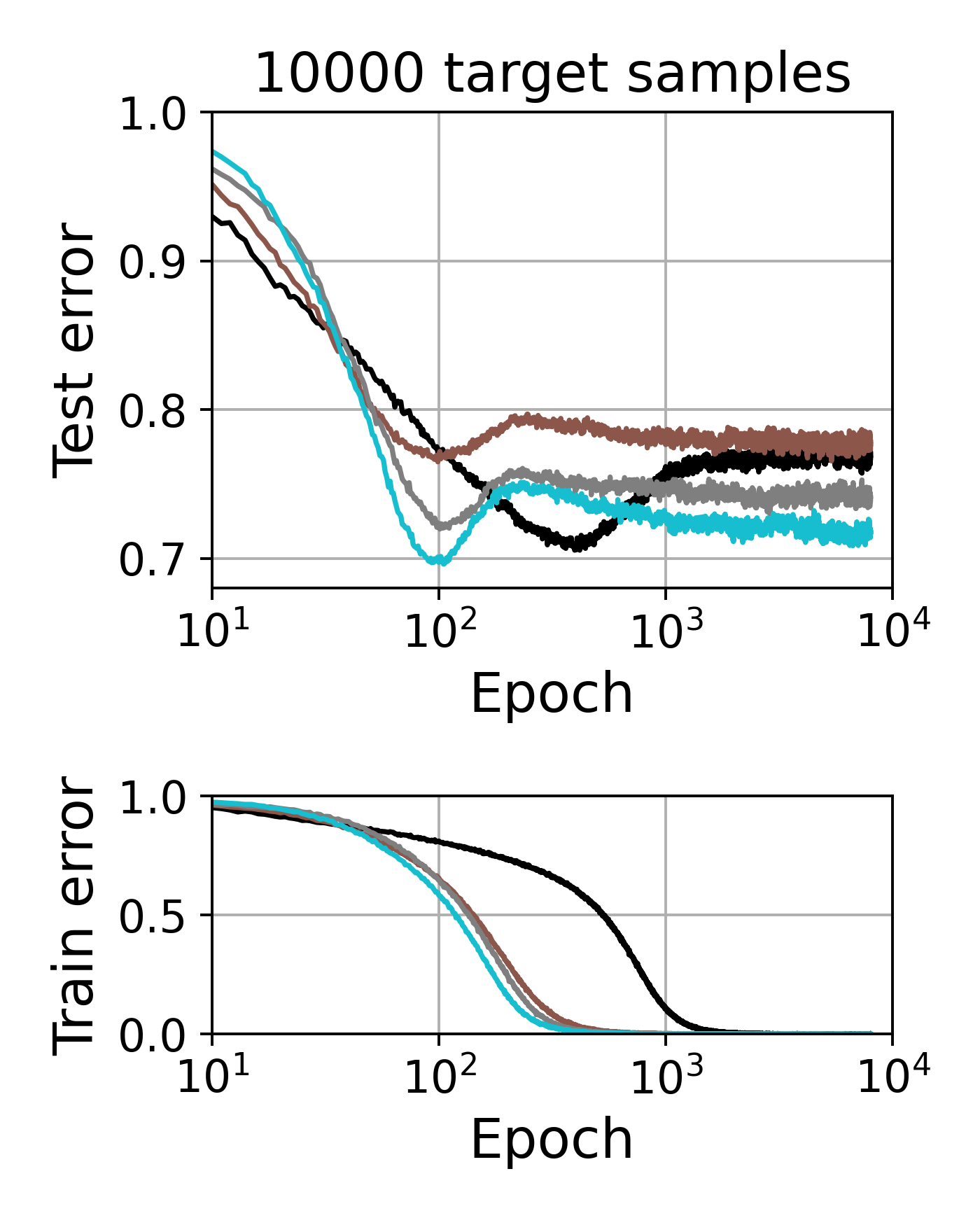}}
   \caption{The effect of the source dataset size on the target model training. Evaluation of \textbf{ViT} training for a target classification task of \textbf{CIFAR-100} (all 100 classes)  with 20\% label noise in the target dataset. The transfer learning is from the source task of 100 Tiny ImageNet classes (input image size 32x32x3). Each curve color corresponds to another size of the source dataset. Recall the different behavior of the ResNet and DenseNet as observed in Fig.~\ref{fig:src_different_dataset_size_100Classes_srcImagenetF100classtgtCIFAR100all_srcNoiselessTgtNoisy0p2}.}
   \label{fig:src_different_dataset_size_100Classes_srcImagenetF100classtgtCIFAR100all_srcNoiselessTgtNoisy0p2_ViT_appendix}
\end{figure*}

\begin{figure*}[t]
  \centering
  \includegraphics[width=0.85\textwidth]{figures/legend_src_different_dataset_size_100Classes_srcImagenetF100classtgtCIFAR100all_srcNoiselessTgtNoisy0p2_wScratch.png}
  \\[-1ex]
  \subfloat[ResNet (width parameter 18)]{\label{appendix:fig:src_different_dataset_size_100Classes_srcImagenetF100classtgtCIFAR100all_srcNoiselessTgtNoisy0p2_resnetW18}
   \includegraphics[width=0.245\textwidth]{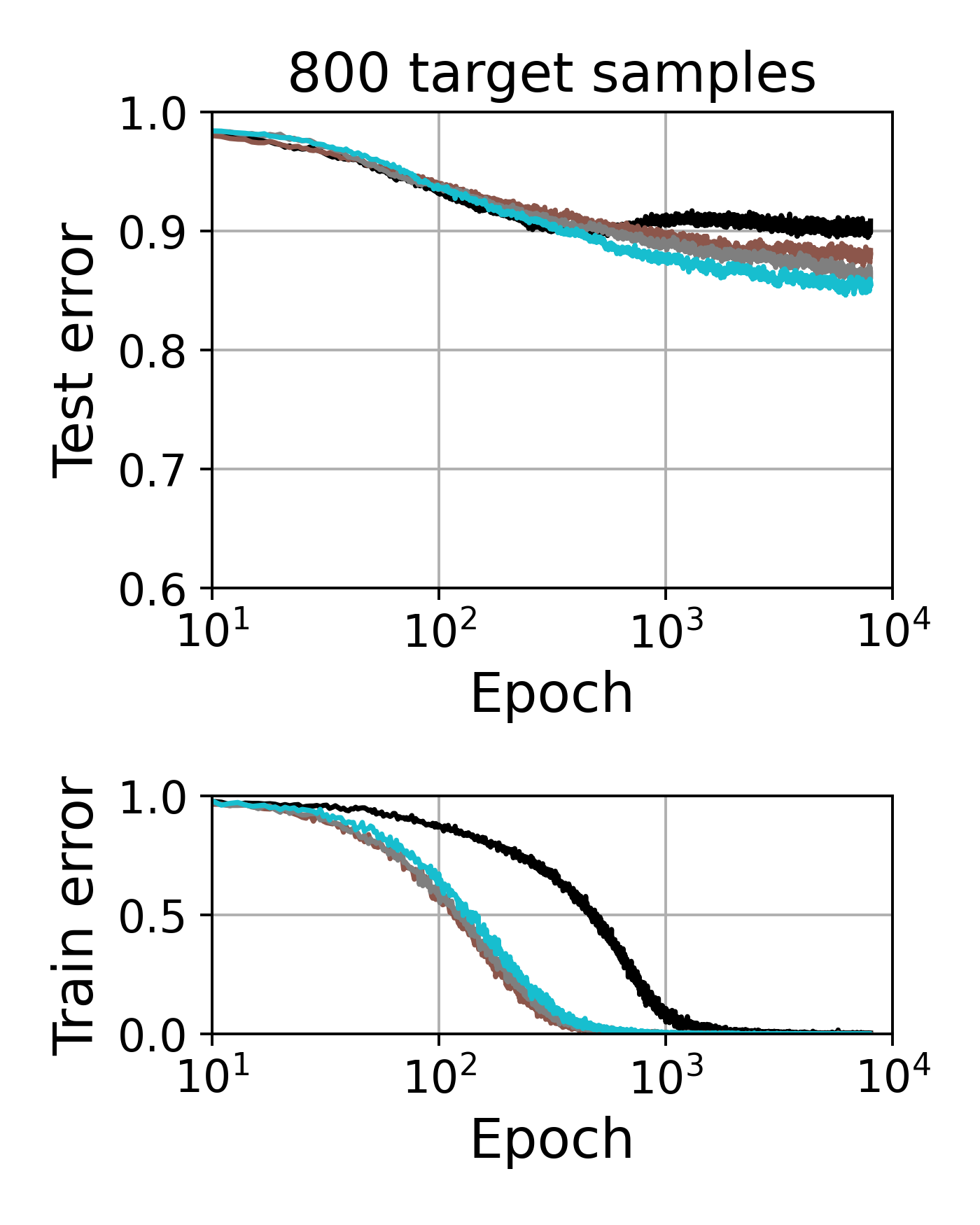}
      \includegraphics[width=0.245\textwidth]{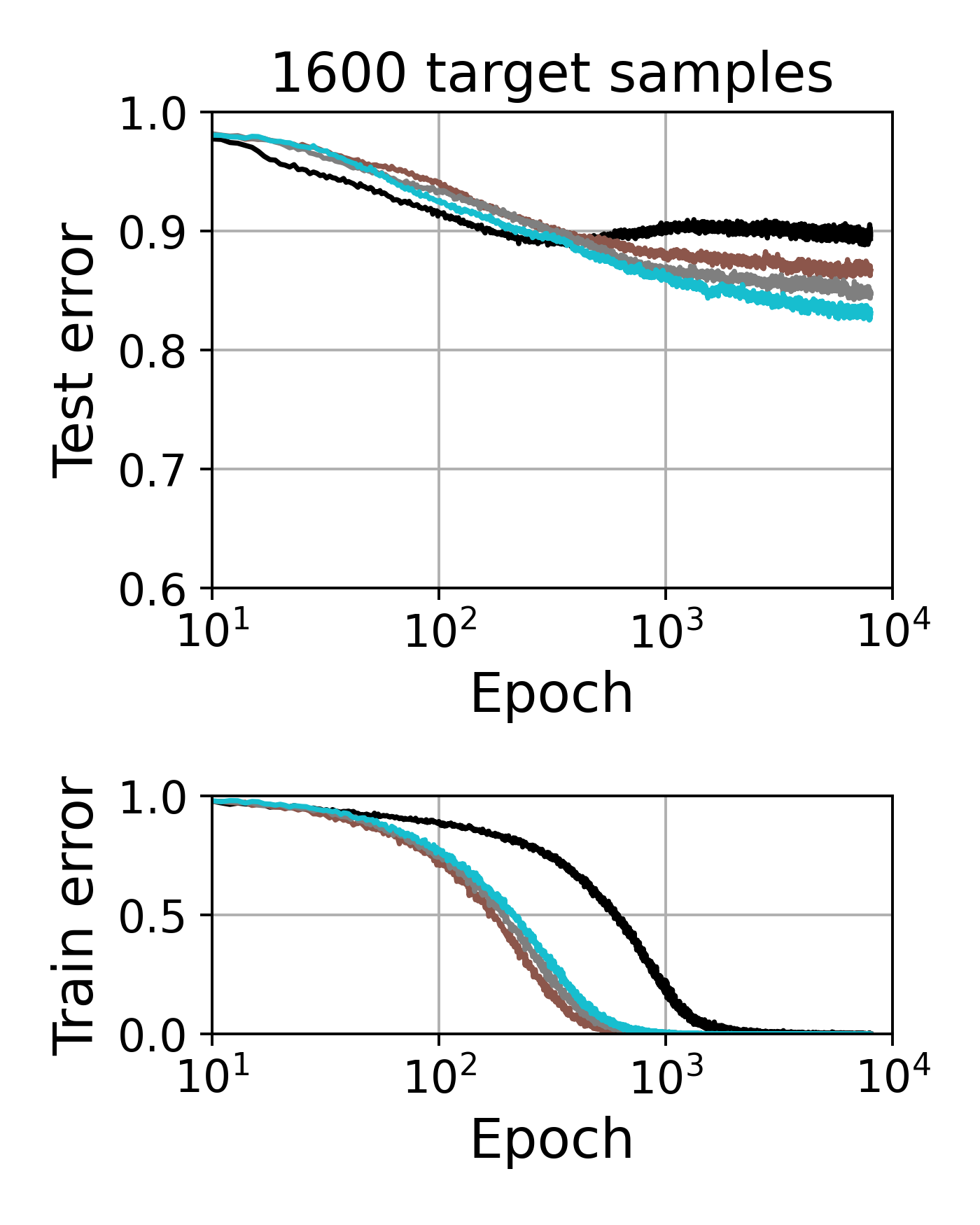}
   \includegraphics[width=0.245\textwidth]{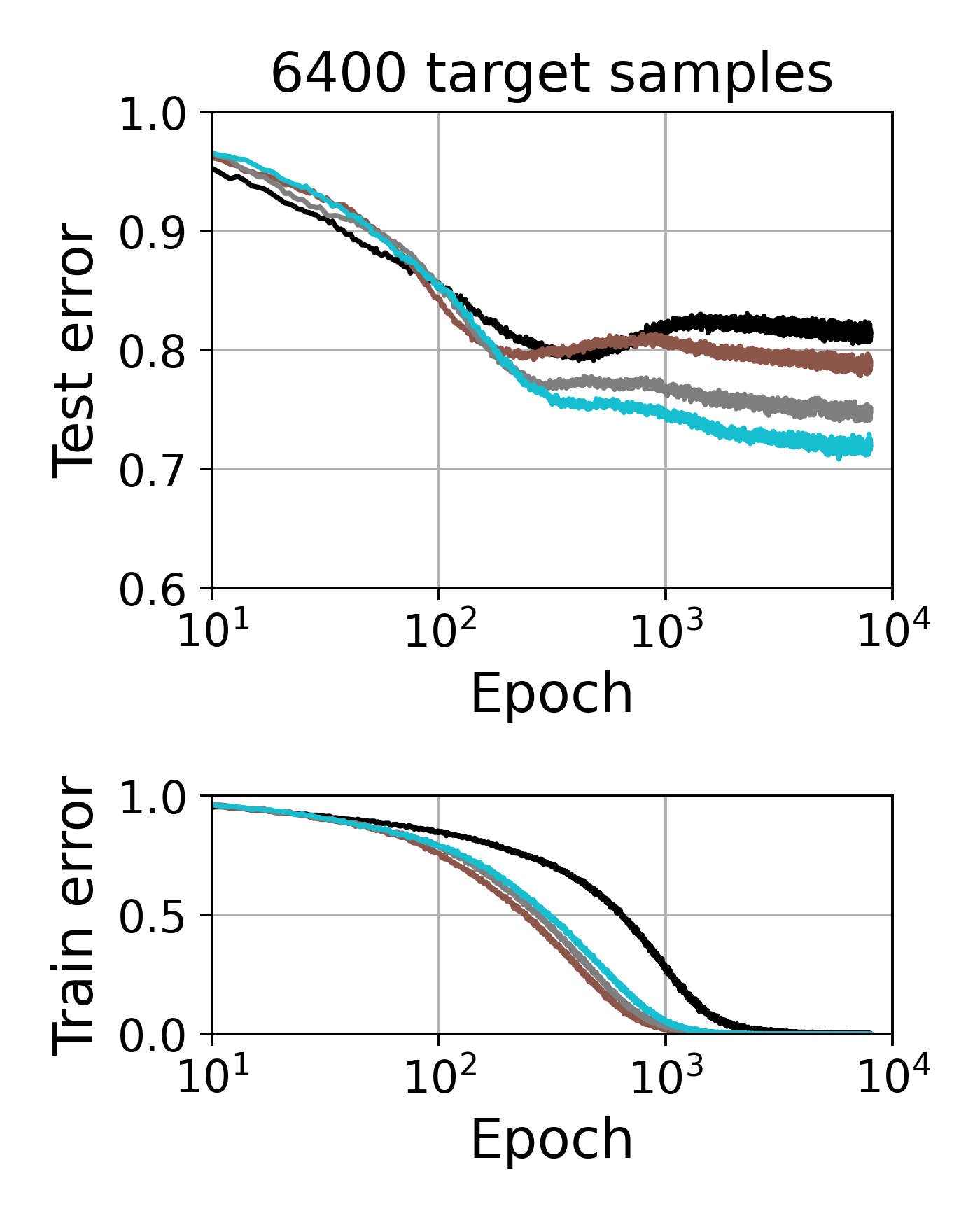}
   \includegraphics[width=0.245\textwidth]{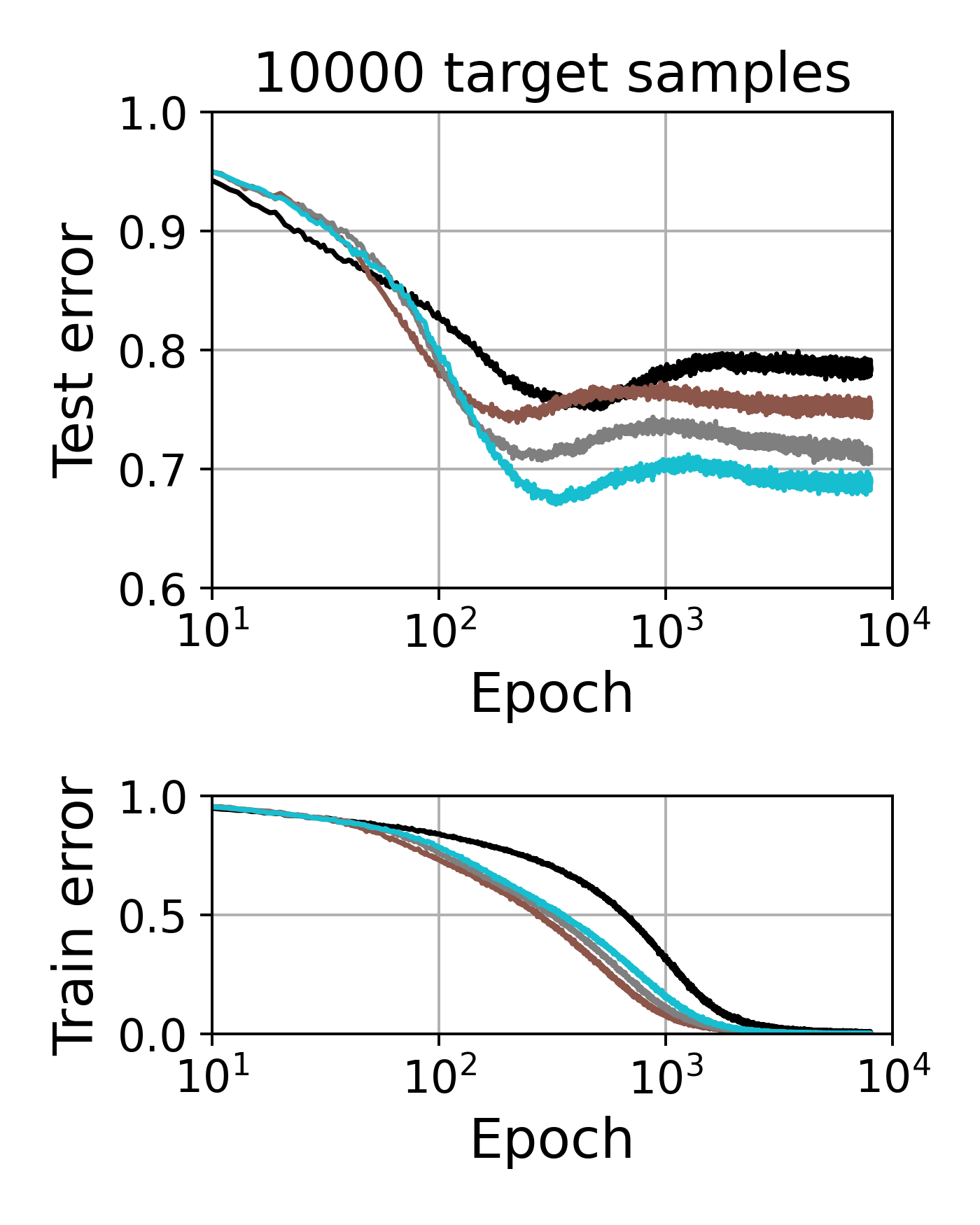} }
   \\
   \subfloat[ResNet (width parameter 24)]{\label{appendix:fig:src_different_dataset_size_100Classes_srcImagenetF100classtgtCIFAR100all_srcNoiselessTgtNoisy0p2_resnetW24}
   \includegraphics[width=0.245\textwidth]{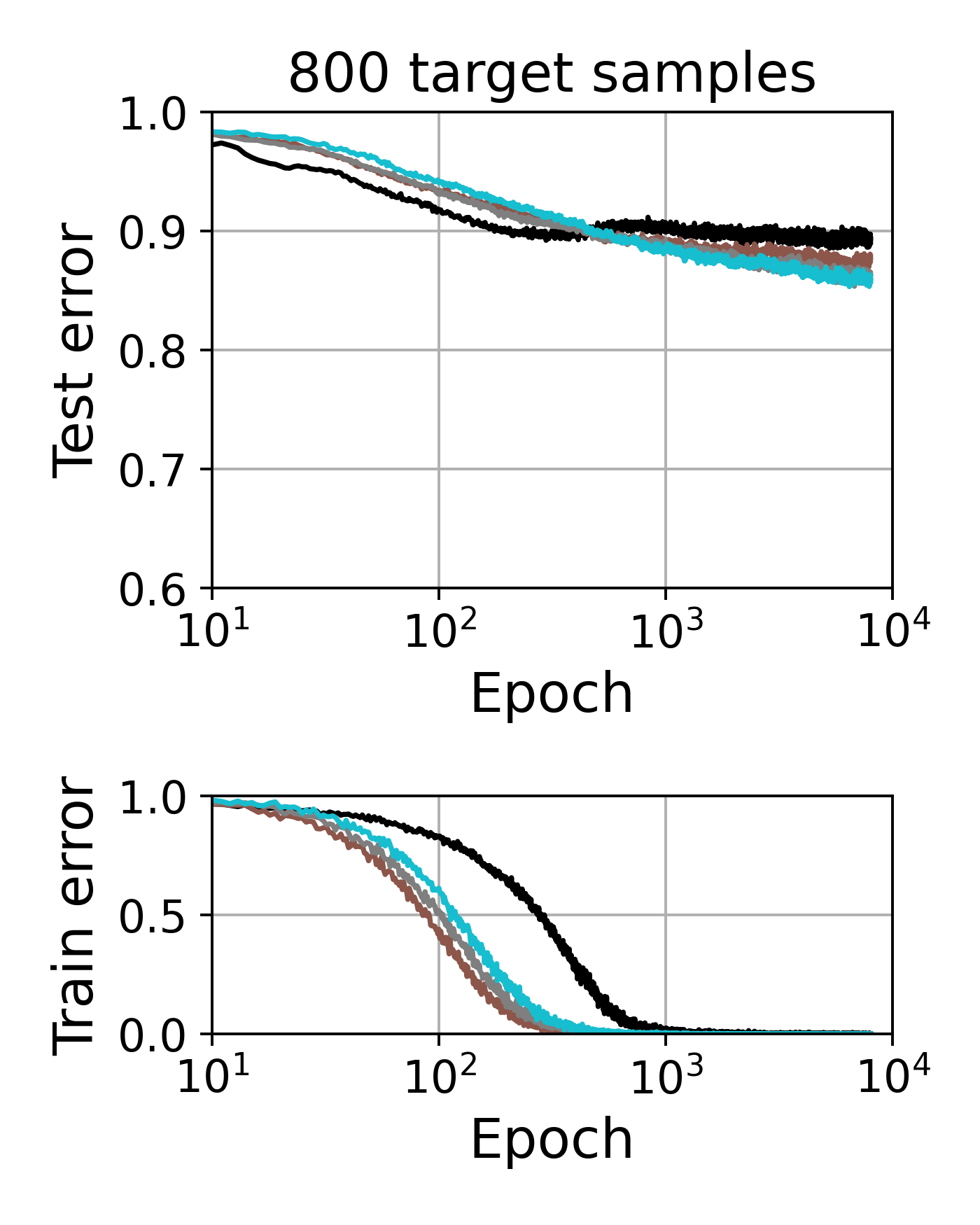}
      \includegraphics[width=0.245\textwidth]{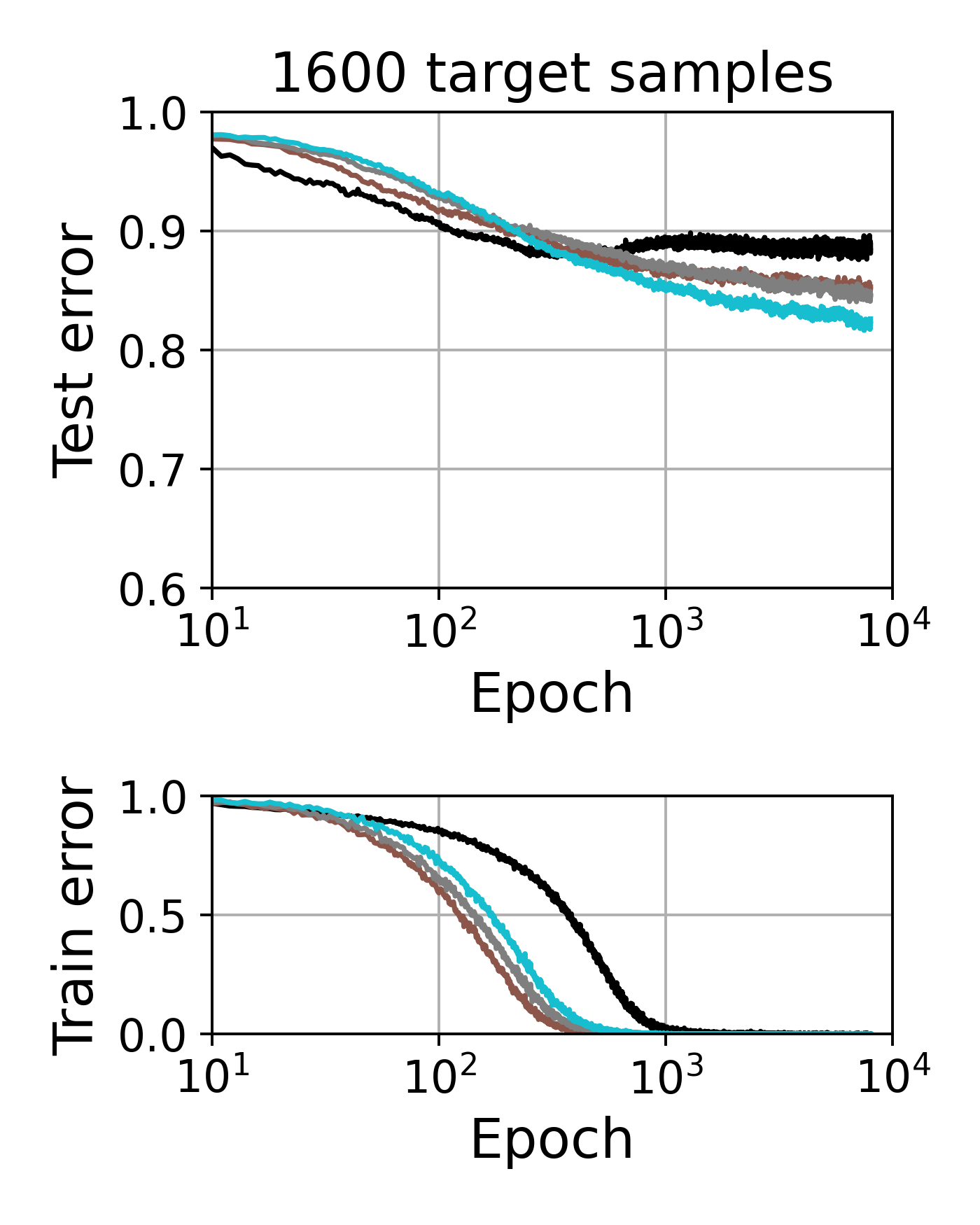}
   \includegraphics[width=0.245\textwidth]{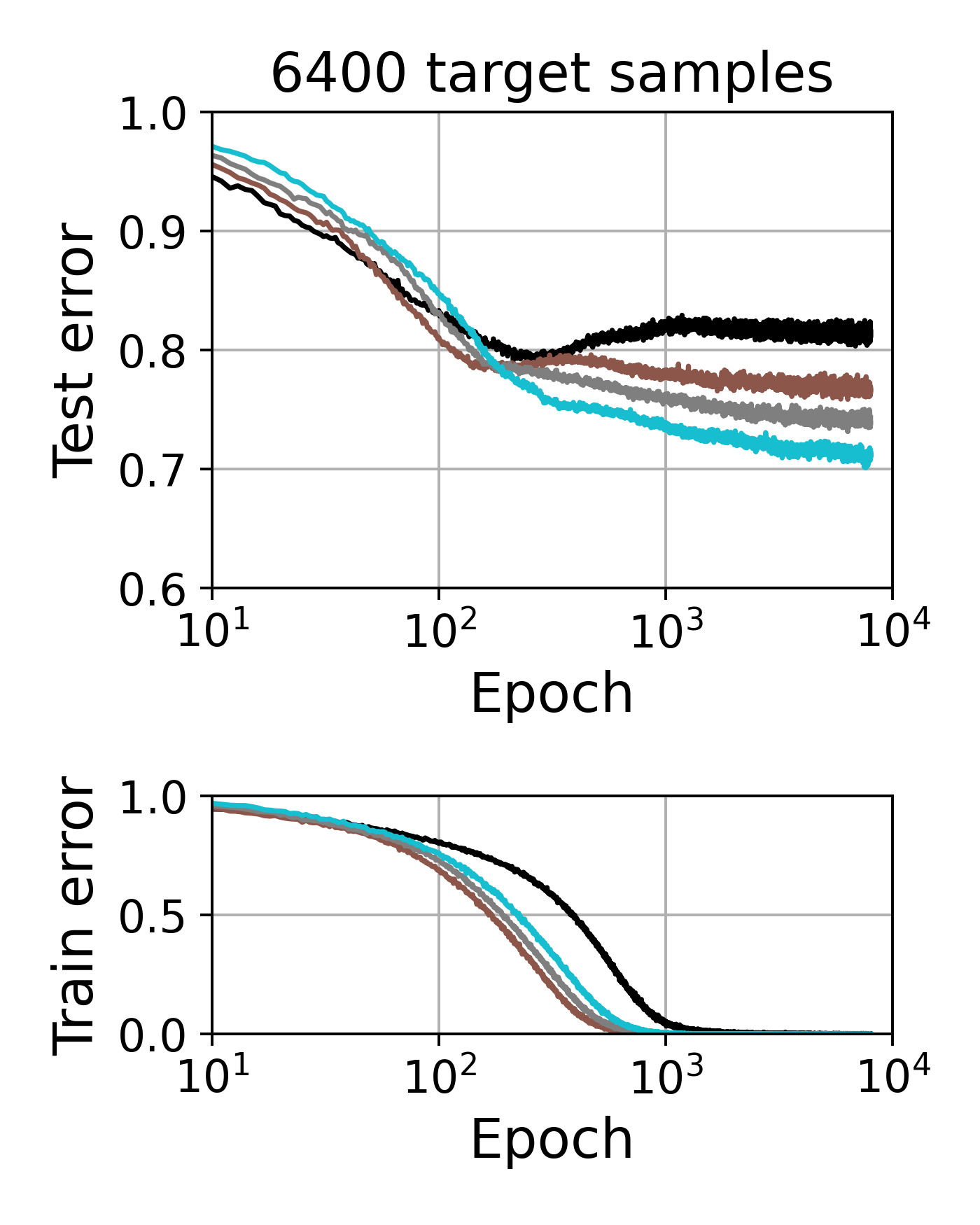}
   \includegraphics[width=0.245\textwidth]{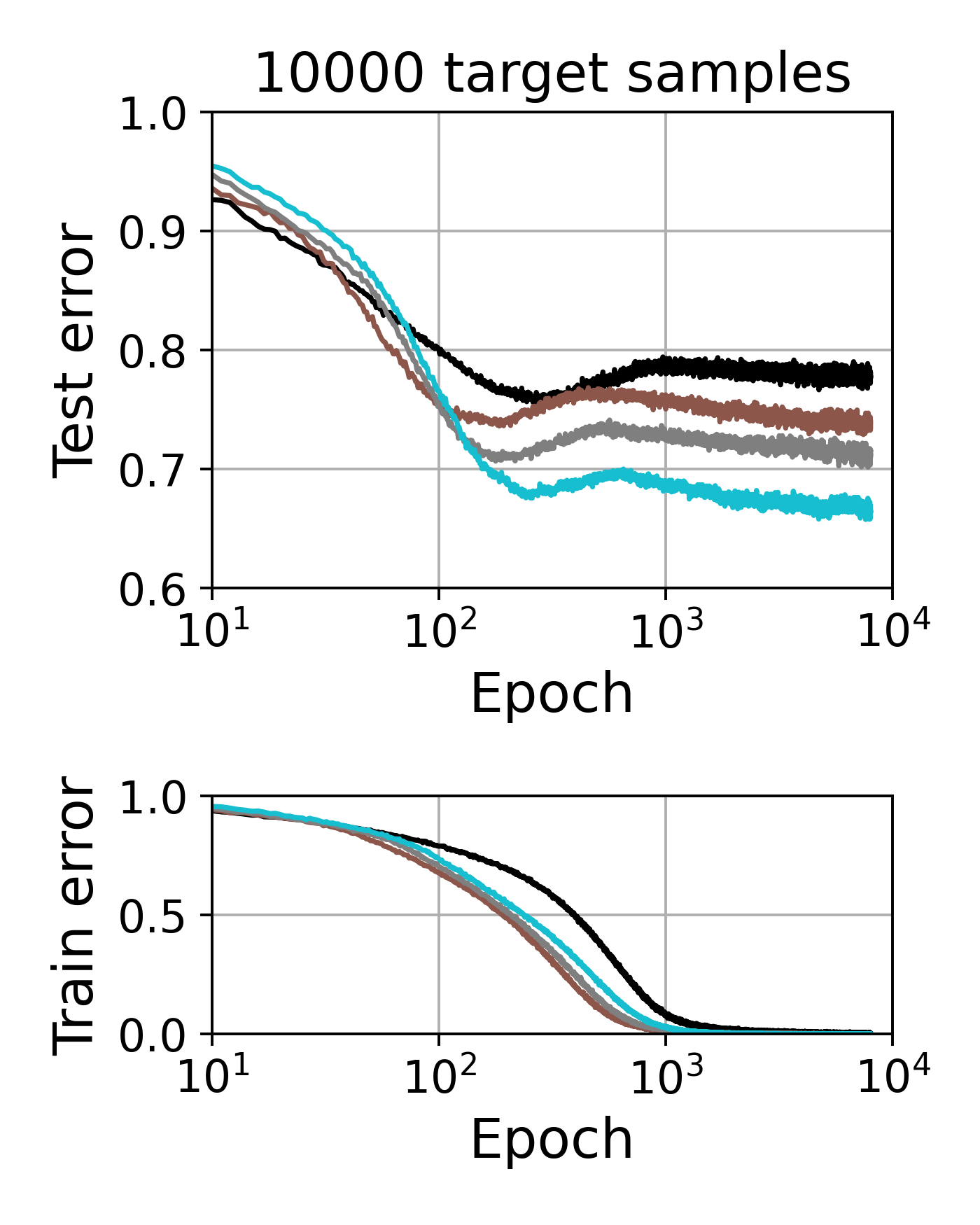} }
   \caption{The effect of the source dataset size on the target model training. Results for \textbf{narrower forms of ResNet-18} (the definition of ResNet-18 width is as in \cite{nakkiran2019deep,somepalli2022can}; width parameter 64 corresponds to the standard ResNet-18 as examined in Fig.~\ref{appendix:fig:src_different_dataset_size_100Classes_srcImagenetF100classtgtCIFAR100all_srcNoiselessTgtNoisy0p2_resnet} and in the rest of this paper). Compare also to Fig.~\ref{appendix:fig:src_different_dataset_size_100Classes_srcImagenetF100classtgtCIFAR100all_srcNoiselessTgtNoisy0p2_resnet} and note that learning from scratch becomes slower (relative to transfer learning) for a narrower DNN. These results evaluate training of a target classification task of \textbf{CIFAR-100} (all 100 classes)  with 20\% label noise in the target dataset. The transfer learning is from the source task of 100 Tiny ImageNet classes (input image size 32x32x3). Each curve color corresponds to another size of the source dataset.  }
   \label{fig:src_different_dataset_size_100Classes_srcImagenetF100classtgtCIFAR100all_srcNoiselessTgtNoisy0p2_narrowResNet}
\end{figure*}

 In Section \ref{subsec:effect of source dataset size} we have considered the effect of the \textit{source} dataset size on the evolution of the error curves in the transfer learning of the \textit{target} DNN.
 Here we provide additional results for various datasets and dataset sizes. 
 Figures \ref{fig:src_different_dataset_size_100Classes_srcImagenetF100classtgtFood100all_srcNoiselessTgtNoisy0p2}, \ref{fig:src_different_dataset_size_10Classes_srcCIFAR10classtgtFoodF10_srcNoiselessTgtNoisy0p2} show results for transfer learning of ResNet-18 where the source and target tasks have the same number of classes and input image size and, hence, all the source DNN layers (including the last layer) are transferred and used for the target DNN initialization; these results clearly demonstrate the slower arrival to interpolation of the target dataset for a larger source dataset (also note, in Figs.~\ref{fig:src_different_dataset_size_100Classes_srcImagenetF100classtgtFood100all_srcNoiselessTgtNoisy0p2}, \ref{fig:src_different_dataset_size_10Classes_srcCIFAR10classtgtFoodF10_srcNoiselessTgtNoisy0p2}, the corresponding delay in the double descent peak when the target dataset is sufficiently large). 

The results for DenseNet and ViT in Figures \ref{appendix:fig:src_different_dataset_size_100Classes_srcImagenetF100classtgtCIFAR100all_srcNoiselessTgtNoisy0p2_densenet} and \ref{fig:src_different_dataset_size_100Classes_srcImagenetF100classtgtCIFAR100all_srcNoiselessTgtNoisy0p2_ViT_appendix}, respectively, show slower arrival to interpolation of the target dataset for a larger source dataset --- but mainly for the smaller target datasets. This demonstrates the effect of the architecture and is a potential consequence of the better transfer learning ability of the convolutional form of ResNet, e.g., compared to ViT \cite{raghu2021do}. 

In Fig.~\ref{fig:src_different_dataset_size_100Classes_srcImagenetF100classtgtCIFAR100all_srcNoiselessTgtNoisy0p2_narrowResNet} we evaluate the training of ResNet-18 forms that are narrower than the standard ResNet-18. These results emphasize the effect of the parameterization level of the model on the transfer learning speed, and its relation to the speed of learning from scratch. We use the same definition of variable-width ResNet-18 as in \cite{nakkiran2019deep,somepalli2022can}.

\section{Additional Results for Section 4}
\label{appendix:sec:Additional Experimental Results for Section 4}

\begin{figure*}[t]
  \centering
  \includegraphics[width=0.9\textwidth]{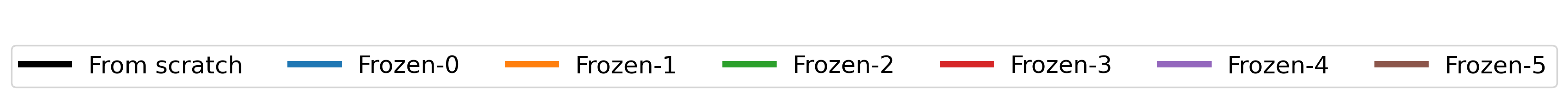}
  \\[-3ex]
    \subfloat[]{
   \includegraphics[width=0.245\textwidth]{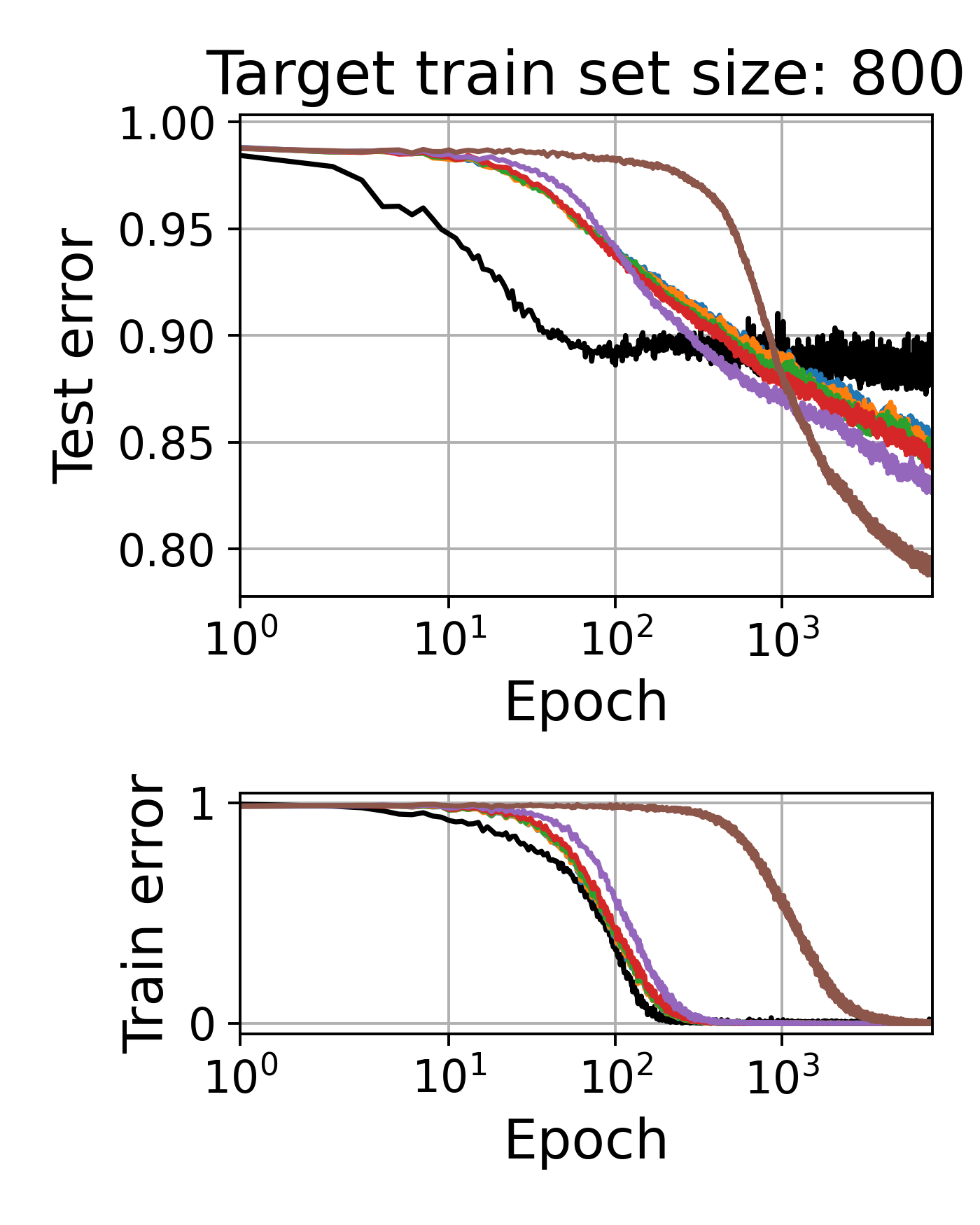}
   \label{fig:test_error_vs_epoch_for_freezing_levels_100Classes_srcImagenetF100classtgtCIFAR100all_dataset800_srcNoiselessTgtNoisy0p2}}
  \subfloat[]{
   \includegraphics[width=0.245\textwidth]{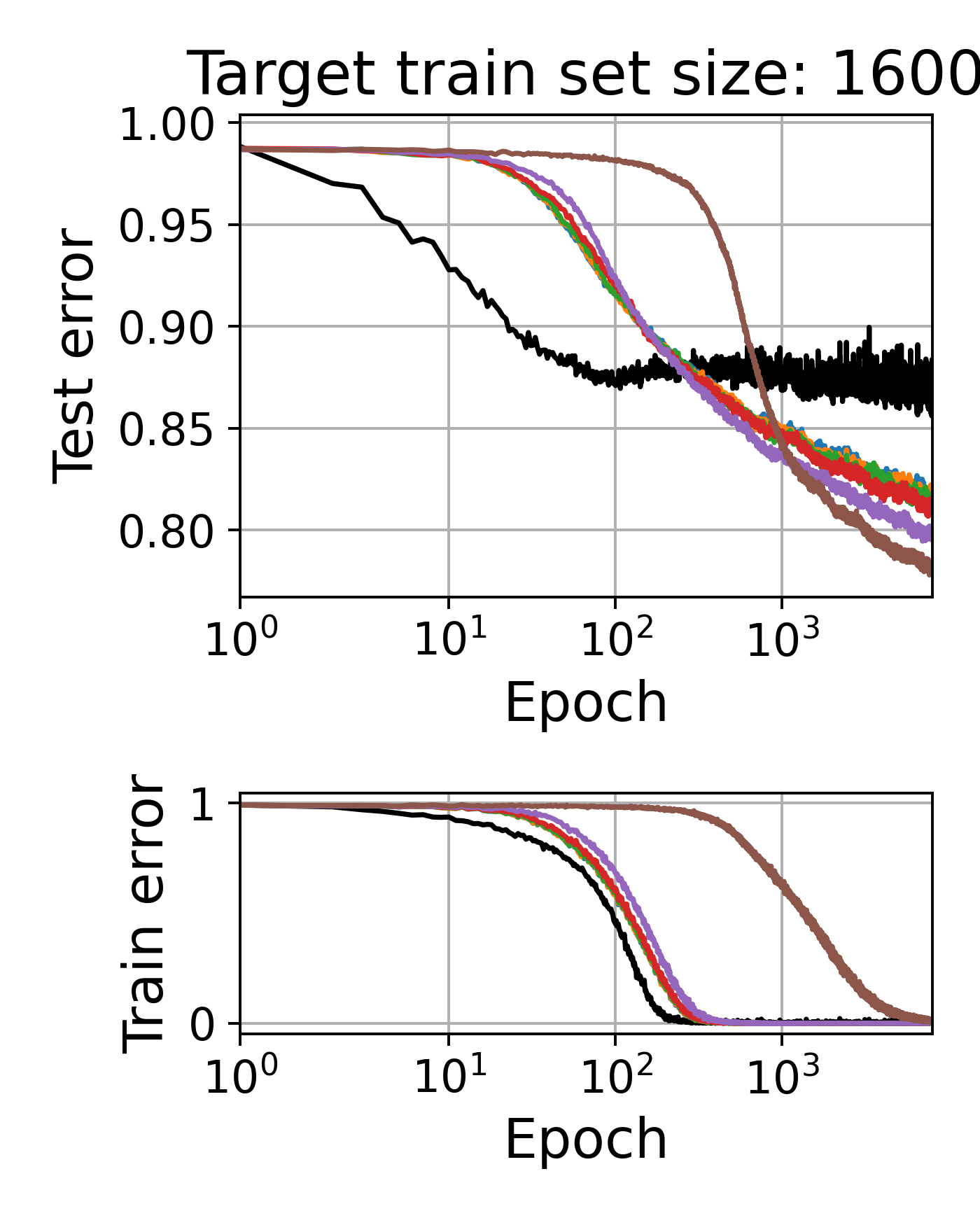}
   \label{fig:test_error_vs_epoch_for_freezing_levels_100Classes_srcImagenetF100classtgtCIFAR100all_dataset1600_srcNoiselessTgtNoisy0p2}}
  \subfloat[]{
   \includegraphics[width=0.245\textwidth]{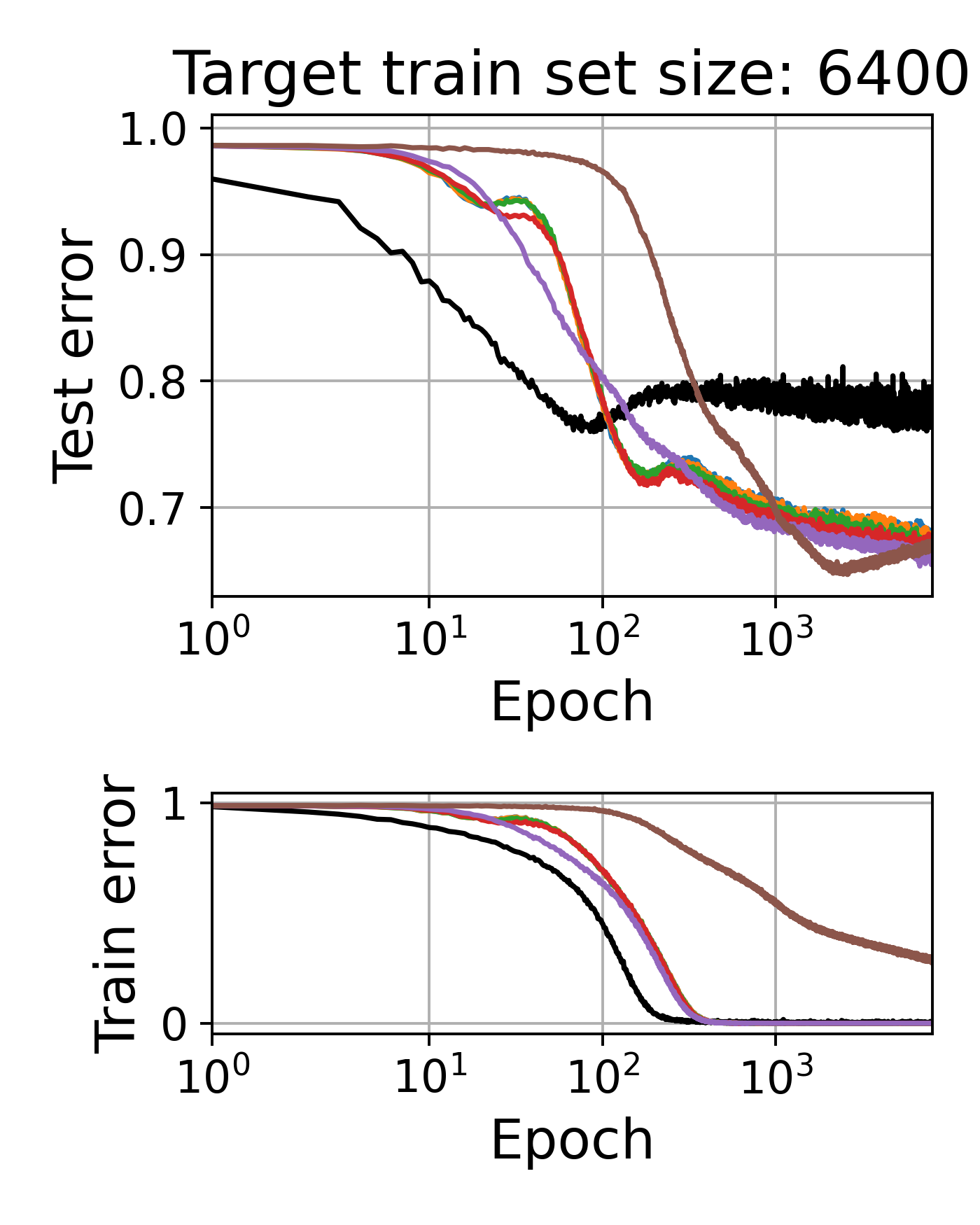}
   \label{fig:test_error_vs_epoch_for_freezing_levels_100Classes_srcImagenetF100classtgtCIFAR100all_dataset6400_srcNoiselessTgtNoisy0p2}}
   \subfloat[]{
   \includegraphics[width=0.245\textwidth]{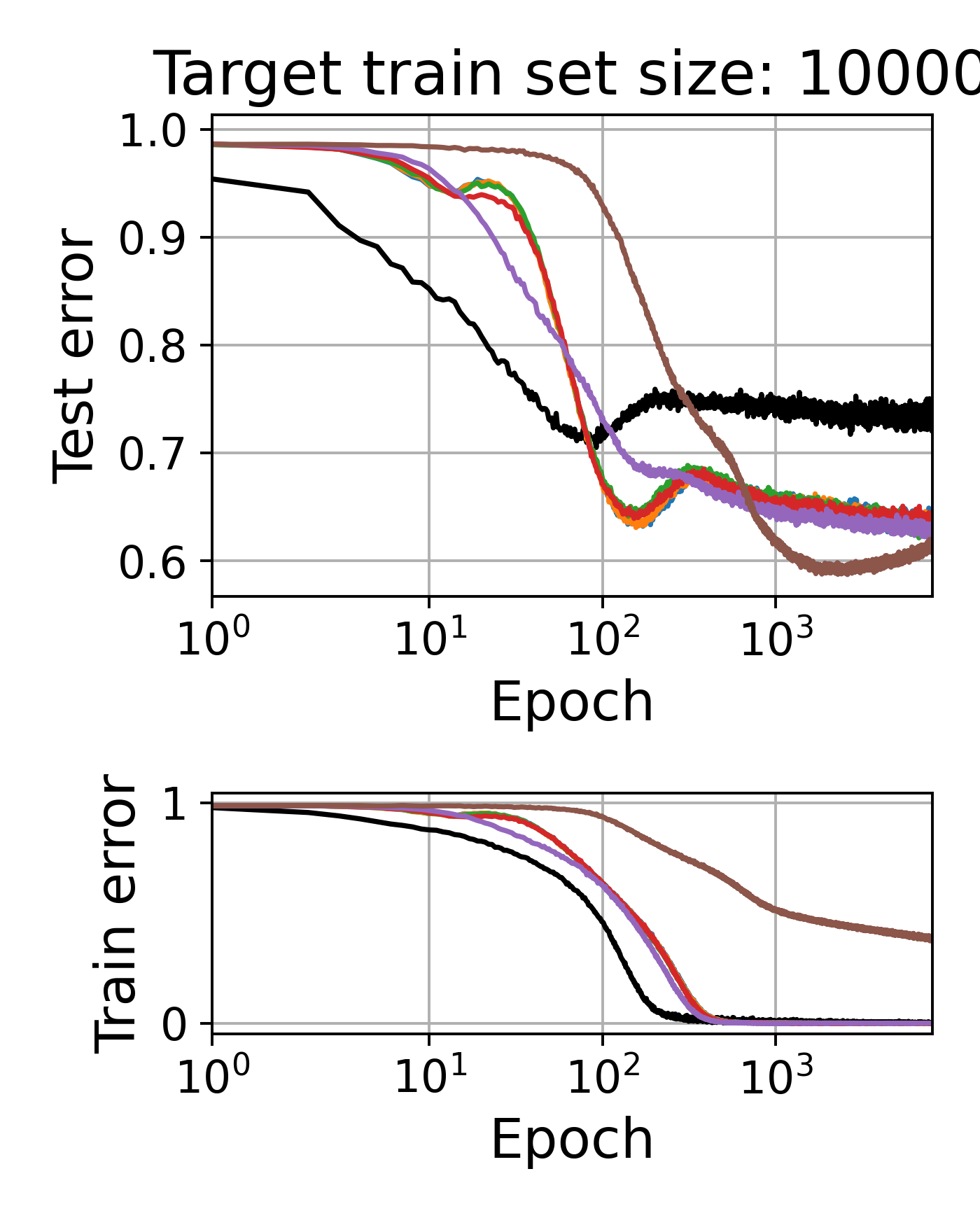}
   \label{fig:test_error_vs_epoch_for_freezing_levels_100Classes_srcImagenetF100classtgtCIFAR100all_dataset10000_srcNoiselessTgtNoisy0p2}}
   \caption{Evaluation of \textbf{ResNet} transfer learning at various freezing levels, for a target classification task of CIFAR-100 (all 100 classes, 20\% label noise in target datasets). The transfer learning is from the source task of 100 Tiny ImageNet classes (input image size 32x32x3) with source dataset of 50k training samples. Note that, for relatively large target datasets, freezing layers can eliminate double descent.}
   \label{fig:test_error_vs_epoch_for_freezing_levels_resnet_100Classes_srcImagenetF100classtgtCIFAR100all_srcNoiselessTgtNoisy0p2}
\end{figure*}

\begin{figure*}[t]
  \centering
  \includegraphics[width=0.9\textwidth]{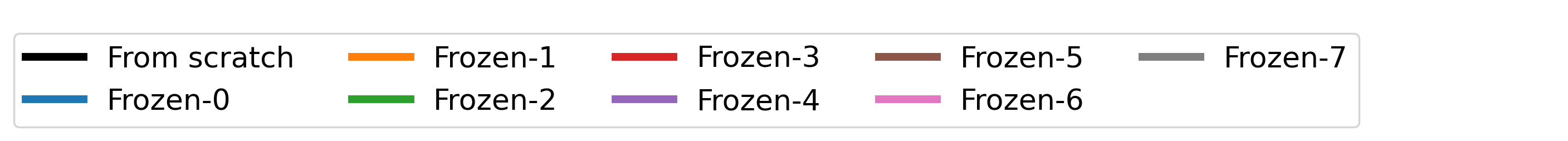}
  \\[-3ex]
    \subfloat[]{
   \includegraphics[width=0.245\textwidth]{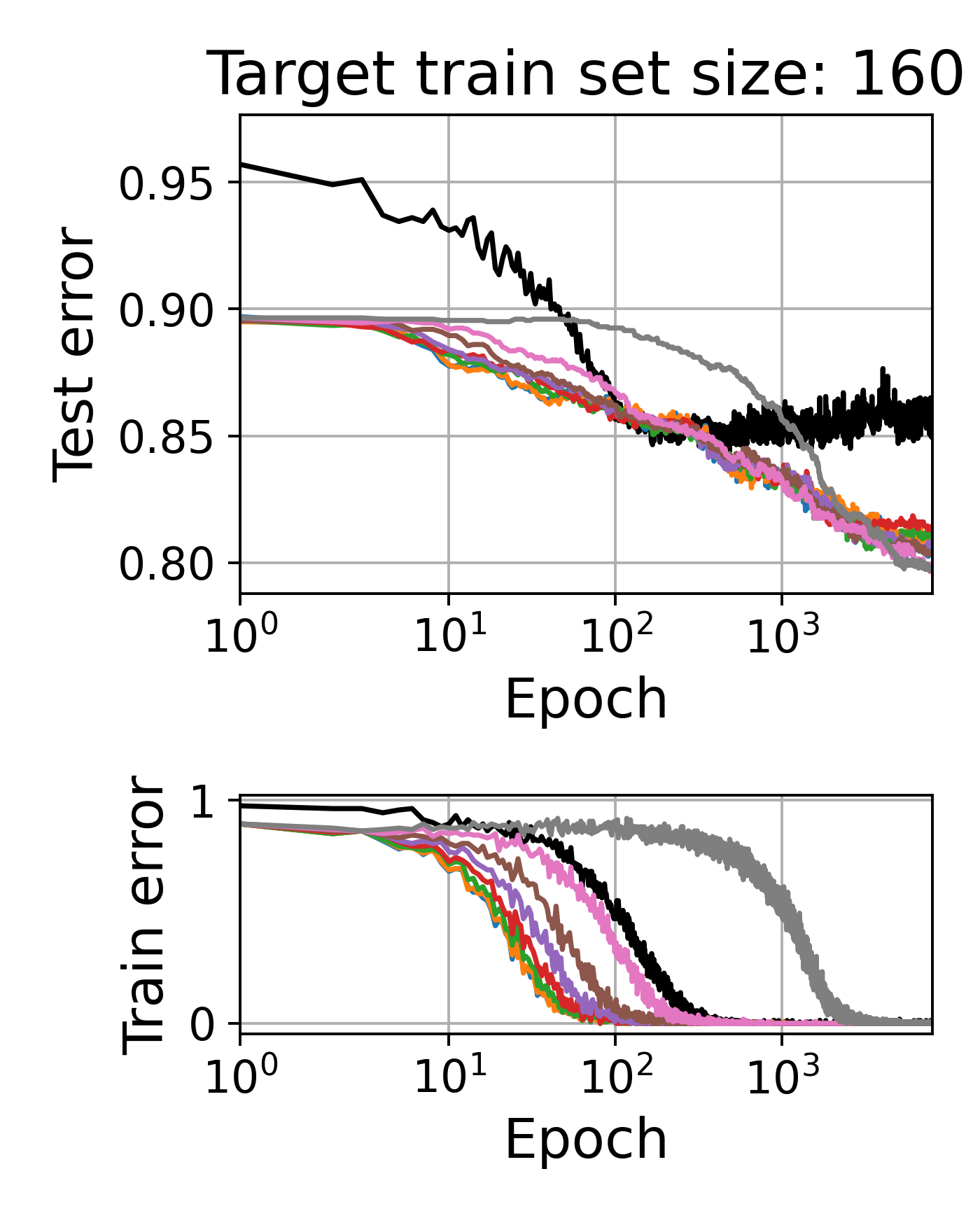}
   \label{fig:test_error_vs_epoch_for_freezing_levels_vit_tinyimagenet32_src40SimilarClasstgt40class_dataset160_srcNoiselessTgtNoisy0p2}}
  \subfloat[]{
   \includegraphics[width=0.245\textwidth]{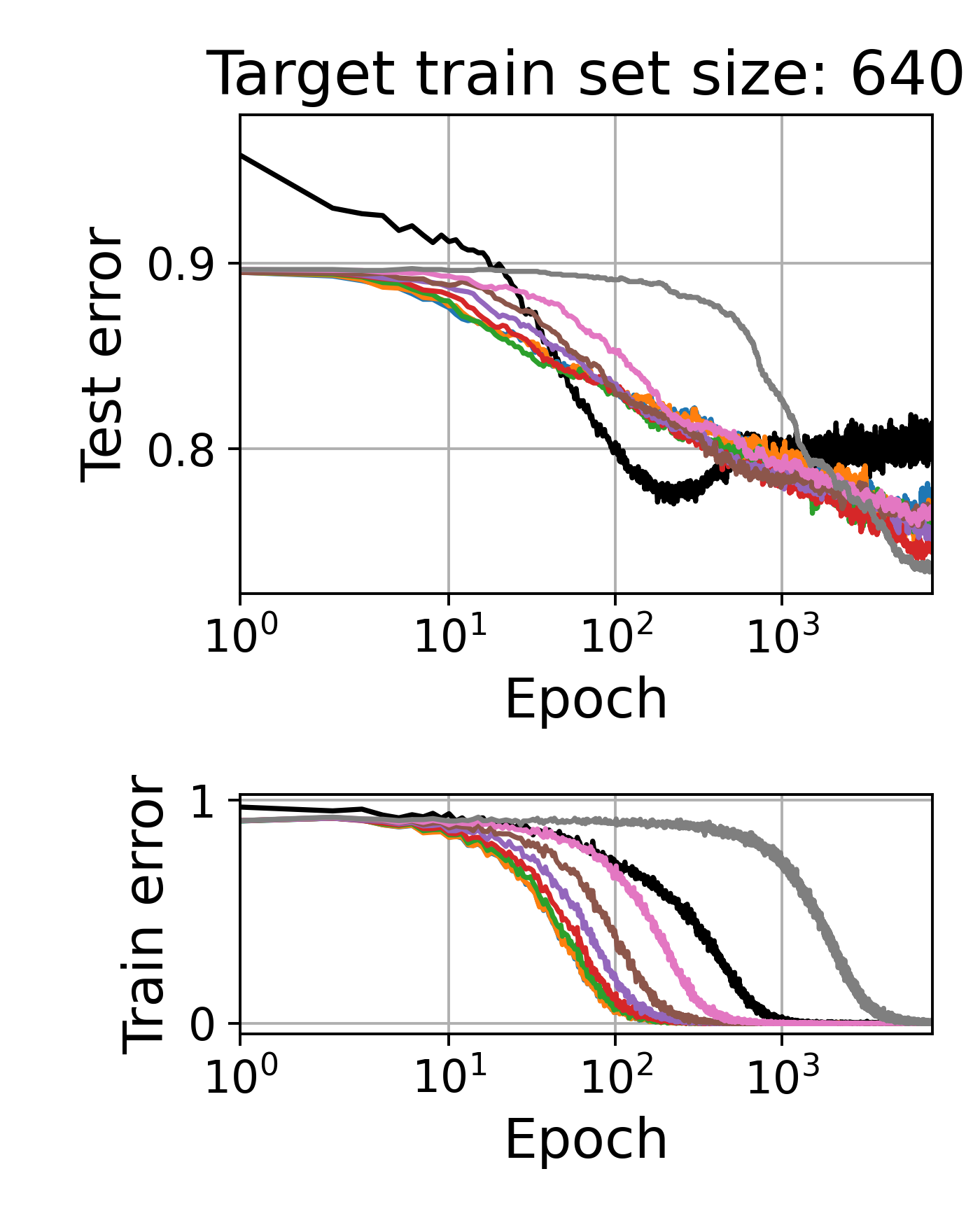}
   \label{fig:test_error_vs_epoch_for_freezing_levels_vit_tinyimagenet32_src40SimilarClasstgt40class_dataset640_srcNoiselessTgtNoisy0p2}}
  \subfloat[]{
   \includegraphics[width=0.245\textwidth]{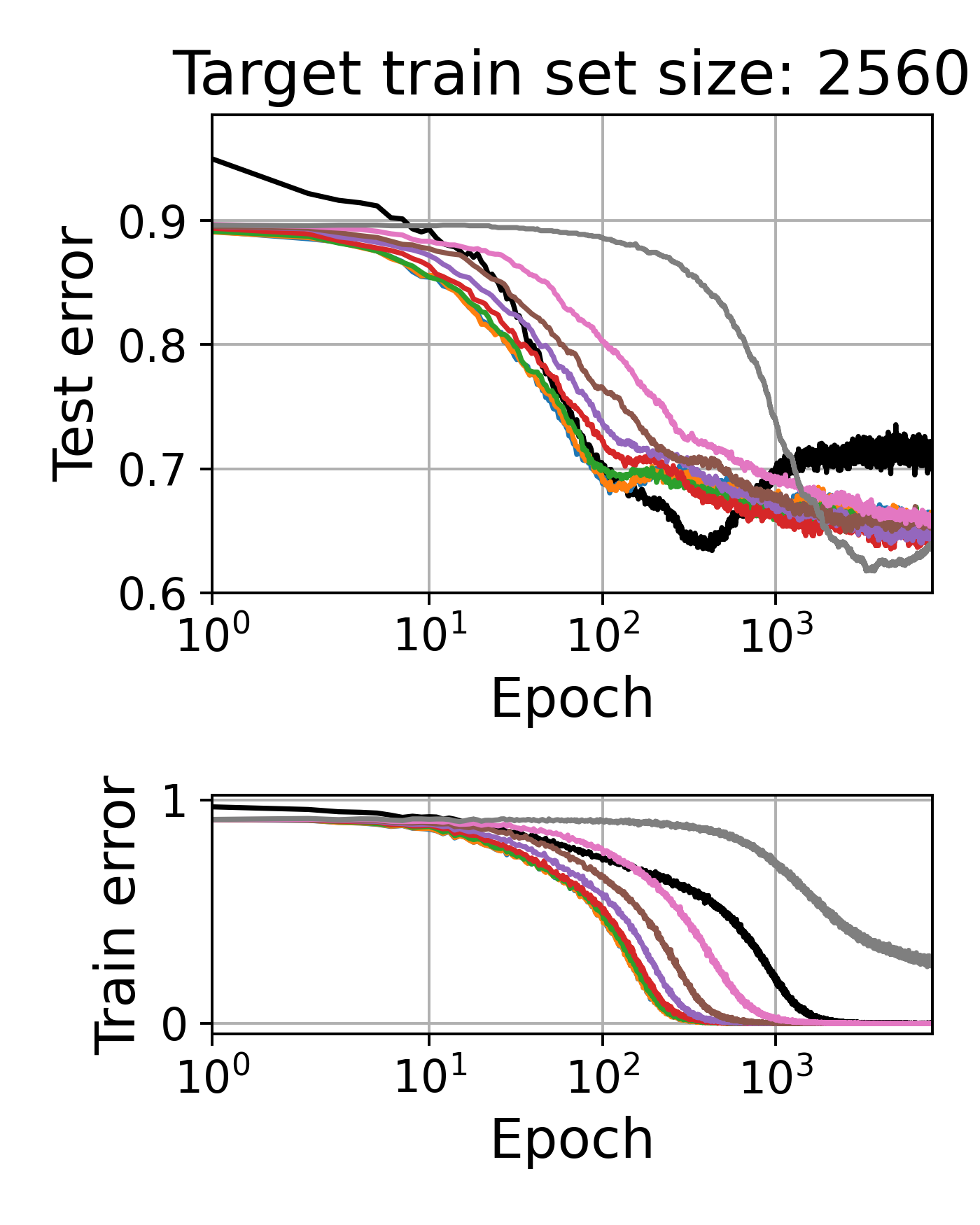}
   \label{fig:test_error_vs_epoch_for_freezing_levels_vit_tinyimagenet32_src40SimilarClasstgt40class_dataset2560_srcNoiselessTgtNoisy0p2}}
   \subfloat[]{
   \includegraphics[width=0.245\textwidth]{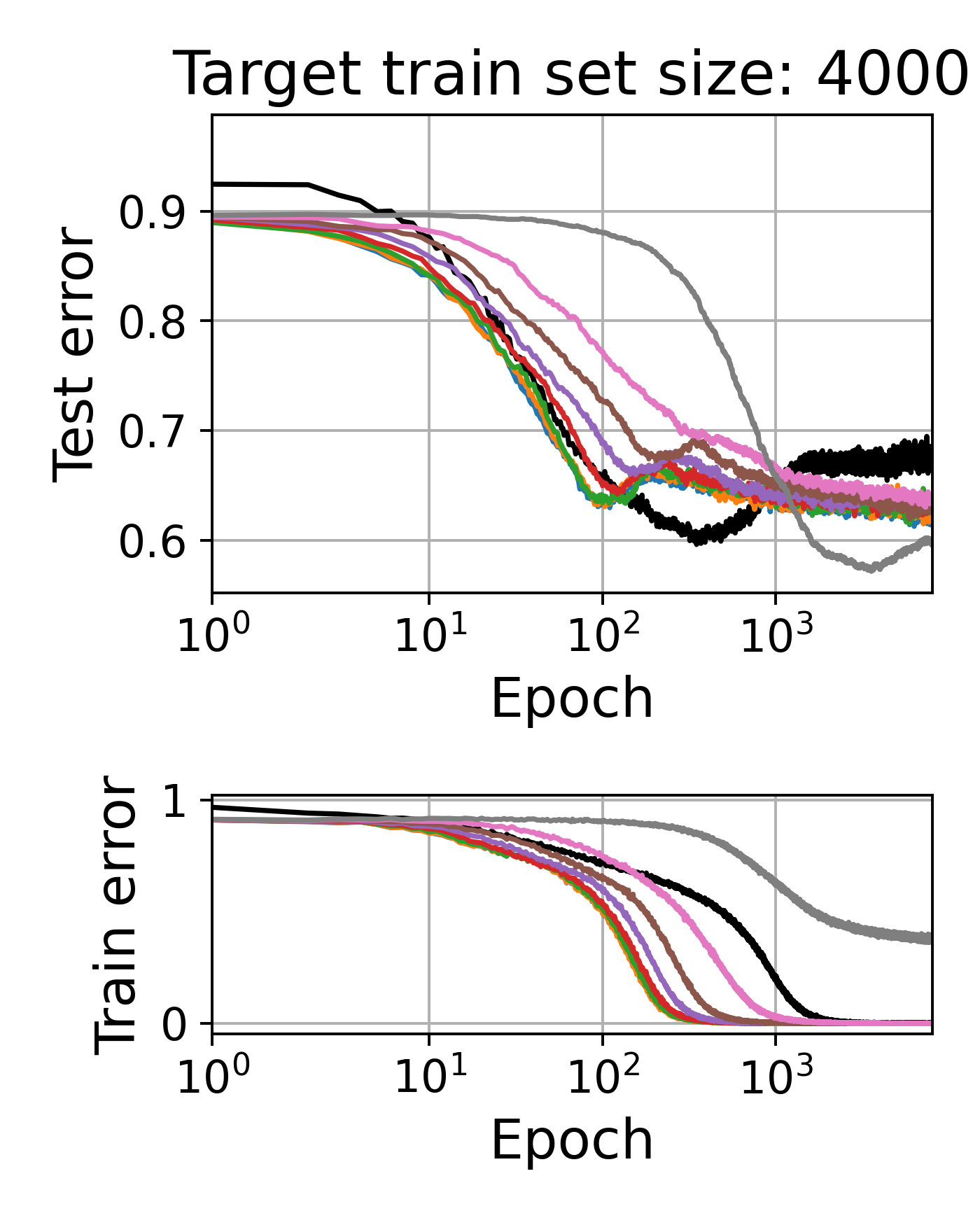}
   \label{fig:test_error_vs_epoch_for_freezing_levels_vit_tinyimagenet32_src40SimilarClasstgt40class_dataset4000_srcNoiselessTgtNoisy0p2}}
   \caption{Evaluation of \textbf{ViT} transfer learning at various freezing levels, for a target classification task of 40 classes from CIFAR-100 (20\% label noise in target datasets). The transfer learning is from the source task of 40 Tiny ImageNet classes that are similar to the target task classes (input image size 32x32x3) with source dataset of 20k training samples. Note that, for relatively large target datasets, freezing layers can eliminate double descent.}
   \label{fig:test_error_vs_epoch_for_freezing_levels_vit_tinyimagenet32_src40SimilarClasstgt40class_srcNoiselessTgtNoisy0p2}
\end{figure*}

\begin{figure*}[t]
  \centering
  \includegraphics[width=0.9\textwidth]{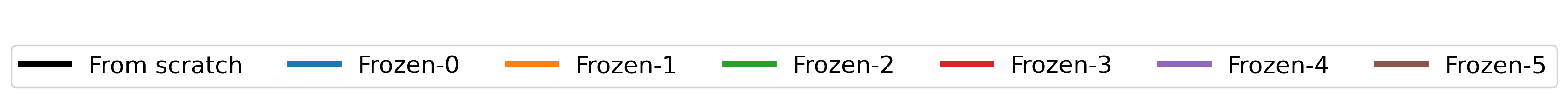}
  \\[-3ex]
  \subfloat[]{
   \includegraphics[width=0.245\textwidth]{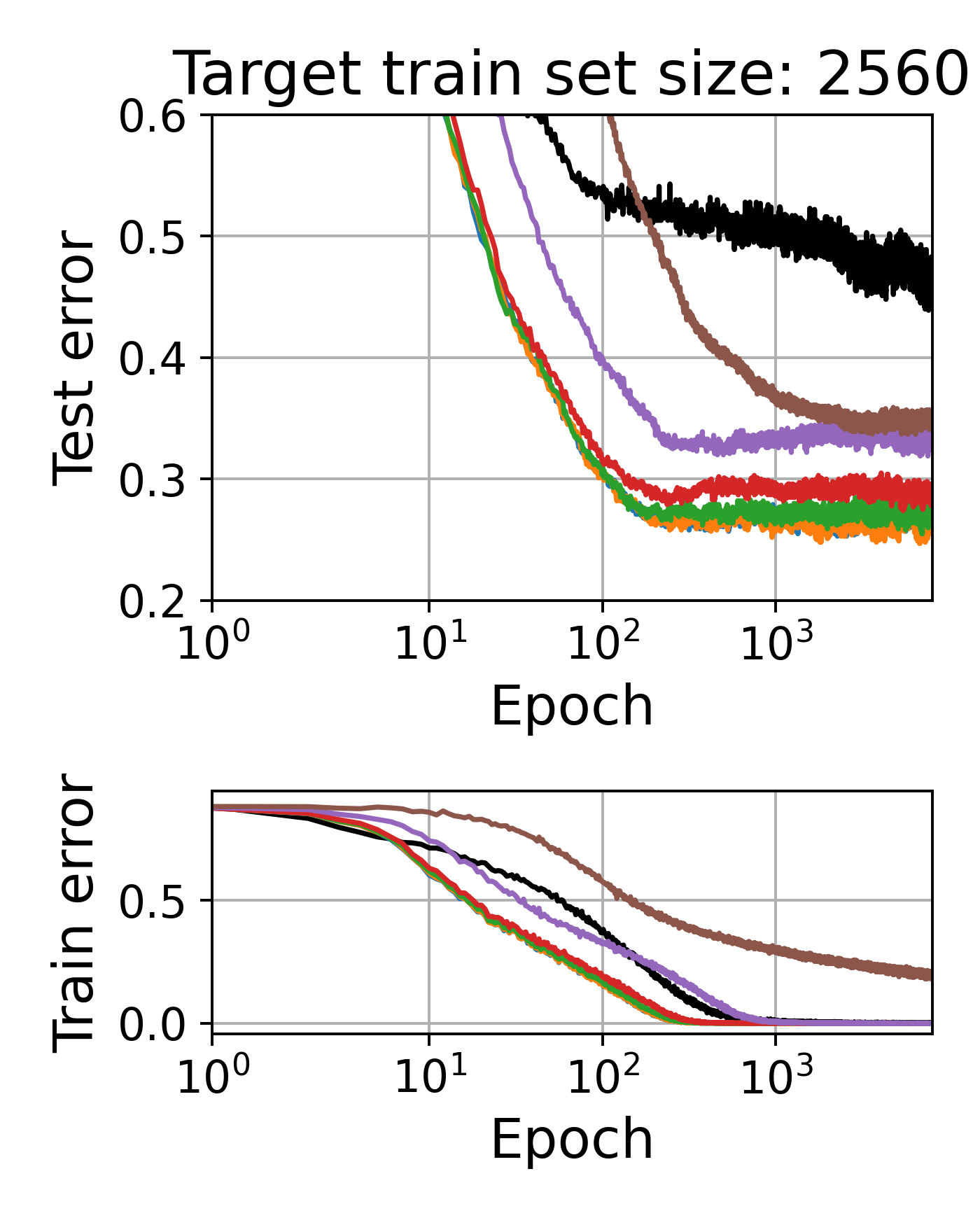}
   \label{fig:test_error_vs_epoch_for_freezing_levels_densenet_src100k_tinyimagenet64_200Classes_src200classtgtCIFAR10class_dataset2560_srcNoiselessTgtNoiseless}}
   \subfloat[]{
   \includegraphics[width=0.245\textwidth]{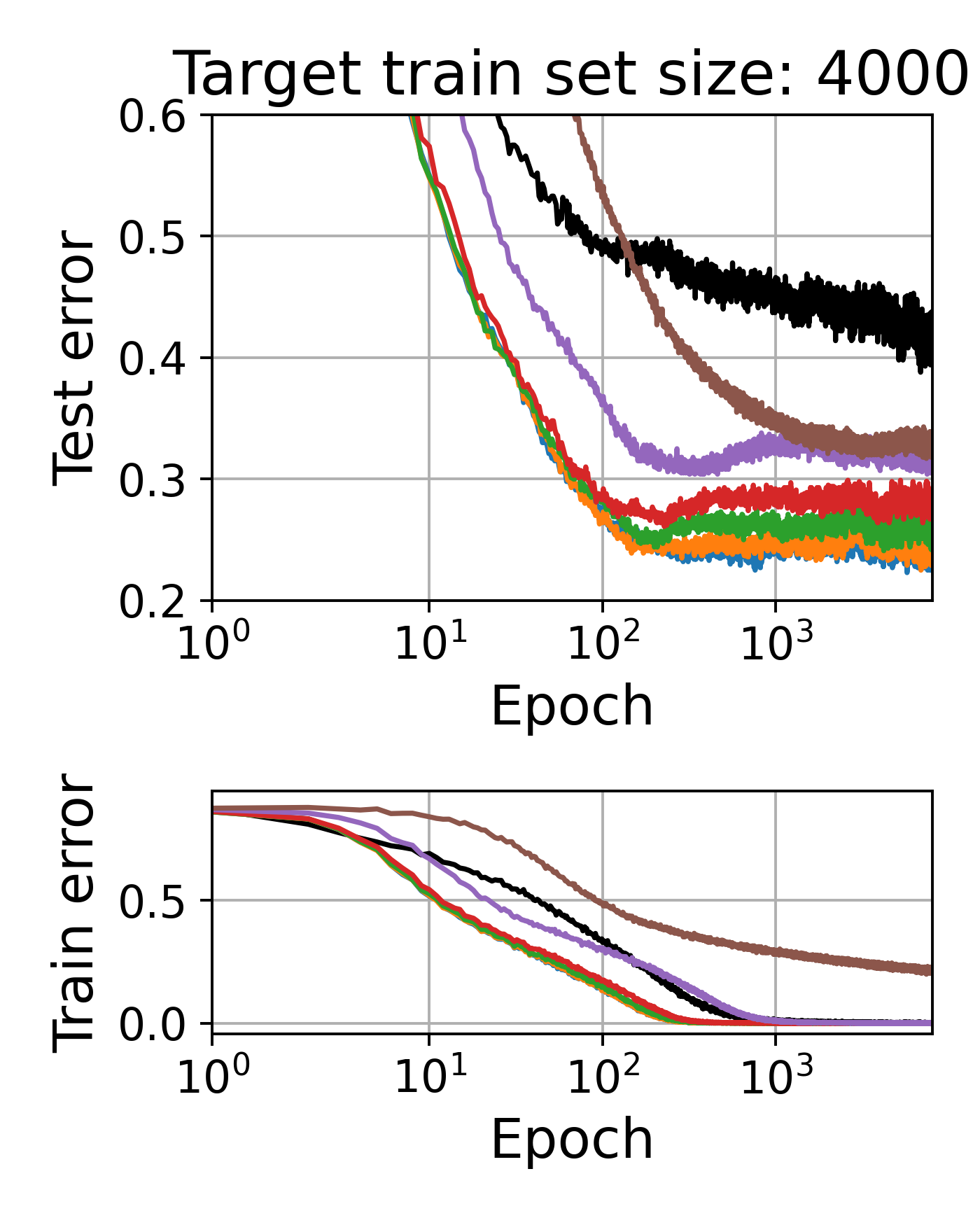}
   \label{fig:test_error_vs_epoch_for_freezing_levels_densenet_src100k_tinyimagenet64_200Classes_src200classtgtCIFAR10class_dataset4000_srcNoiselessTgtNoiseless}}
   \caption{Evaluation of transfer learning errors at various freezing levels. The target task is CIFAR-10 \textbf{without} label noise in target datasets. The subfigures in each row differ in the target dataset size. The architecture is \textbf{DenseNet}. The transfer learning are from the Tiny ImageNet source task (200 classes, input image size 64x64x3). These results show that freezing layers can eliminate double descent, if exists, also in noiseless settings.}
   \label{fig:test_error_vs_epoch_for_freezing_levels_densenet_src100k_tinyimagenet64_200Classes_src200classtgtCIFAR10class_srcNoiselessTgtNoiseless}
\end{figure*}

In Section \ref{sec:Freezing Layers}  we examine the error evolution during training for transfer learning with various levels of freezing (i.e., different number of frozen layers), here we provide additional results. 

\subsection{Additional Results for Section 4.2}
\label{appendix:subsec:Additional Experimental Results for Section 4.2}

In Section \ref{subsec:Freezing Layers Can Eliminate Double Descent for Large Target Datasets} we examine the ability of freezing layers to eliminate the epoch-wise double descent, which occurs in transfer learning with relatively large target datasets. Fig.~4 shows results for ResNet, here we provide additional results with other datasets for ResNet (Fig.~\ref{fig:test_error_vs_epoch_for_freezing_levels_resnet_100Classes_srcImagenetF100classtgtCIFAR100all_srcNoiselessTgtNoisy0p2}), ViT (Fig.~\ref{fig:test_error_vs_epoch_for_freezing_levels_vit_tinyimagenet32_src40SimilarClasstgt40class_srcNoiselessTgtNoisy0p2})  and DenseNet (Fig.~\ref{fig:test_error_vs_epoch_for_freezing_levels_densenet_src100k_tinyimagenet64_200Classes_src200classtgtCIFAR10class_srcNoiselessTgtNoiseless}). 
In all these examples, we again observe that when the target dataset is large enough to induce double descent (see, e.g., the rightmost subfigures), freezing many layers can eliminate the double descent.

\subsection{Additional Results for Section 4.3}
\label{appendix:subsec:Additional Experimental Results for Section 4.3}

Figure \ref{fig:ViT_2d_error_diagrams_freezing_vs_epochs} in the main paper shows the test and training errors over a 2D plane of freezing levels vs.~epochs for transfer learning of a ViT model.  

\begin{figure*}[t]
  \centering
  \subfloat[]{\includegraphics[width=0.48\linewidth]{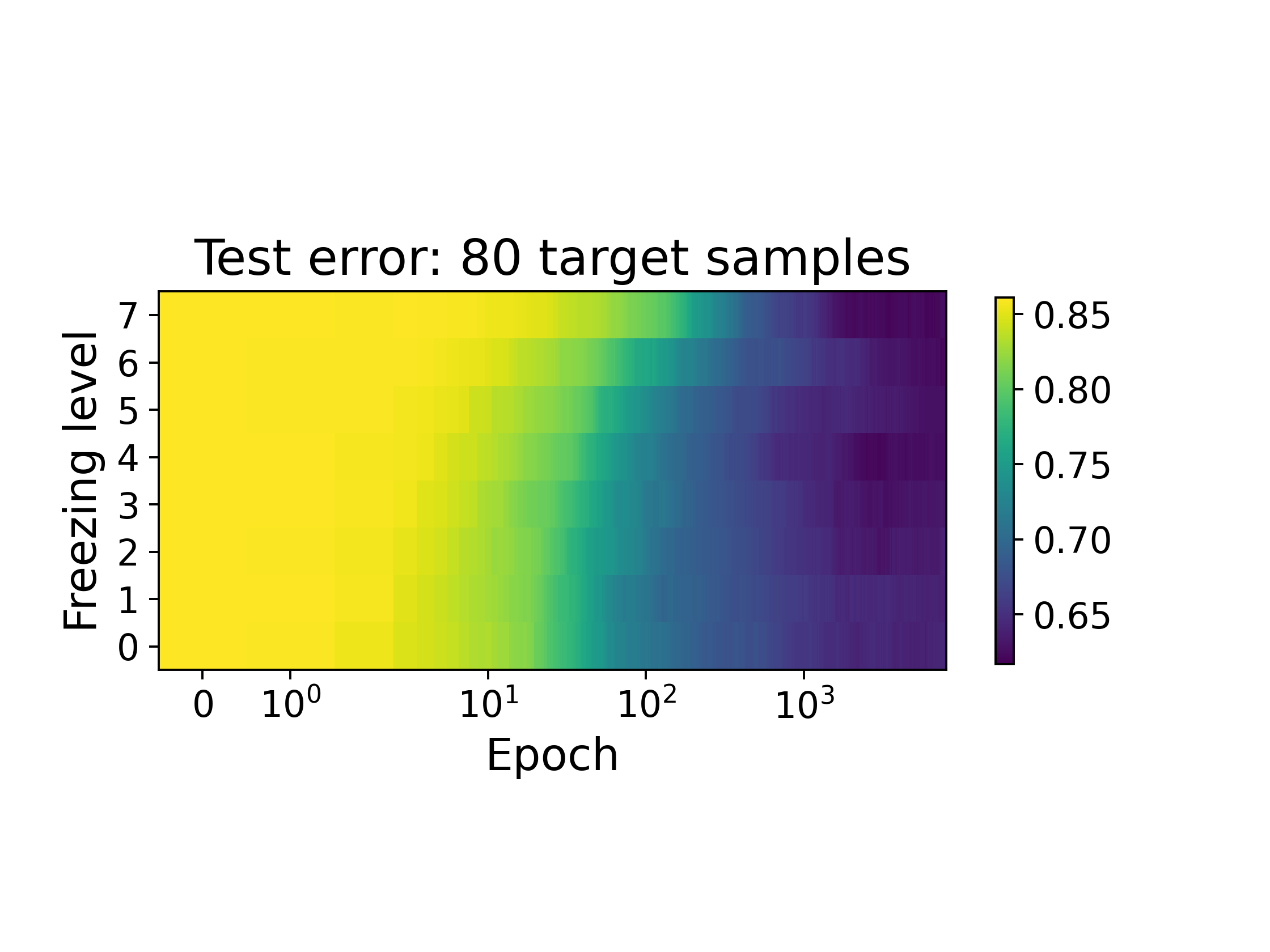}\label{fig:vit_BSP5_src100k_tinyimagenet32_200Classes_src200classtgtCIFAR10class_tgtdataset80_test_err_w} \includegraphics[width=0.48\linewidth]{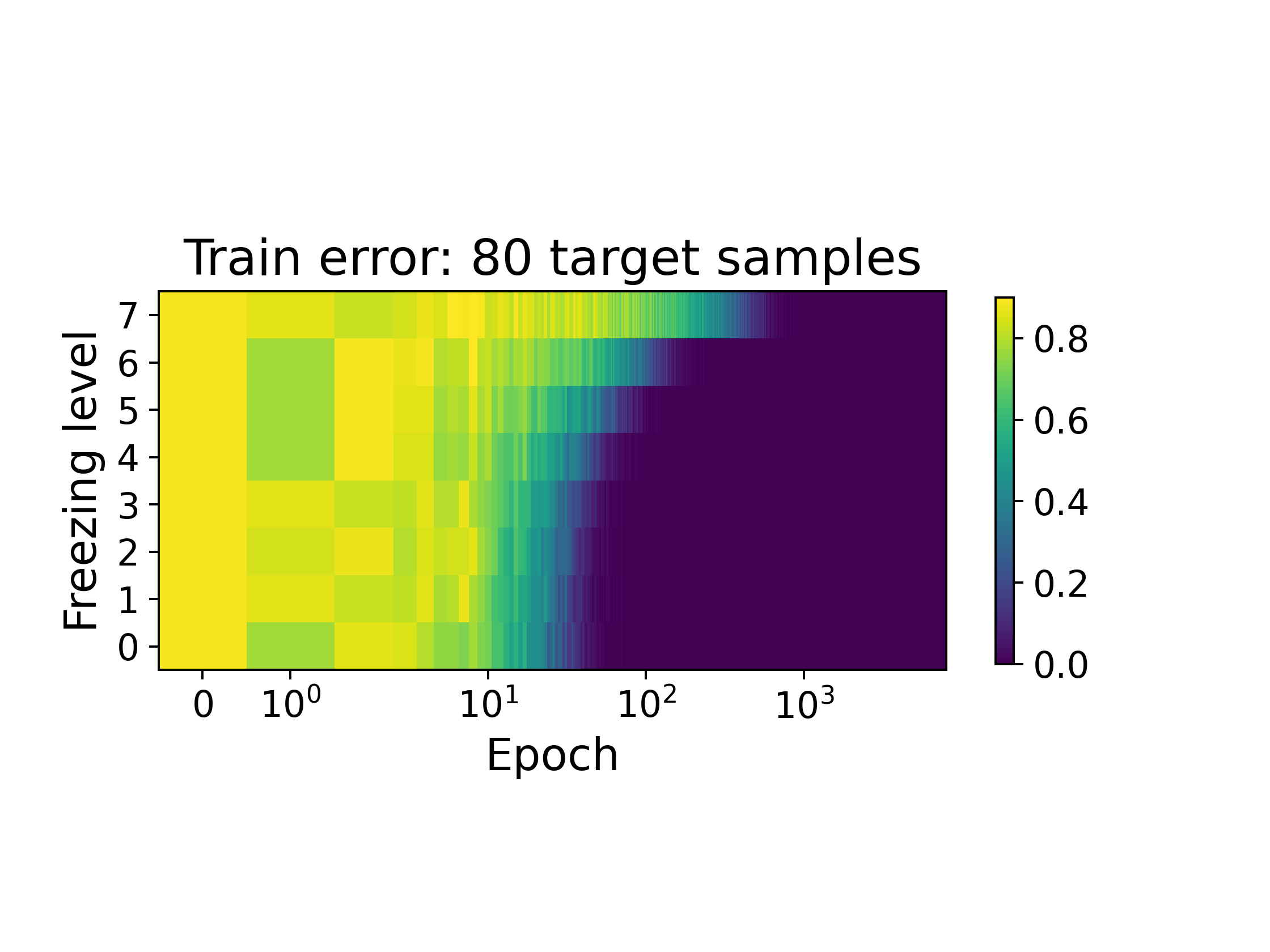}\label{fig:vit_BSP5_src100k_tinyimagenet32_200Classes_src200classtgtCIFAR10class_tgtdataset80_train_err_w}}
  \\[-2ex]
  \subfloat[]{\includegraphics[width=0.48\linewidth]{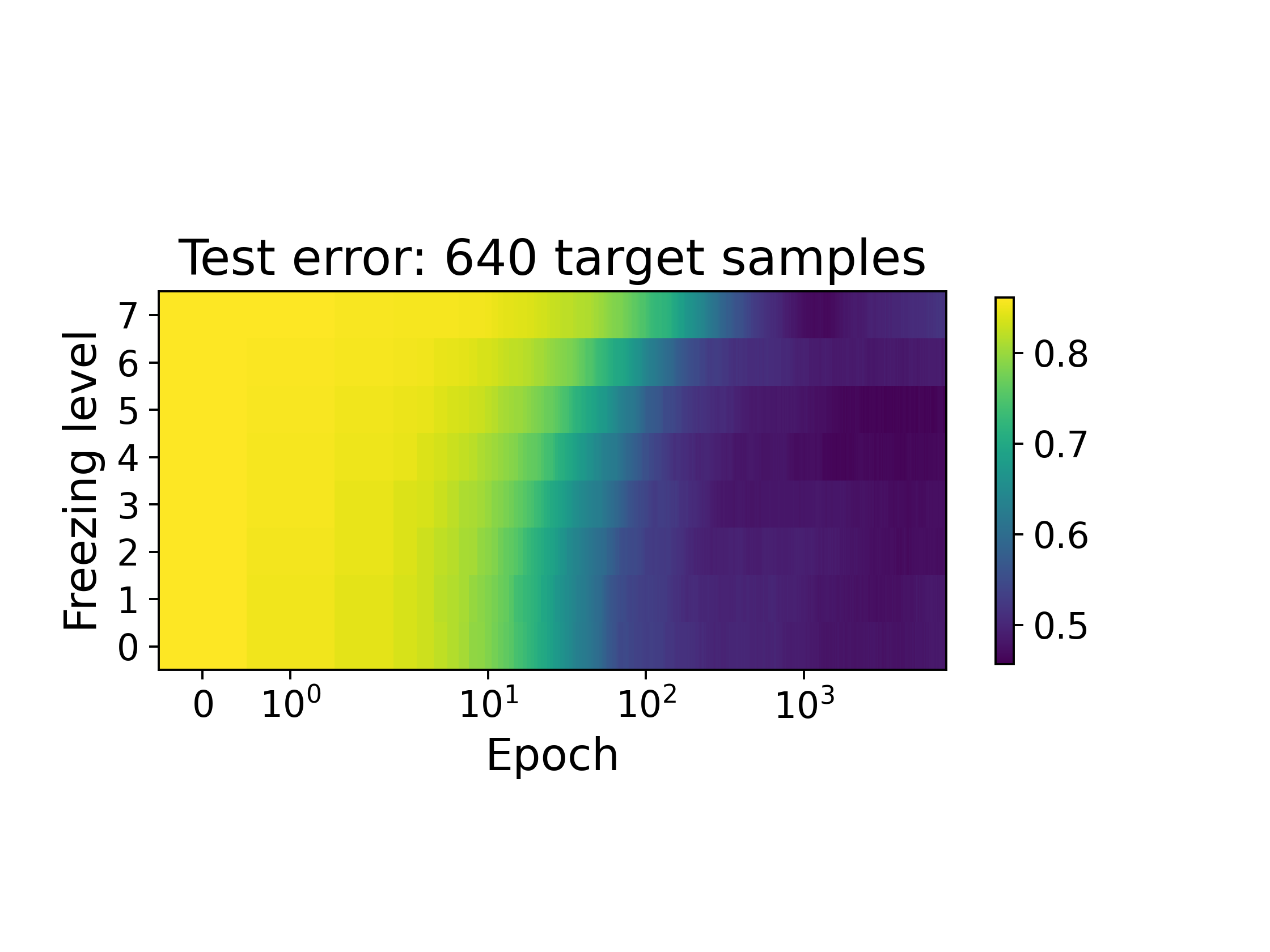}\label{fig:vit_BSP5_src100k_tinyimagenet32_200Classes_src200classtgtCIFAR10class_tgtdataset640_test_err_w} \includegraphics[width=0.48\linewidth]{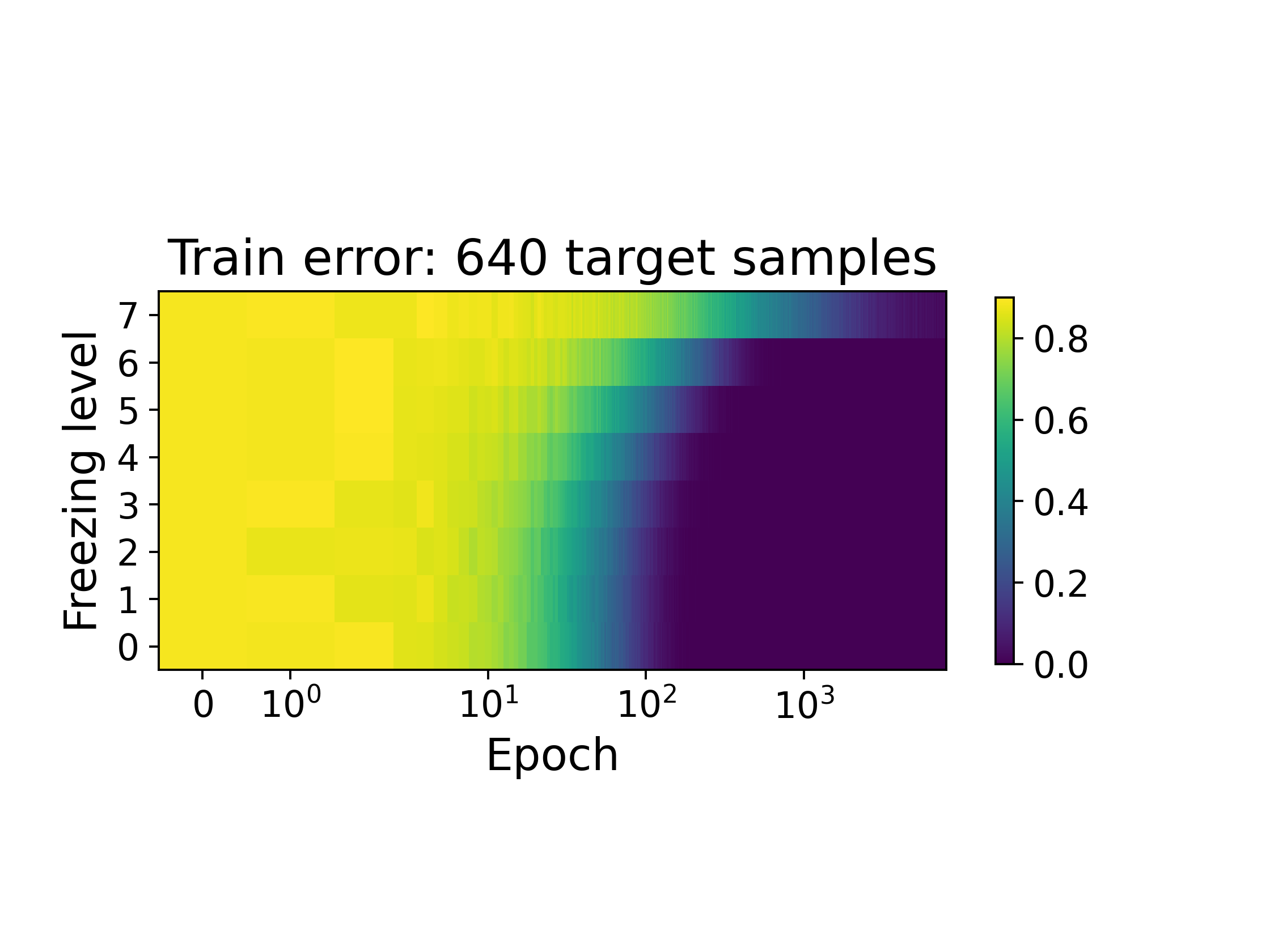}\label{fig:vit_BSP5_src100k_tinyimagenet32_200Classes_src200classtgtCIFAR10class_tgtdataset640_train_err_w}}
  \\[-2ex]
  \subfloat[]{\includegraphics[width=0.48\linewidth]{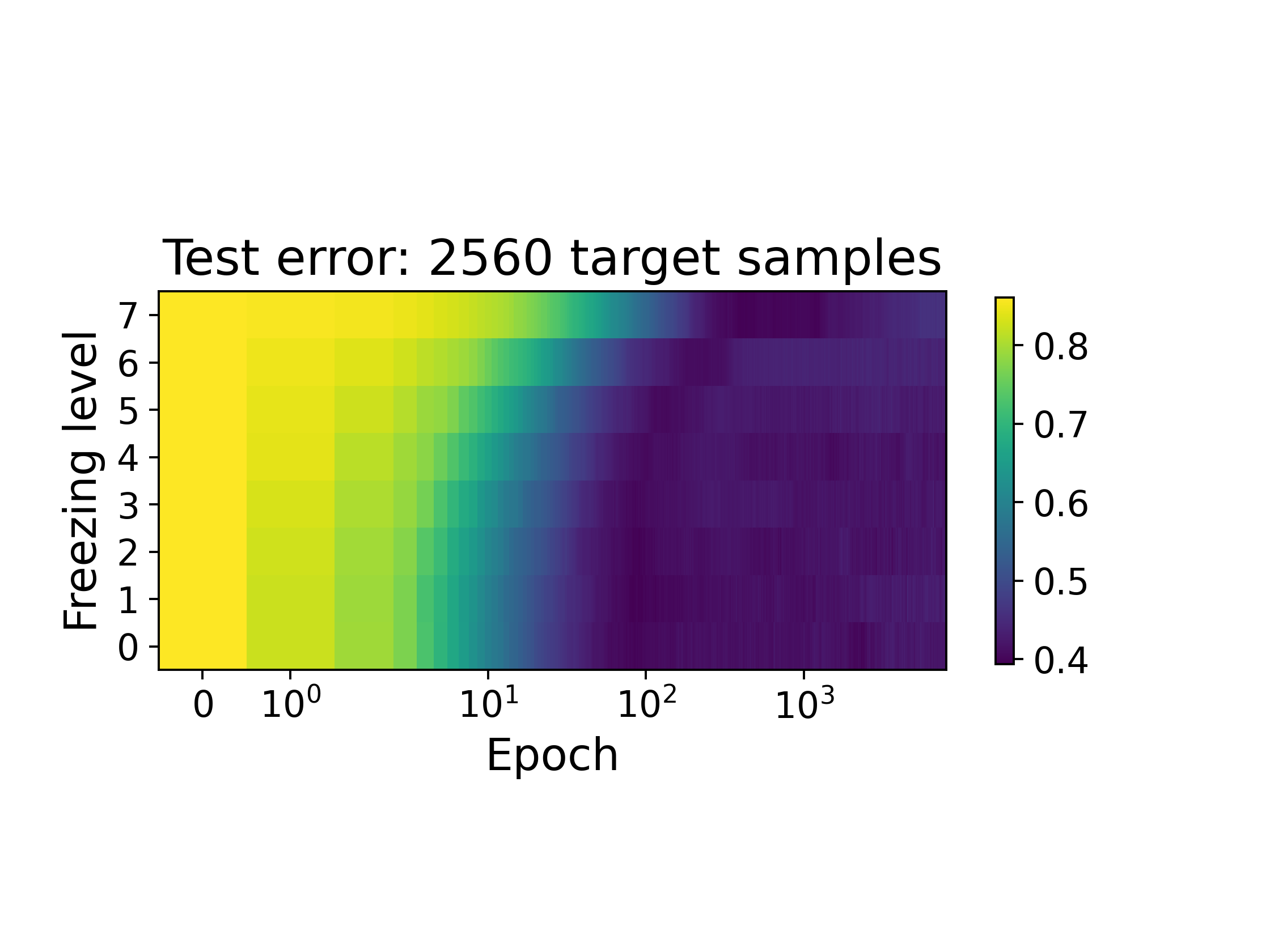}\label{fig:vit_BSP5_src100k_tinyimagenet32_200Classes_src200classtgtCIFAR10class_tgtdataset2560_test_err_w} \includegraphics[width=0.48\linewidth]{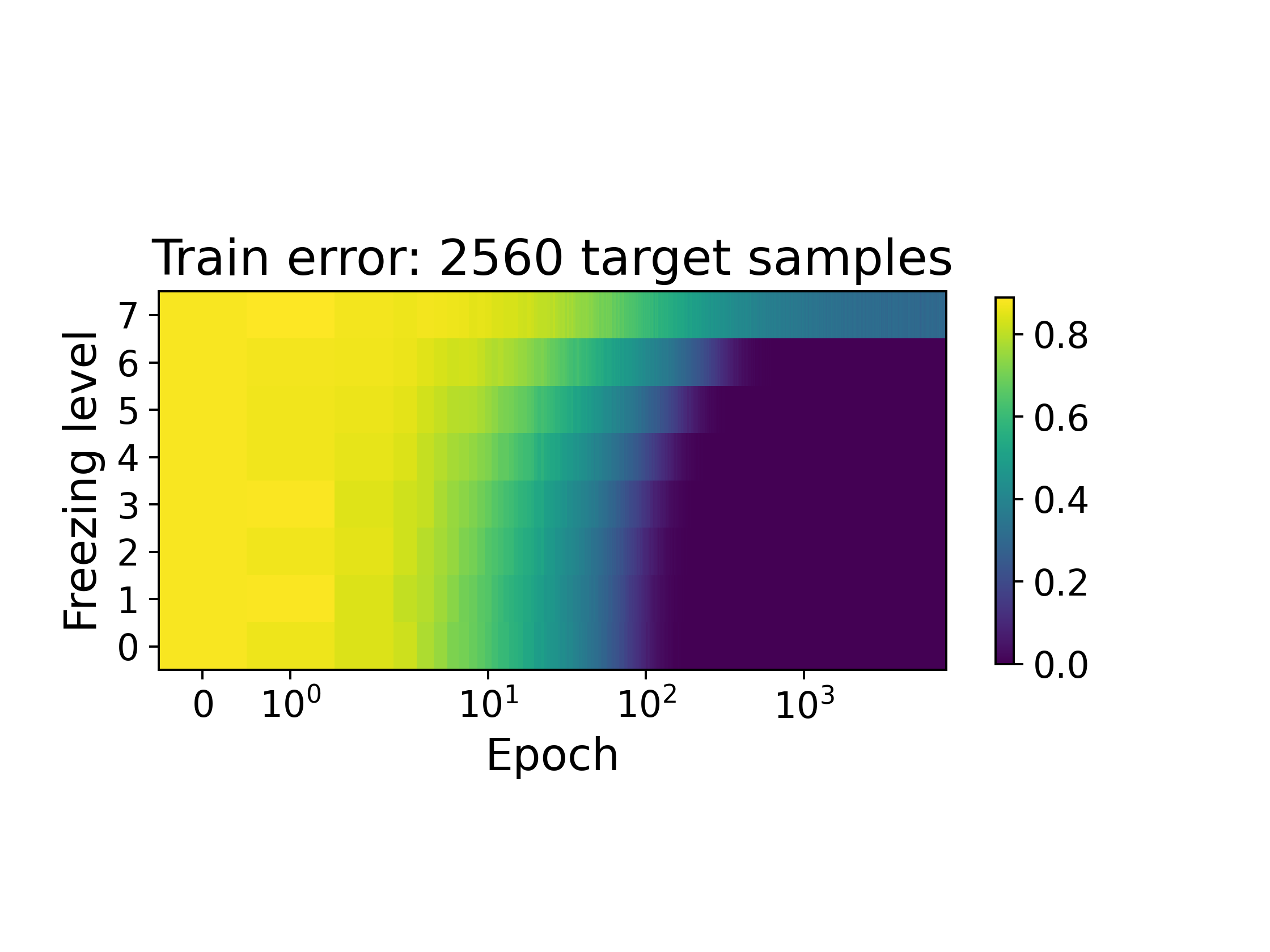}\label{fig:vit_BSP5_src100k_tinyimagenet32_200Classes_src200classtgtCIFAR10class_tgtdataset2560_train_err_w}}
  \\[-2ex]
  \subfloat[]{\includegraphics[width=0.48\linewidth]{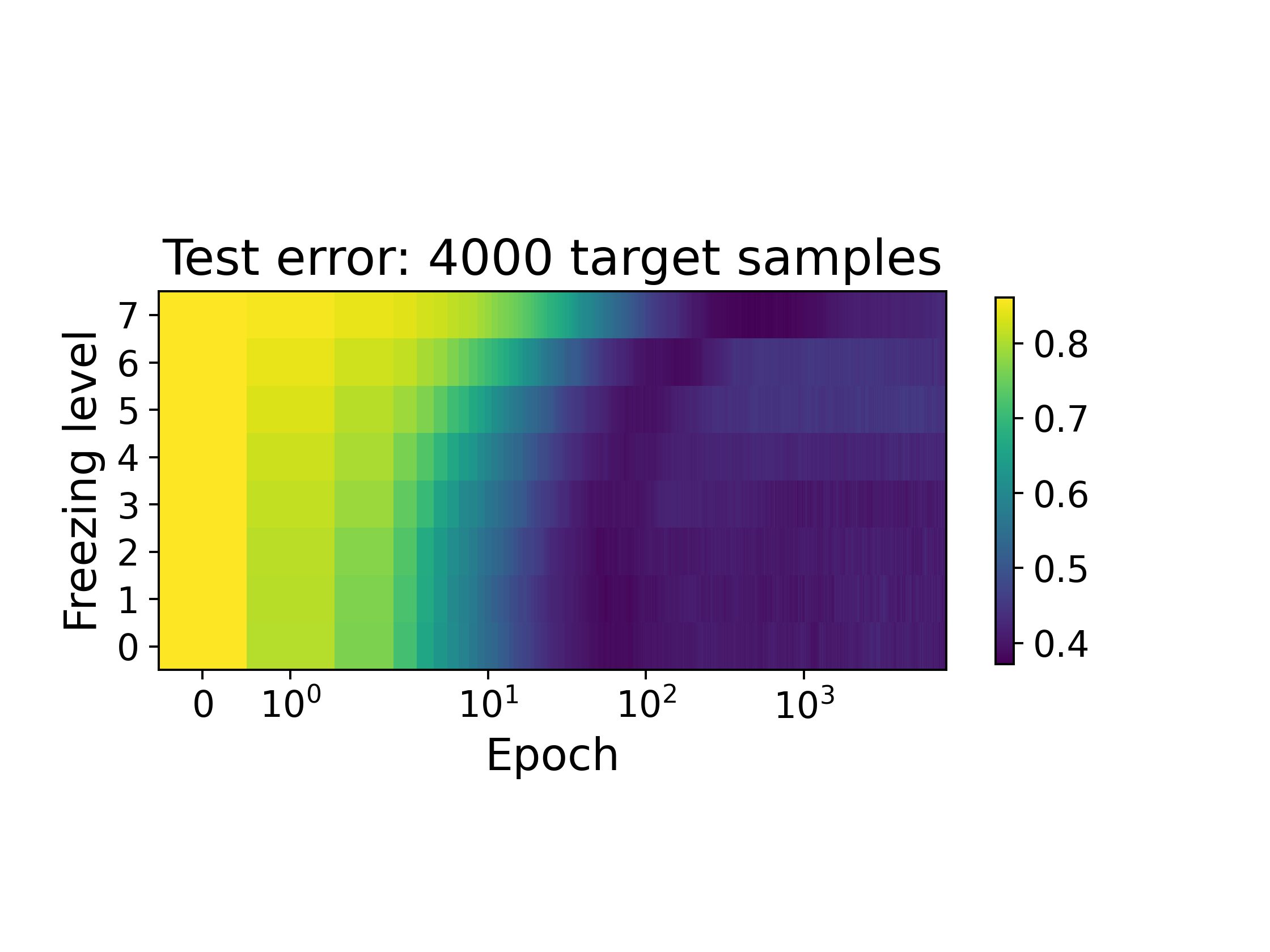}\label{fig:vit_BSP5_src100k_tinyimagenet32_200Classes_src200classtgtCIFAR10class_tgtdataset4000_test_err_w} \includegraphics[width=0.48\linewidth]{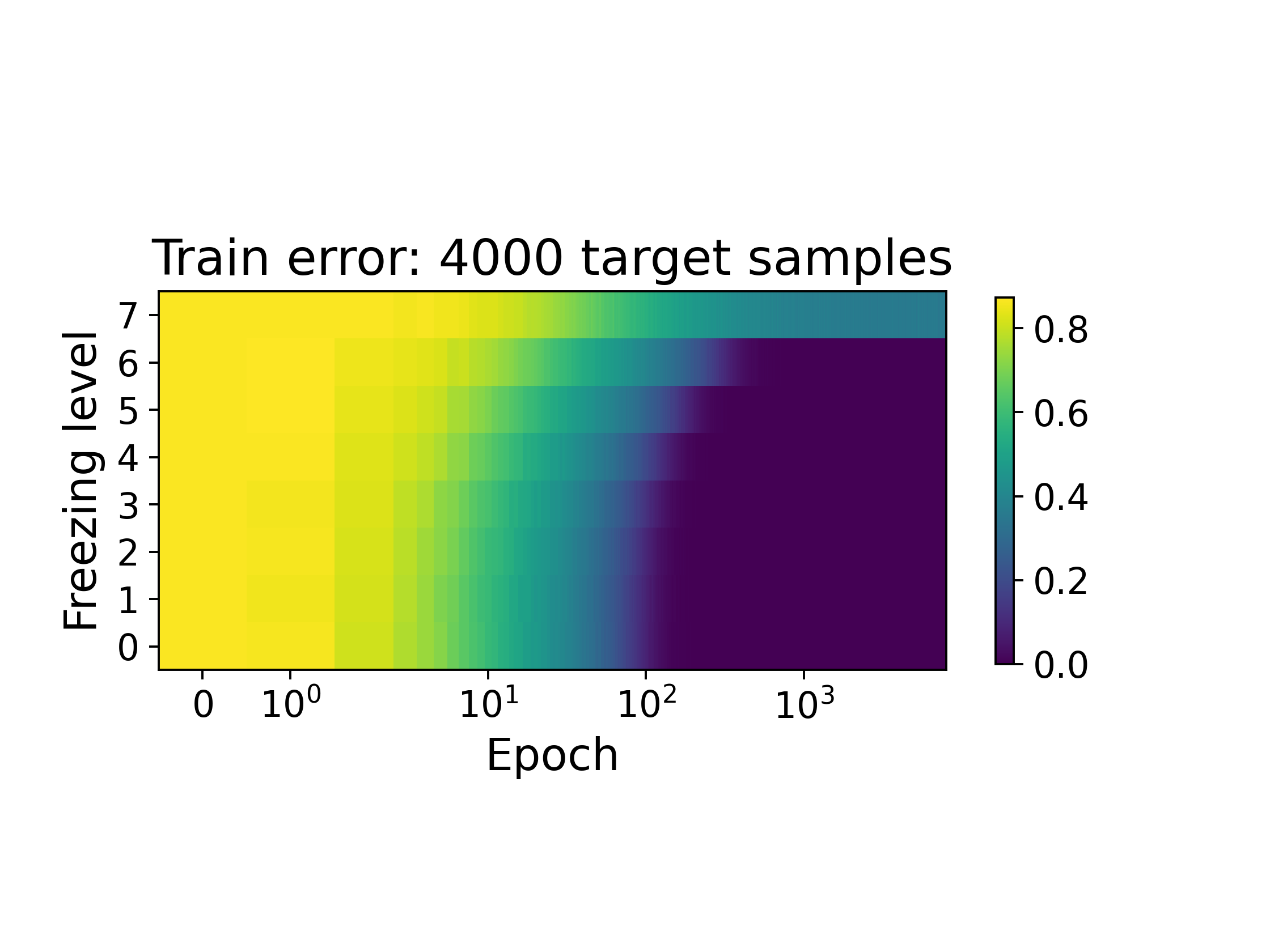}\label{fig:vit_BSP5_src100k_tinyimagenet32_200Classes_src200classtgtCIFAR10class_tgtdataset4000_train_err_w}}
  \caption{Evaluation of transfer learning errors at various freezing levels and training epochs. The target task is \textbf{CIFAR-10} with 20\% label noise in target datasets. The subfigures in each row differ in the target dataset size. The architecture is \textbf{ViT}. The transfer learning are from the \textbf{Tiny ImageNet} source task (200 classes, input image size 32x32x3). }
  \label{fig:ViT_2d_error_diagrams_freezing_vs_epochs_appendix}
\end{figure*}

\begin{figure*}[t]
  \centering
  \subfloat[]{\label{fig:vit_BSP5_src20k_tinyimagenet32_src40SimilarClasstgt40class_tgtdataset160}\includegraphics[width=0.48\linewidth]{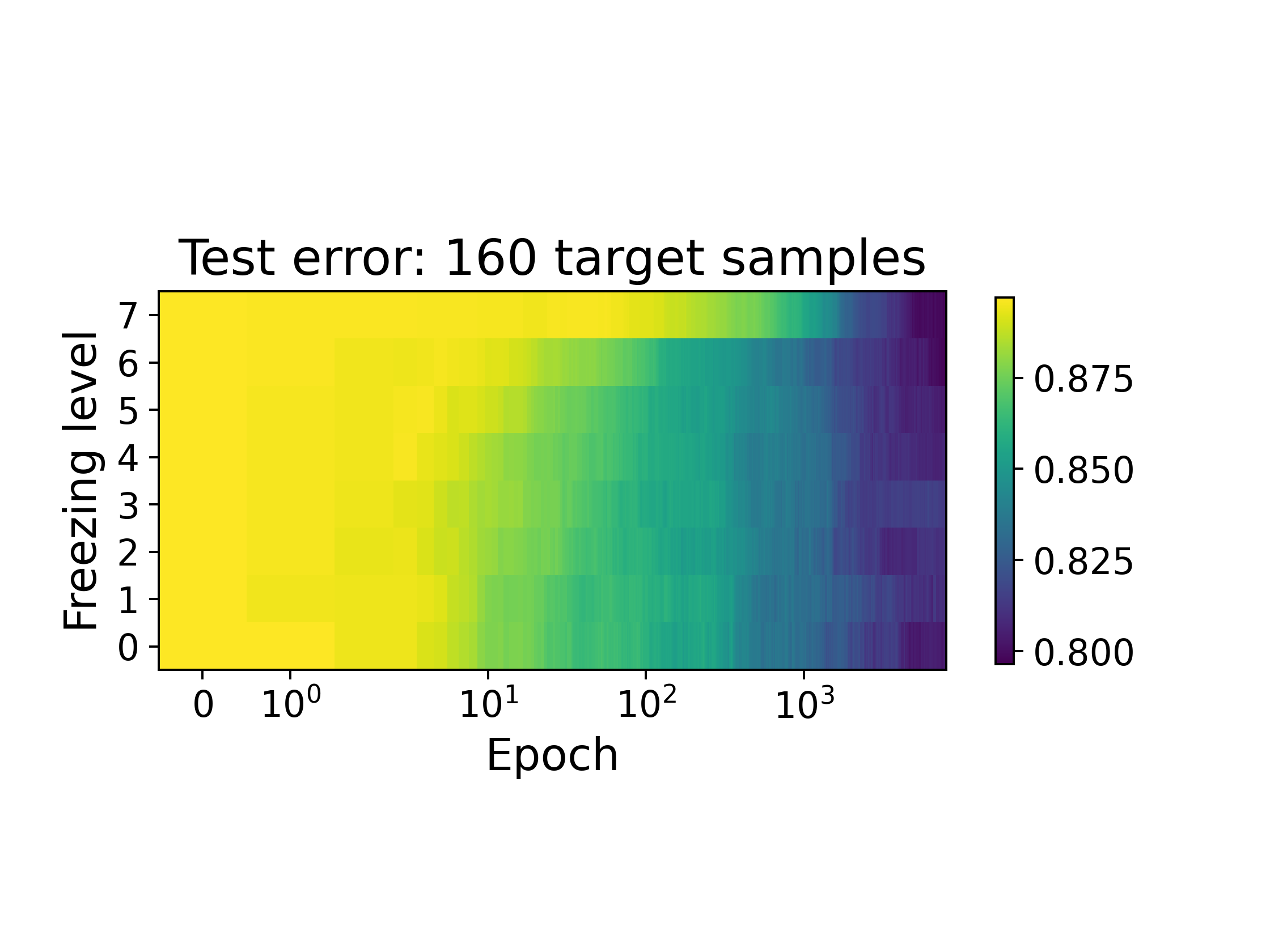} 
  \includegraphics[width=0.48\linewidth]{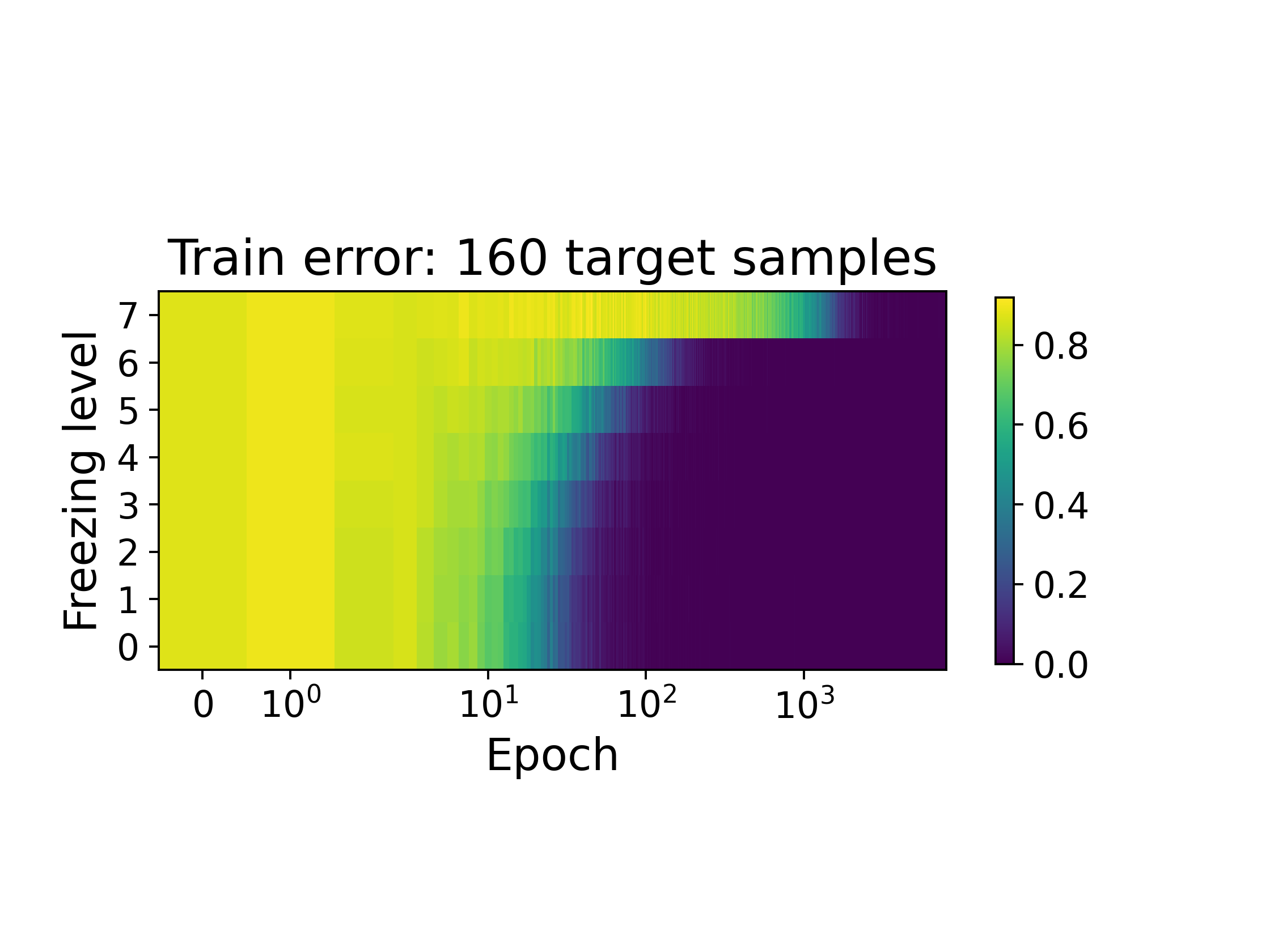}}
  \\[-2ex]
  \subfloat[]{\label{fig:vit_BSP5_src20k_tinyimagenet32_src40SimilarClasstgt40class_tgtdataset640}\includegraphics[width=0.48\linewidth]{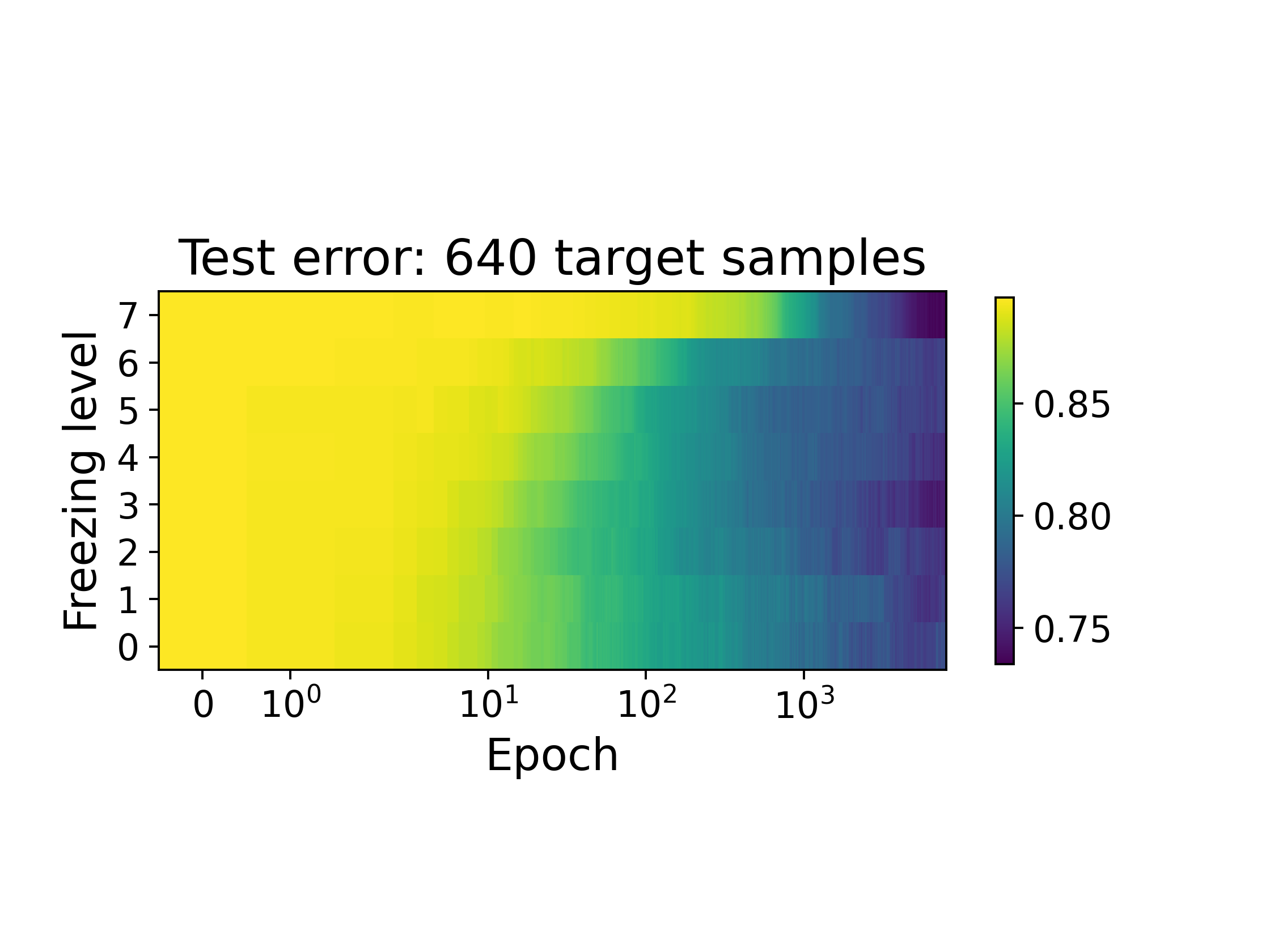} 
  \includegraphics[width=0.48\linewidth]{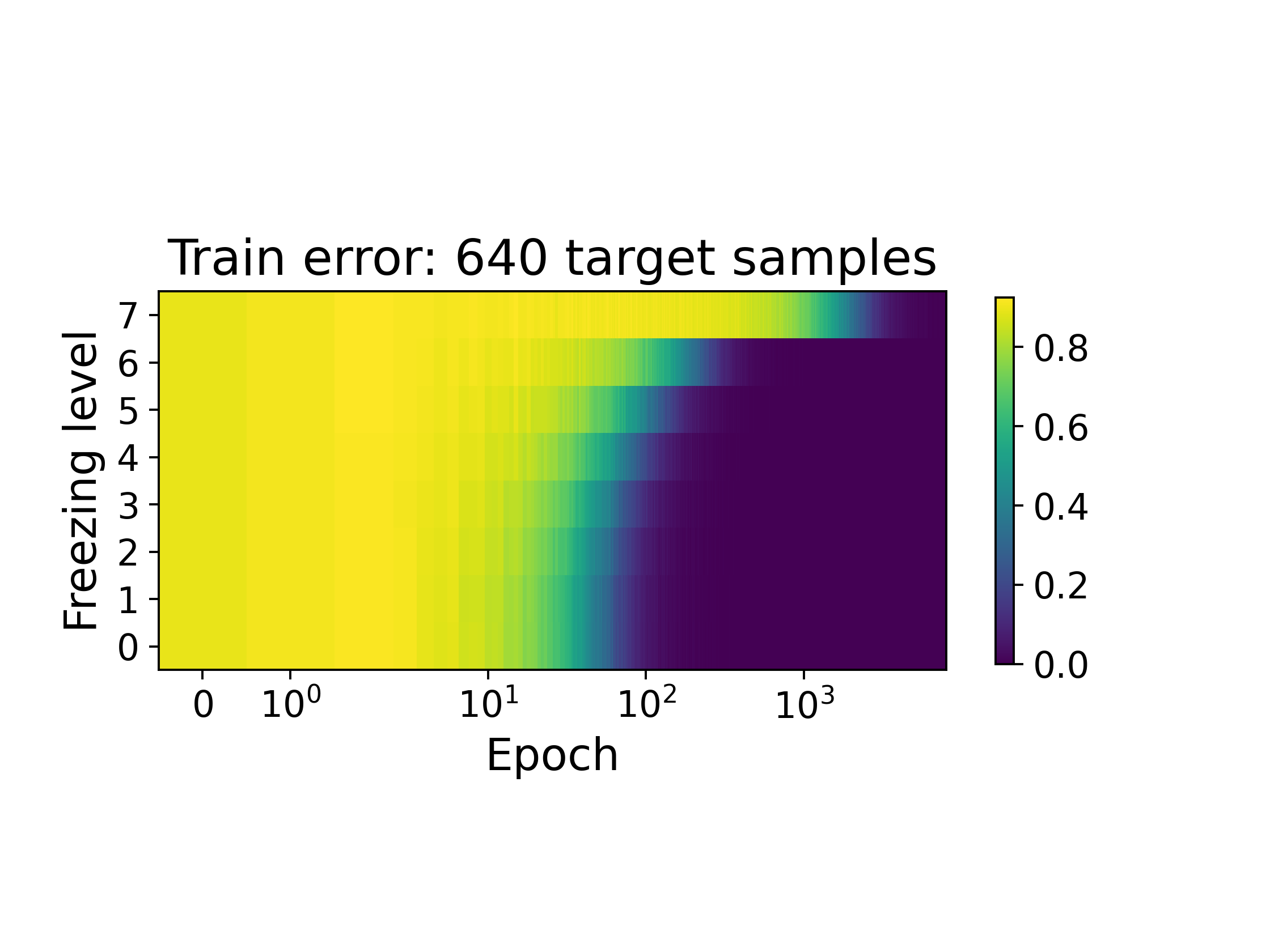}}
  \\[-2ex]
  \subfloat[]{\label{fig:vit_BSP5_src20k_tinyimagenet32_src40SimilarClasstgt40class_tgtdataset2560}\includegraphics[width=0.48\linewidth]{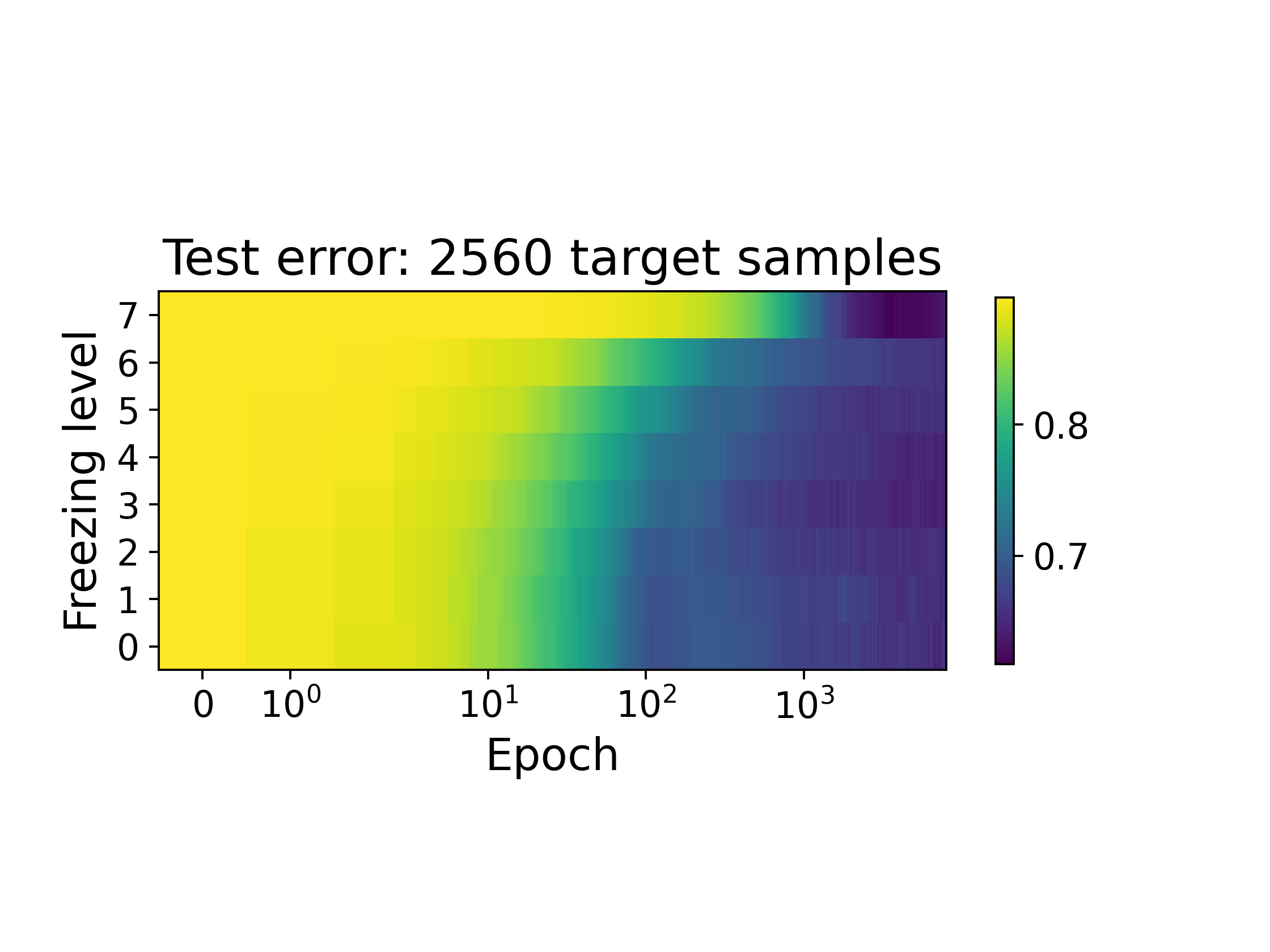} 
  \includegraphics[width=0.48\linewidth]{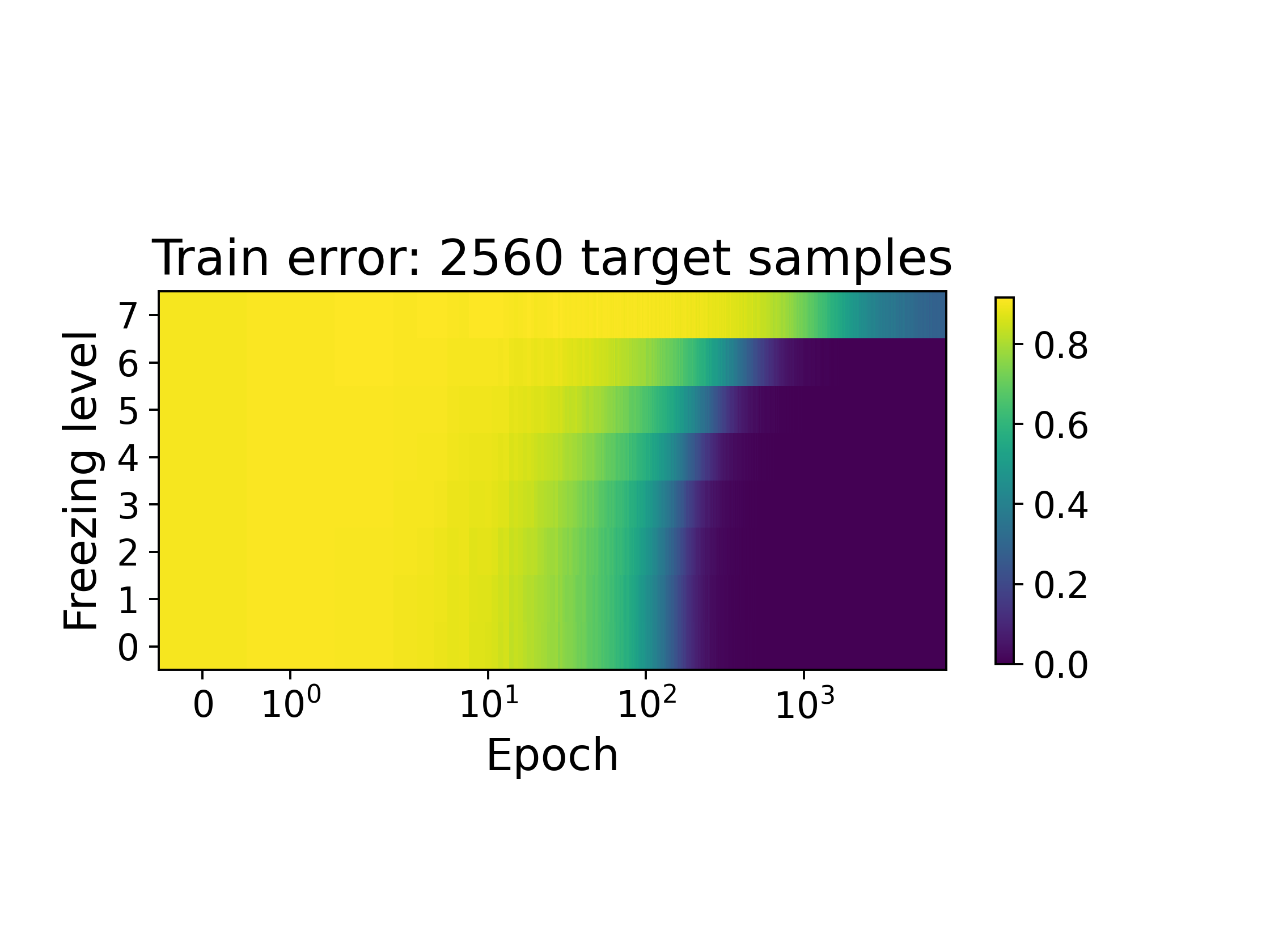}}
  \\[-2ex]
  \subfloat[]{\label{fig:vit_BSP5_src20k_tinyimagenet32_src40SimilarClasstgt40class_tgtdataset4000}\includegraphics[width=0.48\linewidth]{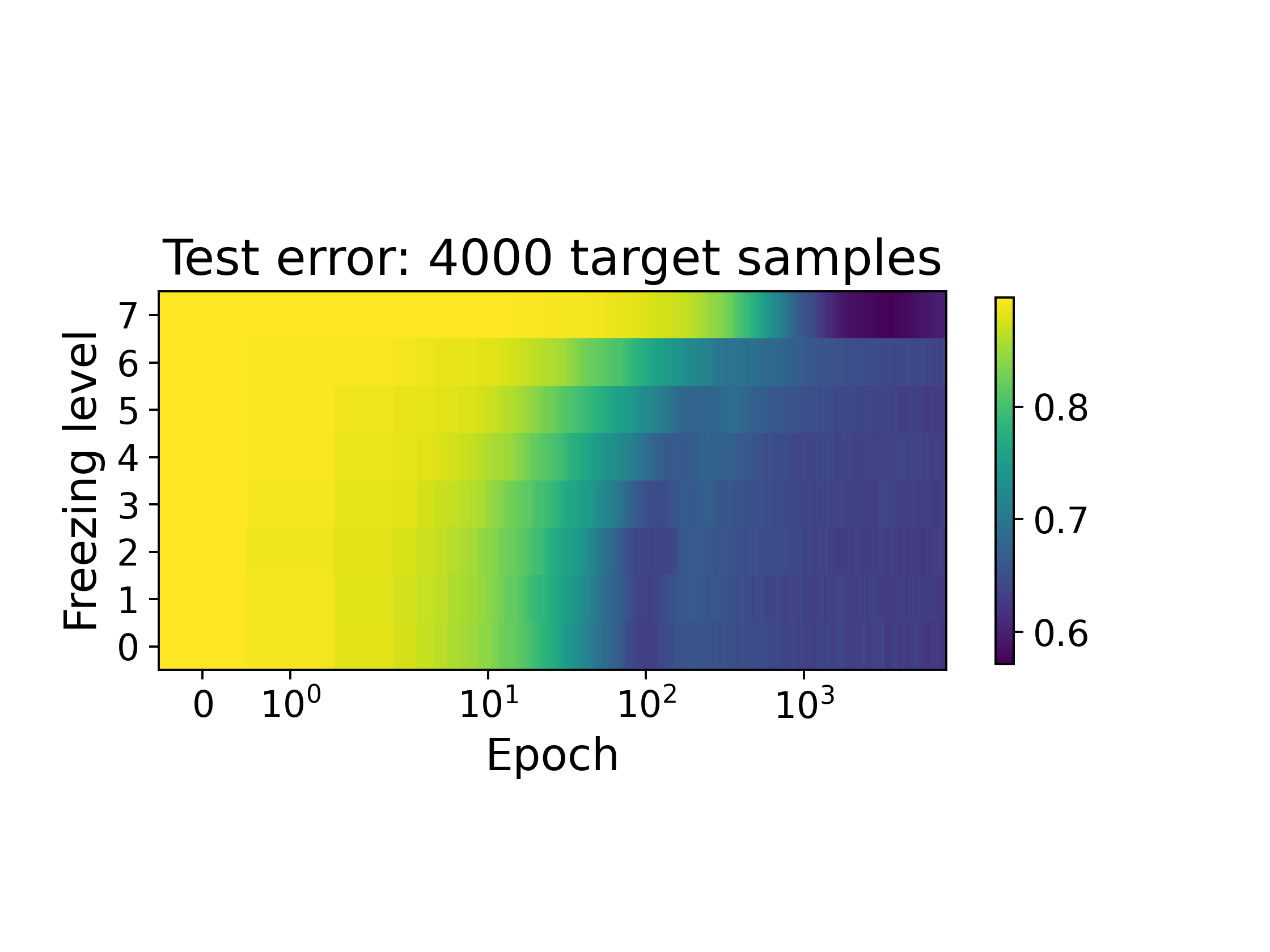} 
  \includegraphics[width=0.48\linewidth]{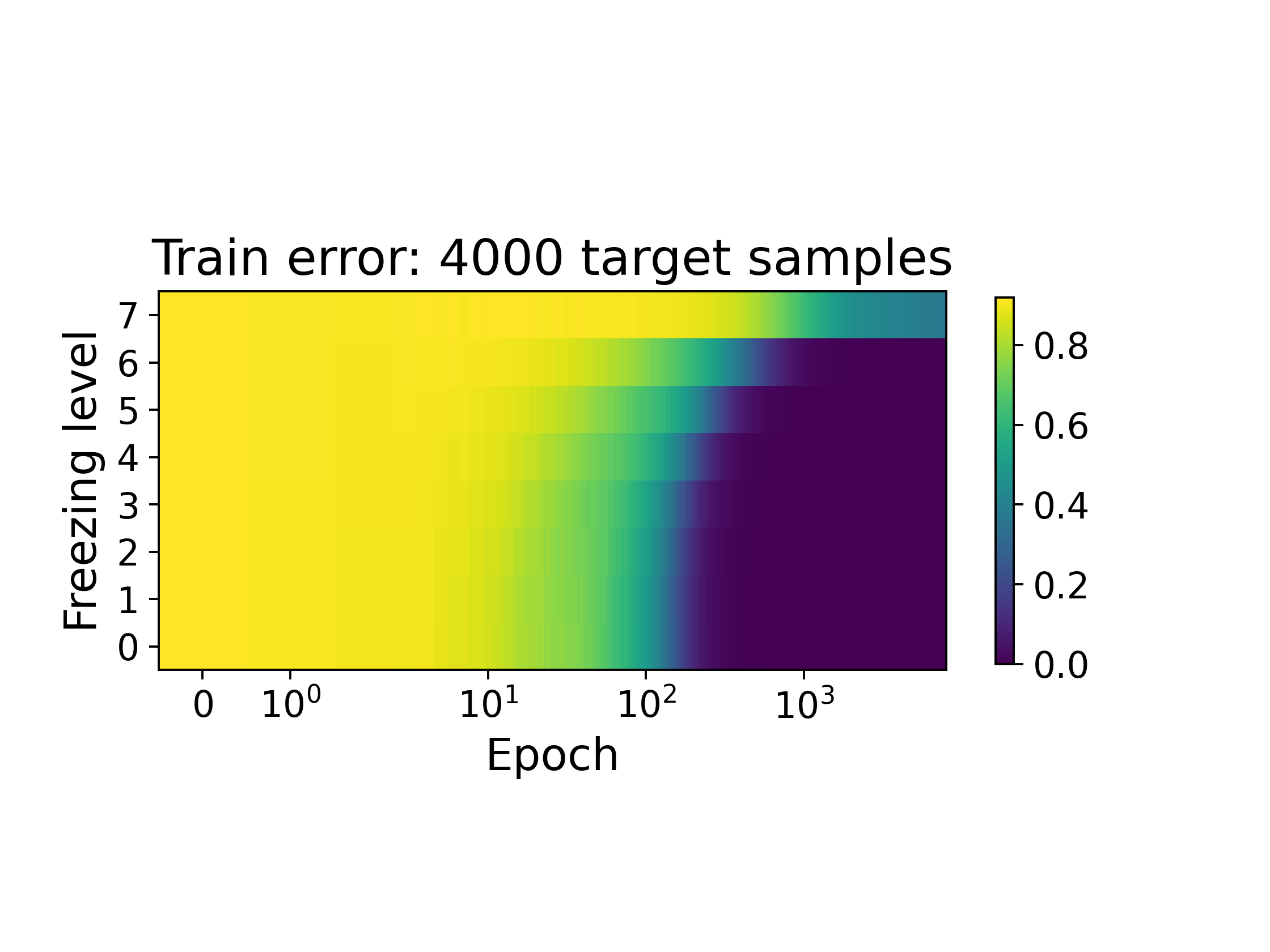}}  
  \caption{Evaluation of transfer learning errors at various freezing levels and training epochs. The target classification task is of \textbf{40 classes from CIFAR-100} with 20\% label noise in target datasets. The subfigures in each row differ in the target dataset size. The architecture is \textbf{ViT}. The transfer learning  is from the source task of \textbf{40 Tiny ImageNet classes} that are similar to the target task classes (input image size 32x32x3) with source dataset of 20k training samples. }
  \label{appendix:fig:ViT_2d_error_diagrams_freezing_vs_epochs_40_class}
\end{figure*}

\begin{figure*}[t]
  \centering
  \subfloat[]{\label{fig:densenet_BSP5_src20k_tinyimagenet32_src40SimilarClasstgt40class_tgtdataset160}\includegraphics[width=0.48\linewidth]{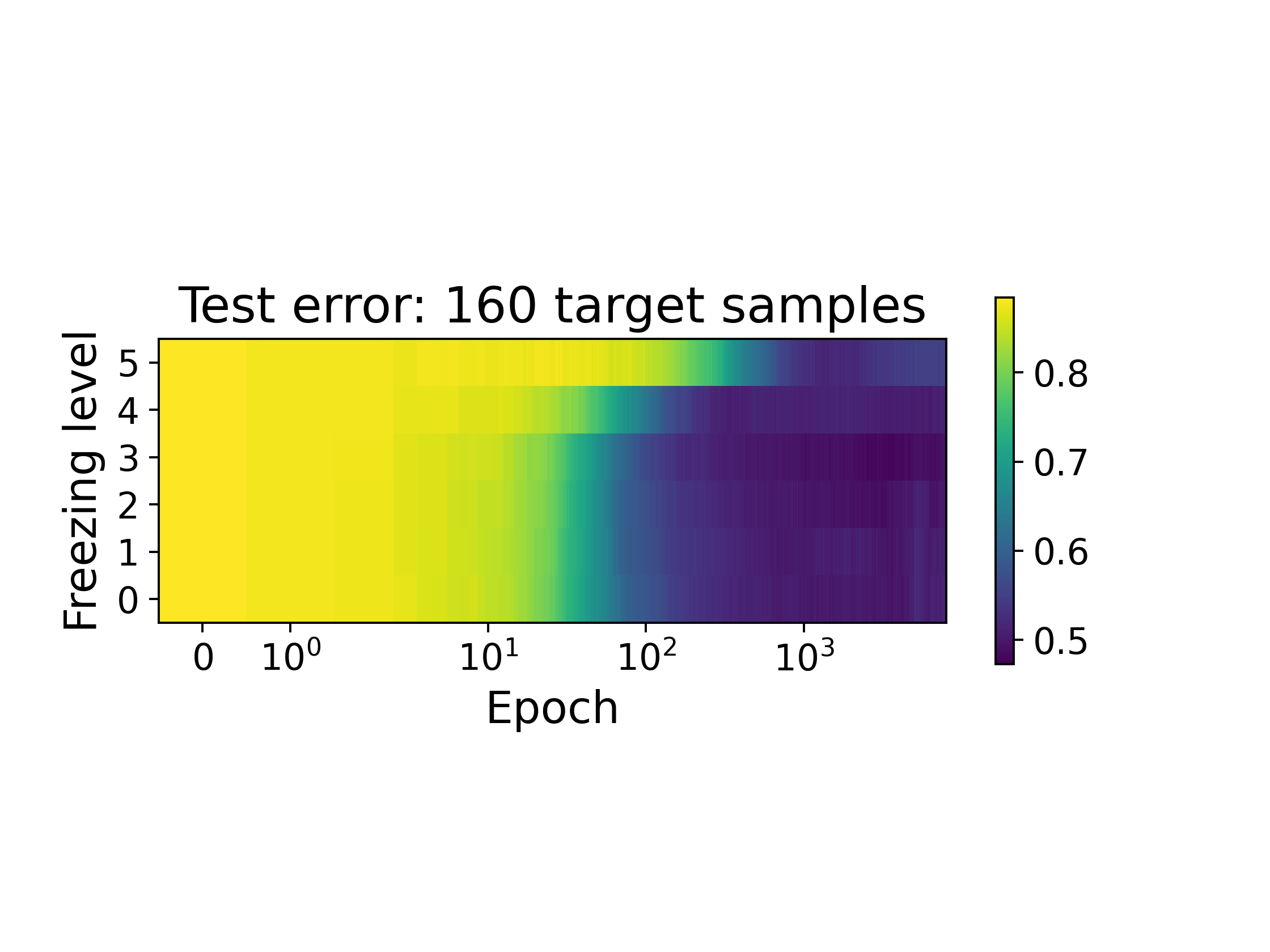} 
  \includegraphics[width=0.48\linewidth]{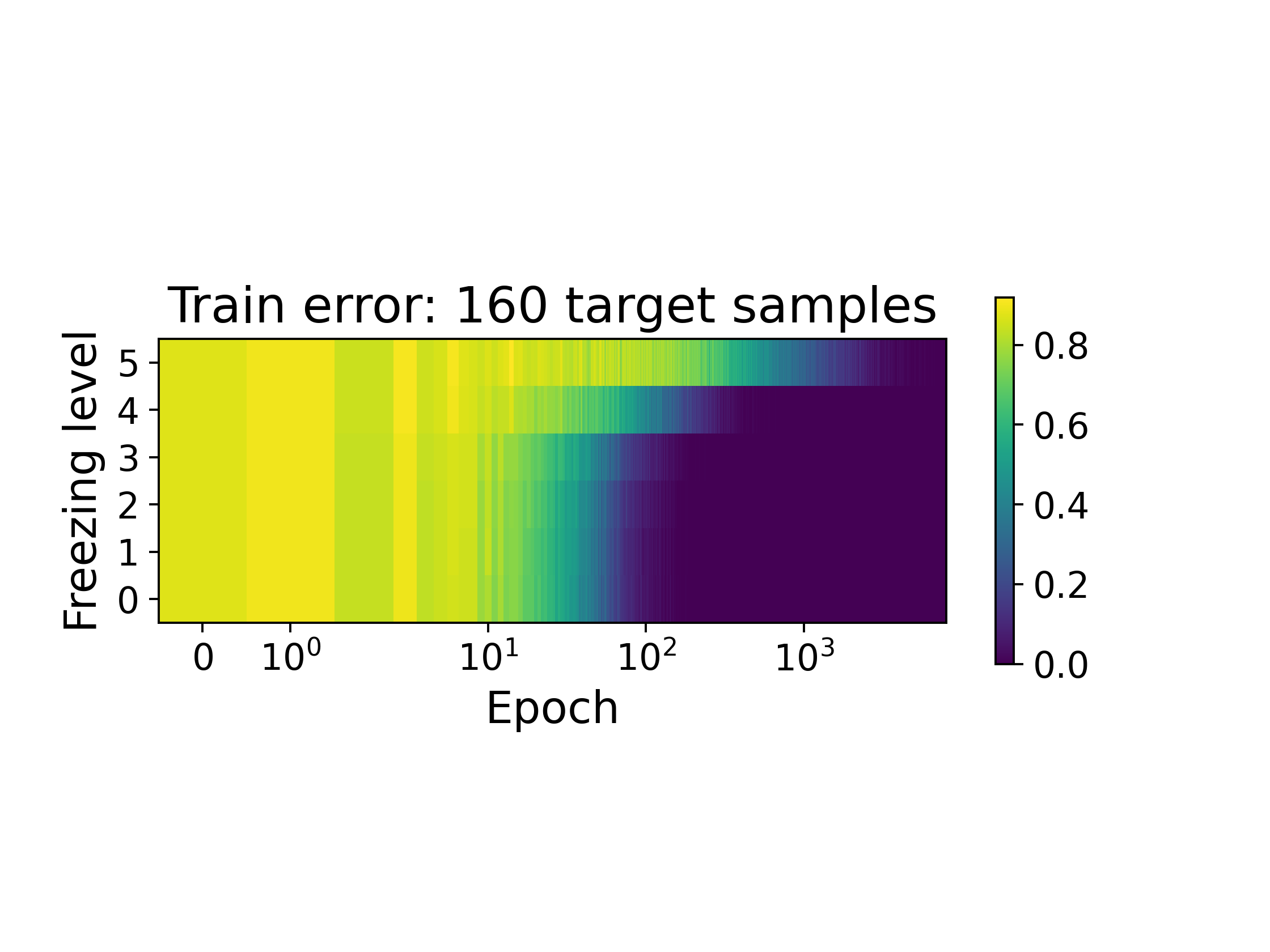}}
  \\[-2ex]
  \subfloat[]{\label{fig:densenet_BSP5_src20k_tinyimagenet32_src40SimilarClasstgt40class_tgtdataset640}\includegraphics[width=0.48\linewidth]{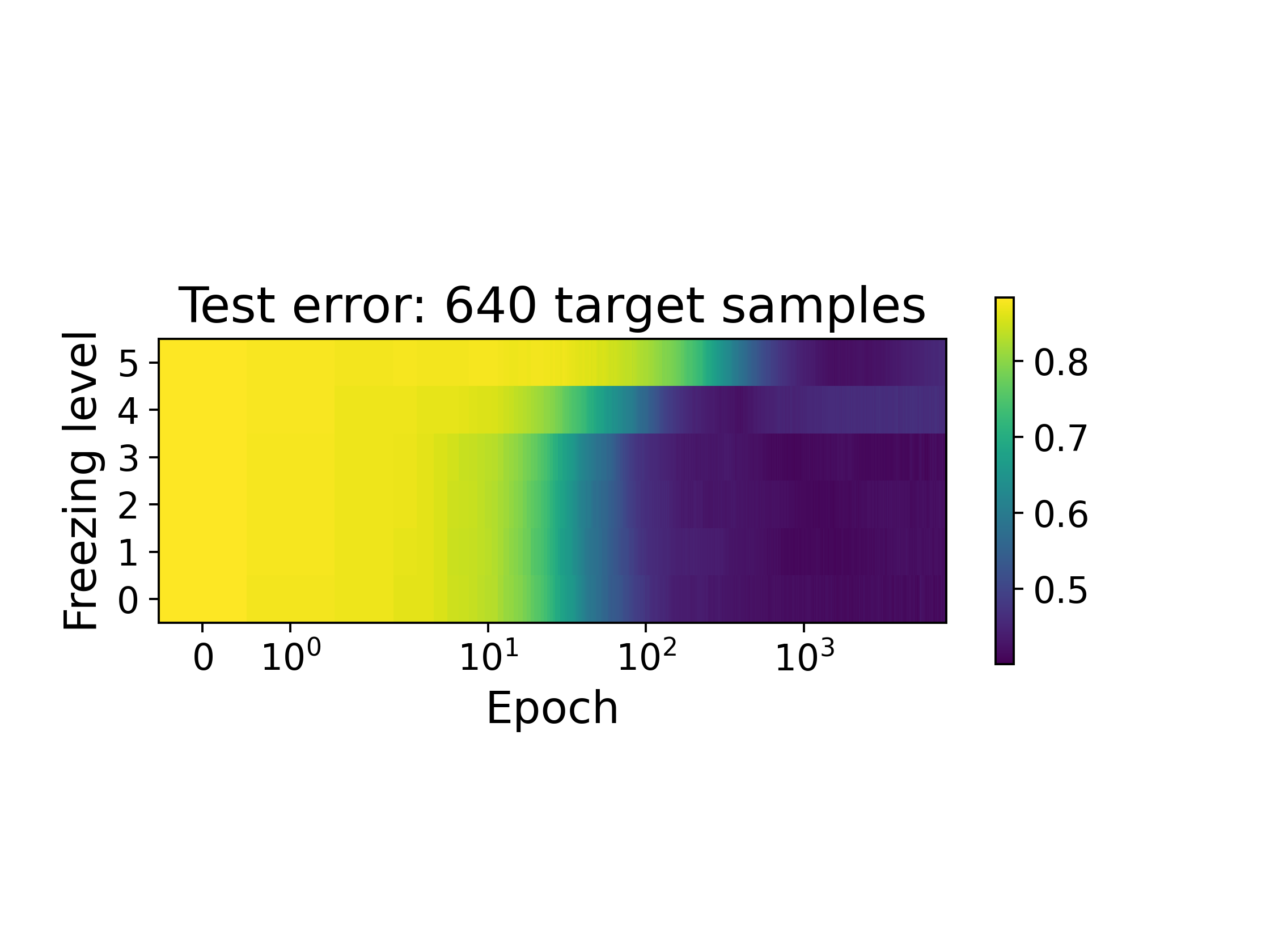} 
  \includegraphics[width=0.48\linewidth]{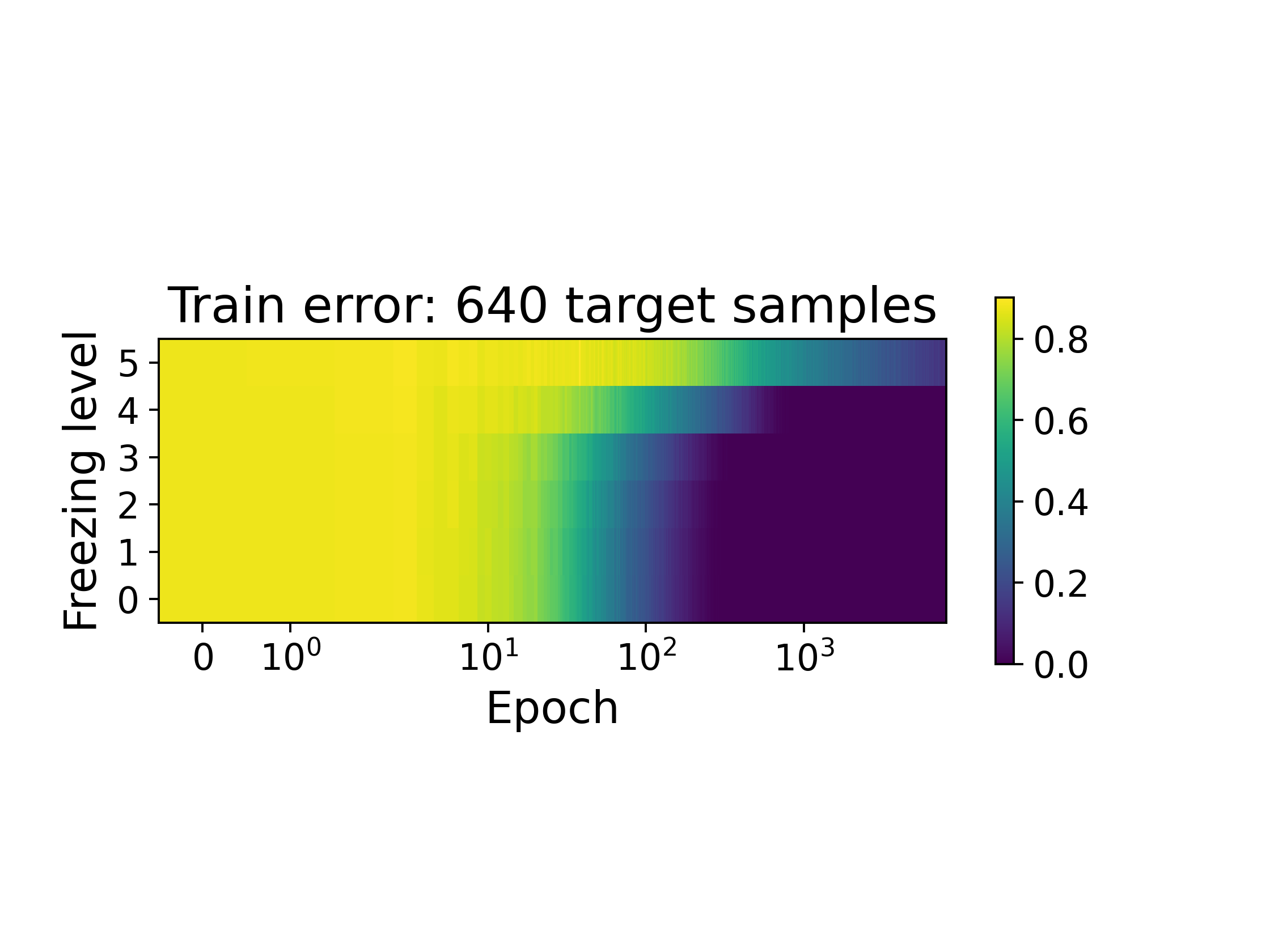}}
  \\[-2ex]
  \subfloat[]{\label{fig:densenet_BSP5_src20k_tinyimagenet32_src40SimilarClasstgt40class_tgtdataset2560}\includegraphics[width=0.48\linewidth]{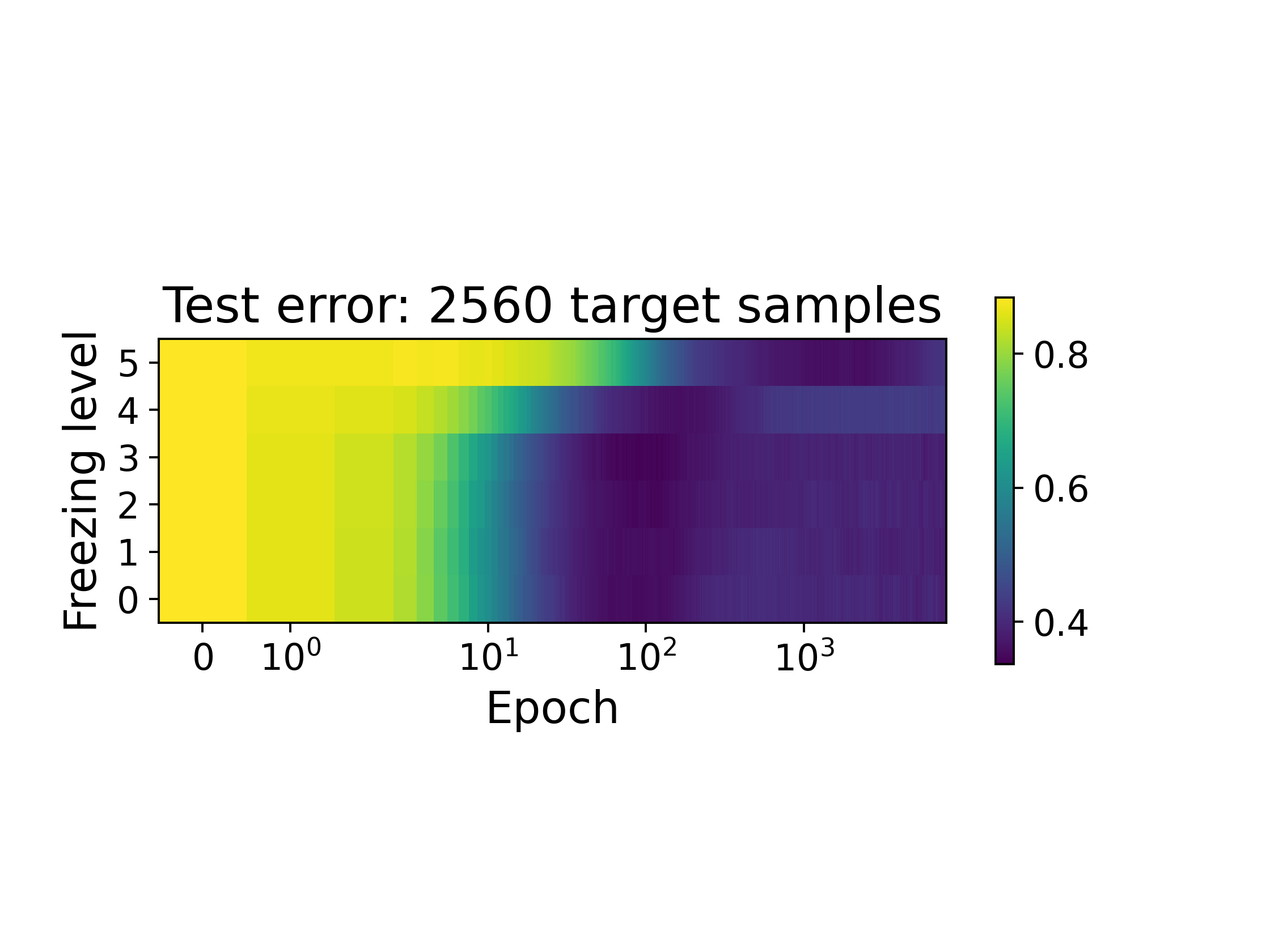} 
  \includegraphics[width=0.48\linewidth]{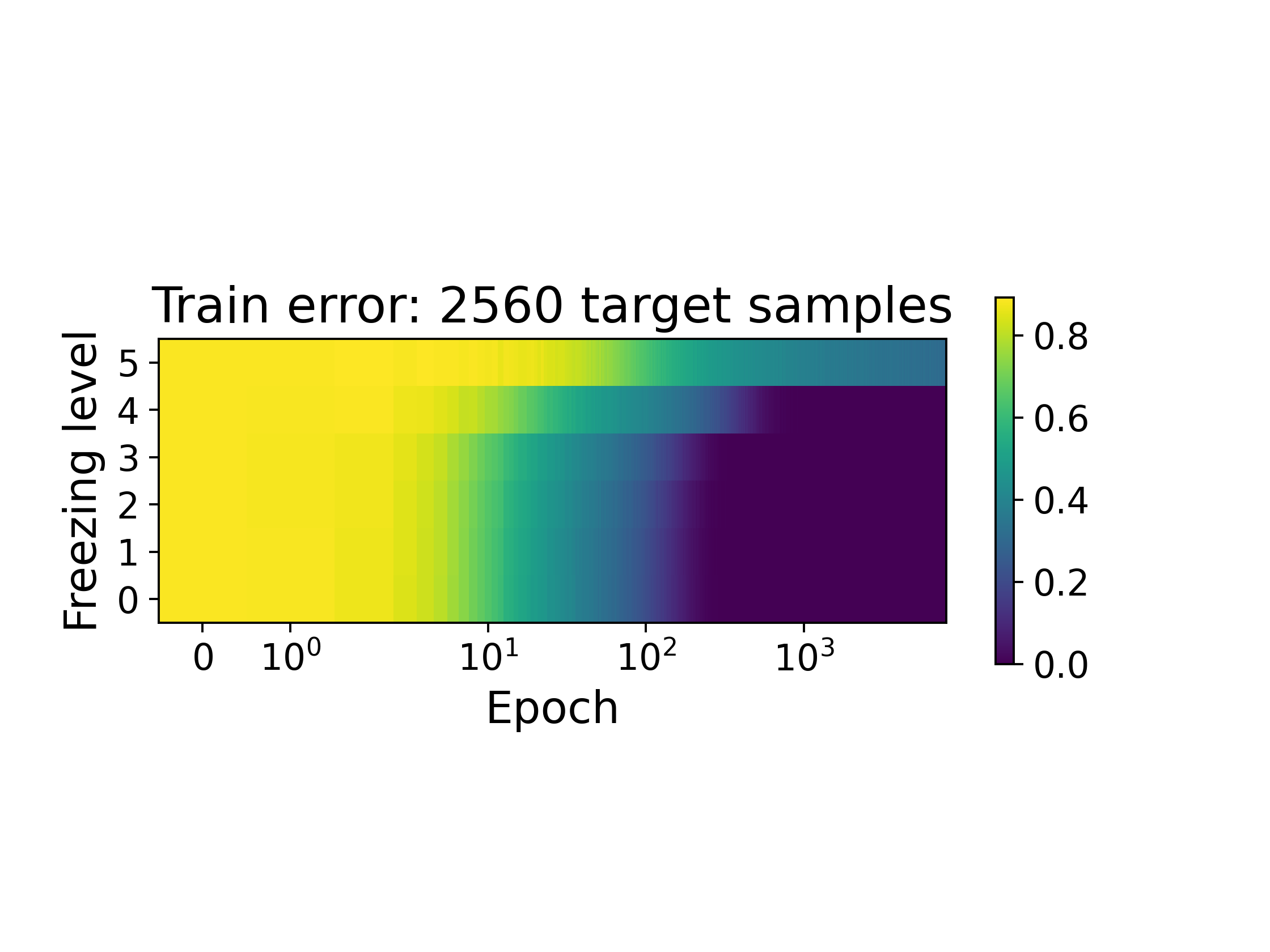}}
  \\[-2ex]
  \subfloat[]{\label{fig:densenet_BSP5_src20k_tinyimagenet32_src40SimilarClasstgt40class_tgtdataset4000}\includegraphics[width=0.48\linewidth]{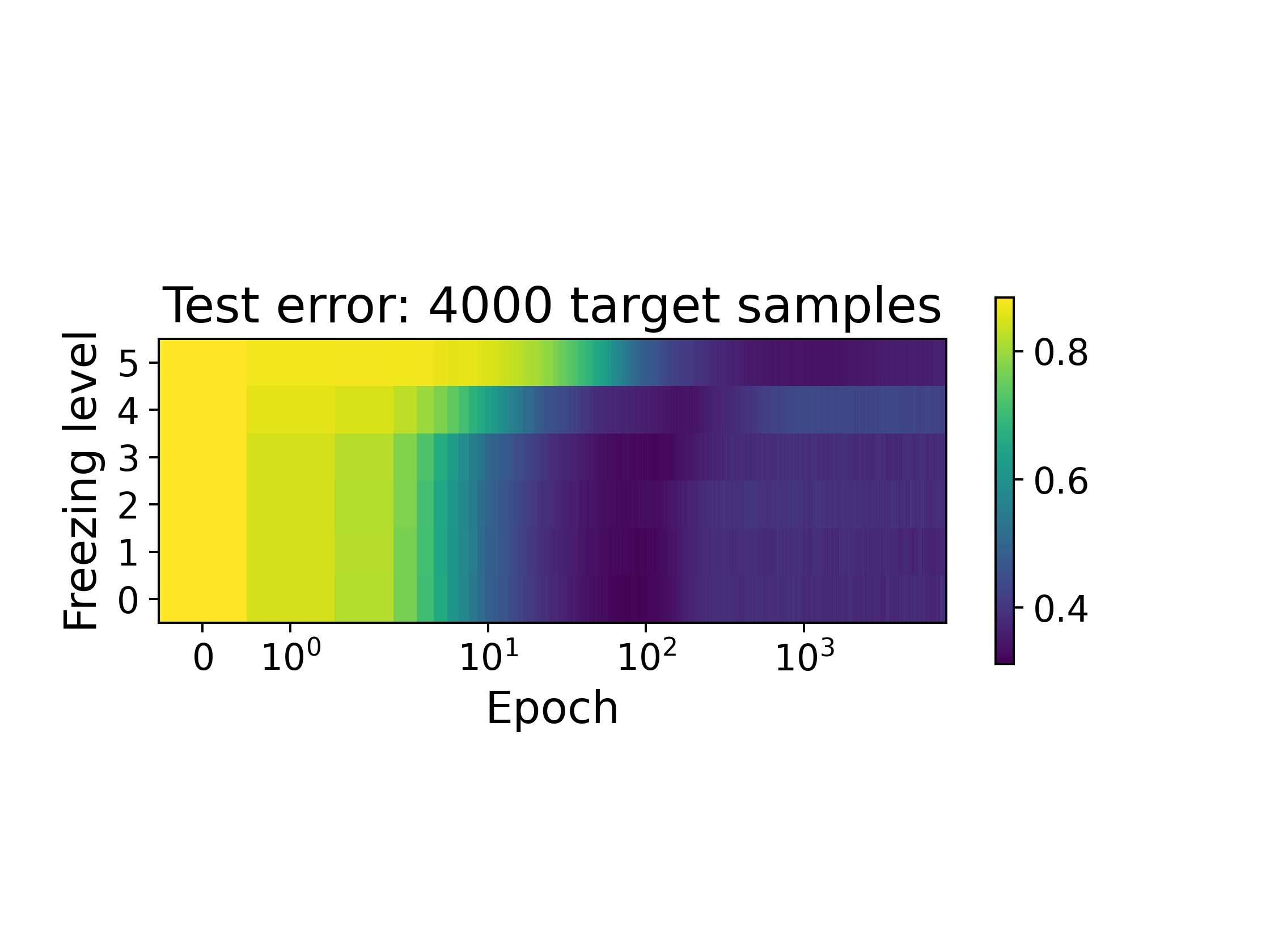} 
  \includegraphics[width=0.48\linewidth]{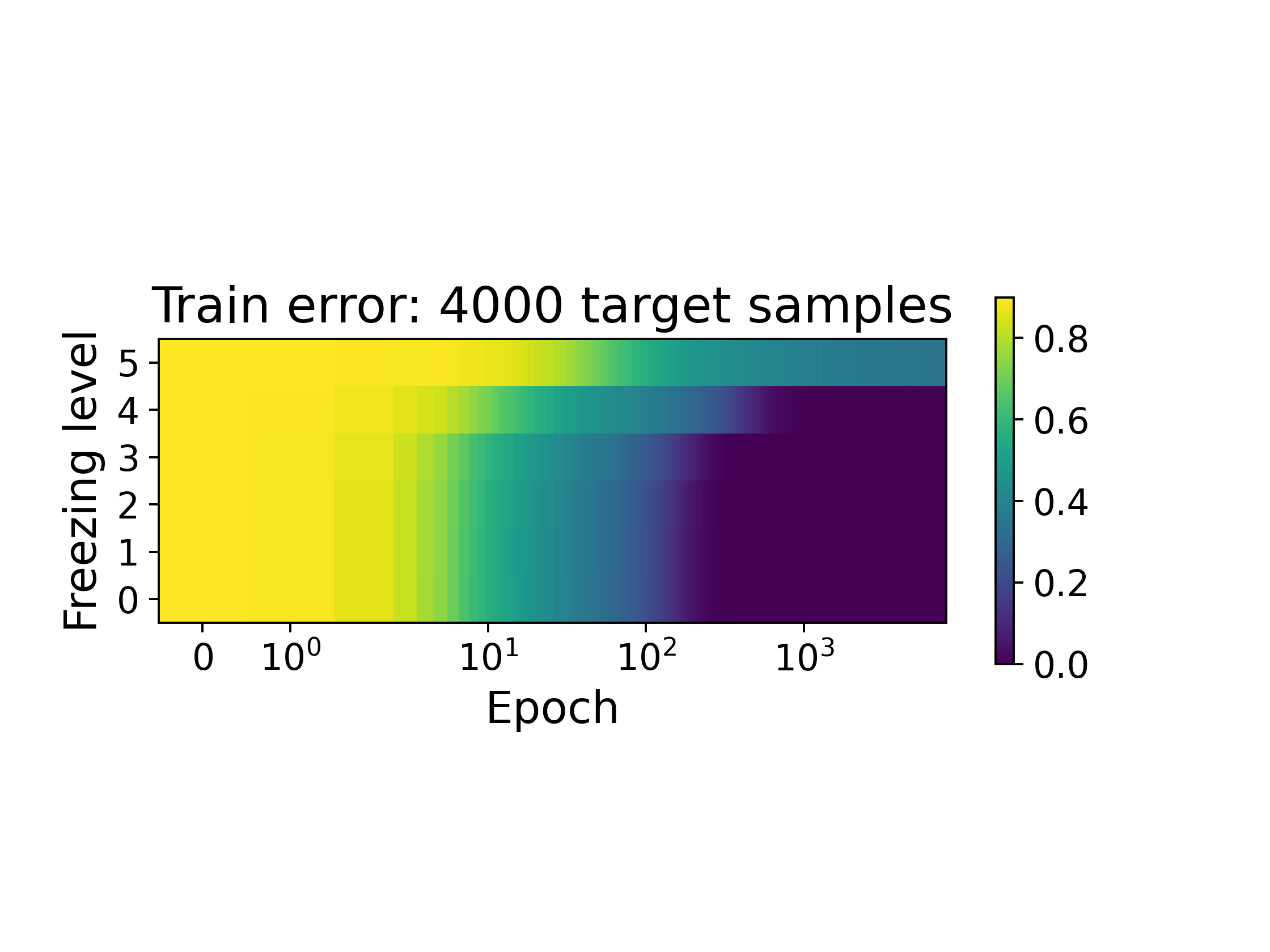}}  
  \caption{Evaluation of transfer learning errors at various freezing levels and training epochs. The target classification task is of \textbf{40 classes from CIFAR-100} with 20\% label noise in target datasets. The subfigures in each row differ in the target dataset size. The architecture is \textbf{DenseNet}. The transfer learning  is from the source task of \textbf{40 Tiny ImageNet classes} that are similar to the target task classes (input image size 32x32x3) with source dataset of 20k training samples. }
  \label{appendix:fig:densenet_2d_error_diagrams_freezing_vs_epochs_40_class}
\end{figure*}

As discussed in Section \ref{subsec:Freezing-wise Double Descent}, the additional results here in Figures \ref{fig:ViT_2d_error_diagrams_freezing_vs_epochs_appendix}, \ref{appendix:fig:ViT_2d_error_diagrams_freezing_vs_epochs_40_class}, \ref{appendix:fig:densenet_2d_error_diagrams_freezing_vs_epochs_40_class} demonstrate that a freezing-wise double descent of the test error (basically, a test error peak around the interpolation threshold of the freezing levels axis) can occur if the target dataset is sufficiently large and the training is sufficiently long (i.e., the epoch number is high enough). Such freezing-wise double descent can be clearly observed in Figures \ref{fig:vit_BSP5_src100k_tinyimagenet32_200Classes_src200classtgtCIFAR10class_tgtdataset640_test_err_w}, \ref{fig:vit_BSP5_src100k_tinyimagenet32_200Classes_src200classtgtCIFAR10class_tgtdataset2560_test_err_w}, \ref{fig:vit_BSP5_src100k_tinyimagenet32_200Classes_src200classtgtCIFAR10class_tgtdataset4000_test_err_w},  \ref{fig:vit_BSP5_src20k_tinyimagenet32_src40SimilarClasstgt40class_tgtdataset2560}, \ref{fig:vit_BSP5_src20k_tinyimagenet32_src40SimilarClasstgt40class_tgtdataset4000}, \ref{fig:densenet_BSP5_src20k_tinyimagenet32_src40SimilarClasstgt40class_tgtdataset640}, \ref{fig:densenet_BSP5_src20k_tinyimagenet32_src40SimilarClasstgt40class_tgtdataset2560}, \ref{fig:densenet_BSP5_src20k_tinyimagenet32_src40SimilarClasstgt40class_tgtdataset4000}.

Whereas Figures \ref{fig:ViT_2d_error_diagrams_freezing_vs_epochs_appendix}, \ref{appendix:fig:ViT_2d_error_diagrams_freezing_vs_epochs_40_class} consider the ViT, Fig.~\ref{appendix:fig:densenet_2d_error_diagrams_freezing_vs_epochs_40_class} shows the 2D error diagrams for the DenseNet. Specifically, Figs.~\ref{fig:densenet_BSP5_src20k_tinyimagenet32_src40SimilarClasstgt40class_tgtdataset640}, \ref{fig:densenet_BSP5_src20k_tinyimagenet32_src40SimilarClasstgt40class_tgtdataset2560}, \ref{fig:densenet_BSP5_src20k_tinyimagenet32_src40SimilarClasstgt40class_tgtdataset4000} demonstrate the existence of the freezing-wise double descent also in DenseNet.

\section{Additional Results for Section 5}
\label{appendix:sec:Additional Experimental Results for Section 5}

\begin{figure*}[t]
  \centering
  \includegraphics[width=0.6\linewidth]{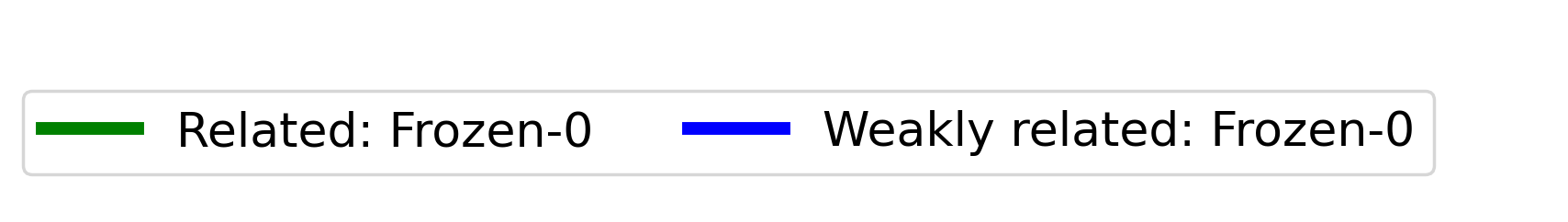}
  \\[-1ex]
    \subfloat[ViT]{
            \includegraphics[width=0.32\linewidth]{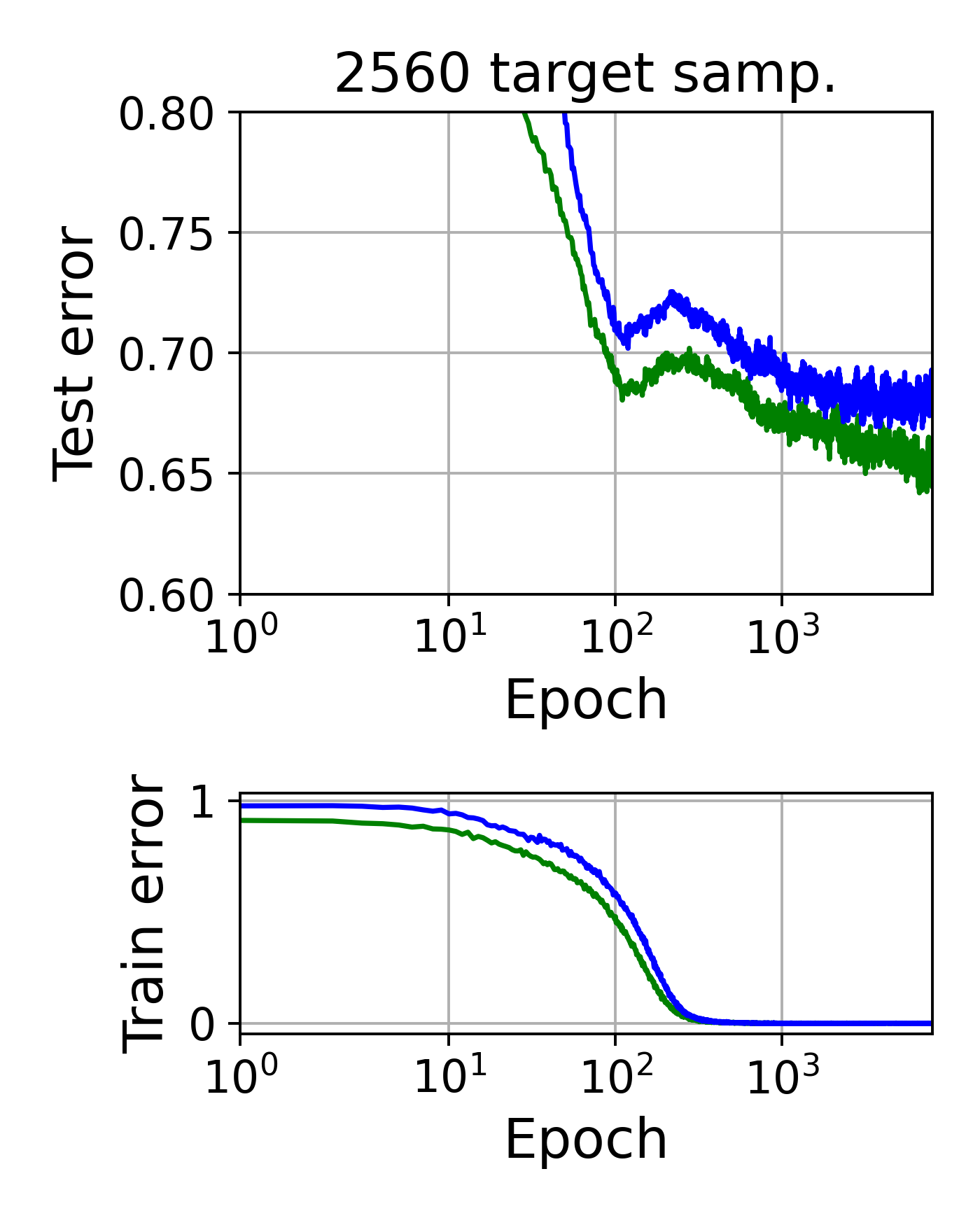}
                \includegraphics[width=0.32\linewidth]{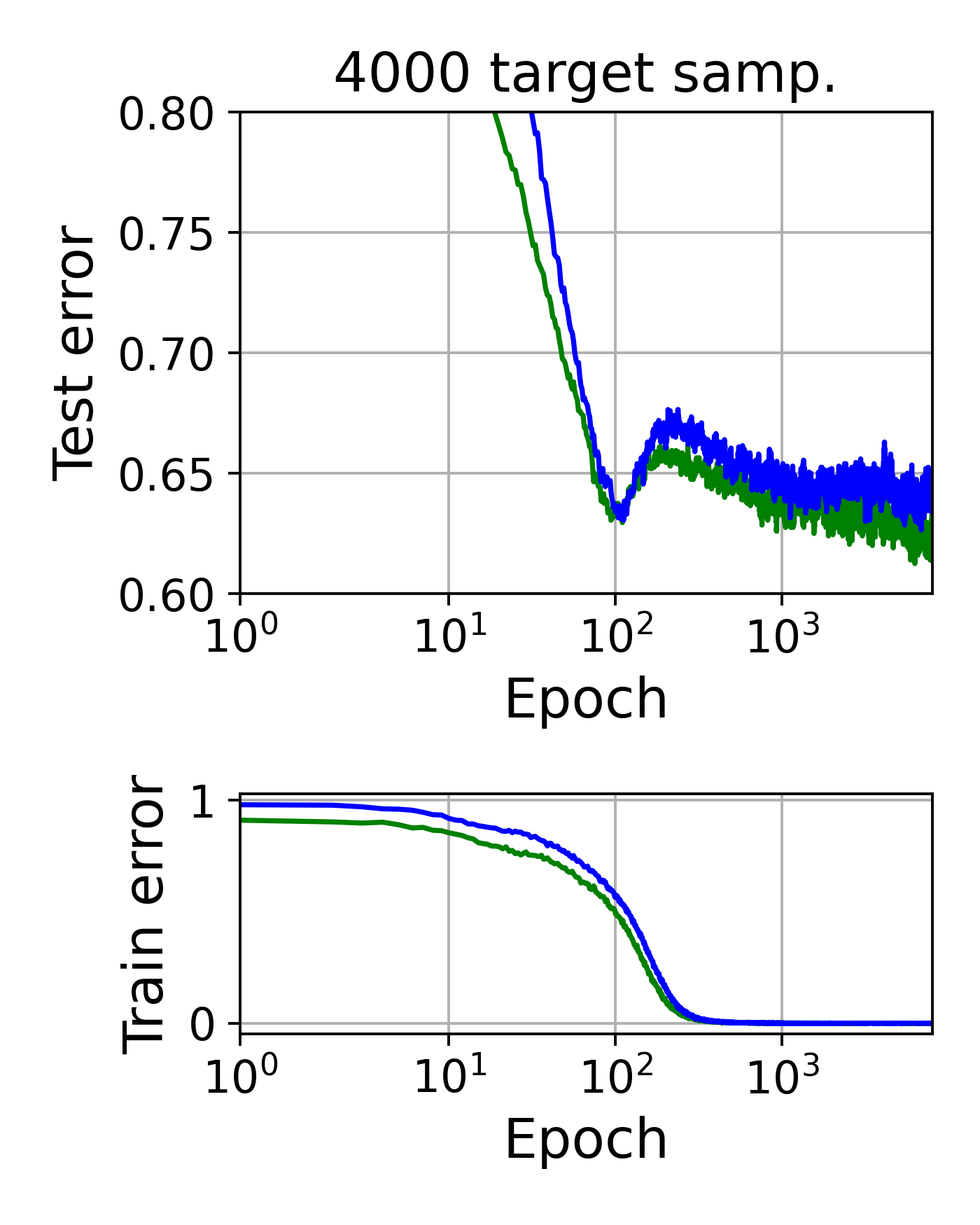}
                    \includegraphics[width=0.32\linewidth]{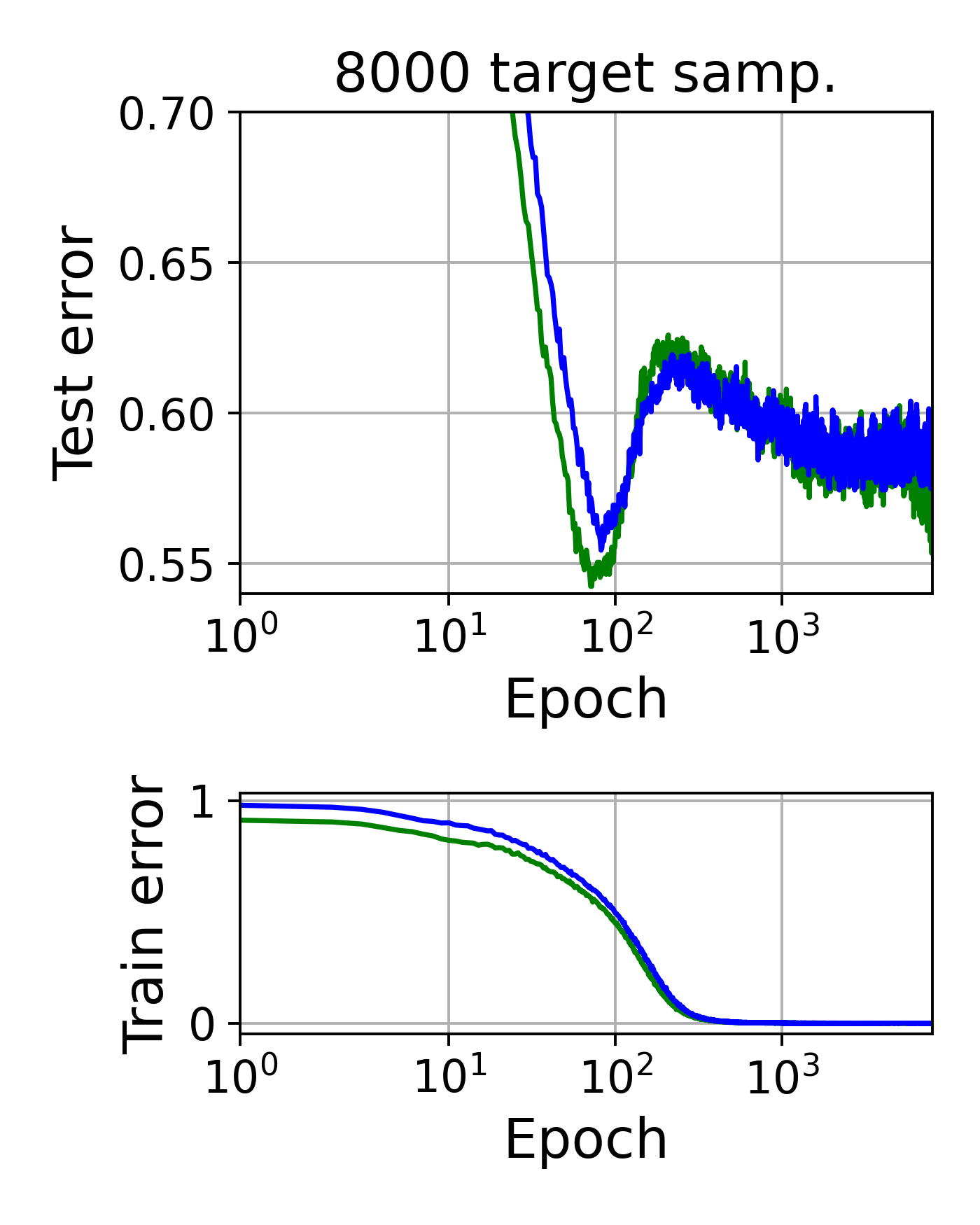}
    \label{subfig:task_similarity_vit_tinyimagenet32First40Classes-vs-tinyimagenet32Similar40Classes__srcNoiselessTgtNoisy0p2} }
    \\
    \includegraphics[width=0.6\linewidth]{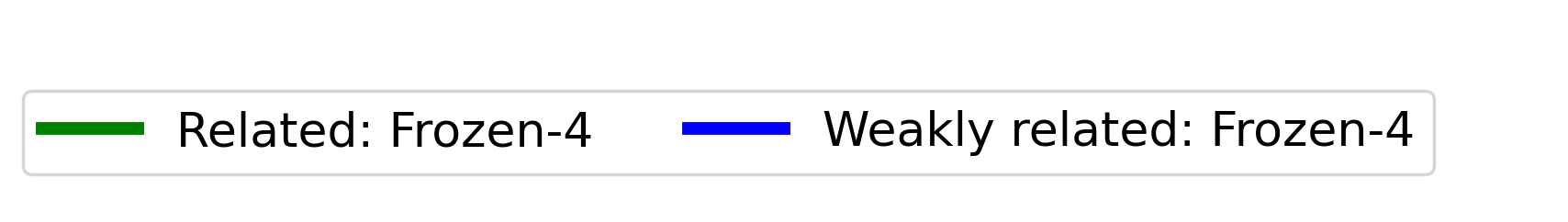}
  \\[-1ex]
    \subfloat[DenseNet]{
            \includegraphics[width=0.32\linewidth]{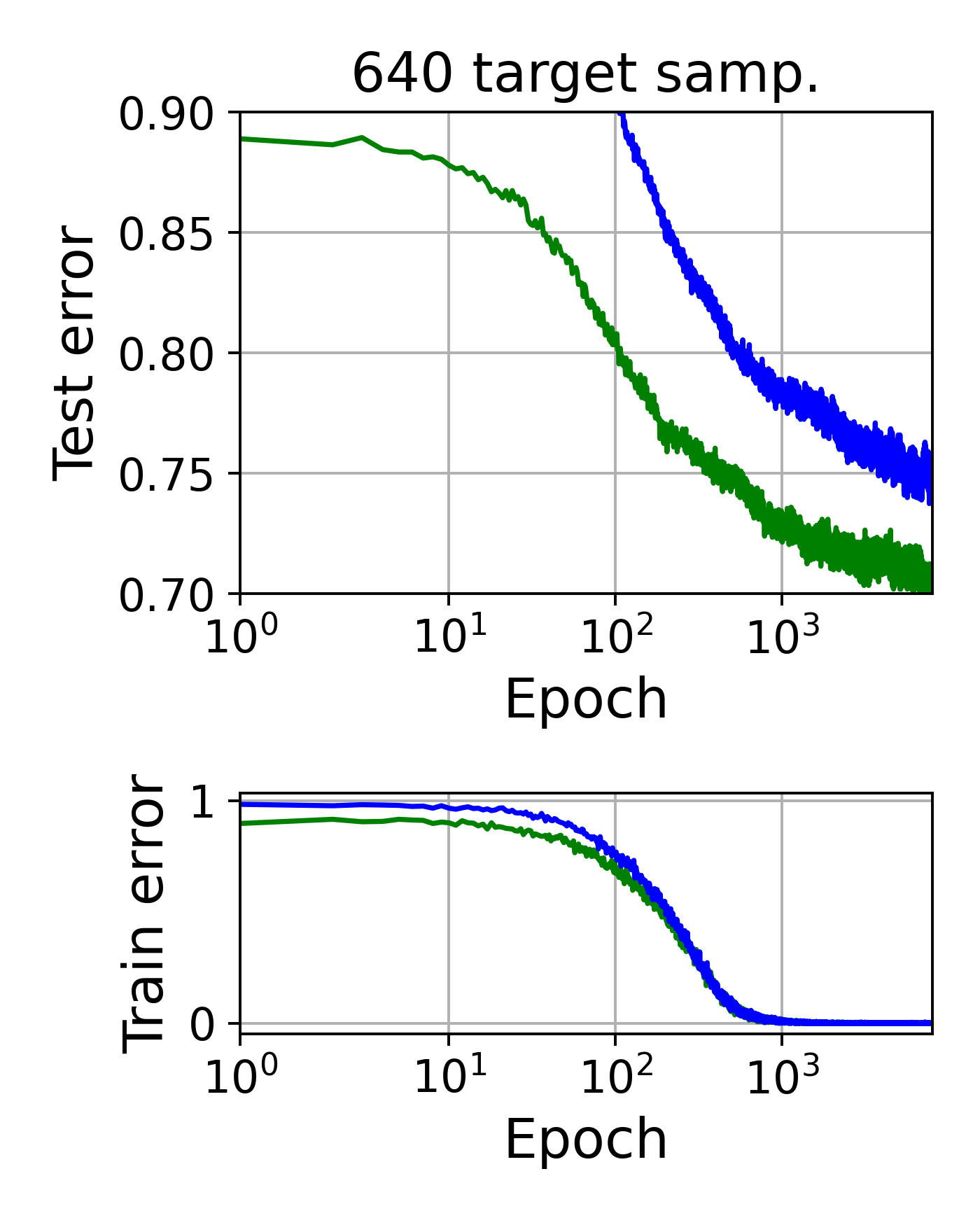}
                \includegraphics[width=0.32\linewidth]{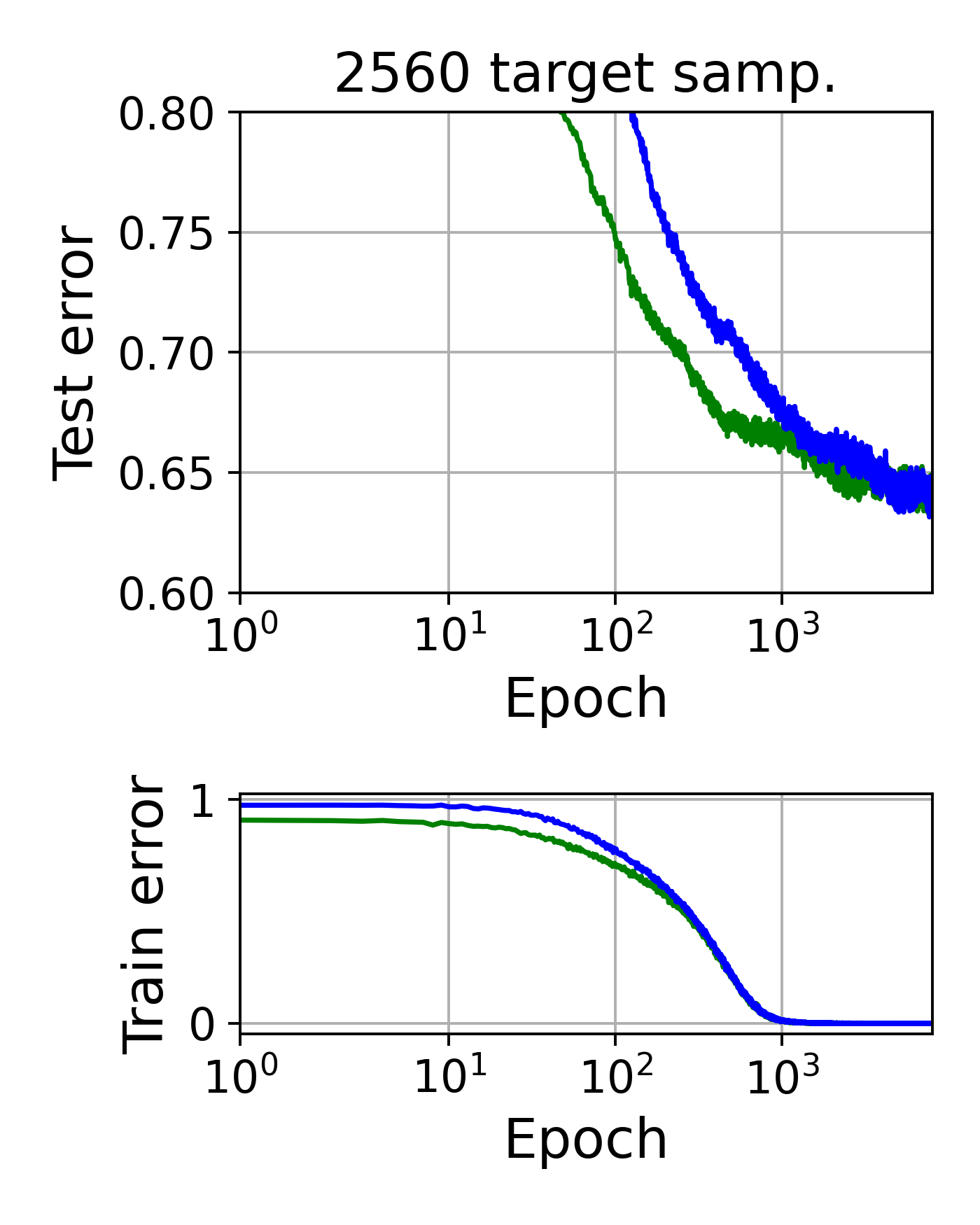}
                    \includegraphics[width=0.32\linewidth]{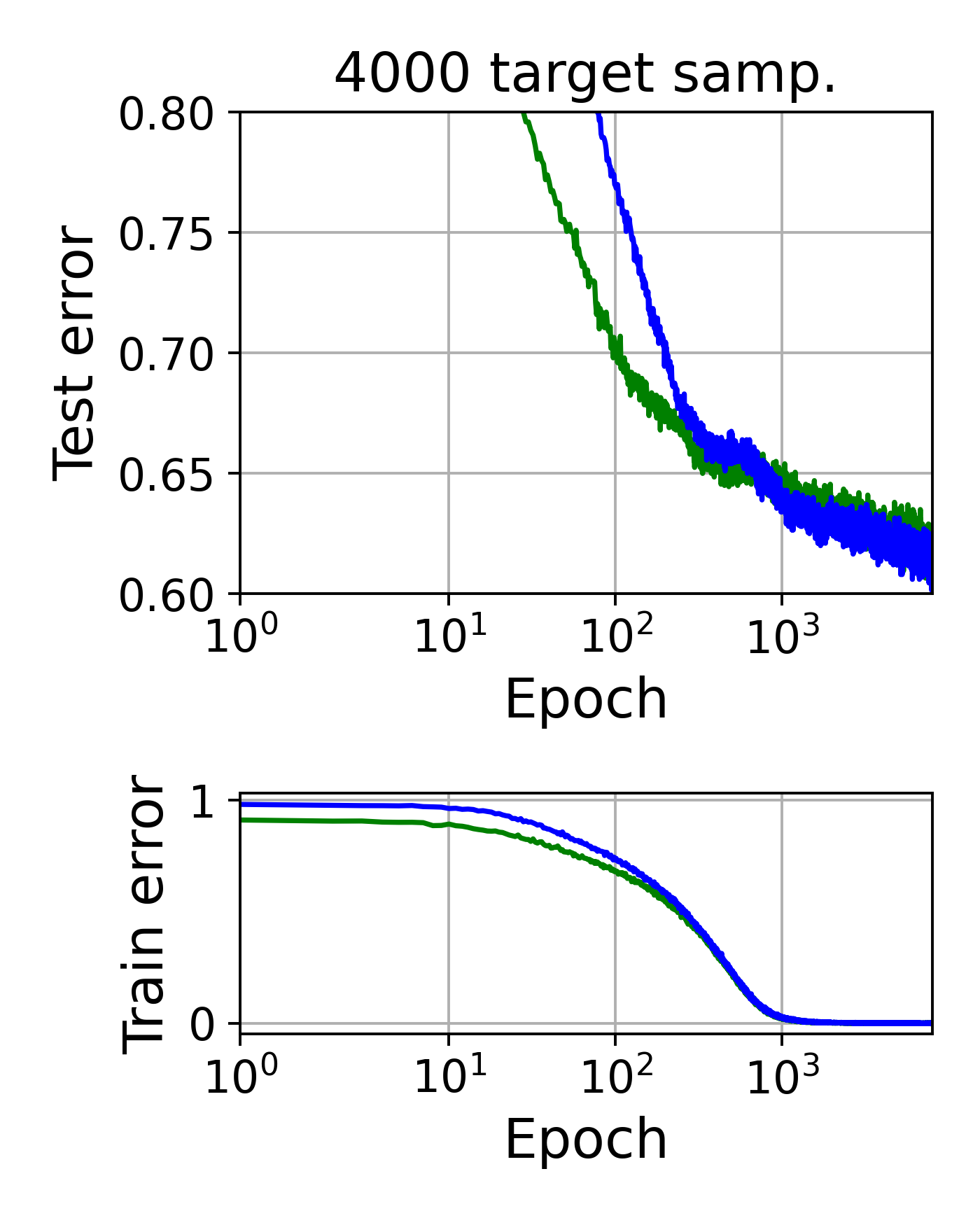}
    \label{subfig:task_similarity_densenet_tinyimagenet32First40Classes-vs-tinyimagenet32Similar40Classes__srcNoiselessTgtNoisy0p2} }
  \caption{Transfer from a more related task can be less beneficial due to double descent. The architectures are (a) ViT, (b) DenseNet. The target task includes 40 classes from CIFAR-100. The comparison is between transfer learning from a related task (40 classes from Tiny ImageNet that are conceptually related to the target task classes) and transfer learning from a weakly related task (40 classes from Tiny ImageNet that are arbitrarily chosen). Target datasets are with 20\% label noise. Each subfigure corresponds to another size of the target dataset. For relatively large target datasets, transfer from a less related task can be on par or better than transfer from a more related task. See Appendix \ref{appendix:sec:Additional Details on the Examined Classification Problems} for a detailed description of the 40 classes in the similar source task.}
  \label{appendix:fig:task_similarity_40_class}
\end{figure*}

In Section \ref{sec:Task Similarity} we examine transfer learning from source tasks at different similarity levels. In Fig.~6 we show the error curves in the learning of a ResNet-18. 
Here, we provide in Fig.~\ref{appendix:fig:task_similarity_40_class} the error curves in the learning of ViT and DenseNet (the considered 40-class tasks are defined in Appendix \ref{appendix:sec:Additional Details on the Examined Classification Problems}). 
These examples show that transfer from a less related task can be on par or better than transfer from a more related task when the target dataset is sufficiently large (see, e.g., the subfigures for the target dataset size of 8000 and 4000 samples in Fig.~\ref{subfig:task_similarity_vit_tinyimagenet32First40Classes-vs-tinyimagenet32Similar40Classes__srcNoiselessTgtNoisy0p2}, Fig.~\ref{subfig:task_similarity_densenet_tinyimagenet32First40Classes-vs-tinyimagenet32Similar40Classes__srcNoiselessTgtNoisy0p2}, respectively).

\end{document}